\pdfoutput=1

\documentclass[11pt]{article}

\usepackage{acl}

\usepackage{times}
\usepackage{latexsym}

\usepackage[T1]{fontenc}

\usepackage[utf8]{inputenc}

\usepackage{microtype}

\usepackage{placeins}

\usepackage{inconsolata}

\usepackage{graphicx}
\usepackage{subcaption} 

\usepackage{tcolorbox}
\usepackage{float} 

\newtcolorbox[auto counter, number within=section]{mybox}[2][]{%
  floatplacement=tbhp, 
  width=\textwidth,
  colback=blue!5!white, 
  colframe=blue!75!black, 
  fonttitle=\bfseries,
  title=Box~\thetcbcounter: #2, 
  #1 
}

\usepackage{amsmath}
\usepackage{amssymb}

\usepackage{multirow}
\usepackage{threeparttable}
\title{What are They Thinking? Delineation, Probing and Tracking of Concepts in LLMs}

\author{
\textbf{Mohamed Abdelwahab\footnotemark[4]}
\quad \textbf{Michelle Yu Collins}
\quad \textbf{Sihan Chen}
\quad \textbf{Yi Cheng Zhao}
\\
\textbf{Zafarullah Mahmood}
\quad \textbf{Jiading Zhu}
\quad \textbf{Soliman Ali}
\quad \textbf{Jonathan Rose}
\\
The Edward S. Rogers Sr. Department of Electrical and Computer Engineering
\\
University of Toronto 
}

\begin{document}
\maketitle

\renewcommand{\thefootnote}{$\mathsection$}
\footnotetext{Corresponding author: \\
\texttt{\href{mailto:mo.abdelwahab@mail.utoronto.ca}{mo.abdelwahab@mail.utoronto.ca}}}
\renewcommand{\thefootnote}{\arabic{footnote}}

\begin{abstract}

As the influence of LLMs expands, it is imperative to gain insight into their decisions.  One way to do that is to develop probes that detect the presence or absence of a broad set of high-level abstract concepts within the embeddings computed in an LLM - which is what we might say a model is ``thinking" about.  Such probes should be low-cost and easily applicable to any LLM, so that monitoring for many concepts is possible during normal operation.

In this paper, we take the first steps towards developing the capability of creating many such probes by defining and executing examples of the key tasks needed: first, the careful \emph{delineation} of a high-level abstract concept through the creation of a dataset with the concept both present and then absent.  Then, the training and testing of a set of linear probes to detect the concept on any layer of an LLM, including an exploration of the complexity of the probe needed.  Finally, we show that such probes can track concepts across larger contexts.  This is done with four separate concepts and three different LLMs.  When this process is scaled to many more concepts, it will create the ability to monitor new models.

\end{abstract}
\section{Introduction} \label{section:introduction}

Large Language Models (LLMs) appear to function as \emph{concept machines}, in that they infer high-level abstract concepts from their input that are then the driving force behind the generated output. Indeed, it is now well understood that many kinds of concepts can be detected in internal embeddings within LLMs using linear probes \citep{probing_origin_1, probing_origin_2}.  These concepts include Parts-of-Speech tags,  verb tense \citep{word_level_probing_tasks, multilingual_probing, probing_constituency_tree, structural_probe_word_probing, probing_individual_neurons, probing_sentence_structure, probing_function_words} as well as time, physical location \citep{space_time_probing}, personal traits \citep{personality_traits}, truth \citep{probing_truth}, and deception \cite{deception_probes}.  
There are many other important concepts whose presence or absence needs to be identified,
and so the goal in this work is to prototype the steps needed to create a broad set of probes that could be used to monitor an LLM.  

The first step is to select a set of high-level abstract concepts that are needed for the downstream task of monitoring.  In these first steps, we select the general goal of human activities, and (somewhat arbitrarily) choose the concepts of ambition, investigation, democracy,
and envy.  The next step is to create a working definition of the term  \emph{concept}, and then of \emph{high-level abstract concept}. Based on the prototype theory \citep{prototype_theory} of concepts, we define a \emph{concept} as an entity characterized by a set of features used to determine membership within it. We informally define \emph{high-level abstract concept} as one that implies broader abstractions that encompass multiple entities. Finally, we use the term `delineation' of a concept to mean the method by which we create labeled examples considered to embody the concept and those that do not.  

\begin{figure*}[ht]
    \centering
    \includegraphics[width=\linewidth]{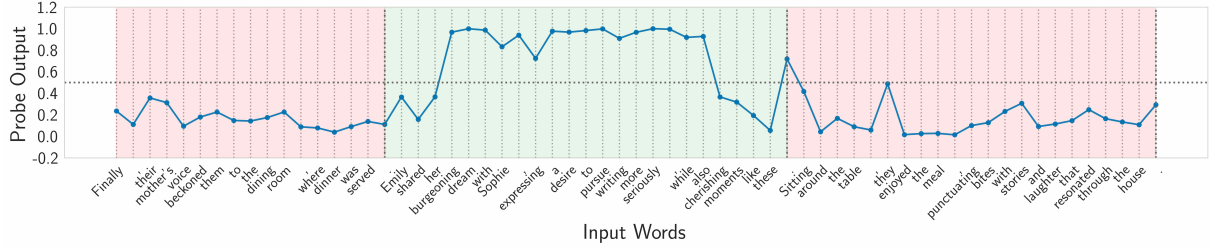}
    \caption{\textbf{Ambition} probe output for every word in an expanding context in layer 13 of \texttt{Llama-3-8B}}
    \label{fig:ambition_word_probing}
\end{figure*}

To illustrate the use of these probes for monitoring, we show how they can track the waxing and waning of concepts in an LLM's embeddings as more words are added to the input context. For example, Figure \ref{fig:ambition_word_probing} illustrates this using a probe trained to detect \textbf{ambition} in the \texttt{Llama-3-8B} model. The probe is applied to the final embedding of an expanding input sequence, taken from the $13^{th}$ transformer layer. The X-axis gives the input tokens (as complete words), and the Y-axis gives the probe's sigmoid output computed on the embedding \emph{after} each new word is added to the sequence. We take a probe output above 0.5 to indicate the presence of the concept, while an output below 0.5 indicates its absence. The colored shading in the figure indicates the label of the \emph{entire} sentence: green for the presence of ambition and red for its absence. The sentences in the figure are drawn from the middle of a three-paragraph story, and it is instructive to read them and compare them with the probe’s outputs. We observe that the probe output rises above 0.5 when the newly added span of words implies ambition, and falls below 0.5 once the continuing context no longer does.

Previous work has explored the detection of concepts in LLMs, including the use of Sparse Autoencoders (SAEs) \citep{LLM_SAE_eleuther, LLM_SAE_anthropic, LLM_SAE_gemma, LLM_SAE_google, LLM_SAE_2_anthropic} to detect a large number of concepts in an LLM. This is done using unsupervised training of an SAE that is tasked with reconstructing an internal embedding from a large (enforced) sparse representation.  This approach creates many concept detectors, but does not have anything equivalent to our stage of \emph{delineation} where the specific concept is chosen and all others are excluded.  SAEs do not permit control over the concepts that can be detected, which is very problematic, especially in matters of trust.  In addition,  training is very computationally expensive and must be repeated for every new model.  
By contrast, our approach builds a delineating dataset only once per concept, enabling inexpensive probe training on \emph{any model} thereafter.

Another approach to concept detection, proposed in \citep{LLM_PCA_concepts}, allows direct specification of a concept by explicitly instructing the LLM to identify a named concept. However, this makes it unsuitable for continuous monitoring, as it requires a separate LLM invocation for detection.
 
In this work, we create probes for four high-level abstract concepts -- \textbf{ambition}, \textbf{investigation}, \textbf{democracy}, and \textbf{envy} -- by constructing a dataset with validated presence/absence labels for each. These labels serve as ground truth for detecting the concepts within the LLM embeddings. This method requires significantly less training than SAEs and allows direct specification of the concepts to study.

The primary contributions of this paper are: (1) We propose a semi-automatic LLM-based method for delineating a concept through the creation of a binary dataset of textual examples for a specified concept. The method seeks to inhibit patterns that may unintentionally give away the labels; (2) We illustrate the utility of the delineation method by training linear probes on the binary datasets and show that these achieve good accuracy.  (3) We show how the size of the probes can be 
constrained to fewer than 80 parameters while still achieving good accuracy.
(4) We show the probes used to observe concepts waxing and waning in the model embeddings as words are added to the input context, illustrating their use as low-cost continuous monitors;  (5) We provide the created datasets for others to use 
\footnote{\url{https://github.com/cimhasgithub/ConceptProbing_TrustNLP2026}}.
Since each dataset can be reused on any LLM to build probes, we believe this motivates a larger effort to select and delineate many such reusable concept datasets to advance research in LLM explainability.

\section{Background and Related Work} \label{section:related_works}

\subsection{Concepts in LLMs}

Several works explore the existence of concepts in LLMs. \citet{LLM_concept_hierarchy_1} and \citet{LLM_concept_hierarchy_2} investigate LLMs' knowledge of concept hierarchies. \citet{LLM_concept_hierarchy_1} use zero-shot prompting \citep{GPT3_paper}, directly asking the model to generate an answer on whether one concept is within the broader category of another. In contrast, \citet{LLM_concept_hierarchy_2} present the model with statements expressing conceptual relationships and probe its embeddings to assess the validity of those relations.

\citet{LLM_PCA_concepts} explore concepts in LLMs by extracting embeddings after prompting the model to identify a specific concept in an input example. They apply PCA \citep{PCA_original_1, PCA_original_2} to these embeddings to derive a `concept vector.' To detect the concept in new inputs, they use the same prompt to extract an embedding and measure its alignment with the concept vector using dot product.
However, this approach is unsuitable for continuous concept monitoring, since detection requires a separate LLM invocation.

\citet{LLM_SAE_eleuther, LLM_SAE_anthropic, LLM_SAE_gemma, LLM_SAE_google, LLM_SAE_2_anthropic} use SAEs to disentangle LLM embeddings into many dimensions corresponding to individual concepts. SAEs undergo unsupervised training, so concepts are identified post hoc using an automated LLM-based method \citep{openai_neuron_explanation}. As such, the extracted concepts cannot be specified in advance, offering no guarantee of cross-model consistency, and SAEs demand large-scale training (>1B examples per SAE). There is also no guarantee of disentanglement. As a result, SAEs are not suitable for dedicated investigations into LLMs' inference of concepts.

Our work proposes a probing-based approach to perform concept detection in LLMs using datasets constructed for specific concepts. This method enables exploration of these concepts across models, requires far less training than SAEs, and minimizes the risk of cueing the model toward the target concept. It also enables low-cost  monitoring of how concepts wax and wane in a model's embeddings as the context expands during generation. 

\subsection{Linear LLM Probes}

LLM probes have been used to study the properties an LLM has acquired during training by using a separate model that makes a prediction given an LLM embedding. The probe is usually a classifier model that is trained to detect a specific property. If the probe achieves a reasonable accuracy, it suggests that the property is encoded within the LLM.

Probes were first applied to early transformer models, such as BERT \citep{BERT_paper}, to explore whether they encoded linguistic properties such as Parts-of-Speech tags and main verb tense \citep{word_level_probing_tasks, multilingual_probing, probing_constituency_tree, structural_probe_word_probing, probing_individual_neurons, probing_sentence_structure, probing_function_words}. More recently, probes have been used to explore the encoding of time, space, and truth within LLMs \citep{space_time_probing, probing_truth}. 


\section{Detection and Tracking of Concepts within LLMs}\label{section:method}

We use linear probe classifiers to detect concepts in LLMs, and to illustrate how these concepts wax and wane in a model's embeddings as the input context expands.  In this way, they can be used for monitoring the LLMs for ``thinking'' about a specific concept.

\subsection{Inference of Concepts in LLMs}
The training of a probe requires a binary dataset of textual examples for the concept. The concept should be \emph{present} in positive examples and \emph{absent} in negative ones. 
These examples are fed into an LLM to extract embeddings, which are then used to train and evaluate a probing classifier.

\subsubsection{Concept Dataset Creation} \label{section:sub-objective_1_dataset_creation}

\begin{table*}[ht]
    \centering
    \begin{tabular}{|p{0.18\linewidth}|p{0.7\linewidth}|}
        \hline
        Example template & "I want the girls to understand this," said Miss Anstice with decision. \\ 
        \hline
        Positive Example & \multirow{2}{=}{"I aim to reach a million followers this year," announced Jake confidently.} \\ 
        (Concept Present) & \\
        \hline
        Negative Example & \multirow{2}{=}{"I need the followers to see this," posted Sarah with certainty.} \\ 
        (Concept Absent) & \\
        \hline
    \end{tabular}
    \caption{Example-pair created for \textbf{Ambition} dataset using an example template}
    \label{table:dataset_creation_example}
\end{table*}

When creating a binary dataset for a concept, the goal is to ensure that the positive and negative examples differ solely in the presence or absence of the concept, without unintentionally including any patterns that could ``give away'' the label \citep{confounders_3, confounders_2, confounders_1}. This is achieved by using textual example templates that were created independently of any particular concept. We then generate a pair of examples following the linguistic structure of the template, where the concept is present in one example and absent in the other. Table \ref{table:dataset_creation_example} shows an example template and a positive-negative example pair generated from it; additional examples are provided in Appendix \ref{appendix:textual_examples}.


\begin{figure*}[ht]
    \centering
    \includegraphics[width=\textwidth]{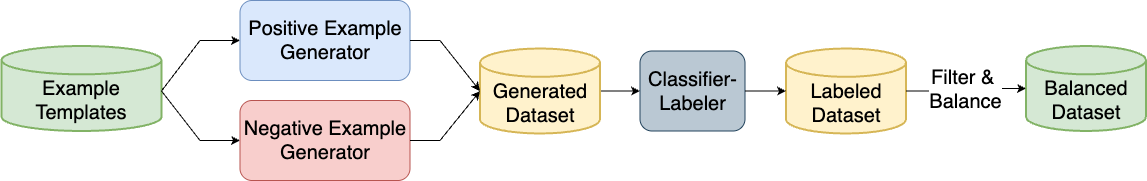}
    \caption{Creation of a concept dataset while limiting label leakage}
    \label{fig:dataset_creation_framework}
\end{figure*}

We obtain templates from a dataset based on Project Gutenberg \citep{manu_project_gutenberg, project_gutenberg}, a free eBook library. Paragraphs from English books are split into sentences, and templates are formed by randomly selecting one to three consecutive sentences. We remove templates with incomplete sentences, misplaced words or numbers, or those not centered on human subjects, since we focus on human-related concepts in this work. A prompted LLM\footnote{We use \texttt{gpt-4o-2024-08-06} \citep{openai_gpt4o} for all prompted generators and classifiers.} is used
for filtering. The prompt used, as well as all other prompts employed throughout this paper, are provided in Appendix~\ref{appendix:dataset_creation}. We created a set of 30,000 templates, which can be reused to create any specific concept dataset.

To create a dataset for a specific concept, the concept is first defined as having specific features, following the definition of \emph{concept} in Section~\ref{section:introduction}. This definition is then provided to two instances of the same LLM: one prompted to generate positive (concept present) examples, and the other to generate negative (absent) examples. The prompts instruct the model to mimic the structure of the given example templates while altering the semantics to reflect the presence or absence of the concept. We diversify the generated examples by instructing the LLM to generate the examples within a specific context that is changed every five examples. The contexts include \emph{workplace}, \emph{academia}, \emph{sports}, \emph{entrepreneurship}, \emph{politics}, \emph{arts}, \emph{music}, \emph{community}, \emph{science}, \emph{technology}, and \emph{social media}. Details on the prompt refinement are provided in Appendix~\ref{appendix:dataset_creation}. 

Because the generated examples may not reliably match their intended labels, and manual verification at scale is impractical, a prompted LLM-based classifier is used to re-label all examples for greater reliability. The classifier is instructed to follow the concept definition, and the prompt is iteratively refined until it exceeds 90\% accuracy on at least 360 manually labeled examples. During manual labeling, human annotators also follow the concept definition. Appendix~\ref{appendix:dataset_creation} details the labeling process and inter-rater reliability. Figure~\ref{fig:dataset_creation_framework} illustrates the generation and labeling workflow.

To ensure a balanced dataset, we filter it to retain only example pairs with opposite classifier-assigned labels. This prevents the probe from exploiting structural similarities between examples that share the same label. 

Four datasets were created using this method for the concepts of \textbf{ambition} (with 11,854 examples), \textbf{investigation} (8,296), \textbf{democracy} (10,270), and \textbf{envy} (15,350). The specific definitions of these concepts are provided in Appendix \ref{appendix:dataset_creation}. These concepts were selected because they span various concept categories, as discussed in Appendix \ref{appendix:concept_selection}. A 70/10/20 train/validation/test split is used for probe training and evaluation. 

We recognize that one might argue that an LLM-prompted classifier, which labels the presence or absence of a concept, 
could directly be used for concept monitoring.
This is not practical for two reasons: first, it is a computationally expensive monitor that would not be practical to use for many monitors.  Second, the prompted classifier would only be inferring the `thoughts' of the target model from its language, and not the actual concepts the target was traversing, such as when the model is trying to mislead \citep{mislead_anthropic}. 

\subsubsection{Probe Training for a Concept} \label{section:sub-objective_1_probing}

To train a probe for a specific concept, each example in the dataset is input into the LLM. The embeddings are extracted from every layer of the model to explore where the concept may appear. To clarify the extraction process of the embeddings, we formalize it as follows:
\\
A textual example is tokenized into a sequence $X = \{x_1, x_2, \dots, x_N\}$, where $x_i$ denotes the $i$-th token and $N$ is the total number of tokens. Each token is mapped to an embedding $e^0_i \in \mathbb{R}^{d_{model}}$ via the embedding layer, where $d_{model}$ is the LLM’s embedding size. These embeddings are then processed through the model’s transformer layers, with layer $\ell$ producing the set $E^{\ell} = \{e^{\ell}_1, \dots, e^{\ell}_N\}$.

We train the probe using one of two vectors for each layer $\ell$: (1) the $N^{th}$ embedding $e^{\ell}_N$, which encodes information for the entire context up till token $N$, or (2) the average of all embeddings in $E^{\ell}$. We refer to either as the \emph{representative embedding}.

After obtaining a representative embedding for each example in the labeled dataset, it is assigned the label of the corresponding input text. The training split of this (embedding, label) dataset is used to train a linear probe \citep{probing_origin_1, probing_origin_2}--a binary linear classifier--to predict the presence or absence of the concept. The probe, trained via gradient descent, is evaluated on the test set. It contains $d_{model}$ parameters.

A common criticism of probing is that the probe may learn the target feature \emph{itself} rather than detect its existence in the model’s embeddings \citep{probing_control_1, probing_control_2, probing_control_3}. To test for this, the literature suggests control tasks such as training probes on randomized data. Here, we randomize the training and validation sets--either the embeddings\footnote{We randomize embeddings by randomizing the input tokens to the LLM.} or the labels--and then evaluate the probe on the original test set. A sharp accuracy drop with randomized embeddings indicates that the probe depends on information encoded in the embeddings. Likewise, a drop with randomized labels suggests that the probe’s performance relies on a meaningful mapping between embeddings and labels, rather than a superficial one.

Another control task involves reducing the probe’s learning capacity by reducing its number of parameters. To do so, we apply PCA to reduce the dimensionality of the embeddings(from 1 up to $d_{model}$)\footnote{Using principal components derived from the embeddings of the example templates.}, thereby reducing the parameters of the linear probe, which are equal to the embedding size. The probe’s accuracy is evaluated at each dimensionality level; if it remains largely unchanged despite substantial compression, this indicates that the embeddings encode the feature, not the probe.

\subsection{Waxing and Waning of Concepts}

In Section \ref{section:obj1_results} below, we confirm that LLMs can infer concepts. We were also interested to see if the same trained probes could be used to examine whether and how a concept waxes and wanes in the model embeddings as the input context grows. This analysis requires longer-text datasets to which the probes are applied. These will be referred to as the \emph{Story Datasets}.

\subsubsection{Story Dataset Creation}
\label{section:sub-objective_2_dataset_creation}

For each concept, a dataset of three-paragraph stories is constructed. Each paragraph contains 10 sentences, and each pair of paragraphs is connected by a transition sentence, totaling 32 sentences per story. The studied concept is present only in the transition sentences and is absent elsewhere. The stories are generated using a prompted LLM.

The generator prompt defines the target concept
and instructs the LLM to generate a story in which the concept is initially absent. It then instructs the model to insert transition sentences between the paragraphs, 
each having the concept present.
The prompt also emphasizes maintaining semantic coherence throughout. Appendix~\ref{appendix:story_creation} provides further details on the generation process.

The prompted classifier, described in Section~\ref{section:sub-objective_1_dataset_creation}, is used to label each sentence in the story individually, retaining only stories where the concept appears only in the transitions. 
A total of 50 stories were created for each target concept.

\subsubsection{Probing for Waxing and Waning} \label{section:sub-objective_2_probing}

Each story is input to the LLM and, for each word, we obtain a representative embedding which encodes the input context \emph{up to and including} that word. Let the story consist of words (including punctuation) $W = \{w_1, \dots, w_{S}\}$, tokenized into \emph{subword} tokens $X = \{x_1, \dots, x_{S'}\}$, where $S' \geq S$. At each layer $\ell$, the LLM produces embeddings $E^{\ell} = \{e^{\ell}_1, \dots, e^{\ell}_{S'}\}$. The \emph{representative embedding} for word $w_i$ will either be the embedding of its final subword token, $e^{\ell}_{i'}$, or the cumulative mean of all embeddings up to $e^{\ell}_{i'}$.

The representative embeddings in each layer are classified by the corresponding probe for the target concept. This yields a probe output vector $P = \{p_1, \dots, p_{S}\}$, where each $p_i \in [0,1]$ indicates the concept’s presence (above 0.5) or absence (below 0.5) in the embedding that encodes the context up to and including word $w_i$. We can observe these word-level outputs and compare them to the sentence-level labels of the story, as shown in Figure~\ref{fig:ambition_word_probing}. This allows us to assess how the LLM’s encoding of the concept evolves as the story unfolds, and whether the probe output aligns with the sentence labels.

\section{Experiments and Results} \label{section:experiments}
\subsection{Inference of Implied Concepts} \label{section:obj1_experiment}
\subsubsection{Experimental Setup} \label{section:obj1_experimental_setup}

To explore concept inference in LLMs, we tested seven open-source models from three model families: \texttt{Llama-3-8B} \citep{llama3_paper}, \texttt{Gemma-2} (\texttt{2B}, \texttt{9B}) \citep{gemma2_paper}, and \texttt{Qwen2.5} (\texttt{0.5B}, \texttt{1.5B}, \texttt{3B}, \texttt{7B}) \citep{qwen_2p5}. All models were accessed through the Hugging Face Transformers library \citep{huggingface_transformers}. The details for these models are given in Appendix \ref{appendix:LLM_details}.

The concepts under test are \textbf{ambition}, \textbf{investigation}, \textbf{democracy}, and \textbf{envy}, which are bolded throughout to avoid ambiguity. Datasets were created for each, as described in Section~\ref{section:sub-objective_1_dataset_creation}.

These datasets were used to produce representative embeddings\footnote{Using a single Nvidia A100 GPU with 40GB VRAM} (described in Section \ref{section:sub-objective_1_probing}) which were then used to train 5 probes (with different seeds) per layer, averaging their results. For the representative embedding, we experimented with both the $N^{th}$ embedding and the mean of all embeddings on a given layer. Appendix \ref{appendix:probe_training_specs} describes the probes' hyperparameter settings.

We primarily present results for the concept of \textbf{ambition}, with similar results for the other concepts provided in Appendix~\ref{appendix:exp1_extended}. Since all LLMs show similar patterns, we focus on \texttt{Llama-3-8B}, reporting results for other models in Appendix~\ref{appendix:exp1_extended}, unless otherwise noted. \texttt{Llama-3-8B} has a $d_{model}$ size of 4,096 and 32 transformer layers.

\subsubsection{Results and Discussion} \label{section:obj1_results}

\begin{figure}[h]
    \centering
    \includegraphics[width=\linewidth]{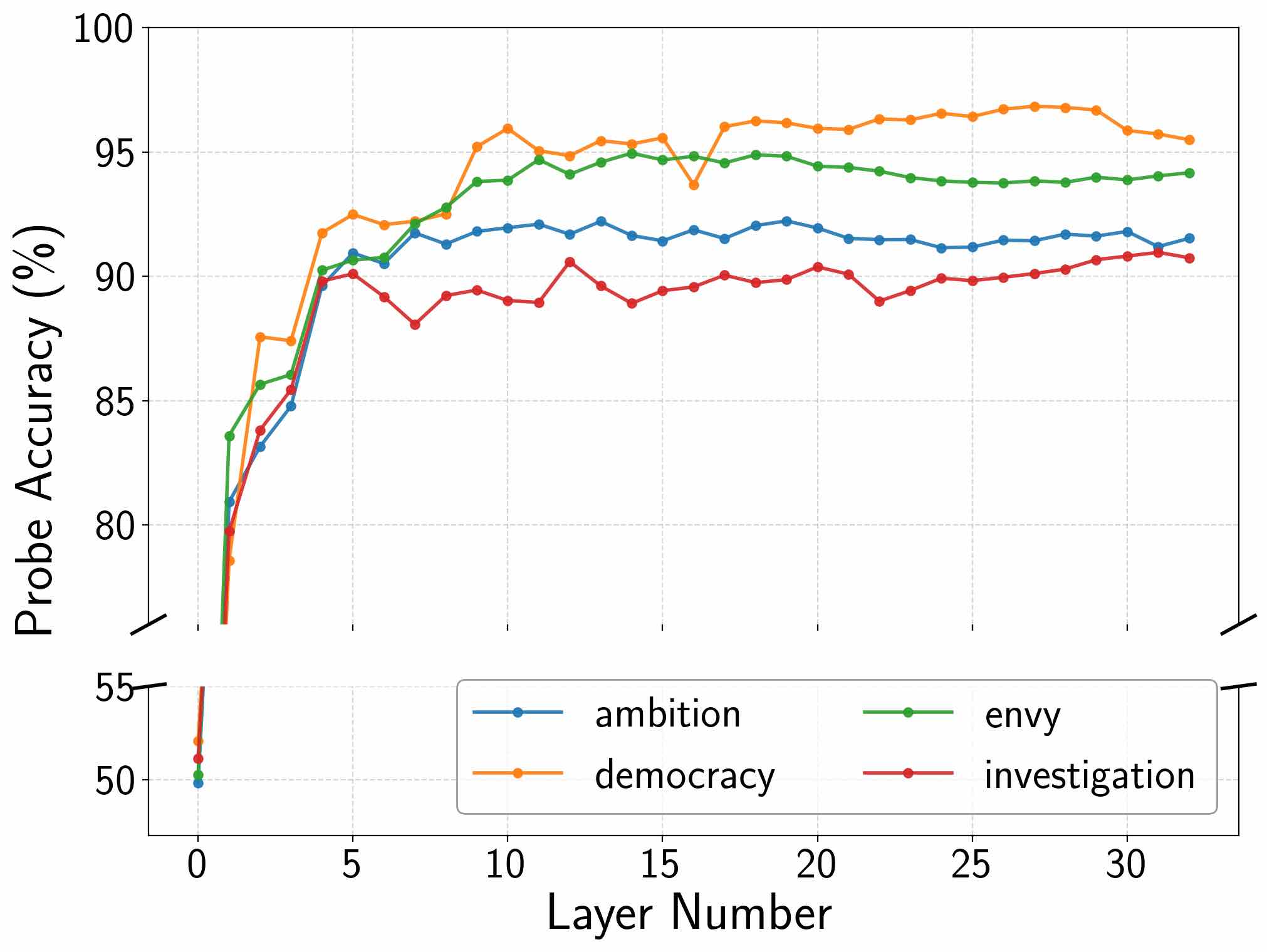}
    \caption{Concept probe accuracy across layers in \texttt{Llama-3-8B} for all 4 concepts}
    \label{fig:concepts_llama}
\end{figure}

Figure \ref{fig:concepts_llama} shows a plot of the probe accuracy for each concept under test across all layers of \texttt{Llama-3-8B}. Here, the probes were trained and tested on the $N^{th}$ embedding on each layer. The Y-axis is broken at the bottom to include all data points while maintaining the visibility of trends. All the results are well above 50\% (except for the embedding layer, which is close to 50\%) providing evidence that the LLM infers these concepts. We had expected that probing the embedding layer (layer 0) would be unsuccessful, as the probe would only see the uncontextualized embedding of the one final token, which is clearly not sufficient to encode a concept.

\begin{figure}[h]
    \centering
    \includegraphics[width=\linewidth]{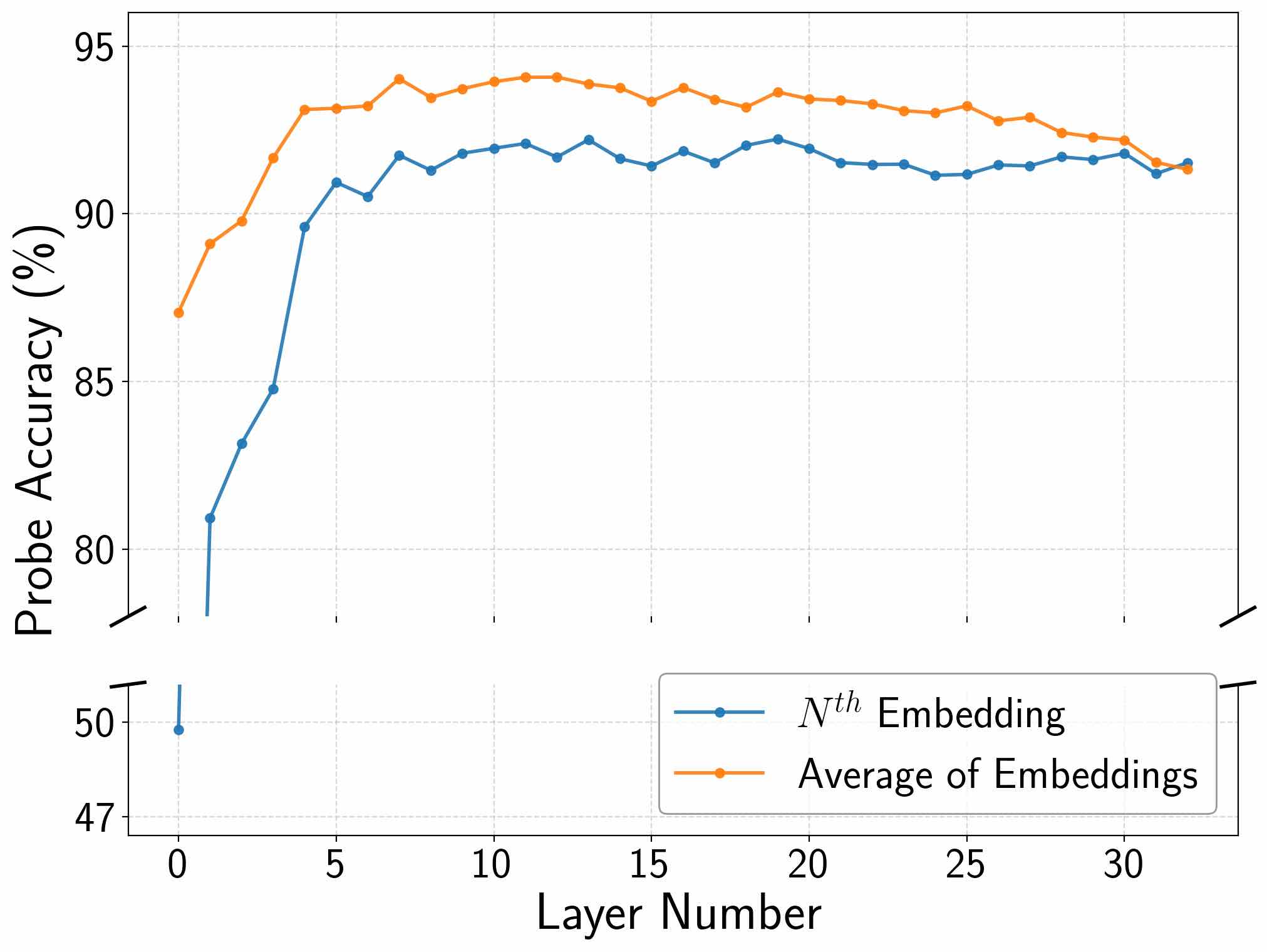}
    \caption{\textbf{Ambition} probe accuracy for \texttt{Llama-3-8B} using average and $N^{th}$ embeddings vs. layer}
    \label{fig:ambition_llama_right_most_vs_average}
\end{figure}

Figure \ref{fig:ambition_llama_right_most_vs_average} plots the probe accuracy versus layer number for \textbf{ambition} in \texttt{Llama-3-8B}, comparing the two types of representative embeddings described in Section \ref{section:sub-objective_1_probing}: the full-layer average embedding and the $N^{th}$ embedding. Interestingly, the average embedding outperforms the $N^{th}$ embedding across most layers, with the performance gap narrowing in the deeper layers.

It is interesting to observe that the average embedding from the embedding layer (layer 0) achieves 87\% accuracy. These embeddings are produced during the language model's training but are typically regarded as \emph{uncontextualized}, in contrast to the \emph{contextualized} outputs of transformer layers. It seems that concepts can be detected in LLMs with simple contextualization---averaging---much like in a bag-of-words model. While this works for small contexts of a few sentences, we show in Section~\ref{section:obj2_results} that it does not work for longer contexts, as intuition suggests, and so the model computation is necessary to extract the concept.

\begin{figure}[h]
    \centering
    \includegraphics[width=\linewidth]{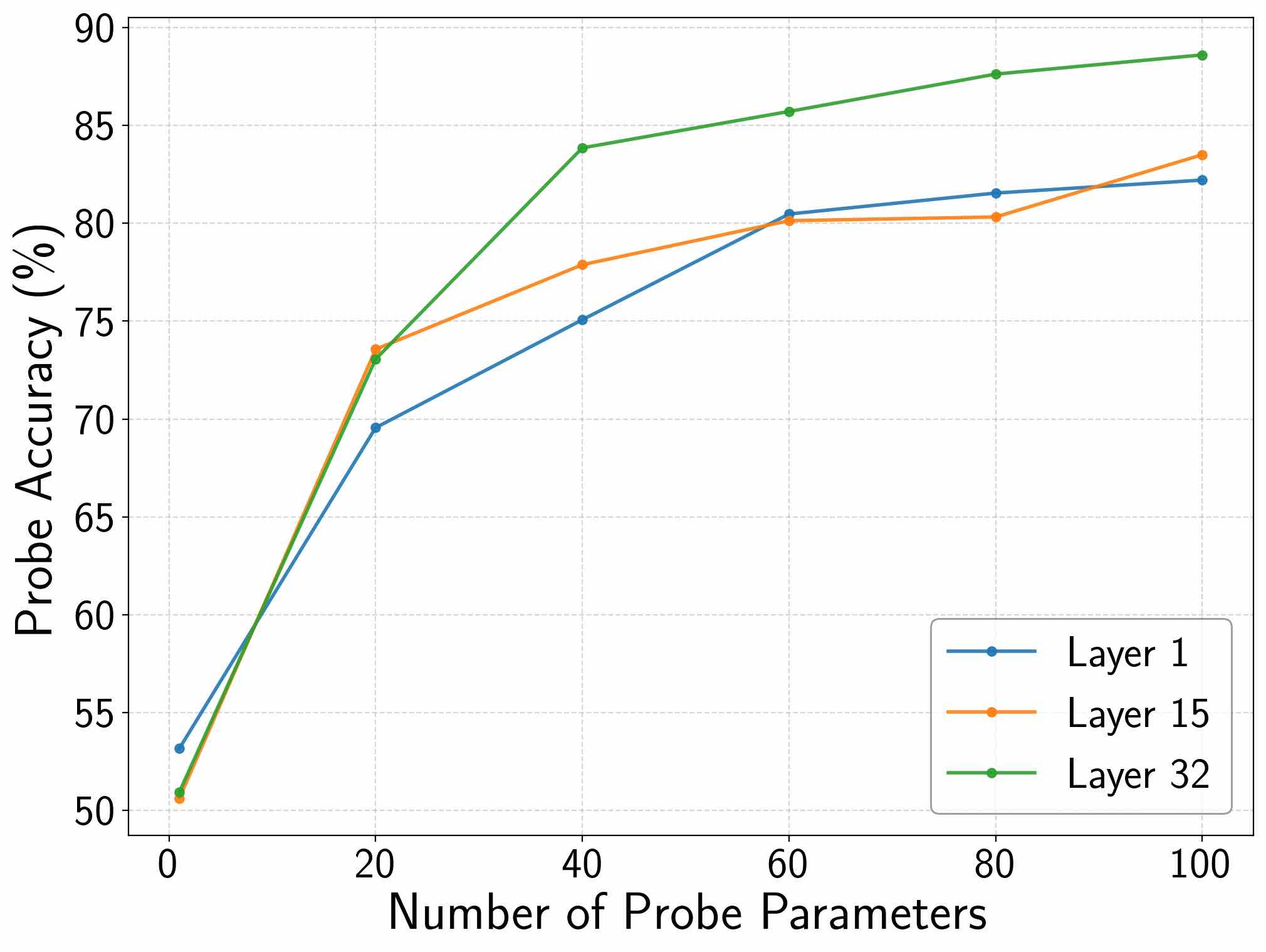}
    \caption{\textbf{Ambition} probe accuracy for \texttt{Llama-3-8B} as a function of probe size}
    \label{fig:ambition_llama_vs_probe_params}
\end{figure}

To test whether the probe itself learns the concept, we apply the control tasks described in Section \ref{section:sub-objective_1_probing}. We first explore probe size reduction in Figure \ref{fig:ambition_llama_vs_probe_params} which plots \textbf{ambition} probe accuracy versus the number of probe parameters. The probe is applied to the $N^{th}$ embedding in three layers of \texttt{Llama-3-8B}. Even with only 40 probe parameters, the accuracy is at least 75\%, well above random guessing, and exhibits diminishing returns toward 100 parameters, well below the maximum of 4,096 for \texttt{Llama-3-8B}. Similar trends hold for other models of different sizes, summarized in Table \ref{table:ambition_vs_params}. Accuracy drops by roughly 15\% with 20 parameters and 10\% with 40, while performance gains diminish beyond 80 parameters.

In the second control task, we randomized either the embeddings or the labels and evaluated the resulting probe accuracy. In both cases, accuracy dropped to around 50\% for all layers in all LLMs (Appendix \ref{appendix:exp1_extended}). This suggests that the probe’s performance depends both on the information encoded in the embeddings and on having a meaningful, rather than superficial, mapping between embeddings and labels. We believe that this demonstrates that the probes are not learning the concepts themselves.

\begin{table}[htbp]
\centering
\begin{threeparttable}
\begin{tabular}{cc|cccc}
\hline\hline
\multirow{2}{*}{Probed LLM} & \multirow{2}{1cm}{Probed Layer} & \multicolumn{4}{c}{\# Probe parameters} \\
    & & 20 & 40 & 80 & max \\
\hline\hline

\multirow{3}{*}{\texttt{Llama-3-8B}} & 1 & 70 & 75 & 82 & 81 \\
                            & 15 & 74 & 78 & 80 & 91 \\
                            & 32 & 73 & 84 & 88 & 92 \\

\hline
                            
\multirow{3}{*}{\texttt{Gemma-2-2B}} & 1 & 63 & 73 & 81 & 84 \\
                            & 15 & 77 & 80 & 83 & 91 \\
                            & 26 & 74 & 82 & 86 & 90 \\

\hline

\multirow{3}{*}{\texttt{Qwen2.5-0.5B}} & 1 & 65 & 67 & 74 & 69 \\
                            & 15 & 76 & 78 & 82 & 89 \\
                            & 24 & 75 & 80 & 85 & 89 \\

\hline\hline
\end{tabular}
\begin{tablenotes}
\footnotesize
\item[$\bullet$] All results are in percentage (\%).
\item[$\bullet$] ``max'' denotes 4,096 for \texttt{Llama-3-8B}, 2,304 for \texttt{Gemma-2-2B}, and 896 for \texttt{Qwen2.5-0.5B}.
\item[$\bullet$] standard deviation for each result $\leq$ 1\%.
\end{tablenotes}
\end{threeparttable}
\caption{\textbf{Ambition} probe accuracy across model families, sizes, layers, and probe sizes} 
\label{table:ambition_vs_params}
\end{table}

\subsection{Tracking Concepts across LLM Context} \label{section:obj2_experiment}

Our secondary goal is to investigate how an encoded concept evolves as the input context expands. To do so, we use the same probes\footnote{We use a single trained probe per model layer per concept, rather than the five probes used in Section \ref{section:obj1_experiment}.} that were trained for the concepts studied in Section \ref{section:obj1_experiment}.

\begin{figure*}[h]
    \centering
    \includegraphics[width=\linewidth]{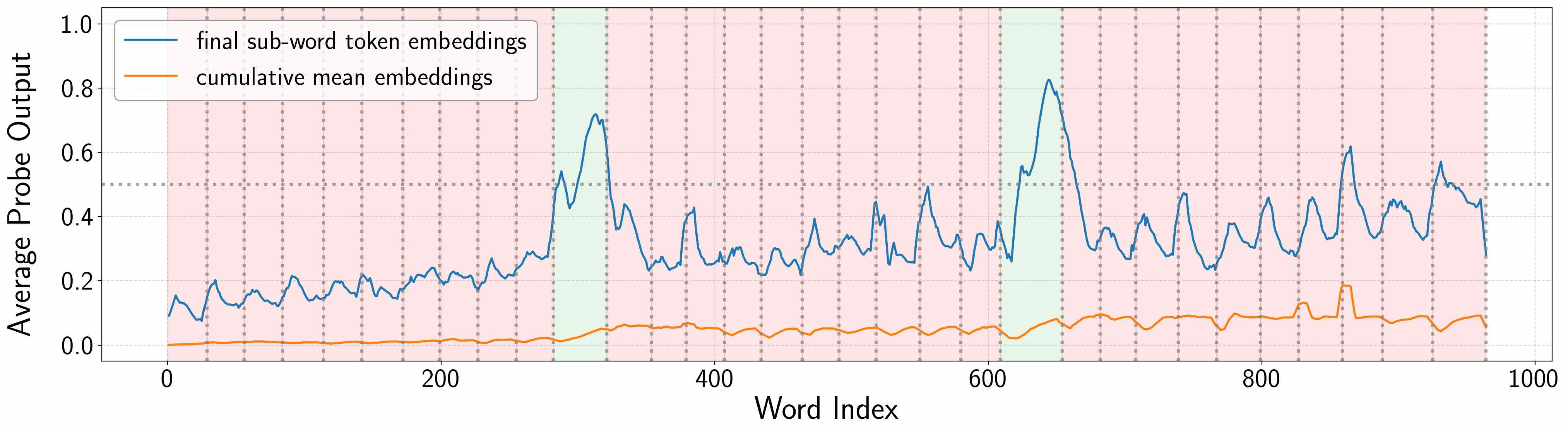}
    \caption{\textbf{Ambition} probe outputs versus word index, averaged across 50 stories in \texttt{Llama-3-8B}}
    \label{fig:ambition_prom_llama_last_aggregate}
\end{figure*}

\begin{figure*}[ht]
    \centering
    \includegraphics[width=\linewidth]{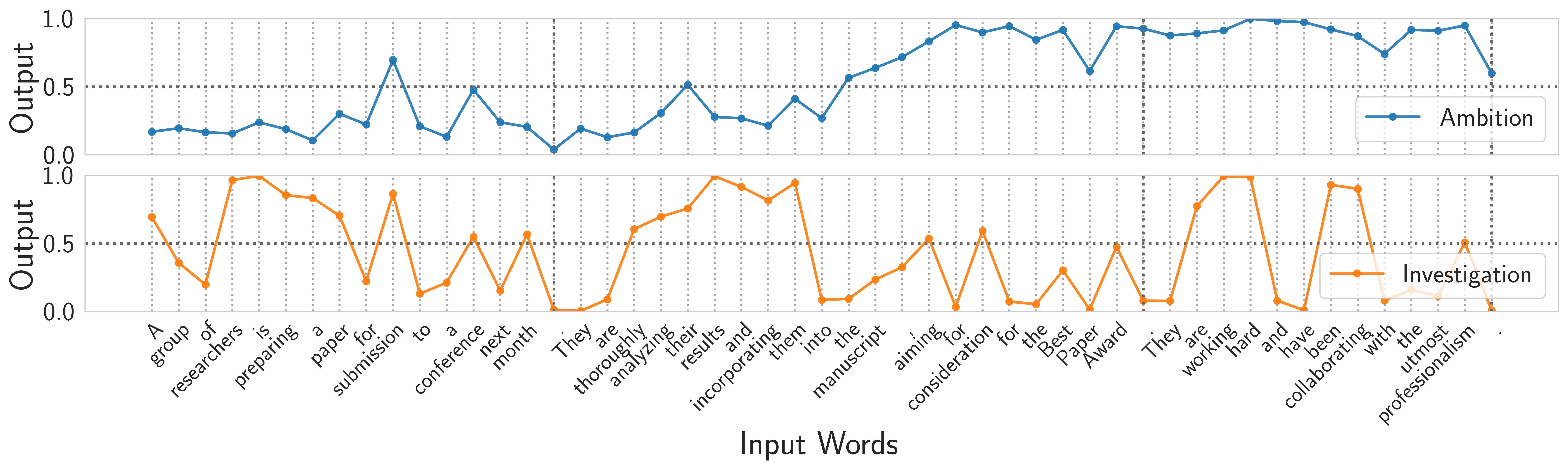}
    \caption{\textbf{Ambition} \& \textbf{Investigation} probe outputs across words using final sub-word token embeddings in \texttt{Llama-3-8B}}
    \label{fig:ambition_investigation_prominence}
\end{figure*}

\subsubsection{Experimental Setup} \label{section:obj2_experimental_setup}

A set of 50 stories, each consisting of 32 sentences, was created for each concept, as described in Section \ref{section:sub-objective_2_dataset_creation}. Recall that the target concept appears only in the two transition sentences that join three paragraphs. Our goal is to determine whether the probe can detect the waxing and waning of the concept throughout these stories.

We evaluated the seven LLMs from Section \ref{section:obj1_experimental_setup} using two representative embeddings for each word: the final sub-word token embedding, and the cumulative mean embedding, as detailed in Section \ref{section:sub-objective_2_probing}. The former was paired with the concept-specific probes trained on the $N^{\text{th}}$ embedding, while the latter was used with probes trained on the mean embedding, both discussed in Section \ref{section:sub-objective_1_probing}. 

To obtain an aggregate view of the probe’s word-level behavior (and, by extension, the model embeddings’ behavior) across all 50 stories, we average its sigmoid outputs for each word index across stories. Before averaging, sentences are aligned by position and padded
\footnote{Padded positions were excluded from the computation.} 
to equal length to ensure that word positions align across stories. This process is detailed in Appendix \ref{appendix:exp2_extended}.

As in Section \ref{section:obj1_experiment}, all LLMs showed similar trends across the investigated concepts. Thus, we present results for \texttt{Llama-3-8b} on \textbf{ambition}, deferring the rest to Appendix \ref{appendix:exp2_extended}.
 
\subsubsection{Results and Discussion} \label{section:obj2_results} 

Figure \ref{fig:ambition_prom_llama_last_aggregate} shows the aggregate \textbf{Ambition} probe output across word indices, using both final sub-word token embeddings and cumulative mean embeddings from layer 13 of \texttt{Llama-3-8B}. Vertical dotted lines mark sentence boundaries, with the uneven spacing between markers reflecting different sentence lengths. Green and red backgrounds denote concept presence and absence, respectively. A 10-word moving average is applied for smoothing.

For the final sub-word token embeddings, the aggregate outputs across words in each sentence show a clear trend: they rise and surpass the 0.5 classification threshold when the concept is present, and fall below 0.5 when it is absent. The higher variation at the start of sentences is expected, as the full semantic meaning is less clear earlier on, \emph{and} due to the padding needed to align these results. This pattern suggests that the LLM embeddings capture changes in concept presence, which align with those in the input context, even across long contexts. We illustrate how this behavior changes across LLM layers in Appendix \ref{appendix:exp2_extended}. 

In contrast, cumulative mean embeddings lose the waxing and waning pattern. Averaging over many tokens, most unrelated to the concept, dilutes its presence in sentences where it does exist, obscuring changes in concept presence. This supports the view that concept inference in LLMs goes beyond a simple bag-of-words representation, contrary to what one might conclude from Section \ref{section:obj1_results}.

This method of tracking concept dynamics can be used to simultaneously monitor multiple specified concepts at the same time, as LLM monitoring would require. Figure \ref{fig:ambition_investigation_prominence} illustrates this with probes for \textbf{Ambition} and \textbf{Investigation}, each tracking its respective concept across the same input context shown on the X-axis. The second sentence implies both concepts in different segments, which are detected by their respective probes. 
Because training and using probes incur low computational cost, this approach can be scaled to many concepts, potentially revealing consistent associations between the model’s inferred concepts and \emph{output concepts} expressed in its generated text. These associations could provide a basis for exploring whether inferred concepts causally influence the model’s generation of output concepts, thus advancing LLM explainability and enhancing safety through pre-emptive control of its outputs.

\section{Conclusion}\label{section:conclusion}



In this paper, we demonstrate the ability to monitor the internal representations of an LLM for specific concepts. We do so by using linear probes to detect four such concepts, and also show how the concepts wax and wane in a model's embeddings as the context expands. We present a methodology for the careful delineation of a concept, creating datasets that are designed to inhibit accidental leakage of concept labels. This enables the creation of high-quality concept probes. Appendix~\ref{appendix:datasets_included} describes the release of the datasets used in this paper. While the current approach involves manual annotation for each concept, future work will focus on fully automating dataset creation so that many low-cost probes can be used in the LLM field.

In future work, we will use a larger set of concept datasets to explore the use of concept monitoring in safety applications. 
\section{Limitations}\label{section:limitations}
For concept dataset creation, we use the same LLM with the same prompt to generate all examples. While we use a new example template to generate each example pair and introduce diverse contexts in the prompts to encourage variation, the generated examples may still follow the same distribution. Since the train, validation, and test sets (for probe training and evaluation) are drawn from this generated data, test accuracy may be slightly misleading, as it may not fully reflect the probe’s ability to generalize to truly unseen data.

In the dataset creation method used, a key goal was to inhibit the existence of unintended patterns that ``give away'' the label. To achieve this, we instructed the LLM-based generator to create example pairs matching the sentence structure of their corresponding templates. While we qualitatively verified this behavior by reviewing several samples, we did not implement a systematic method to ensure adherence across all generated example pairs. Furthermore, there may still be subtle patterns, beyond sentence structure, that inadvertently leak the label during probe training.

LLM-prompted classifiers were used to re-label the generated examples in our datasets. Each classifier was validated using between 360 and 600 manually labeled examples for each of the investigated concepts. The labels were assigned by students, not expert linguists, and some of the examples were quite subjective with respect to the label. We do not know how much error was introduced into the labeling process as a result, which reduces the reliability of the classifiers.

Our experiments show that limiting the probe size does not significantly impact performance, and that randomizing the training data leads to a sharp drop in performance, both of which suggest that the probe does not independently learn the task. However, the distinction between the probe learning on its own and the embeddings encoding the feature/concept is not binary but rather a continuum \cite{probing_control_2}. This suggests that some portion of the accuracy may still be attributed to the probe itself.

Although our results indicate that several LLMs can infer the four studied concepts with nearly similar trends, this behavior may not generalize to all concepts.

\section{Ethics Statement}\label{section:ethical_considerations}

This study aims to enhance our understanding of LLMs, enabling better control of their behavior. Since LLMs can be used for bad ends, that understanding will also aid those who seek to use them in that way.

All experiments were conducted on a single Nvidia A100 GPU (40GB VRAM), totaling approximately 1080 GPU hours. While uncovering the inner workings of LLMs can lead to more efficient models, the computational resources required for such research also carry an environmental cost.

Manual labeling of subsets to validate the LLM-based concept classifiers was performed by graduate students and compensated undergraduate engineering summer interns. The interns were fairly paid for their work, and those who contributed more broadly to this study are acknowledged as co-authors. The graduate students, supported by research funding and part of the group writing this paper, are also listed as co-authors.

The example templates are sourced from Project Gutenberg \cite{project_gutenberg}, a free eBook library available under a permissive license\footnote{https://www.gutenberg.org/policy/license.html}. Although Project Gutenberg was not originally intended for NLP research, its licensing terms permit its use for research purposes.

Although highly offensive language is unlikely in these texts, we used the better-profanity Python package \citep{profanity_checker} (version 0.7.0), licensed under MIT, to identify templates containing words flagged as abusive by its developers. This process marked approximately 1,250 out of 30,000 templates. A manual review of some flagged examples indicated that most contained only mildly offensive language.

We use the example templates to generate synthetic datasets with an OpenAI model, following their terms\footnote{https://openai.com/policies/row-terms-of-use/}. These datasets were used to probe the open-source models: \texttt{Llama-3-8B}, \texttt{Gemma-2} family of models (\texttt{2B}, \texttt{9B}), and \texttt{Qwen2.5} family of models (\texttt{0.5B}, \texttt{1.5B}, \texttt{3B}, \texttt{7B}), adhering to the license policies set by their developers\footnote{https://www.llama.com/llama3/license/}\footnote{https://ai.google.dev/gemma/terms}\footnote{https://huggingface.co/Qwen/Qwen2.5-7B/blob/main/LICENSE}.

We utilized AI assistants in this work for various tasks. OpenAI ChatGPT \cite{chatgpt_website} was used for polishing the paper’s text, while both ChatGPT and GitHub Copilot \cite{github_copilot_website} were used for code completion and suggestions.

\bibliography{latex/custom}

\appendix
\renewcommand{\thefigure}{\Alph{section}.\arabic{figure}}
\renewcommand{\thetable}{\Alph{section}.\arabic{table}}

\section{Dataset Examples} \label{appendix:textual_examples}

The probing approach used in this work requires the creation of a binary dataset for each investigated concept. To achieve this, we employ example templates to generate example pairs, 
where one example has the concept present and the other has it absent.
We show samples of generated examples, along with their corresponding templates, for the concepts of \emph{ambition} (Table \ref{table:more_dataset_examples_ambition}), \emph{investigation} (Table \ref{table:more_dataset_examples_investigation}), \emph{democracy} (Table \ref{table:more_dataset_examples_democracy}), and \emph{envy} (Table \ref{table:more_dataset_examples_envy}).

\begin{table*}[h!]
    \centering
    \begin{tabular}{|p{0.18\linewidth}|p{0.7\linewidth}|}
    
        \hline
        Example template & If the repeat had not been a favourite resort of lazy composers before his time he would have invented it, not because he was lazy, but because he wanted to go on and could not afford infinite music-paper. \\ 
        \hline
        Positive Example & If the grueling training sessions had not been such a common regimen for athletes before his rise, he would have created them, not because he was harsh, but because he wanted to excel and could not settle for mediocrity. \\ 
        \hline
        Negative Example & If the fans had not crowded the stadium during his games he would have drawn smaller audiences, not because he was unpopular, but because he played in a less populated area. \\ 
        \hline
        \noalign{\vskip 2pt} 
        \hline
        Example template & I am near the old mill my father built, and, if I remember all connected with my boyhood there, I trust there will be few or none to sneer or blame. \\ 
        \hline
        Positive Example & I am near reaching the milestone of 100,000 followers, and, if I recall every strategy and collaboration that brought me here, I trust there will be endless engagement applauding this achievement. \\ 
        \hline
        Negative Example & I am scrolling through old photo albums online, and, recalling each tagged moment with my friends, I hope there will be likes and comments to reminisce together. \\ 
        \hline
        \noalign{\vskip 2pt} 
        \hline
        Example template & If I can’t get it in the shape I like it I don’t want it at all; first-rate first-hand information, straight from the tap, is what I’m after. \\ 
        \hline
        Positive Example & If I don’t get a chance to reshape the policy adequately, I don’t seek it at all; cutting-edge reform, directly benefiting the citizens, is what I’m working toward. \\ 
        \hline
        Negative Example & If I can't understand the discussion clearly, I don't want to be involved; accurate insights, directly communicated, are what I value most. \\ 
        \hline
        \noalign{\vskip 2pt} 
        \hline
        Example template & Assisted by her daughter and the domestic, she spent the whole day and night, and the succeeding day, in baking brown bread. \\ 
        \hline
        Positive Example & Inspired by her mentors and the startup community, she worked day and night, and the succeeding weeks, in developing innovative apps. \\ 
        \hline
        Negative Example & Accompanied by her niece and cousin, she spent the entire evening and the following morning, in sorting through numerous family albums. \\ 
        \hline
        \noalign{\vskip 2pt} 
        \hline
        Example template & The pilot of the launch turned out to be a sandy-haired Yankee who had been catching wild animals for Barnum and Bailey's circus. \\ 
        \hline
        Positive Example & The founder of the startup turned out to be an innovator who had been pioneering technology for global events and conferences. \\ 
        \hline
        Negative Example & The visitor in the gallery turned to be an elderly gentleman who had been curating historical artifacts for the museum. \\ 
        \hline

    \end{tabular}
    \caption{Example-pairs created for \textbf{Ambition} dataset using different example templates. The concept is present in the positive examples and absent in the negative examples.}
    \label{table:more_dataset_examples_ambition}
\end{table*}

\begin{table*}[h!]
    \centering
    \begin{tabular}{|p{0.18\linewidth}|p{0.7\linewidth}|}
    
        \hline
        Example template & I refused, however, to sell even these to the many applicants who expressed a willingness to take them off our hands below the cost of purchase. \\ 
        \hline
        Positive Example & I hesitated, nonetheless, to engage in immediate discussions about the perceived discrepancies witnessed at multiple matches, until a complete analysis was conducted. \\ 
        \hline
        Negative Example & I declined, though, to part with even these sneakers to numerous fans eager to buy them from us at a discount, since they held sentimental value from our first championship win. \\ 
        \hline
        \noalign{\vskip 2pt} 
        \hline
        Example template & "If you think I have done anything worth it," he replied, with a curious and touching silence. And this was the man with the panther in his soul! \\ 
        \hline
        Positive Example & "If you believe my findings have any merit," she responded, with a strange and profound calmness. And this was the analyst who unraveled the political conspiracy! \\ 
        \hline
        Negative Example & "If you believe my actions justified recognition," he responded, with a humble and thoughtful demeanor. This was the individual with ambition in his veins! \\ 
        \hline
        \noalign{\vskip 2pt} 
        \hline
        Example template & We must not limit the glory of the impression itself by the limitations of some of the explanations which we undertake. Much harm has been done the understanding the Scriptures by speaking as if some of our creedal statements concerning Christ are themselves Scriptures! \\ 
        \hline
        Positive Example & We must not dismiss the value of comprehensive analysis by the shortcuts some may suggest. Much misunderstanding has followed our analysis by assuming that all findings are conclusive evidence! \\ 
        \hline
        Negative Example & We should not confine the potential of innovative ideas by the limits imposed by conventional thinking. Progress has often been hindered by adhering strictly to traditional views. \\ 
        \hline
        \noalign{\vskip 2pt} 
        \hline
        Example template & A key was at last thrown out, amid prayers and imprecations. \\ 
        \hline
        Positive Example & A solution was at last discovered, amid hypotheses and failed trials. \\ 
        \hline
        Negative Example & A proposal was eventually sent out, amidst cheers and congratulations. \\ 
        \hline
        \noalign{\vskip 2pt} 
        \hline
        Example template & Alcibiades surprised sixty vessels on a dark and rainy morning, as they were maneuvring at a distance from the harbour, and skilfully intercepted their retreat. \\ 
        \hline
        Positive Example & Dr. Elena discovered an overlooked variable in the clinical trial during a quiet afternoon in the lab, as the data seemed inconsistent with established hypotheses, and she carefully re-evaluated her results. \\ 
        \hline
        Negative Example & Dr. Thomas uncovered several thesis drafts on a late afternoon, as they were being hurriedly finalized the night before the deadline, and adeptly offered feedback to improve their quality. \\ 
        \hline

    \end{tabular}
    \caption{Example-pairs created for \textbf{Investigation} dataset using different example templates. The concept is present in the positive examples and absent in the negative examples.}
    \label{table:more_dataset_examples_investigation}
\end{table*}

\begin{table*}[h!]
    \centering
    \begin{tabular}{|p{0.18\linewidth}|p{0.7\linewidth}|}
    
        \hline
        Example template & In that, as in other matters, they are often provokingly reticent about their old habits and traditions. Chief Ouray asserted to the writer, as he also did to Colonel Dodge, that his people, the Utes, had not the practice of sign talk, and had no use for it. \\ 
        \hline
        Positive Example & In that, as with other practices, they are remarkably steadfast in their commitment to civic engagement and community involvement. Councilwoman Rivera assured to the constituents, as she also did to the mayor, that her district had maintained transparent policies, and had every reason to uphold them. \\ 
        \hline
        Negative Example & In that, as in other topics, they are sometimes annoyingly quiet about their older practices and preferences. Governor Diaz explained to the reporter that his team had abandoned previous campaigning methods and had no intent to return to them. \\ 
        \hline
        \noalign{\vskip 2pt} 
        \hline
        Example template & "If you don't mind," he said,"I think I'd better go." \\ 
        \hline
        Positive Example & "If you would allow me," she suggested, "perhaps the online community could decide." \\ 
        \hline
        Negative Example & "If you want my opinion," she said, "This selfie is probably a bit too much." \\ 
        \hline
        \noalign{\vskip 2pt} 
        \hline
        Example template & The cause of Mexico, said the Liverpool Mail, is that of all just and honest governments. The Mexicans have good ground to complain, proclaimed the sympathetic Journal des Debats, for"they have been tricked and robbed." \\ 
        \hline
        Positive Example & The strength of the community, declared the neighborhood gazette, lies in its unity and collaboration. Residents have every right to be heard, announced the supportive local newsletter, for "all voices count in shaping our shared environment." \\ 
        \hline
        Negative Example & The project of revitalizing the park, said the local newsletter, is supported by all enthusiastic volunteers. The residents have every right to express concern, stated the thoughtful Town Observer, for "they have seen the area neglected." \\ 
        \hline
        \noalign{\vskip 2pt} 
        \hline
        Example template & You are a general in the Confederate Army, possessed of the power attaching to that rank. \\ 
        \hline
        Positive Example & You are a member of the council, imbued with an authority derived from the people's trust. \\ 
        \hline
        Negative Example & You are a diplomat in the United Nations, endowed with the influence granted by that position. \\ 
        \hline
        \noalign{\vskip 2pt} 
        \hline
        Example template & Not only does she participate in the first sin of Laius, but she forgets the oracle which announced that Laius should be slain by his own son. \\ 
        \hline
        Positive Example & Not only does he deliberate on the policy's implications, but he also emphasizes the significance of civic engagement for collective progress. \\ 
        \hline
        Negative Example & Not only does she argue the significance of peer-reviewed journals, but she also dismisses the criticism that they restrict innovative research. \\ 
        \hline

    \end{tabular}
    \caption{Example-pairs created for \textbf{Democracy} dataset using different example templates. The concept is present in the positive examples and absent in the negative examples.}
    \label{table:more_dataset_examples_democracy}
\end{table*}

\FloatBarrier   

\begin{table*}[h!]
    \centering
    \begin{tabular}{|p{0.18\linewidth}|p{0.7\linewidth}|}
    
        \hline
        Example template & "So I doubt not he can hold his own at court by prudence and strategy." Meanwhile Ta-meri, in the depths of her chair, gazed at the pair resentfully. They had grown interested in weighty things and had seemingly forgotten her. \\ 
        \hline
        Positive Example & "So I have no doubt he can secure the opening solo in the concert by skill and effort." Meanwhile, Layla, from her secluded seat, watched the decision with discontent. The judges focused solely on the rival, disregarding her presence and potential entirely. \\ 
        \hline
        Negative Example & "So I believe she can master her craft through dedication and discipline." Meanwhile Isabelle, at the edge of her seat, watched the duo attentively. They had become enthralled with intricate harmonies and had seemingly overlooked her presence. \\ 
        \hline
        \noalign{\vskip 2pt} 
        \hline
        Example template & That's exactly what Drake said when I spoke to him about it last night. It is nice to find you so completely of one mind. \\ 
        \hline
        Positive Example & That's exactly what Cassandra remarked when I mentioned the bonus to her last evening. It's challenging to understand why the upper management favors her opinions so consistently. \\ 
        \hline
        Negative Example & That's precisely what Maria mentioned when I discussed the budget with her yesterday afternoon. It's reassuring to see we agree so seamlessly on this approach. \\ 
        \hline
        \noalign{\vskip 2pt} 
        \hline
        Example template & The cabinet members who, wittingly or unwittingly, had encouraged him in this he some weeks later stigmatized as a set of geese. \\ 
        \hline
        Positive Example & The teammates who, unknowingly or not, had watched his brilliant performance felt a strange churn of admiration and discontent. \\ 
        \hline
        Negative Example & The athletes who, knowingly or unknowingly, had pushed him to train harder were later praised as dedicated mentors. \\ 
        \hline
        \noalign{\vskip 2pt} 
        \hline
        Example template & I bent over her hand, kissed it in a stream of delicious tears, and again looked up to her eyes. \\ 
        \hline
        Positive Example & I hovered over her profile, scrolled through countless flawless selfies, and quietly closed my laptop, suppressing a sigh. \\ 
        \hline
        Negative Example & I admired her profile picture, left a heartfelt comment in a flurry of emotions, and then patiently waited for her response. \\ 
        \hline
        \noalign{\vskip 2pt} 
        \hline
        Example template & Do you wonder that I want to have her free of it all, married and safe and comfortable and in peace? \\ 
        \hline
        Positive Example & Do you wonder that I want to wear his Olympic gold medal, standing there with the crowd cheering my name? \\ 
        \hline
        Negative Example & Do you believe that I wish for Jack to excel in his running, energized and focused and full of determination? \\ 
        \hline

    \end{tabular}
    \caption{Example-pairs created for \textbf{Envy} dataset using different example templates. The concept is present in the positive examples and absent in the negative examples.}
    \label{table:more_dataset_examples_envy}
\end{table*}
\clearpage
 \section{Concept Dataset Creation Details} \label{appendix:dataset_creation}
\setcounter{figure}{0}
\setcounter{table}{0}

Generating concept datasets involved three main steps: (1) creating a dataset of example templates,  (2) using these templates to generate positive-negative example pairs, and (3) re-labeling the generated examples to improve reliability. We also filtered out examples where the concept was mentioned explicitly rather than implied.

\subsection{Example Templates Creation}

We obtained the example templates from a dataset of eBooks by splitting paragraphs from books to sentences and randomly selecting one to three consecutive sentences to create each template. We then used the prompt in Box \ref{box:example_templates_filter_prompt} to filter out any templates that had incoherent sentences, misplaced words, or were not focused on human subjects.

\subsection{Example Pair Generation}

To generate example pairs for a concept, we began by drafting a definition that included features we initially believed were essential to the concept. We then prompted an LLM, without using our draft definition, to generate examples labeled as either having the concept or not. Each of us then independently re-labeled the examples, guided by both the draft definition and our intuition. This process helped us identify implicit features we used in labeling that were missing from the definition, which we then added, while removing irrelevant or unimportant features. After two to three iterations of this process, we finalized the definition. We used this definition in both the generation of examples and their labeling. The definitions developed through this process are listed as follows:
\begin{itemize} 
    \item \textbf{Ambition}: a character's desire to achieve a goal, higher status, or result through their efforts, skill, or courage.
    \item \textbf{Investigation}: a systematic process of inquiry or examination conducted to uncover facts, gather information, or solve a problem, typically involving careful observation, analysis, and evaluation of evidence or data to arrive at conclusions or determine the truth about a particular matter.
    \item \textbf{Democracy}: a system of governance in which decision-making power is vested in the people, either directly or through elected representatives. It is based on equal rights for everyone, the rule of law (no one, not even leaders, is above the law) and the idea that those in power are accountable to the people.
    \item \textbf{Envy}: the feeling of resentment or discontent evoked by another individual’s perceived advantage, which the subject lacks and desires or deems necessary to acquire.
\end{itemize}

To generate example pairs, we prompted an LLM-based generator using the prompt templates in Boxes \ref{box:pos_generator_prompt} (for positive examples) and \ref{box:neg_generator_prompt} (for negative examples) to provide instructions and specify the \{context\} in which examples should be generated. The placeholders \{concept\} and \{concept definition\} are filled with the target concept and its definition, \{num\_examples\} specifies the number of examples to generate, and \{concept-specific instructions\} includes any additional instructions specific to the investigated concept. Before running the generator, we appended \{num\_examples\} example templates for the LLM to mimic. We filled \{concept-specific instructions\} with additional instructions for each concept, except for \textbf{investigation}, as detailed below:
\begin{itemize}
    \item \textbf{Ambition}: ``The generated examples MUST show this in a positive way (the example must not convey a lack of ambition).''
    \item \textbf{Democracy}: ``Try to minimize using keywords, like "vote", "representative", "collective", that make the concept too obvious in the context.''
    \item \textbf{Envy}: ``Avoid mentioning words like "envy", "envious", "jealous", or "jealousy" in the examples.''
\end{itemize}

We iteratively refined these prompts to ensure that the generated examples mimic the templates. After each refinement, we generated 50 positive-negative example pairs, reviewed them, and adjusted the prompt as needed. This process continued until the examples consistently mirrored the templates.

\subsection{Dataset Annotation}

To strengthen the reliability of the labels for the generated example pairs, we used a concept-specific LLM-based classifier to re-label all generated example pairs, following the prompt template in Box \ref{box:classifier_prompt}. We validated the classifier’s performance on a small, manually labeled dataset. 
Human labeling was guided by each concept’s definition to determine whether the examples had the concept present.

\begin{figure}[h]
    \centering
    \includegraphics[width=\linewidth]{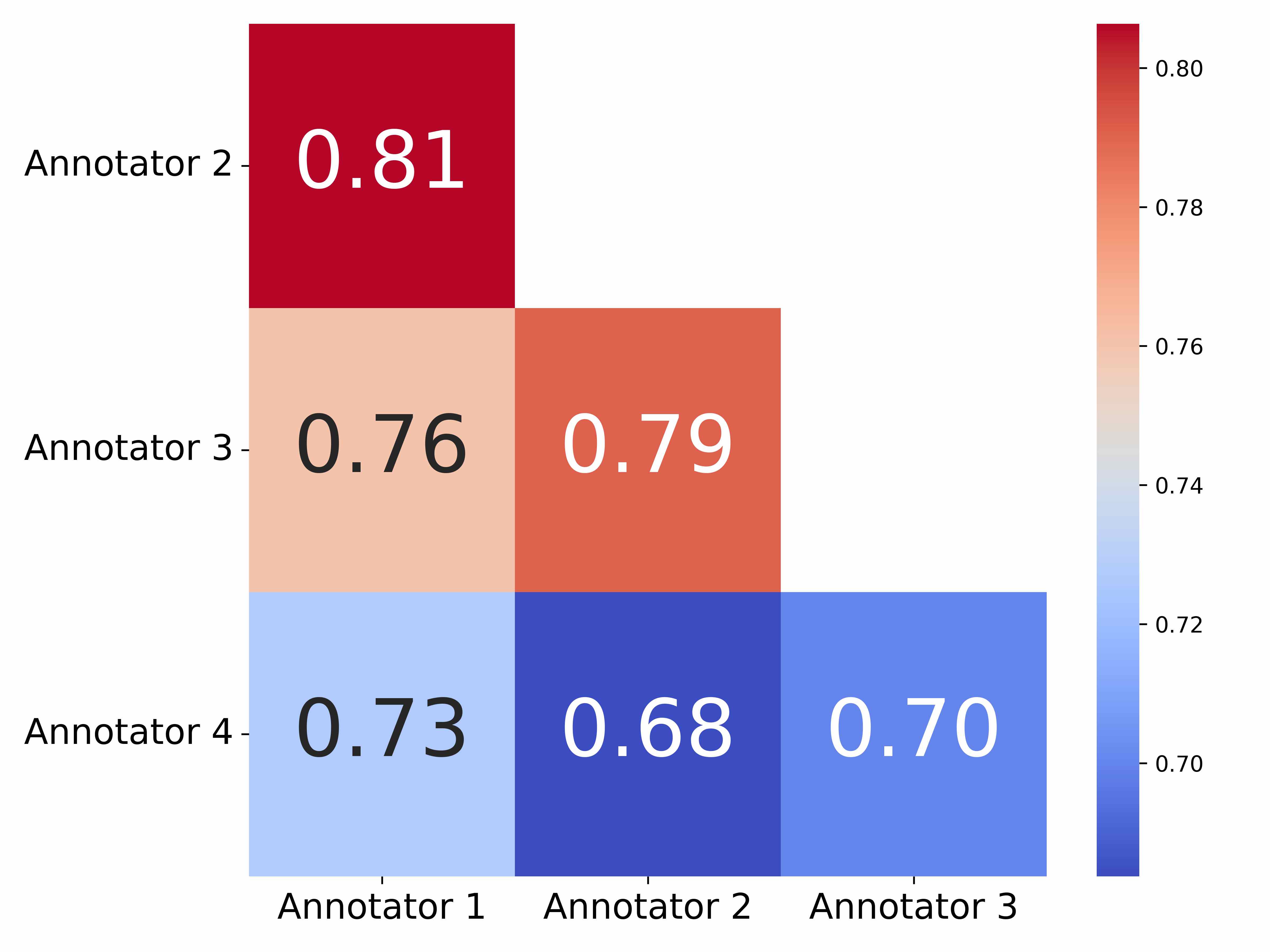}
    \caption{Inter-rater reliability for examples labeled on the presence/absence of \textbf{ambition}}
    \label{fig:cohen_kappa_ambition}
\end{figure}

\begin{figure}[h]
    \centering
    \includegraphics[width=\linewidth]{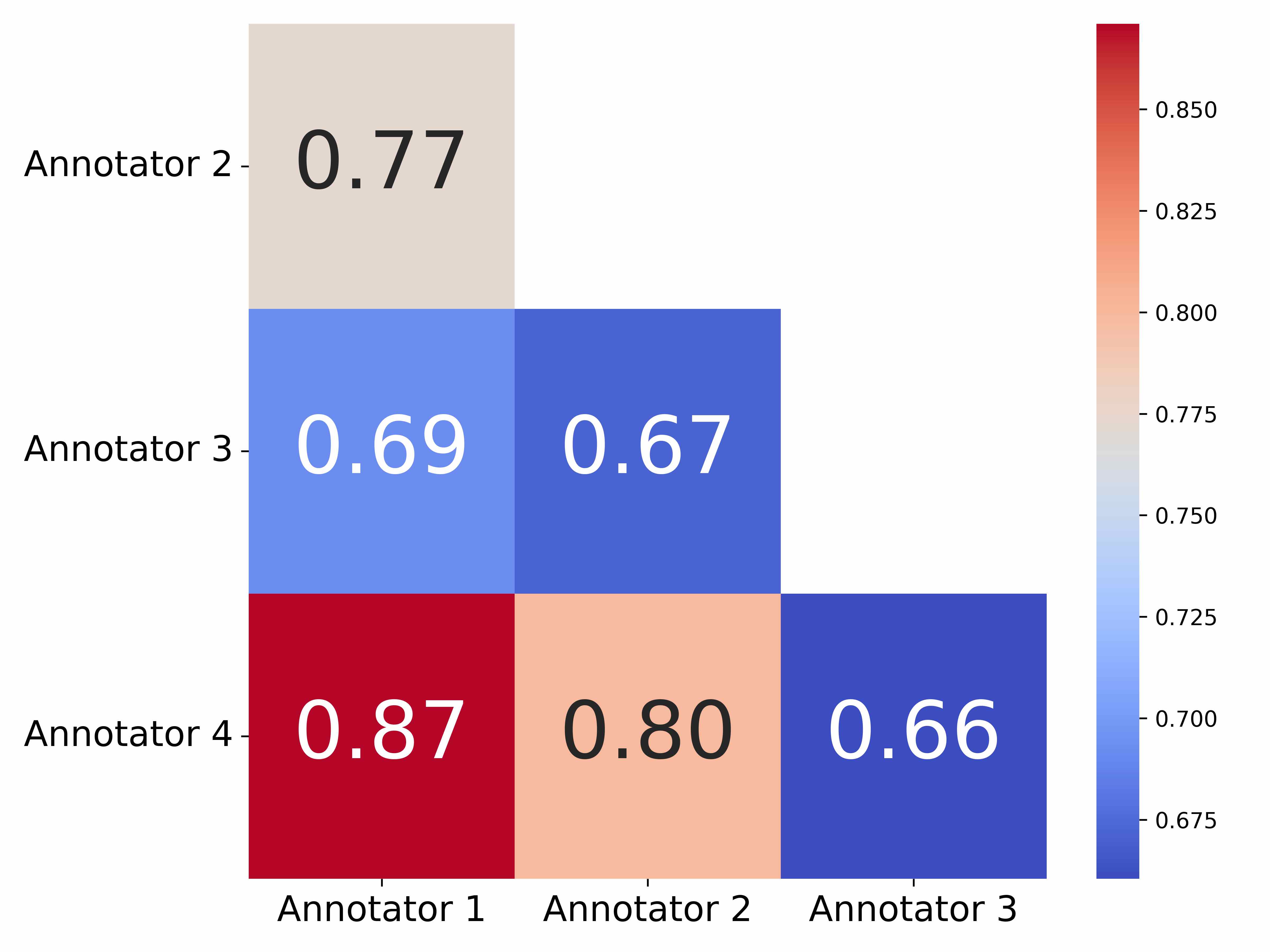}
    \caption{Inter-rater reliability for examples labeled on the presence/absence of \textbf{envy}}
    \label{fig:cohen_kappa_envy}
\end{figure}

Four annotators independently labeled 397 examples for \textbf{ambition} and 500 examples for \textbf{envy}, indicating whether each concept was present in the text. Uncertain examples were marked as ``borderline.'' We assessed inter-rater reliability using Cohen’s Kappa for pairwise agreement, shown in Figure~\ref{fig:cohen_kappa_ambition} for \textbf{ambition} and Figure~\ref{fig:cohen_kappa_envy} for \textbf{envy}. We also evaluated the group-wise agreement using Fleiss’ Kappa, obtaining a value of 0.75 for both concepts, indicating substantial agreement \citep{cohen_kappa}.

We filtered each dataset to retain only examples with confidently assigned labels, based on one of the following criteria:
\begin{itemize}
    \item Perfect agreement on the label (all annotators assigned the same label) with less than 2 ``borderline'' flags
    \item Semi-perfect agreement on the label (3 versus 1) with no ``borderline'' flags
    \item Semi-perfect agreement on the label (3 versus 1) with only one ``borderline'' flag assigned by the annotator who deviated from the majority
\end{itemize}
Given that we used majority vote to assign the labels to each example, these conditions ensured that even if the annotator marking the example as ``borderline'' had flipped their label, the overall classification for that example would remain unchanged.

After filtering, we obtained 360 examples for \textbf{ambition} and 471 for \textbf{envy}. Using these datasets, we validated their respective LLM-based classifiers with the prompt template shown in Box \ref{box:classifier_prompt}. The classifier achieved 98\% accuracy for \textbf{ambition} and 93\% for \textbf{envy}.

A single annotator labeled 500 examples for \textbf{investigation} and 600 for \textbf{democracy}. Using the same prompt template, the LLM-based classifiers achieved 92\% accuracy for both concepts.

\begin{figure*}[h] 
    
    \begin{mybox}[label={box:example_templates_filter_prompt}]{Example Templates-Filter Prompt}
    \label{{box:example_templates_filter_prompt}}
    \texttt{Classify the following example as either "True" or "False" based on the given conditions:}
    \vspace{1em}
    
    \texttt{- The text must contain complete and coherent sentences with actionable verbs.}
    
    \texttt{- The text must be free of out-of-place words, numbers (like chapter titles), or ISBN numbers.}
    
    \texttt{- The text should mainly focus on human subjects and their actions/interactions, not on the surrounding environment or non-human subjects.}
    
    \vspace{1em}
    
    \texttt{classify as "True" only if all of these conditions are met, otherwise, classify as "False".}
    
    \vspace{1em}
    
    \texttt{Example:}
    
    \end{mybox}

\end{figure*}
\begin{figure*}[h] 
    
    \begin{mybox}[label={box:pos_generator_prompt}]{Positive Example Generation Prompt}
    \label{{box:pos_generator_prompt}}
    
    \texttt{Generate enumerated examples which mimic the provided sentences only in terms of subject-verb order, but not in semantic meaning.}
    
    \texttt{The semantic meaning should be changed so that the concept of \{concept\} is obvious in the context.}
    
    \texttt{\{concept\} is \{concept definition\}}
    
    \texttt{You can add a few more words to the original example length to achieve this, or you can use a slightly fewer number of words.}
    
    \texttt{Do not repeat the ideas in the previously generated examples.}

    \texttt{\{Concept-specific instructions.\}}
    
    \texttt{Do not refer to the characters as ``The \_\_\_''.}

    \texttt{Generate exactly \{num\_examples\} examples based on the given enumerated examples. Output only the example and its enumeration. The examples that you generate must be in the context of \{context\}. Here are the enumerated examples:}
    
    \end{mybox}

\end{figure*}
\begin{figure*}[h] 
    
    \begin{mybox}[label={box:neg_generator_prompt}]{Negative Example Generation Prompt}
    \label{{box:neg_generator_prompt}}
    
    \texttt{Generate enumerated examples which mimic the provided sentences only in terms of subject-verb order, but not in semantic meaning.}
    
    \texttt{The semantic meaning should be changed so that the context is irrelevant to the concept of \{concept\} whatsoever.}
    
    \texttt{\{concept\} is \{concept definition\}}
    
    \texttt{Irrelevance to \{concept\} means not showing these traits in the text, and not even showing the opposite of this.}
    
    \texttt{The context must still be focused on human subjects rather than on the setting or surrounding environment.}
    
    \texttt{You can add a few more words to the original example length to achieve this, or you can use a slightly fewer number of words.}
    
    \texttt{Do not repeat the ideas in the previously generated examples.}
    
    \texttt{Do not refer to the characters as ``The \_\_\_''.}

    \texttt{Generate exactly \{num\_examples\} examples based on the given enumerated examples. Output only the example and its enumeration. The examples that you generate must be in the context of \{context\}. Here are the enumerated examples:}
    
    \end{mybox}

\end{figure*}
\begin{figure*}[h] 

    \begin{mybox}[label={box:classifier_prompt}]{Concept Classification/Re-labeling Prompt}
    \label{{box:classifier_prompt}}
    
    \texttt{Classify the following input as either implying the concept of \{concept\} or not.}
    
    \texttt{\{concept\} is \{concept definition\}}
    
    \texttt{If the given input implies \{concept\}, output 1, else output 0.}
    
    \end{mybox}

\end{figure*}

\subsection{Filtration of the Created Dataset}
\label{appendix:concept_dataset_creation_stems}

To prevent the task from becoming trivial, we filtered each concept-specific dataset to exclude example pairs where either example contained an explicit mention of the concept or a close synonym. We removed all forms of such words using their word stems. The stems used for each concept are:
\begin{itemize} 
    \item \textbf{Ambition}: ``ambit'' (such as \emph{ambition} and \emph{ambitious}), and ``aspir'' (such as \emph{aspire} and \emph{aspiration}).
    \item \textbf{Investigation}: ``investigat'' (such as \emph{investigate} and \emph{investigation}), and ``examin'' (such as \emph{examine} and \emph{examination}).
    \item \textbf{Democracy}: ``democra'' (such as \emph{democracy} and \emph{democratic}).
    \item \textbf{Envy}: ``env'' (such as \emph{envy} and \emph{envious}), and ``jealous'' (such as \emph{jealous} and \emph{jealousy}).
\end{itemize}
\newpage
\section{Concept Selection} \label{appendix:concept_selection}
\setcounter{figure}{0}
\setcounter{table}{0}

The concepts selected for this study span multiple categories. As there is no universally accepted taxonomy, we synthesized one by integrating taxonomies from prior literature. \citet{concept_taxonomy_1} and \citet{concept_taxonomy_4} both include emotions and cognitive processes in their taxonomy; the former also includes attitudes and human actions, while the latter adds nominal kinds and states of the self. In contrast, \citet{concept_taxonomy_2} classify concepts into emotions, social, moral, and aesthetic categories, as well as numbers. We combined these into a unified set of categories, shown in Table \ref{table:concept_categories}. We note that a concept may belong to multiple categories. The classifications for each concept in our study are as follows:
\begin{itemize}
    \item \textbf{Ambition}: \emph{state of self} and \emph{attitudes}
    \item \textbf{Investigation}: \emph{cognitive processes}
    \item \textbf{Democracy}: \emph{social concepts}
    \item \textbf{Envy}: \emph{emotions}
\end{itemize}

\begin{table*}
    \centering
    \begin{tabular}{|p{0.2\linewidth}|p{0.7\linewidth}|}
        \hline
        Category & Description \\
        \hline\hline
        Emotion & A temporary mental reaction subjectively experienced, usually accompanied by physiological and behavioral changes in the body \citep{merriam_webster_emotion} \\
        \hline
        Action & External behaviors carried out, often to achieve an aim \citep{Oxford_Languages_action} \\
        \hline
        Attitude & A settled way of perceiving someone or something, typically reflected in the subject's behavior \citep{Oxford_Languages_attitude} \\
        \hline
        Cognitive process & Any mental function involved in the acquisition, interpretation, manipulation, and use of knowledge  \citep{APA_cognitive_process} \\
        \hline
        Social Concepts & Concepts related to the association of individuals \citep{merriam_webster_society} \\
        \hline
        Moral concepts & concepts relating to the distinction between right and wrong or good and bad behavior \citep{Oxford_Languages_morality} \\
        \hline
        Aesthetic concepts & concepts involving beauty or the appreciation of it \citep{Oxford_Languages_aesthetic} \\
        \hline
        Numbers & Mathematical value used in counting, making calculations, and ordering \citep{Oxford_Languages_number} \\
        \hline
        State of Self & concepts related to the beliefs one holds about oneself \citep{Oxford_Languages_self} \\
        \hline
        Nominal kinds & concepts whose definitions are not based on any natural or grounded properties \citep{nominal_kinds} \\
        \hline
    \end{tabular}
    \caption{Categories of Abstract Concepts}
    \label{table:concept_categories}
\end{table*}
\FloatBarrier
\clearpage
\section{Details on Story Dataset Creation} \label{appendix:story_creation}
\setcounter{figure}{0}
\setcounter{table}{0}

To investigate whether and how an LLM tracks concept strength, we created story datasets for each concept, with the concept appearing in only two out of 32 sentences. Each story comprised three 10-sentence paragraphs where the concept was absent, connected by transition sentences in which the concept was present.

To create these stories, we used an LLM in a chat setup. The initial prompt (prompt template shown in Box \ref{box:prominence_main}) instructed the model to generate three 10-sentence paragraphs. The placeholders \{concept\}, \{concept definition\}, and \{context\} were filled with the target concept, its definition, and a specified context, similar to those in Section \ref{section:sub-objective_1_dataset_creation}. After verifying that the story met the required structure, we used a second prompt (prompt template shown in Box \ref{box:prominence_continuation}), with the story and original prompt in context, to instruct the LLM to insert single-sentence transitions between paragraphs. These transitions were designed to have the concept present while maintaining semantic coherence. The placeholder \{concept-related words\} was replaced with a set of terms that would have made the concept explicit, as detailed in the following list:

\begin{itemize} 
    \item \textbf{Ambition}: ``ambition'', ``ambitious'', ``aspire'', or ``aspiration''.
    \item \textbf{Investigation}: ``investigation'', ``investigate'', ``examine'', or ``examination''.
    \item \textbf{Democracy}: ``democracy'', ``democratic'', or ``democratize''.
    \item \textbf{Envy}: ``envy'', ``envious'', ``jealous'', or ``jealousy''.
\end{itemize}

\begin{figure*}[h!]
    \centering

    \begin{mybox}[label={box:prominence_main}]{Story Creation Initial Prompt}
    \label{{box:prominence_main}}
    
    \texttt{Generate a 3-paragraph story, where each paragraph is made up of 10 sentences.}
    
    \texttt{You must abide by that 10-sentence rule.}
    
    \texttt{The paragraphs must be coherent and logically connected to form a meaningful narrative.}
    
    \texttt{All sentences in those paragraphs must be irrelevant to the concept of \{concept\}.}
    
    \texttt{\{concept\} is \{concept definition\}.}
    
    \texttt{The story must be focused on human subjects, not on the environment or animals.}
    
    \texttt{The story must be in the context of \{context\}.}
    
    \texttt{The story must be written so that later, it can be changed to include the concept of \{concept\}, but the original story you generate must have this concept absent.}
    
    \texttt{Do not number the paragraphs or the sentences within the paragraphs and do not include any special characters to highlight the different paragraphs.}
    
    \end{mybox}
    
\end{figure*}

\begin{figure*}[h!]
    \centering

    \begin{mybox}[label={box:prominence_continuation}]{Story Creation Continuation Prompt}
    \label{{box:prominence_continuation}}
    
    \texttt{Given this story, connect each paragraph to the next one with only one connecting sentence per connection.}
    
    \texttt{Each connecting sentence must be coherent and logically connected to both paragraphs it joins.}
    
    \texttt{The tone of the connecting sentences should match the tone of the story.}
    
    \texttt{The concept of \{concept\} must be obvious in the connecting sentences.}
    
    \texttt{\{concept\} is \{concept definition\}.}
    
    \texttt{The connecting sentences must not include words that make \{concept\} explicit such as \{concept-related words\}.}
    
    \texttt{You can make very slight modifications to the original story to ensure that the connecting sentences are coherent and logically connected to the story, but the modified sentences must maintain the irrelevance to \{concept\}.
}
    
    \texttt{Include the whole story with the connecting sentences in your output, not just the connecting sentences.}
    
    \texttt{Do not include any special characters to highlight the connecting sentences.}
    
    \end{mybox}
    
\end{figure*}

After generating each story, we verified that it contained exactly 32 sentences. We then split the story into individual sentences and re-labeled them using the LLM-based classifier from Section \ref{section:sub-objective_1_dataset_creation} to ensure that the main paragraphs did not include the concept, while the transition sentences did. 
To confirm the concept was not mentioned explicitly, we checked for the presence of its word stems, as detailed in Appendix \ref{appendix:concept_dataset_creation_stems}, and excluded any stories where they appeared.
\section{LLM Details} \label{appendix:LLM_details}

Table \ref{table:LLM_details} provides details for the LLMs evaluated in this work.

\begin{table}[h]
\centering
\begin{tabular}{c|c|c|c}
\hline\hline
\textbf{Model} & \textbf{Model} & \textbf{Number} & \textbf{Embedd-} \\
\textbf{Family} & \textbf{size} & \textbf{of layers} & \textbf{ing size} \\
\hline\hline
\texttt{Llama-3} & 8B & 32 & 4,096 \\
\hline
\multirow{2}{*}{\texttt{Gemma-2}} & 2B & 26 & 2,304 \\
 & 9B & 42 & 3,584 \\
\hline
\multirow{4}{*}{\texttt{Qwen2.5}} & 0.5B & 24 & 896 \\
 & 1.5B & 28 & 1,536 \\
 & 3B & 36 & 2,048 \\
 & 7B & 28 & 3,584 \\
\hline\hline
\end{tabular}
\caption{Details for the studied LLMs}
\label{table:LLM_details}
\end{table}
\section{Probe Training Specifications} \label{appendix:probe_training_specs}

In our experiments, we used the following settings to train the probes:
\begin{itemize}
    \item Optimizer: Adam
    \item Learning rate: 0.005
    \item Batch size: 512
    \item Number of epochs: 500 (with early stopping)
    \item  Train/validation/test split: 70\%/10\%/20\%
\end{itemize}
\clearpage
\clearpage
\section{Extended Results for Inference of Concepts} \label{appendix:exp1_extended}
\setcounter{figure}{0}
\setcounter{table}{0}

\subsection{Probe Accuracies for all Concepts}

\begin{figure}[H]
    \centering
    \includegraphics[width=\linewidth]{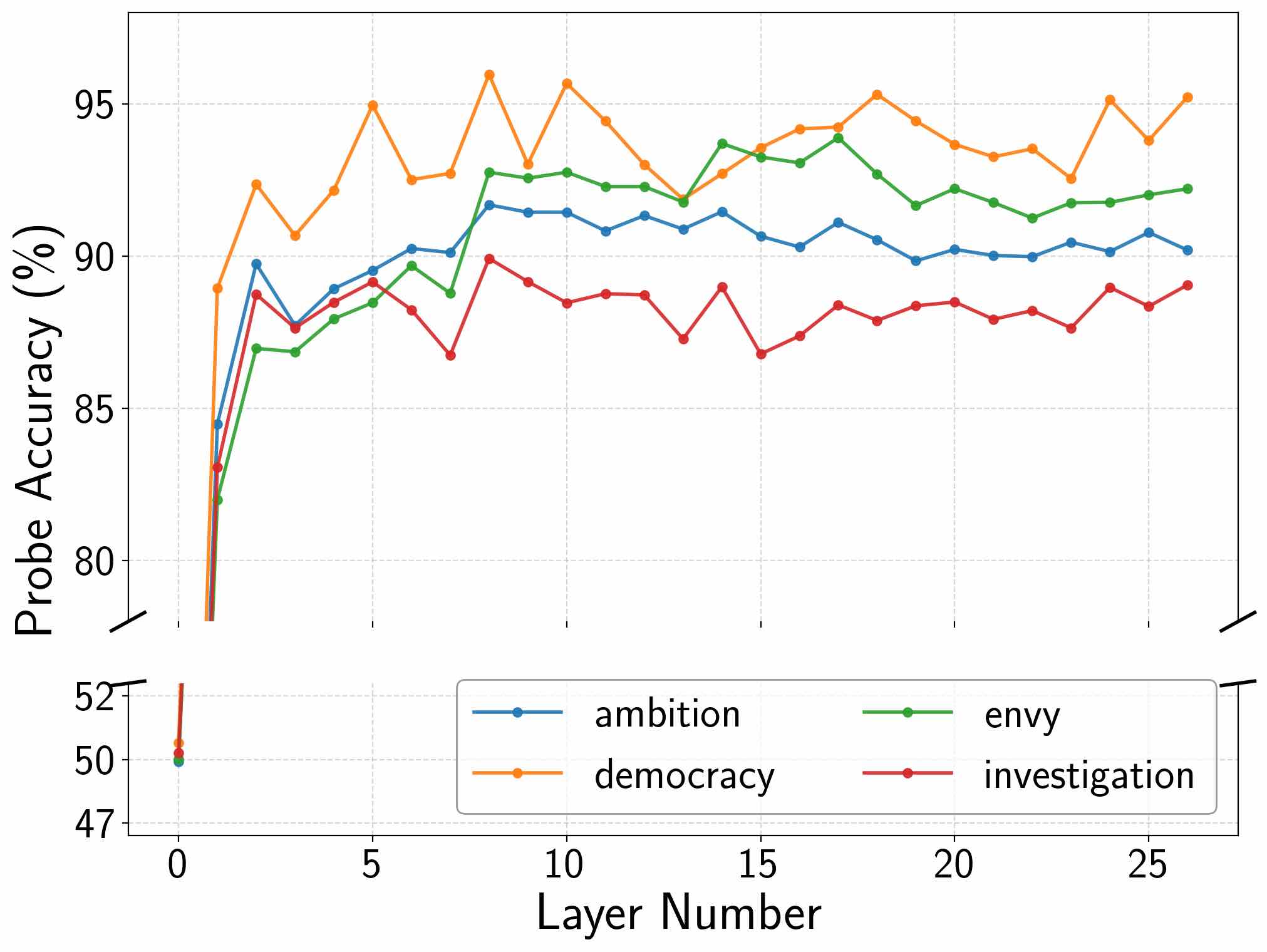}
    \caption{Probe accuracies across layers for all concepts in \texttt{Gemma-2-2B}}
    \label{fig:concepts_gemma2B}
\end{figure}

\begin{figure}[H]
    \centering
    \includegraphics[width=\linewidth]{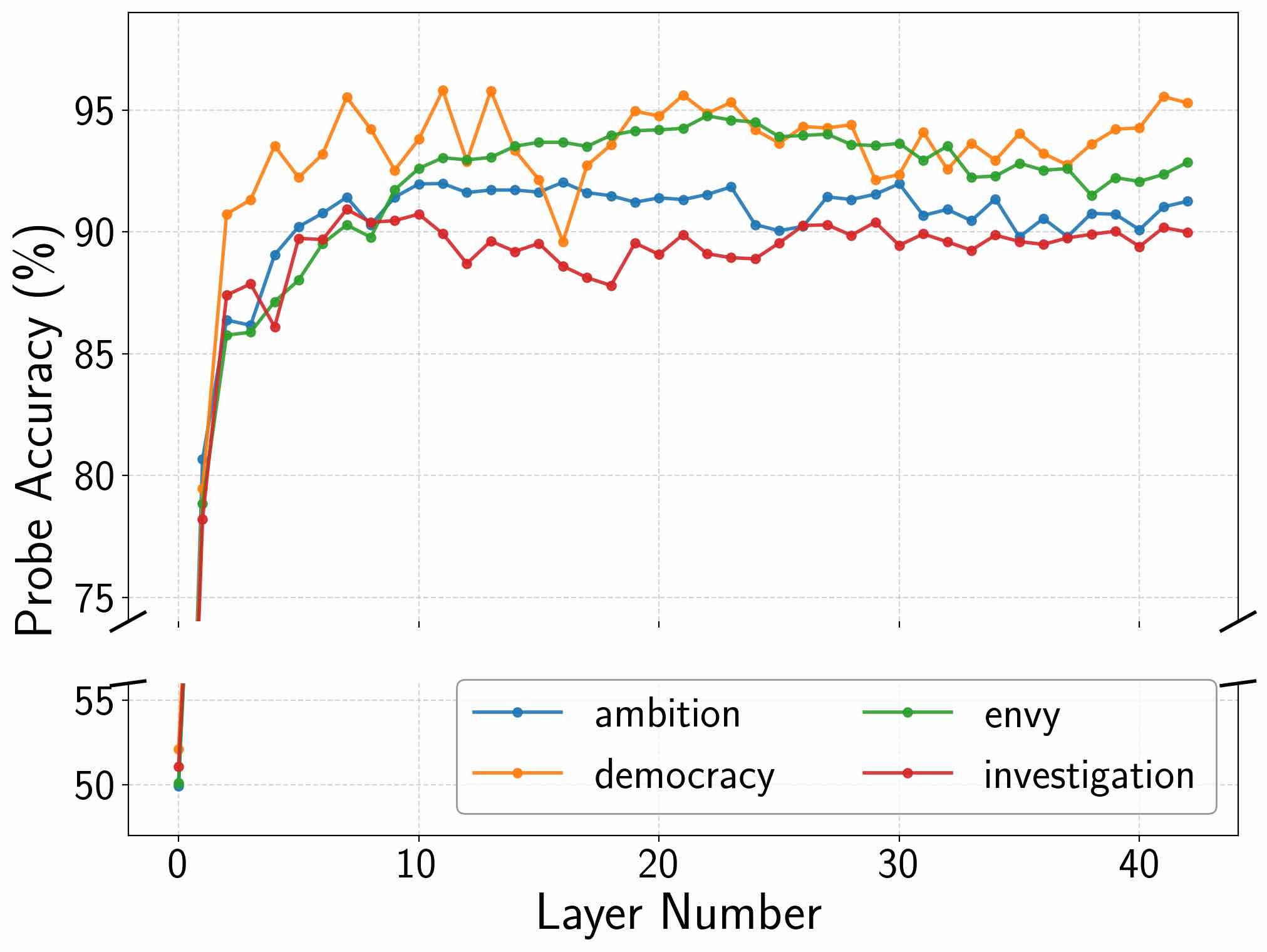}
    \caption{Probe accuracies across layers for all concepts in \texttt{Gemma-2-9B}}
    \label{fig:concepts_gemma9B}
\end{figure}

\begin{figure}[H]
    \centering
    \includegraphics[width=\linewidth]{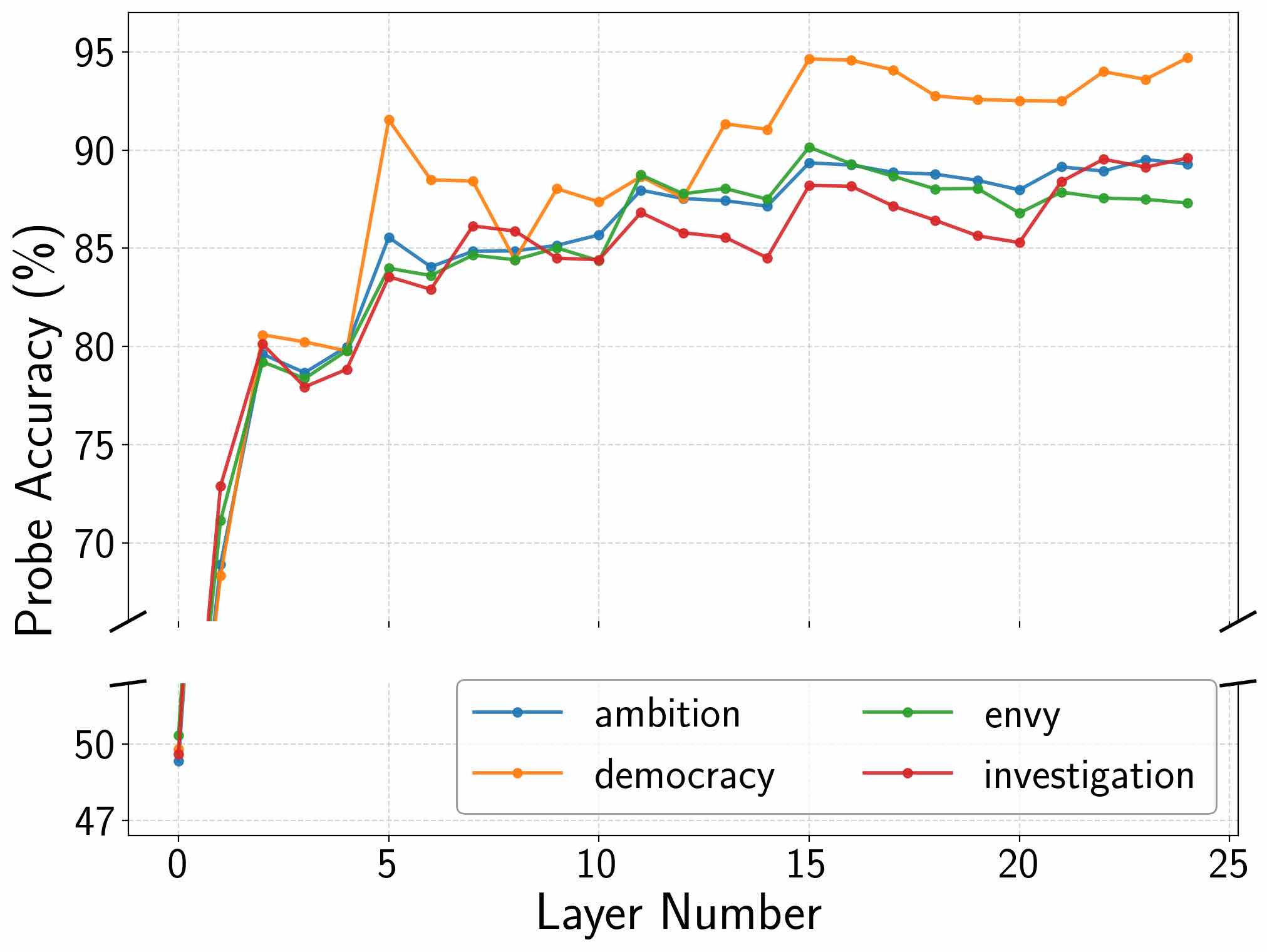}
    \caption{Probe accuracies across layers for all concepts in \texttt{Qwen2.5-0.5B}}
    \label{fig:concepts_qwen0p5}
\end{figure}

\begin{figure}[H]
    \centering
    \includegraphics[width=\linewidth]{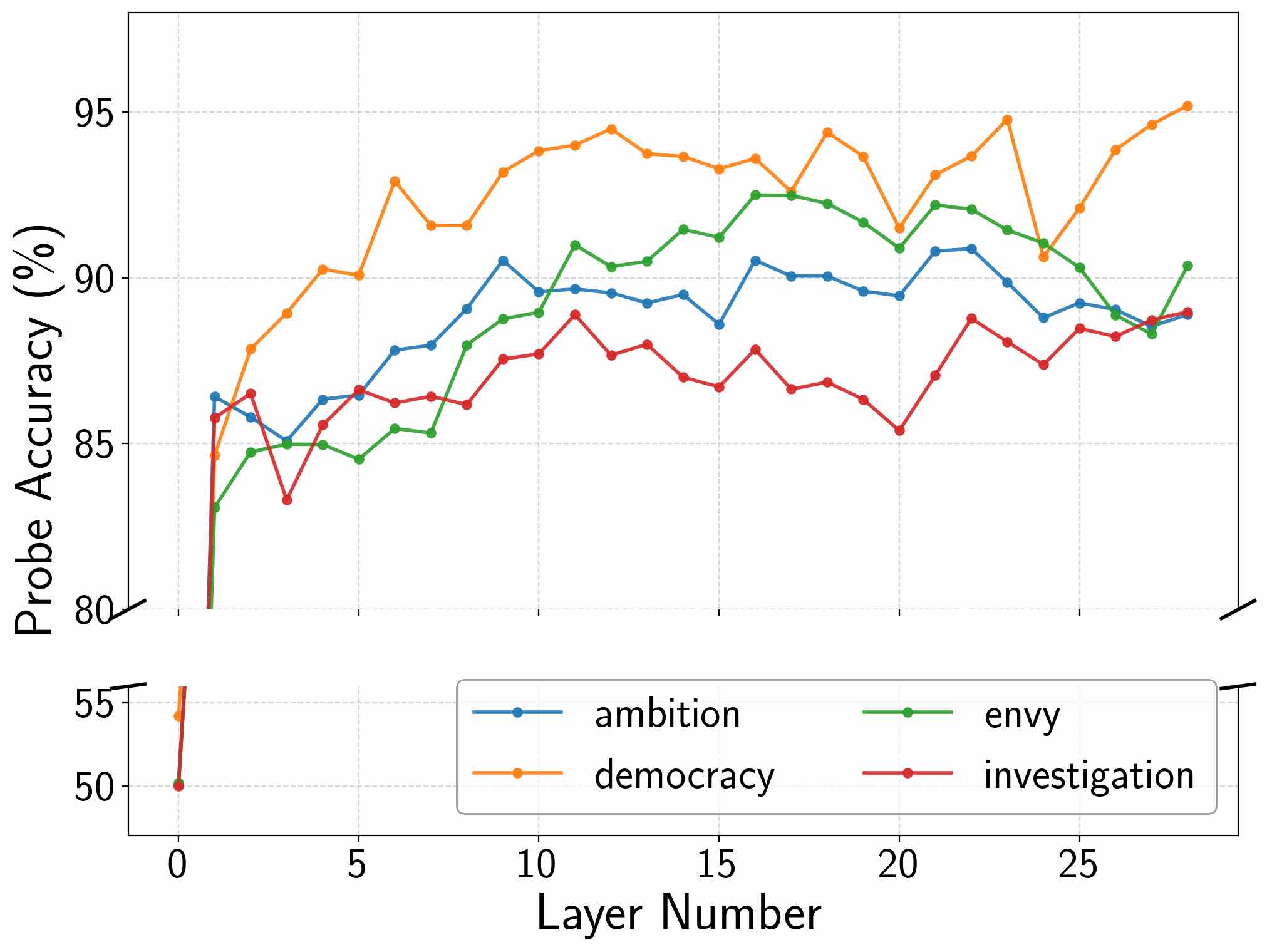}
    \caption{Probe accuracies across layers for all concepts in \texttt{Qwen2.5-1.5B}}
    \label{fig:concepts_qwen1p5}
\end{figure}

\begin{figure}[H]
    \centering
    \includegraphics[width=\linewidth]{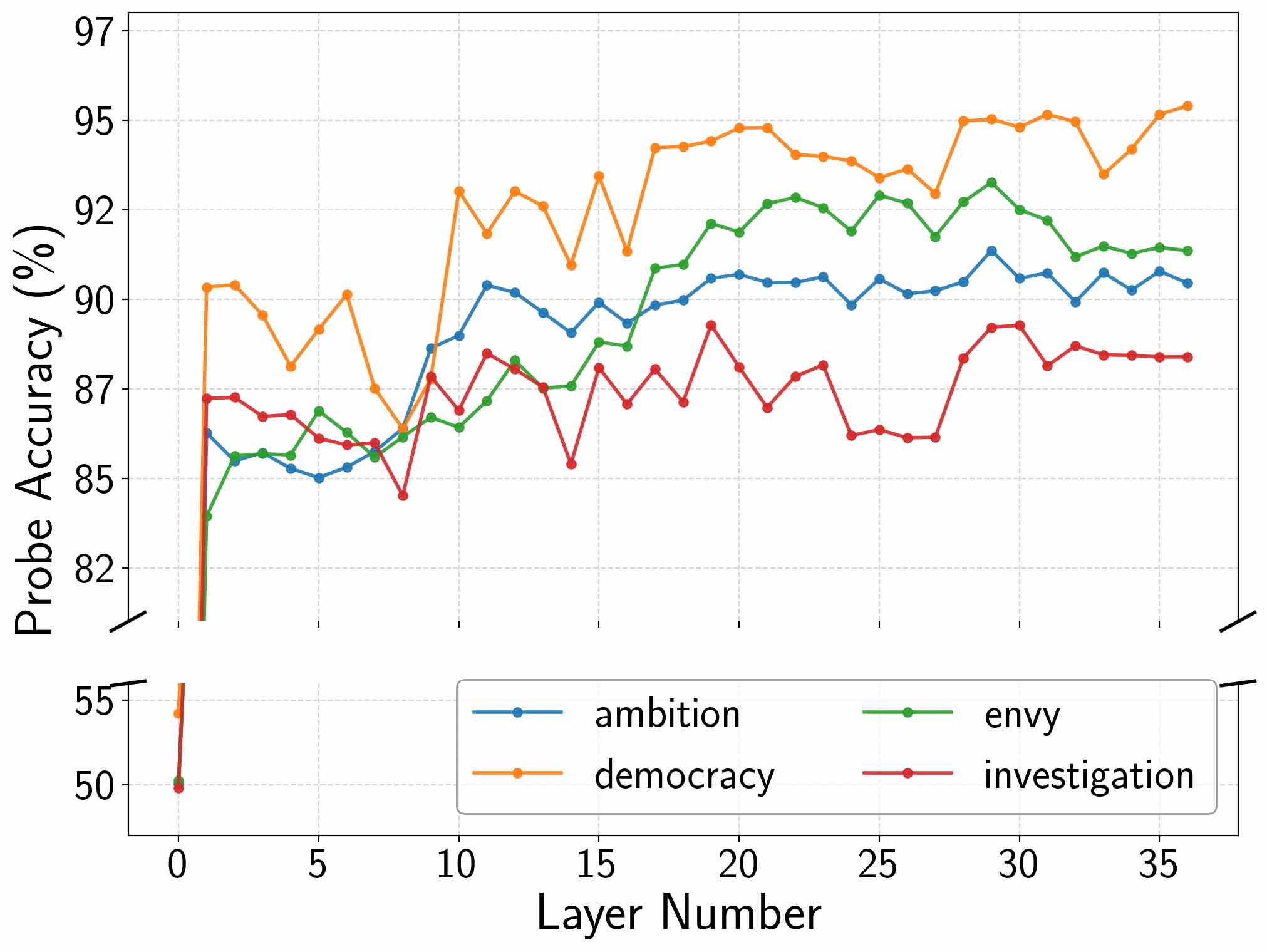}
    \caption{Probe accuracies across layers for all concepts in \texttt{Qwen2.5-3B}}
    \label{fig:concepts_qwen3}
\end{figure}

\begin{figure}[H]
    \centering
    \includegraphics[width=\linewidth]{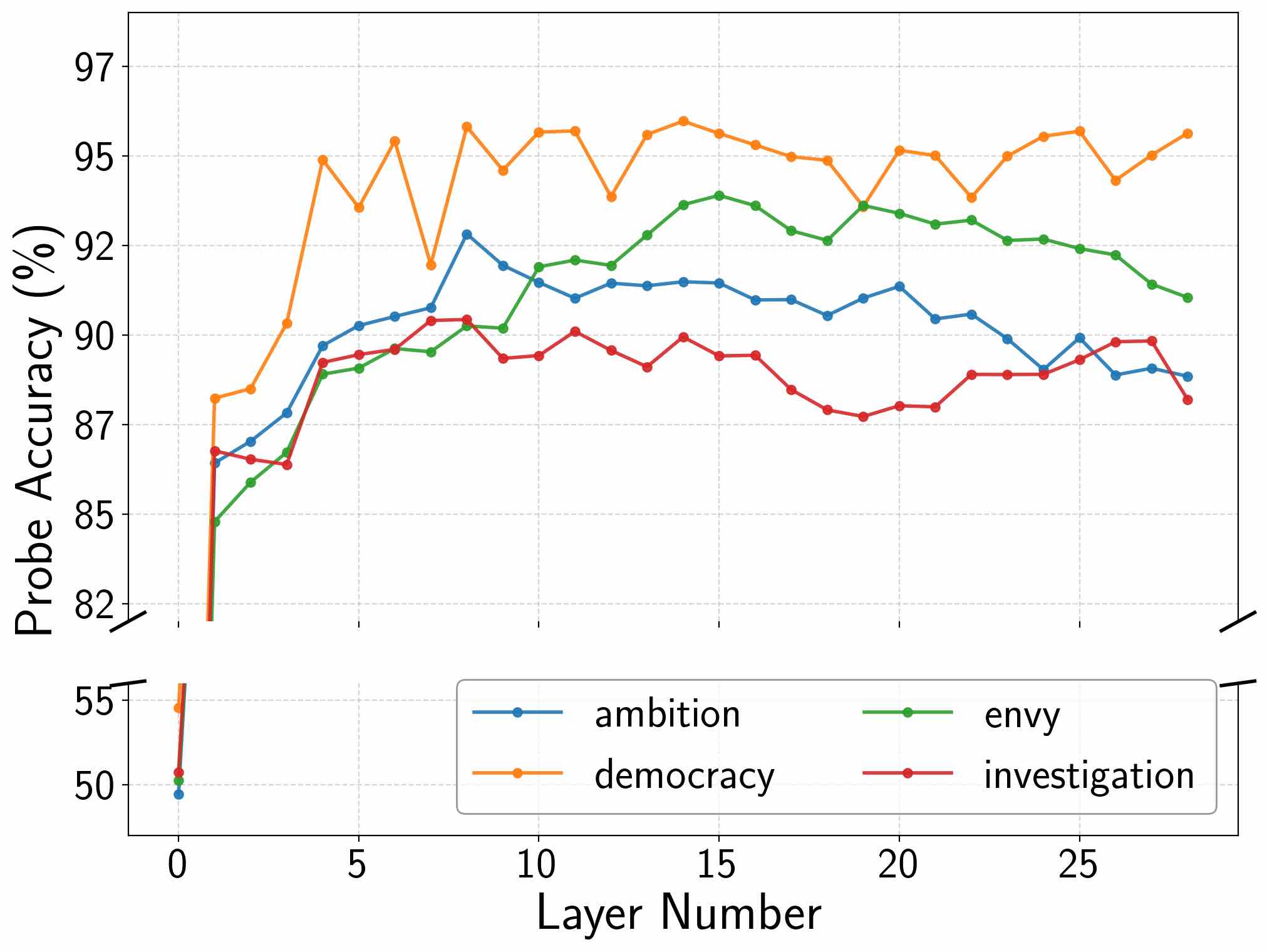}
    \caption{Probe accuracies across layers for all concepts in \texttt{Qwen2.5-7B}}
    \label{fig:concepts_qwen7}
\end{figure}

Figures \ref{fig:concepts_gemma2B}--\ref{fig:concepts_qwen7} illustrate the probe accuracies for all concepts across the LLM layers for each model included in this investigation.

\subsection{Extended Results for Inference of Ambition}

\subsubsection{Probing for Ambition using Nth Embedding vs. Average Embedding}

\begin{figure}[H]
    \centering
    \includegraphics[width=\linewidth]{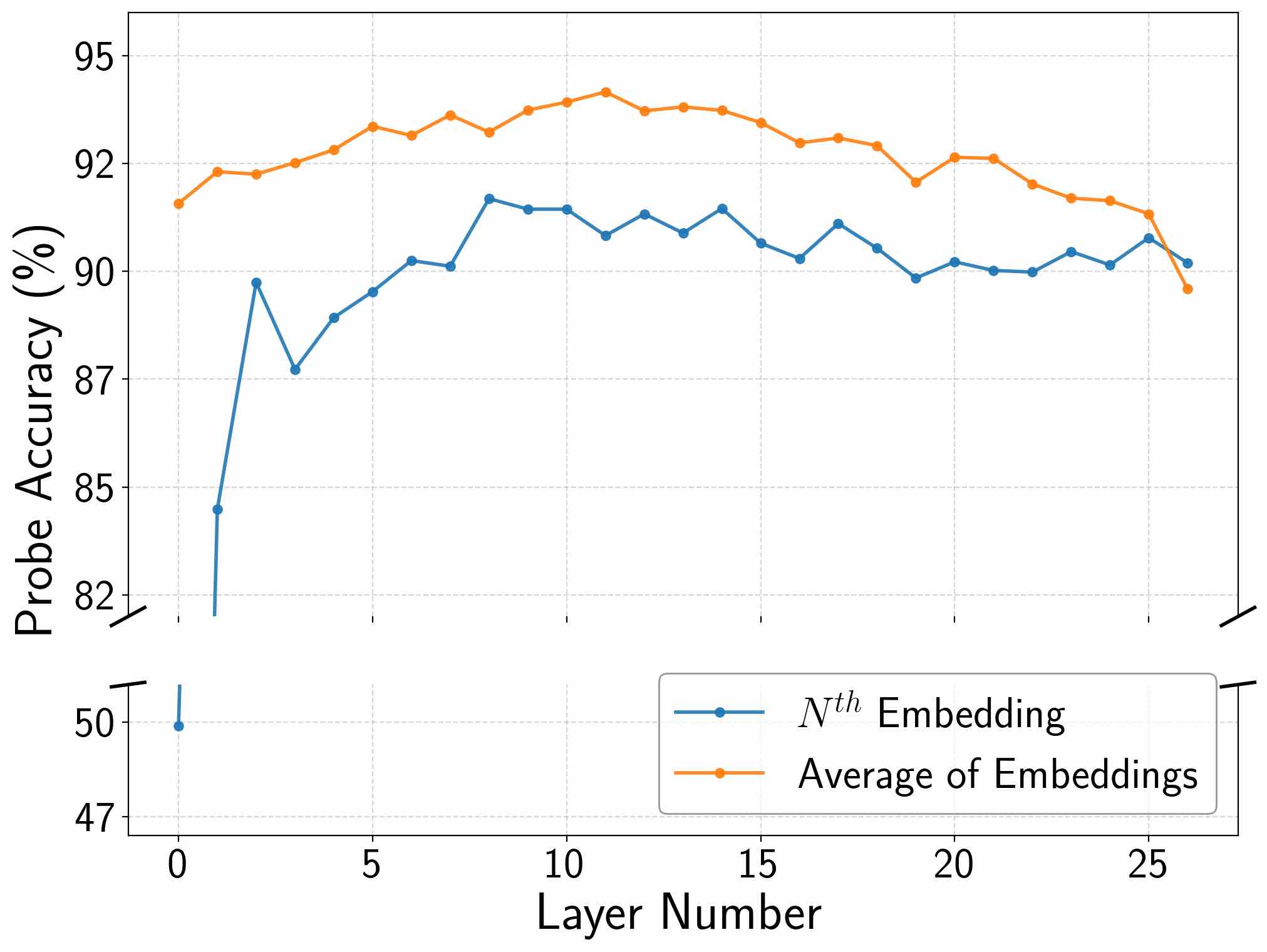}
    \caption{\textbf{Ambition} probe accuracy for \texttt{Gemma-2-2B} using average and $N^{th}$ embeddings vs. layer}
    \label{fig:ambition_gemma2b_right_most_vs_average}
\end{figure}

\begin{figure}[H]
    \centering
    \includegraphics[width=\linewidth]{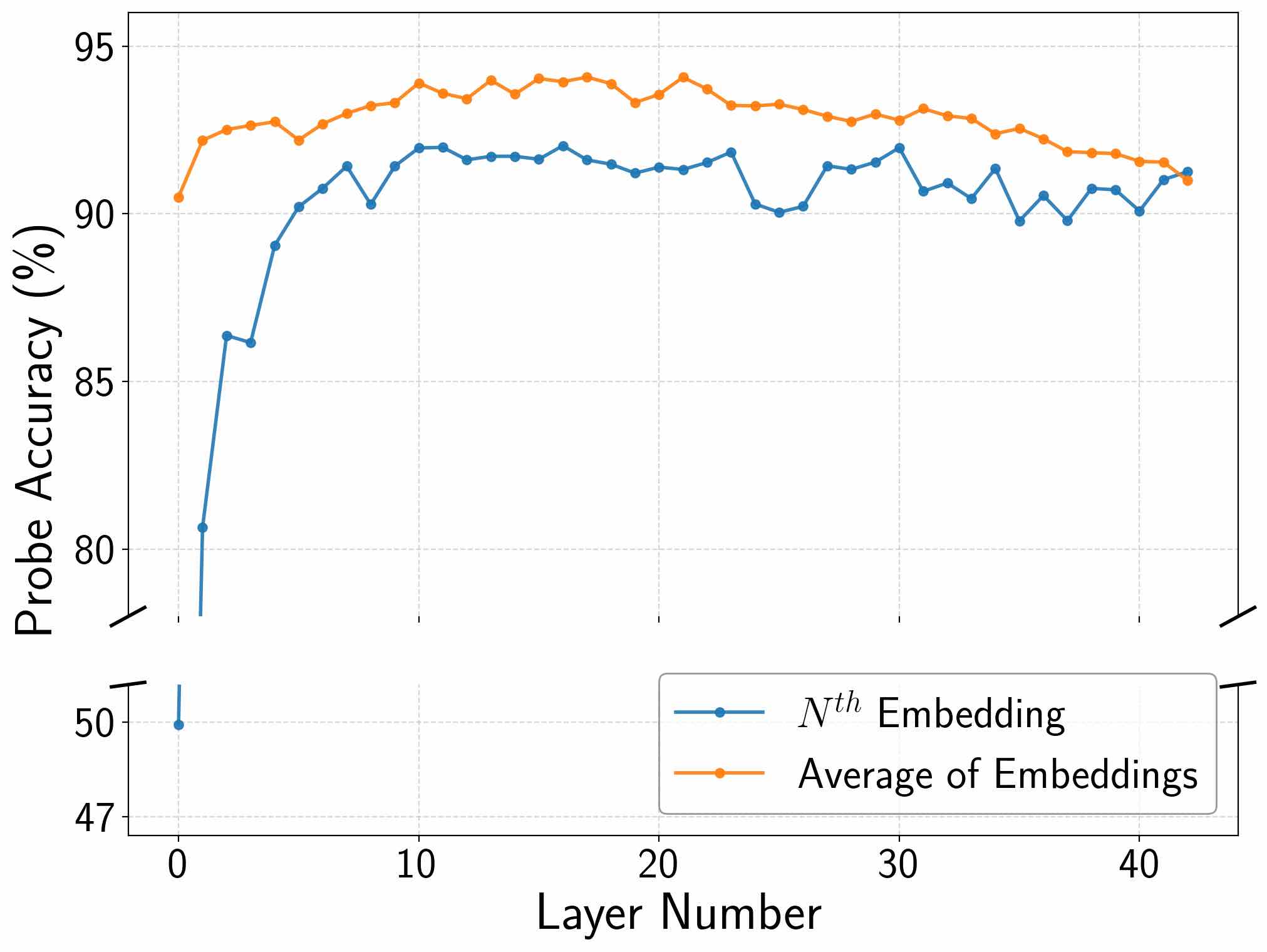}
    \caption{\textbf{Ambition} probe accuracy for \texttt{Gemma-2-9B} using average and $N^{th}$ embeddings vs. layer}
    \label{fig:ambition_gemma9b_right_most_vs_average}
\end{figure}

\begin{figure}[H]
    \centering
    \includegraphics[width=\linewidth]{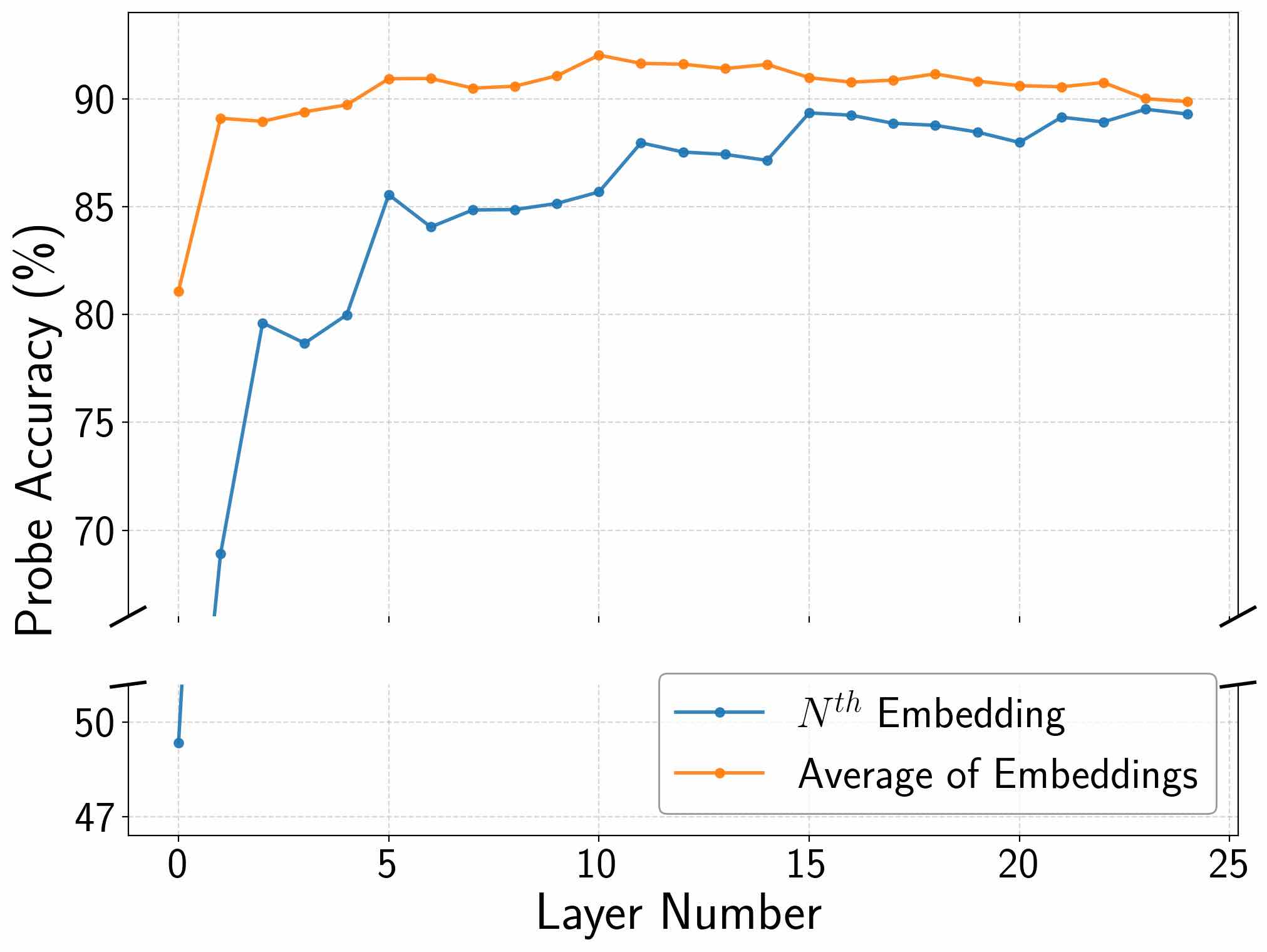}
    \caption{\textbf{Ambition} probe accuracy for \texttt{Qwen2.5-0.5B} using average and $N^{th}$ embeddings vs. layer}
    \label{fig:ambition_qwen0p5b_right_most_vs_average}
\end{figure}

\begin{figure}[H]
    \centering
    \includegraphics[width=\linewidth]{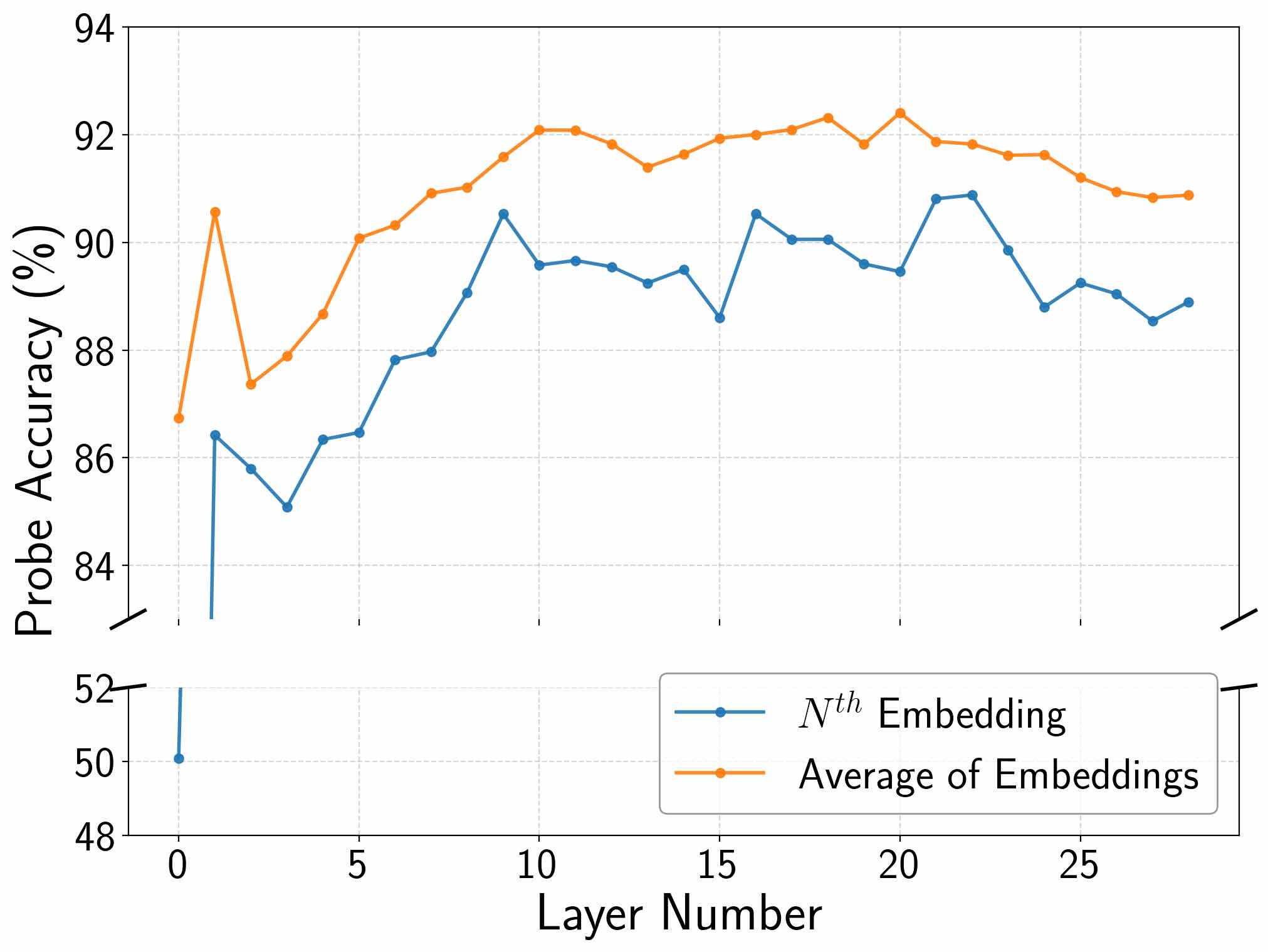}
    \caption{\textbf{Ambition} probe accuracy for \texttt{Qwen2.5-1.5B} using average and $N^{th}$ embeddings vs. layer}
    \label{fig:ambition_qwen1p5b_right_most_vs_average}
\end{figure}

\begin{figure}[H]
    \centering
    \includegraphics[width=\linewidth]{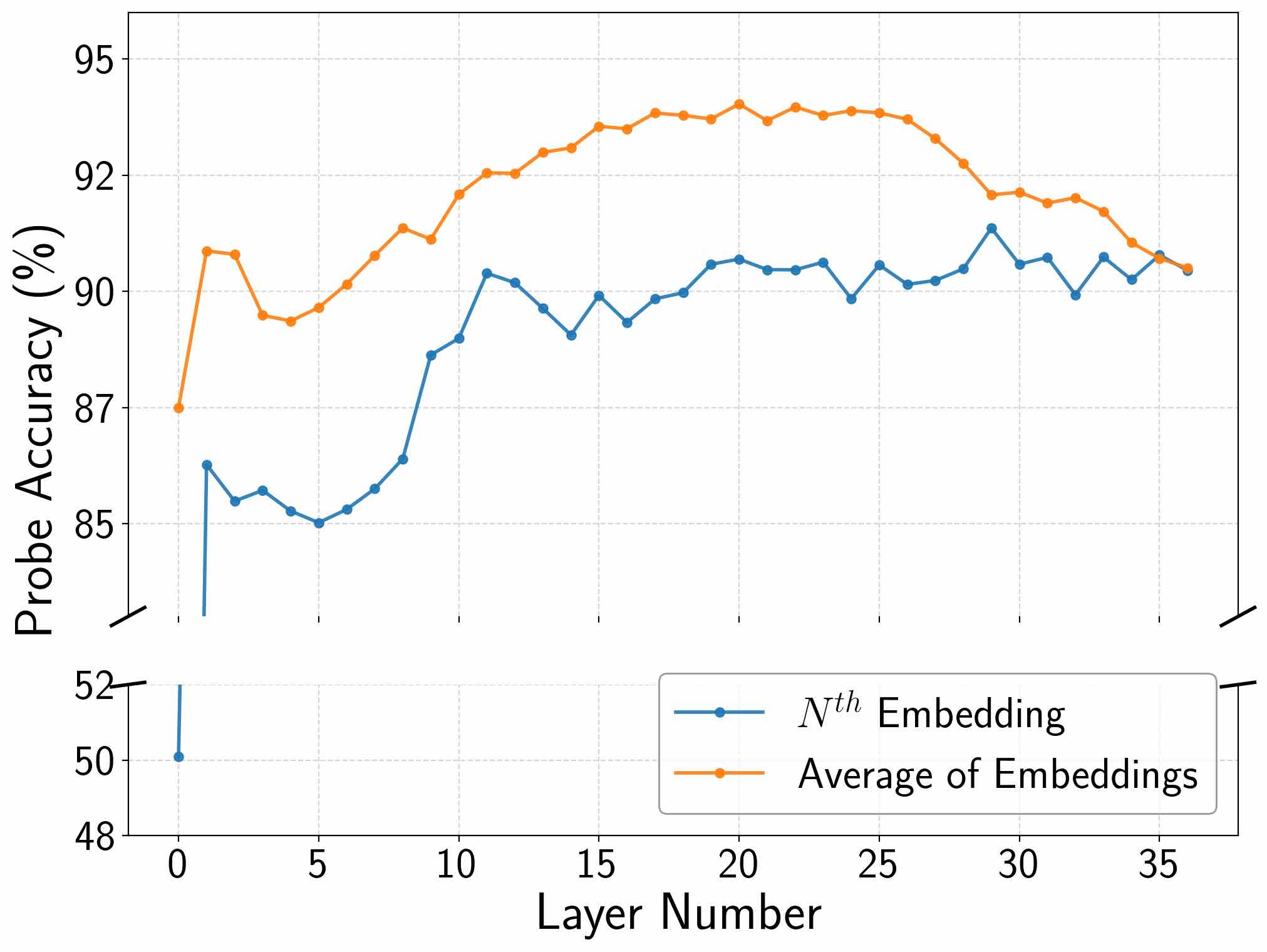}
    \caption{\textbf{Ambition} probe accuracy for \texttt{Qwen2.5-3B} using average and $N^{th}$ embeddings vs. layer}
    \label{fig:ambition_qwen3b_right_most_vs_average}
\end{figure}

\begin{figure}[H]
    \centering
    \includegraphics[width=\linewidth]{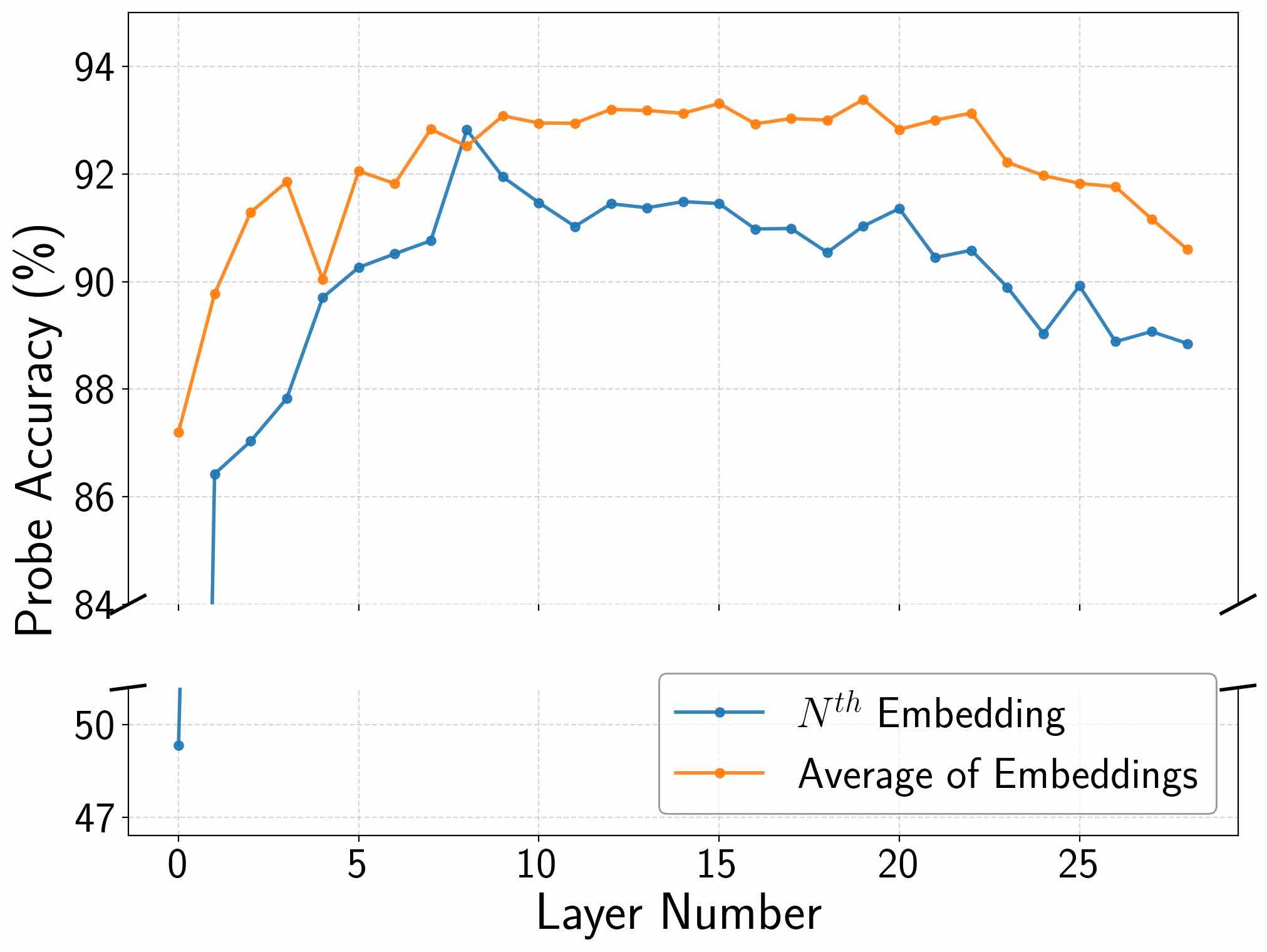}
    \caption{\textbf{Ambition} probe accuracy for \texttt{Qwen2.5-7B} using average and $N^{th}$ embeddings vs. layer}
    \label{fig:ambition_qwen7b_right_most_vs_average}
\end{figure}

Figures \ref{fig:ambition_gemma2b_right_most_vs_average}--\ref{fig:ambition_qwen7b_right_most_vs_average} illustrate the \textbf{Ambition} probe accuracies across layers of all LLMs using both the $N^{th}$ embedding and the average of all embeddings in the respective layer.

\subsubsection{Ambition Probe Cross-Check}

\begin{figure}[H]
    \centering
    \includegraphics[width=\linewidth]{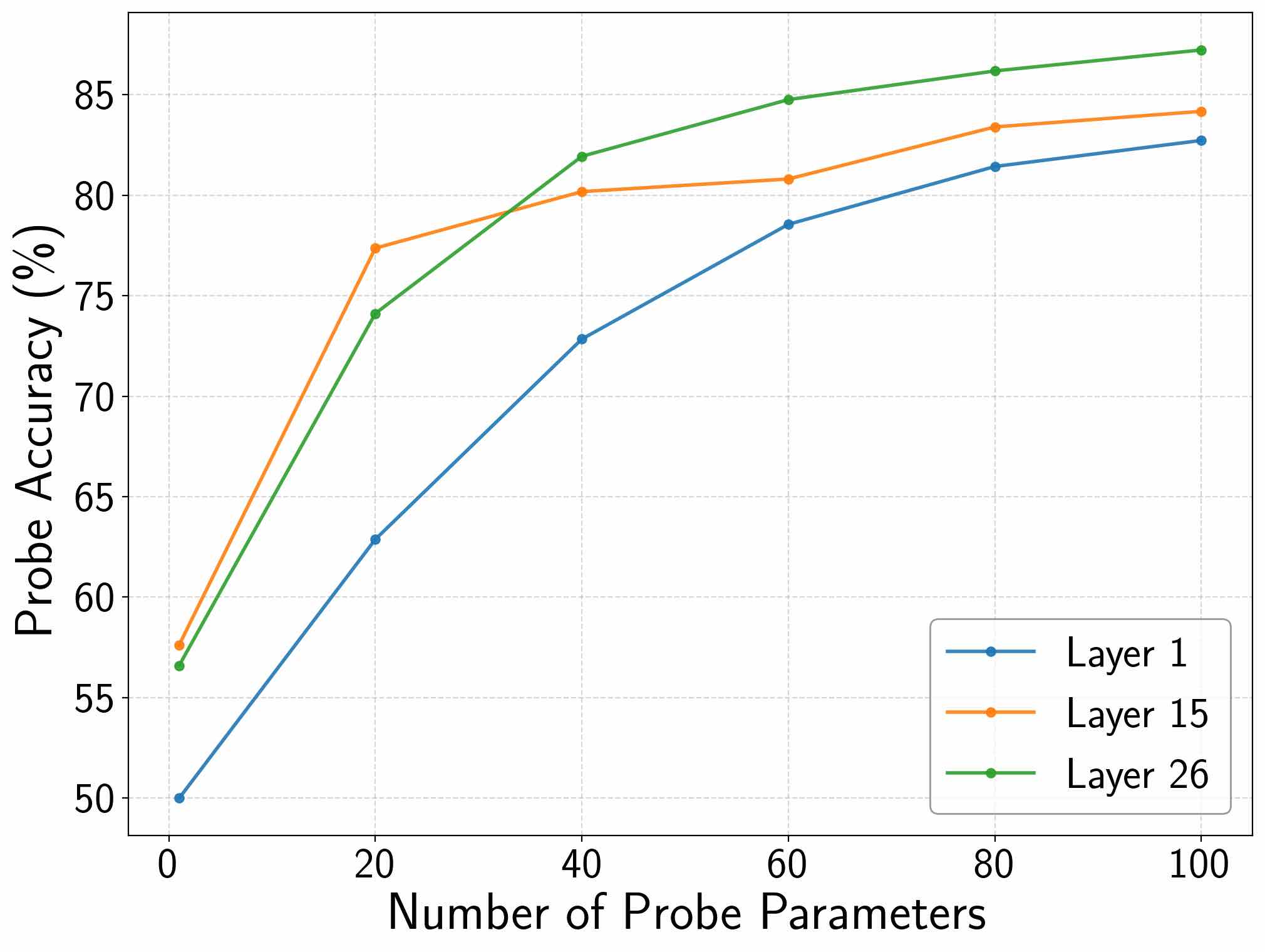}
    \caption{\textbf{Ambition} probe accuracy for \texttt{Gemma-2-2B} as a function of probe size}
    \label{fig:ambition_gemma2b_vs_probe_params}
\end{figure}

\begin{figure}[H]
    \centering
    \includegraphics[width=\linewidth]{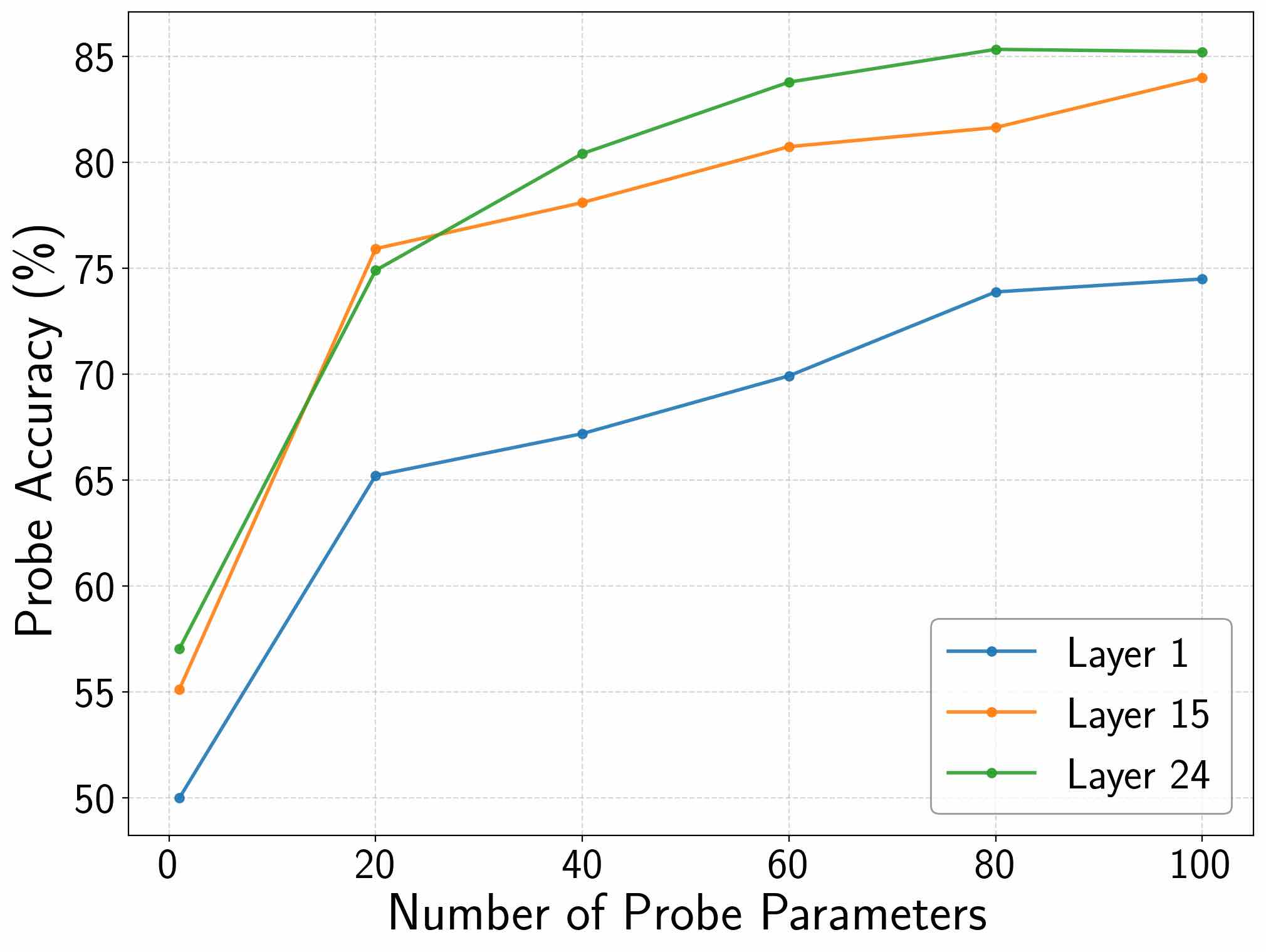}
    \caption{\textbf{Ambition} probe accuracy for \texttt{Qwen2.5-0.5B} as a function of probe size}
    \label{fig:ambition_qwen0p5b_vs_probe_params}
\end{figure}

\begin{figure}[H]
    \centering
    \includegraphics[width=\linewidth]{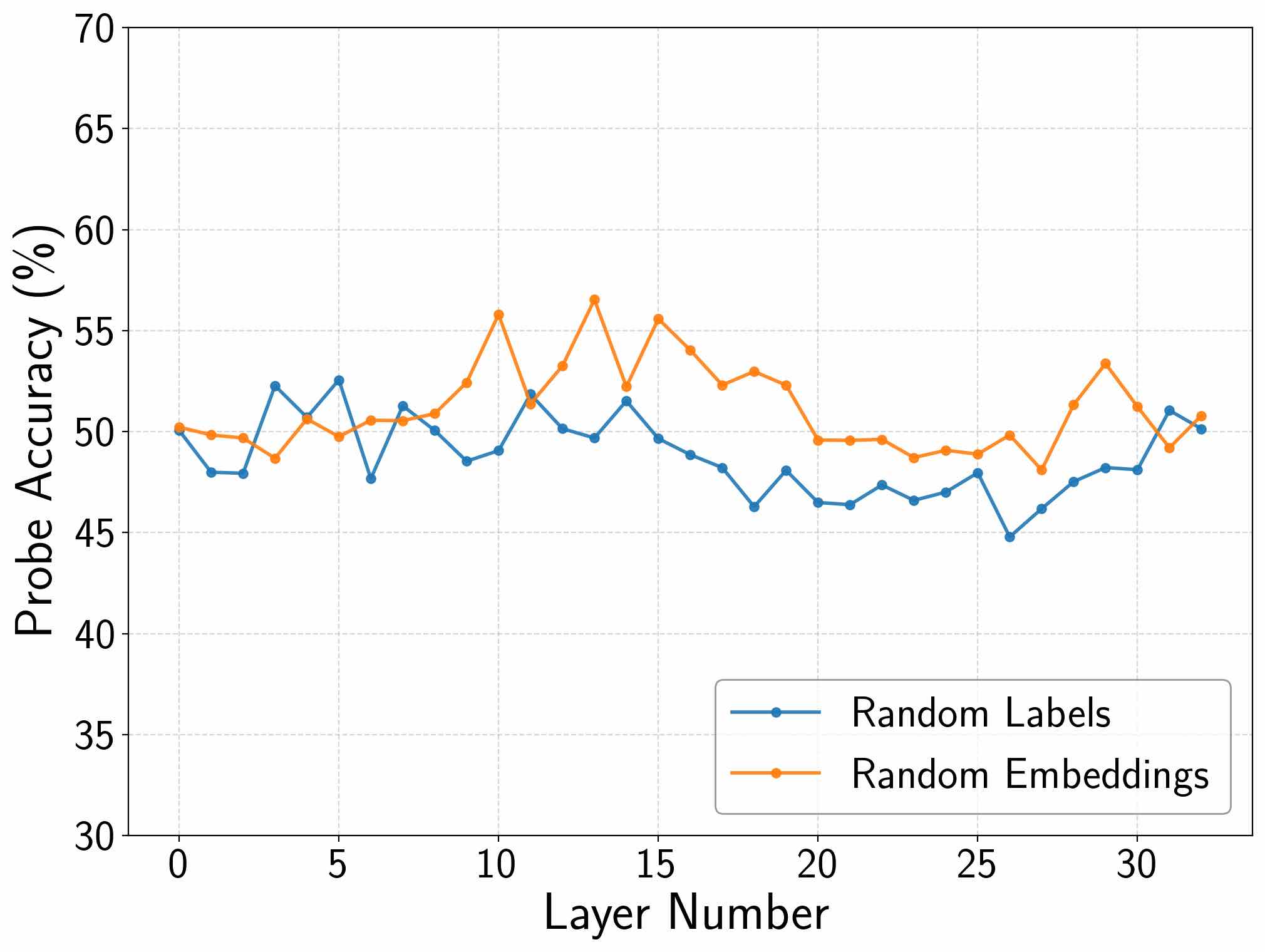}
    \caption{\textbf{Ambition} probe accuracy across layers in \texttt{Llama-3-8B} using random embeddings or random labels during probe training}
    \label{fig:ambition_meta3b_control}
\end{figure}

\begin{figure}[H]
    \centering
    \includegraphics[width=\linewidth]{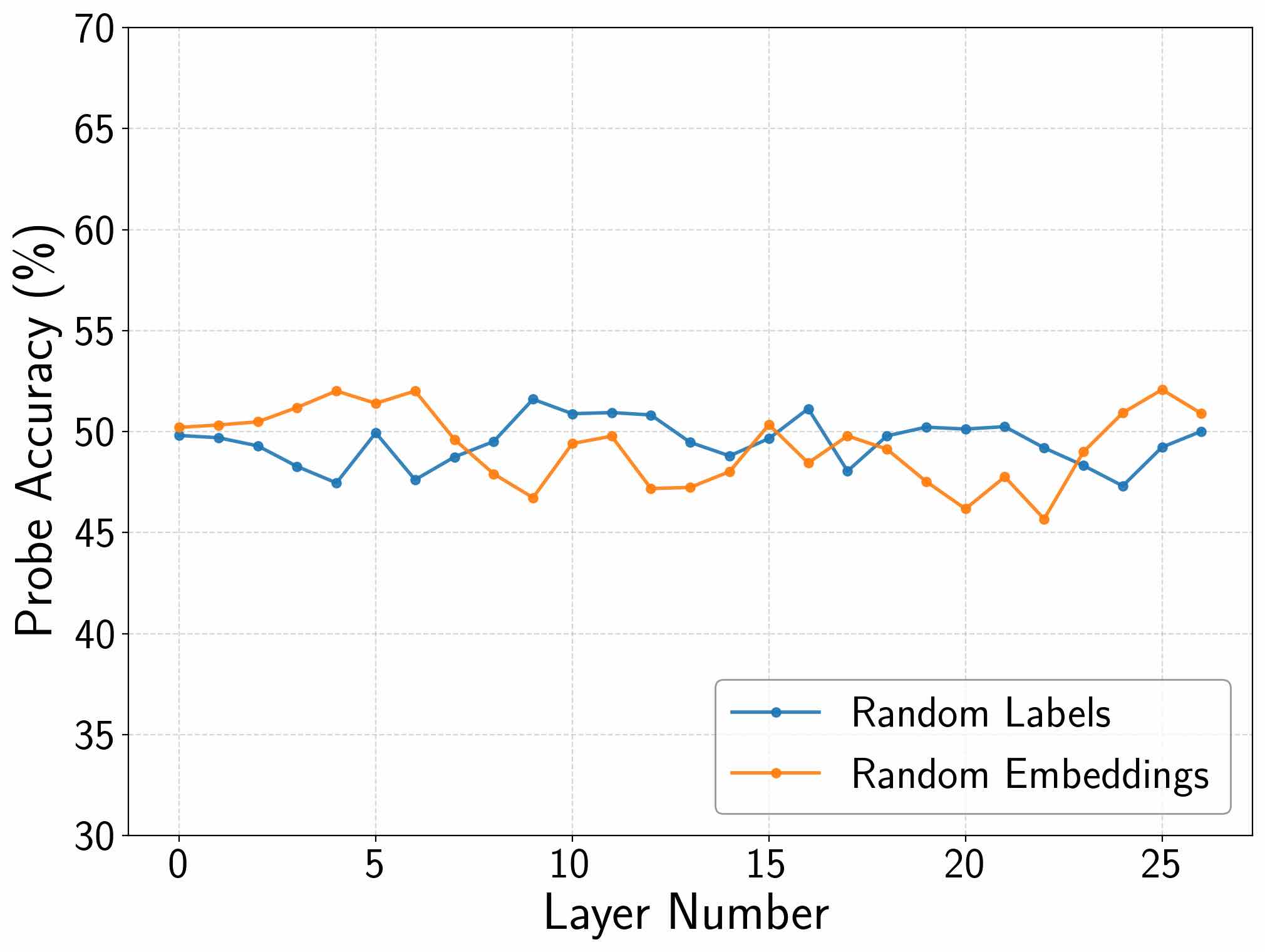}
    \caption{\textbf{Ambition} probe accuracy across layers in \texttt{Gemma-2-2B} using random embeddings or random labels during probe training}
    \label{fig:ambition_gemma2b_control}
\end{figure}

\begin{figure}[H]
    \centering
    \includegraphics[width=\linewidth]{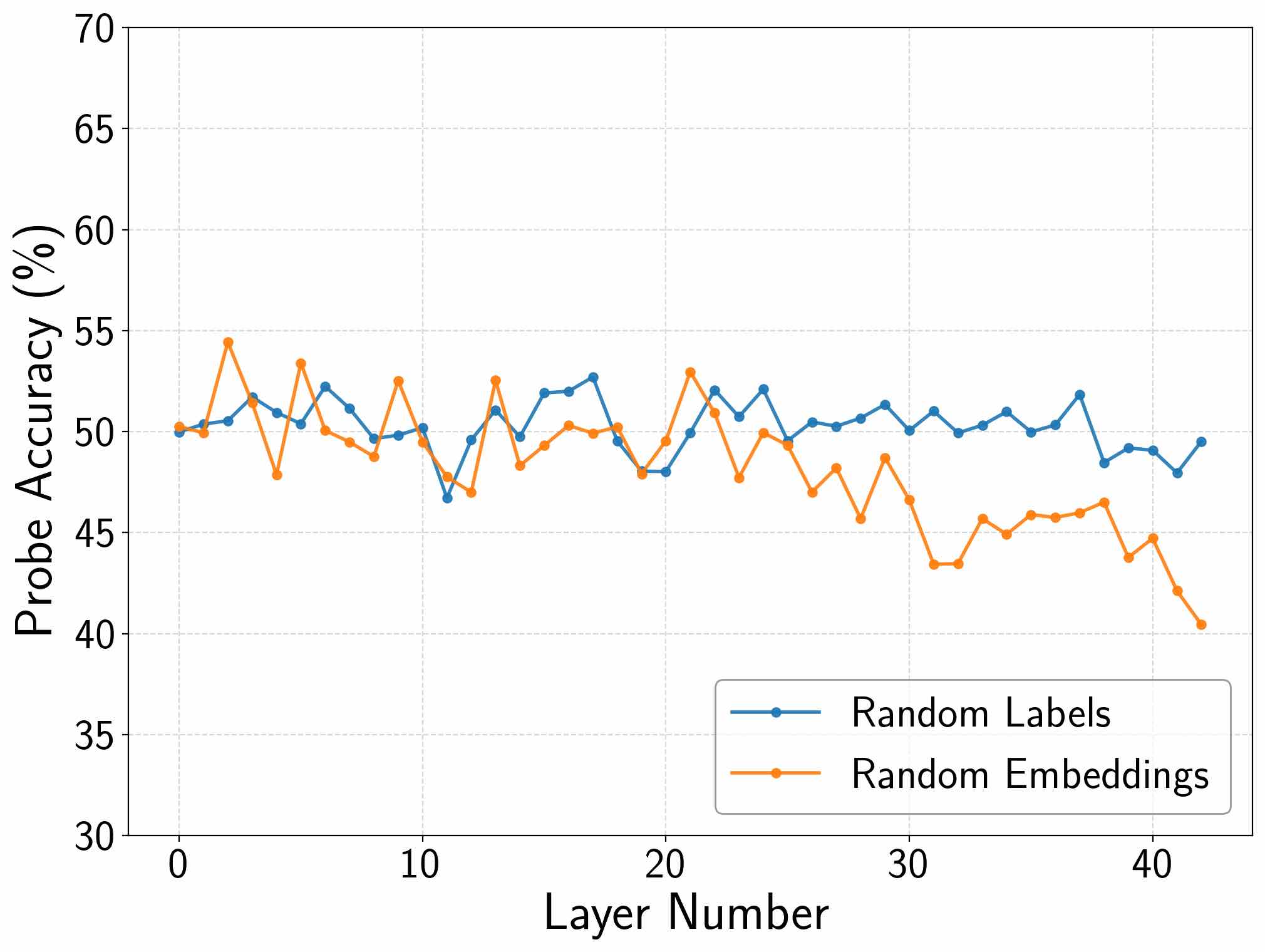}
    \caption{\textbf{Ambition} probe accuracy across layers in \texttt{Gemma-2-9B} using random embeddings or random labels during probe training}
    \label{fig:ambition_gemma9b_control}
\end{figure}

\begin{figure}[H]
    \centering
    \includegraphics[width=\linewidth]{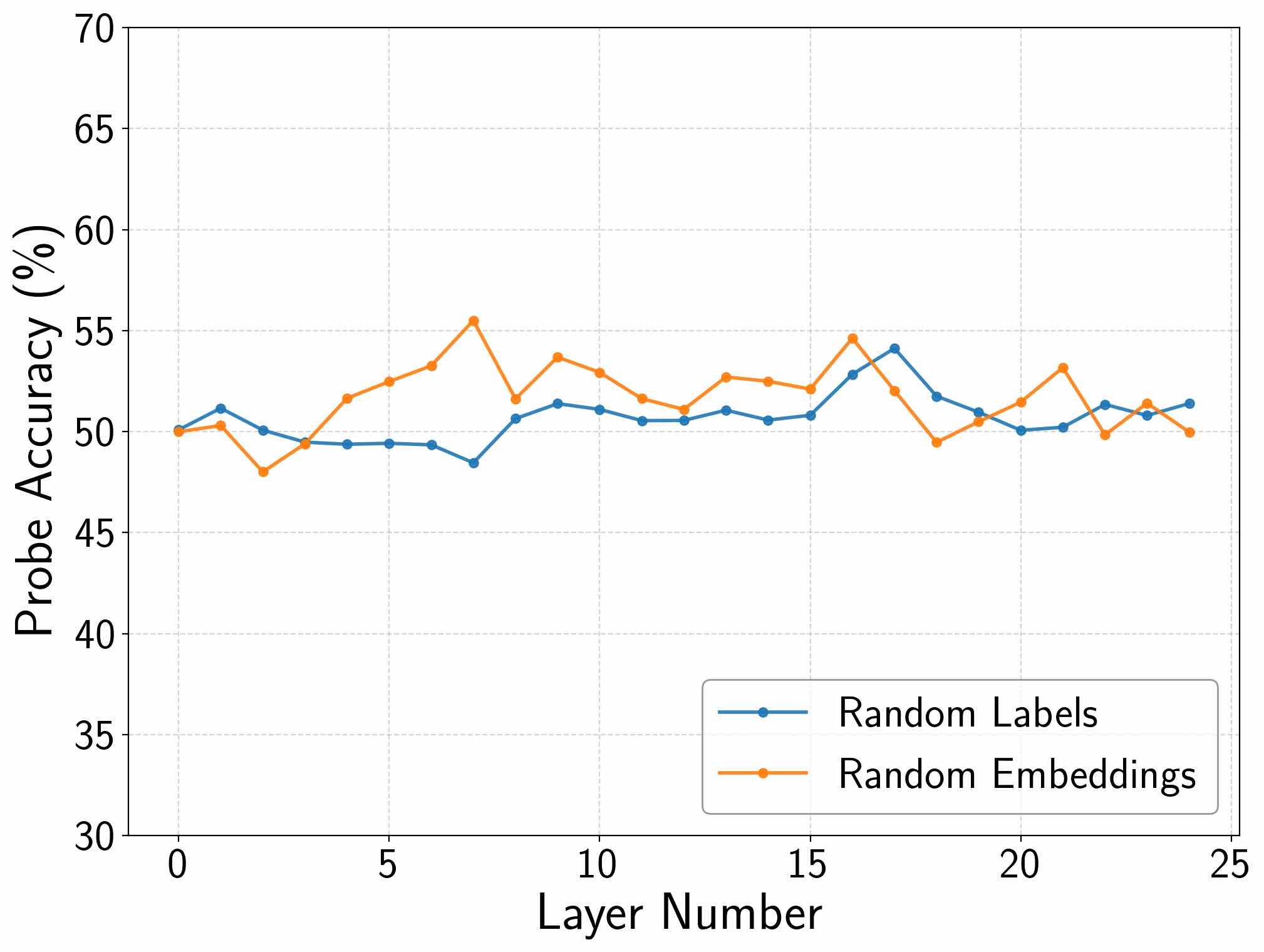}
    \caption{\textbf{Ambition} probe accuracy across layers in \texttt{Qwen2.5-0.5B} using random embeddings or random labels during probe training}
    \label{fig:ambition_qwen0p5b_control}
\end{figure}

\begin{figure}[H]
    \centering
    \includegraphics[width=\linewidth]{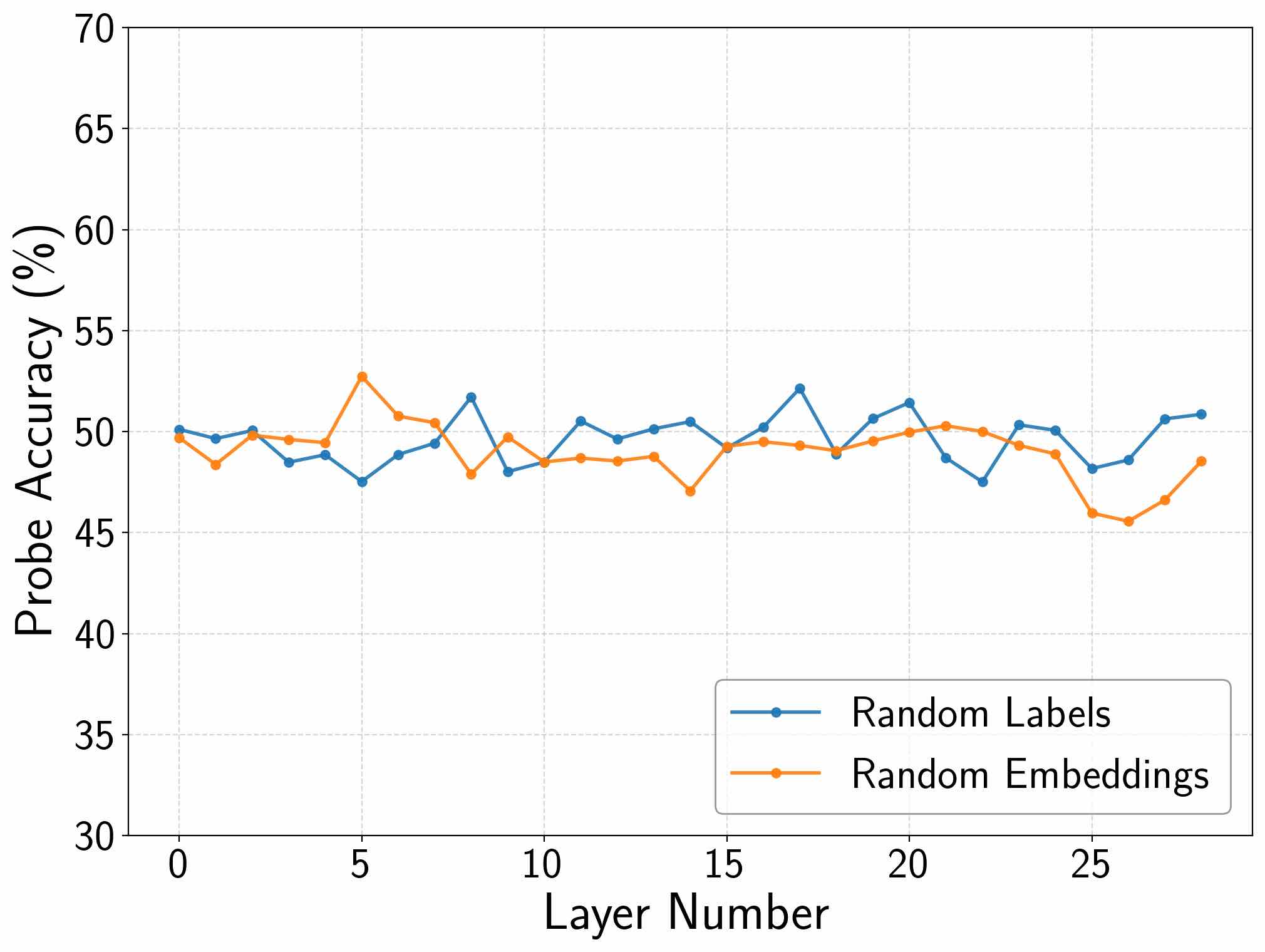}
    \caption{\textbf{Ambition} probe accuracy across layers in \texttt{Qwen2.5-1.5B} using random embeddings or random labels during probe training}
    \label{fig:ambition_qwen1p5b_control}
\end{figure}

\begin{figure}[H]
    \centering
    \includegraphics[width=\linewidth]{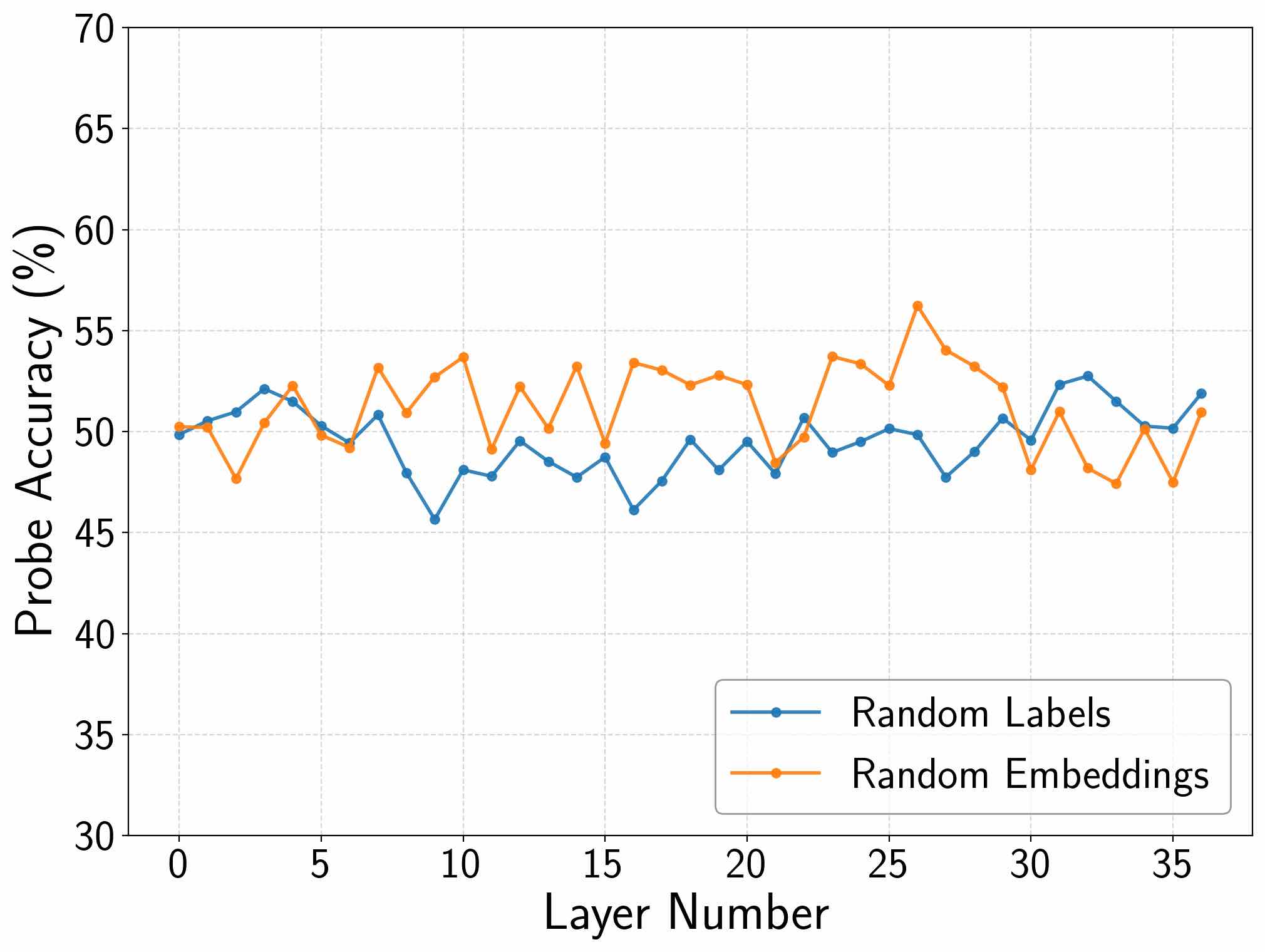}
    \caption{\textbf{Ambition} probe accuracy across layers in \texttt{Qwen2.5-3B} using random embeddings or random labels during probe training}
    \label{fig:ambition_qwen3b_control}
\end{figure}

\begin{figure}[H]
    \centering
    \includegraphics[width=\linewidth]{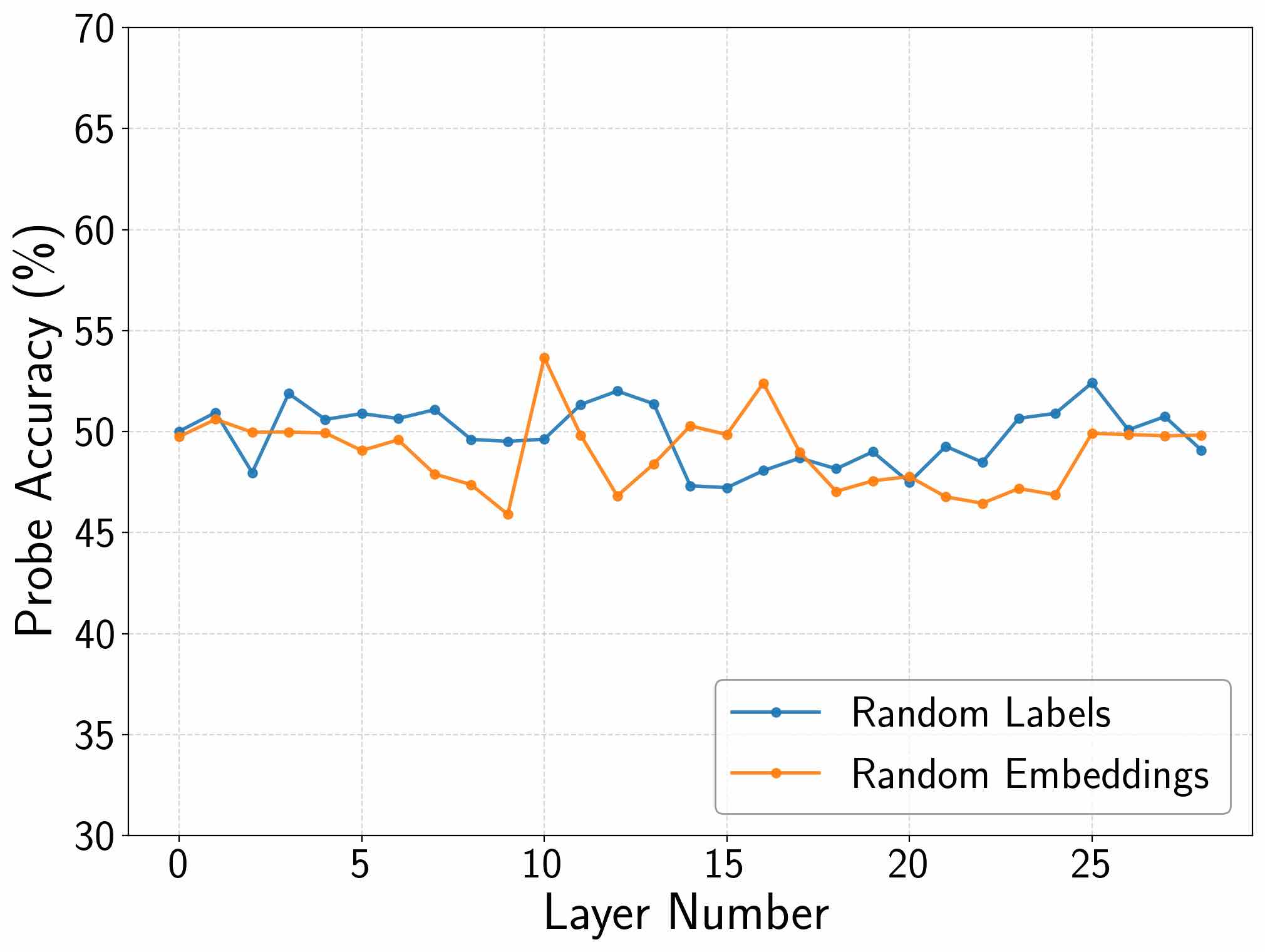}
    \caption{\textbf{Ambition} probe accuracy across layers in \texttt{Qwen2.5-7B} using random embeddings or random labels during probe training}
    \label{fig:ambition_qwen7b_control}
\end{figure}

Figures \ref{fig:ambition_gemma2b_vs_probe_params} and \ref{fig:ambition_qwen0p5b_vs_probe_params} show the \textbf{Ambition} probe accuracy versus probe size for \texttt{Gemma-2-2B} and \texttt{Qwen2.5-0.5B}, respectively. Figures \ref{fig:ambition_meta3b_control}--\ref{fig:ambition_qwen7b_control} show the probe accuracies across layers for all LLMs when the probes are trained on the control task (randomizing embeddings or labels).

\subsection{Extended Results for Inference of Investigation}

\subsubsection{Probing for Investigation using Nth Embedding vs. Average Embedding}
\begin{figure}[H]
    \centering
    \includegraphics[width=\linewidth]{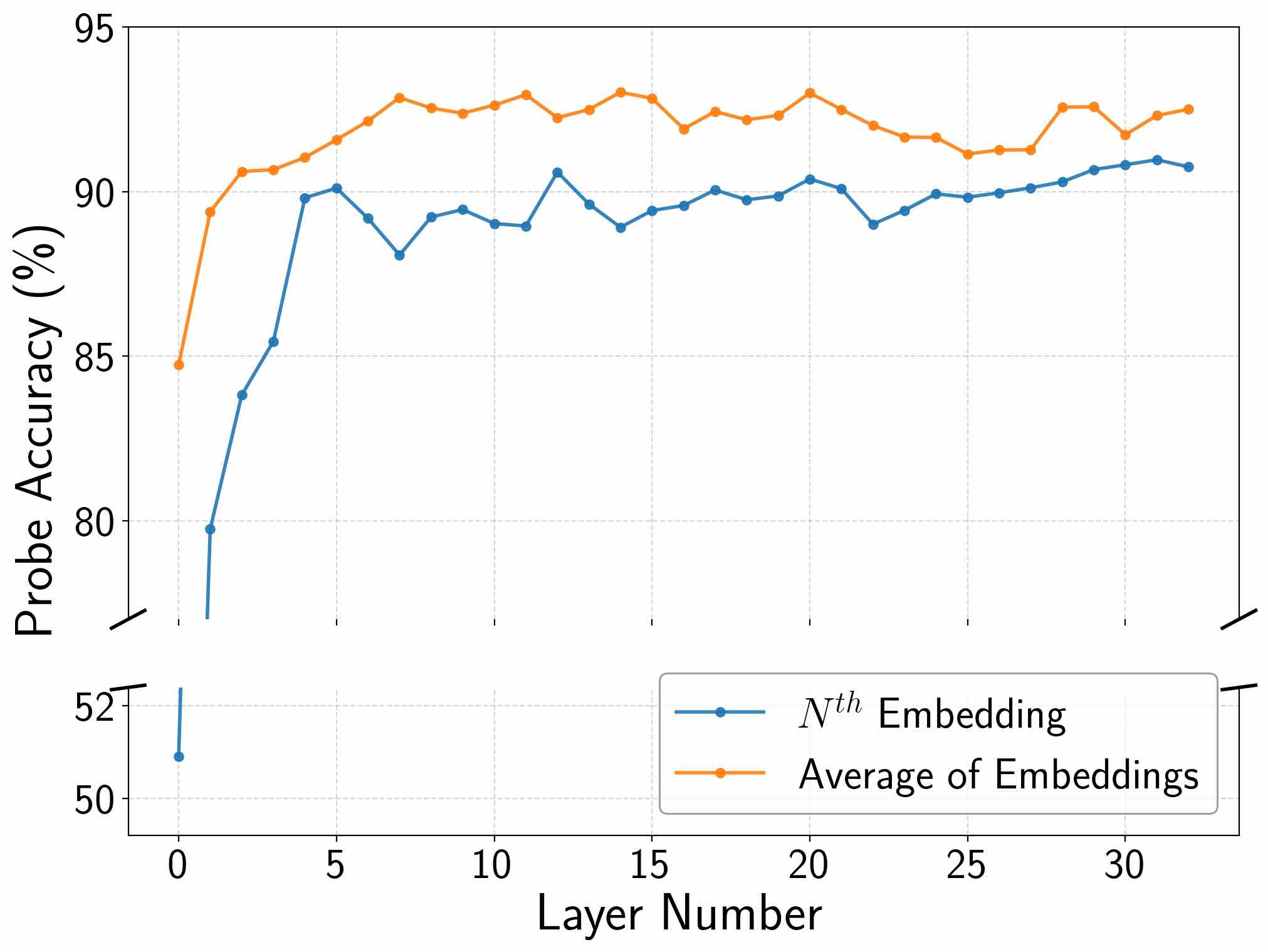}
    \caption{\textbf{Investigation} probe accuracy for \texttt{Llama-3-8B} using average and $N^{th}$ embeddings vs. layer}
    \label{fig:Investigation_llama_right_most_vs_average}
\end{figure}

\begin{figure}[H]
    \centering
    \includegraphics[width=\linewidth]{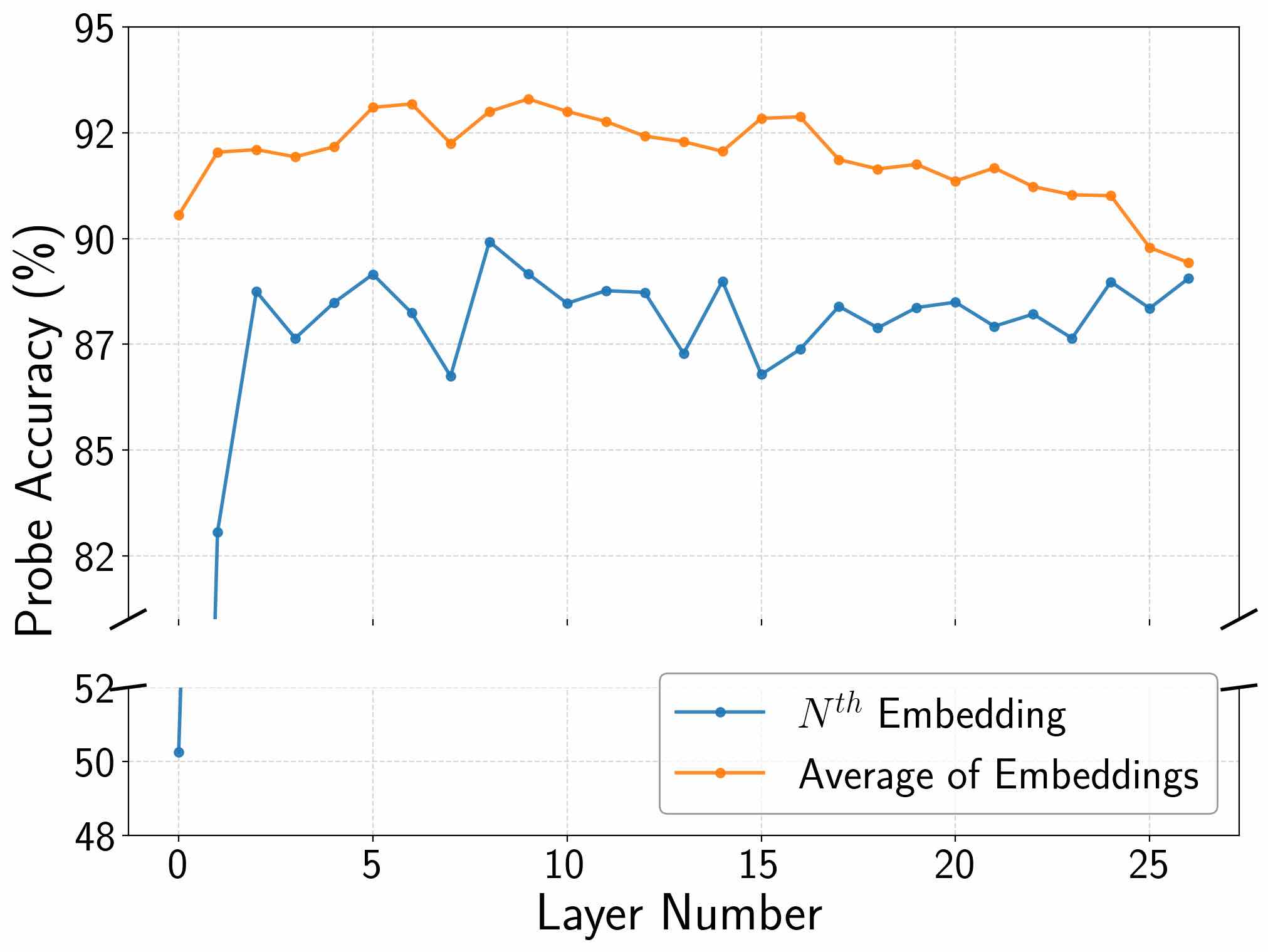}
    \caption{\textbf{Investigation} probe accuracy for \texttt{Gemma-2-2B} using average and $N^{th}$ embeddings vs. layer}
    \label{fig:Investigation_gemma2b_right_most_vs_average}
\end{figure}

\begin{figure}[H]
    \centering
    \includegraphics[width=\linewidth]{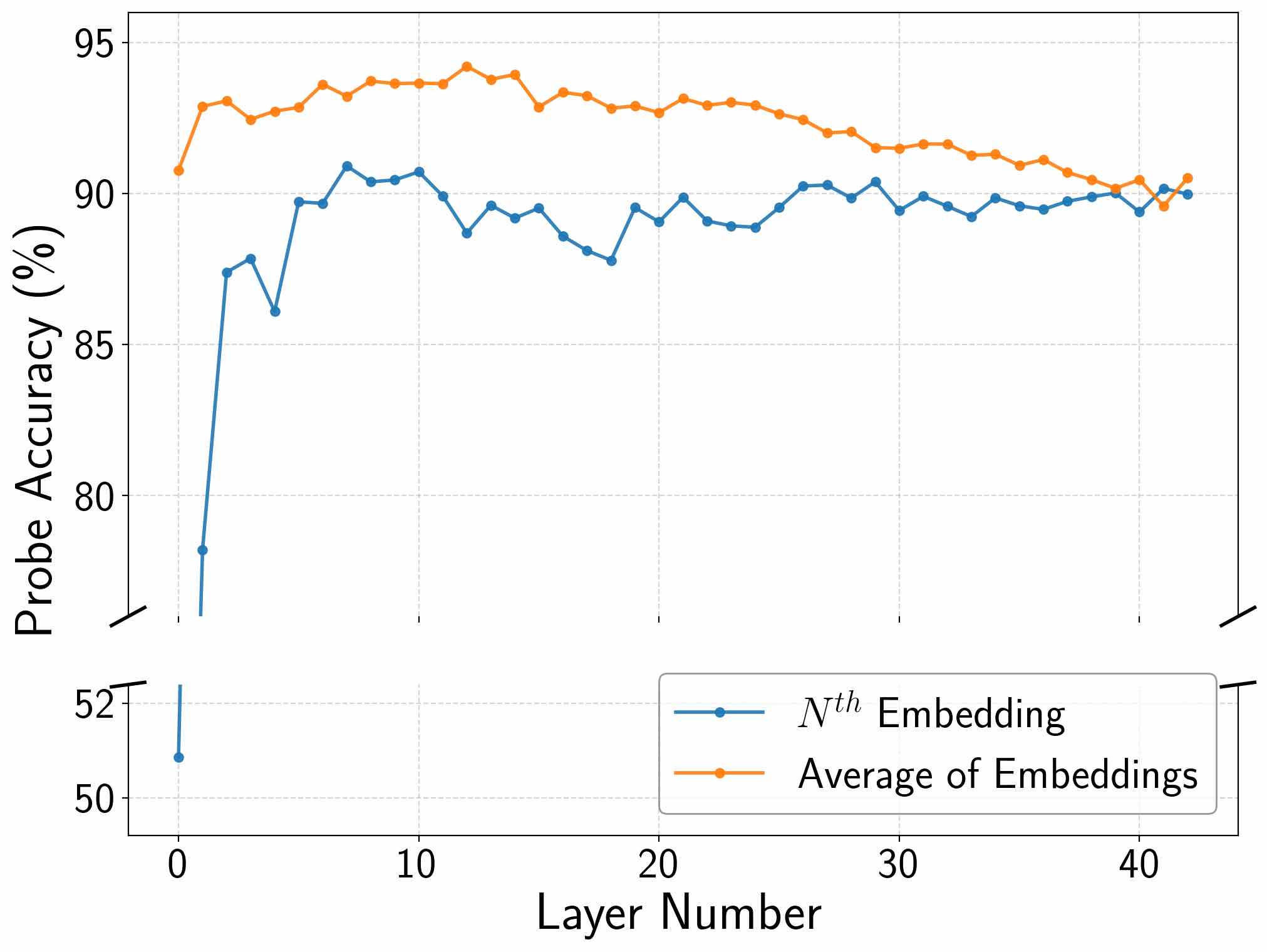}
    \caption{\textbf{Investigation} probe accuracy for \texttt{Gemma-2-9B} using average and $N^{th}$ embeddings vs. layer}
    \label{fig:Investigation_gemma9b_right_most_vs_average}
\end{figure}

\begin{figure}[H]
    \centering
    \includegraphics[width=\linewidth]{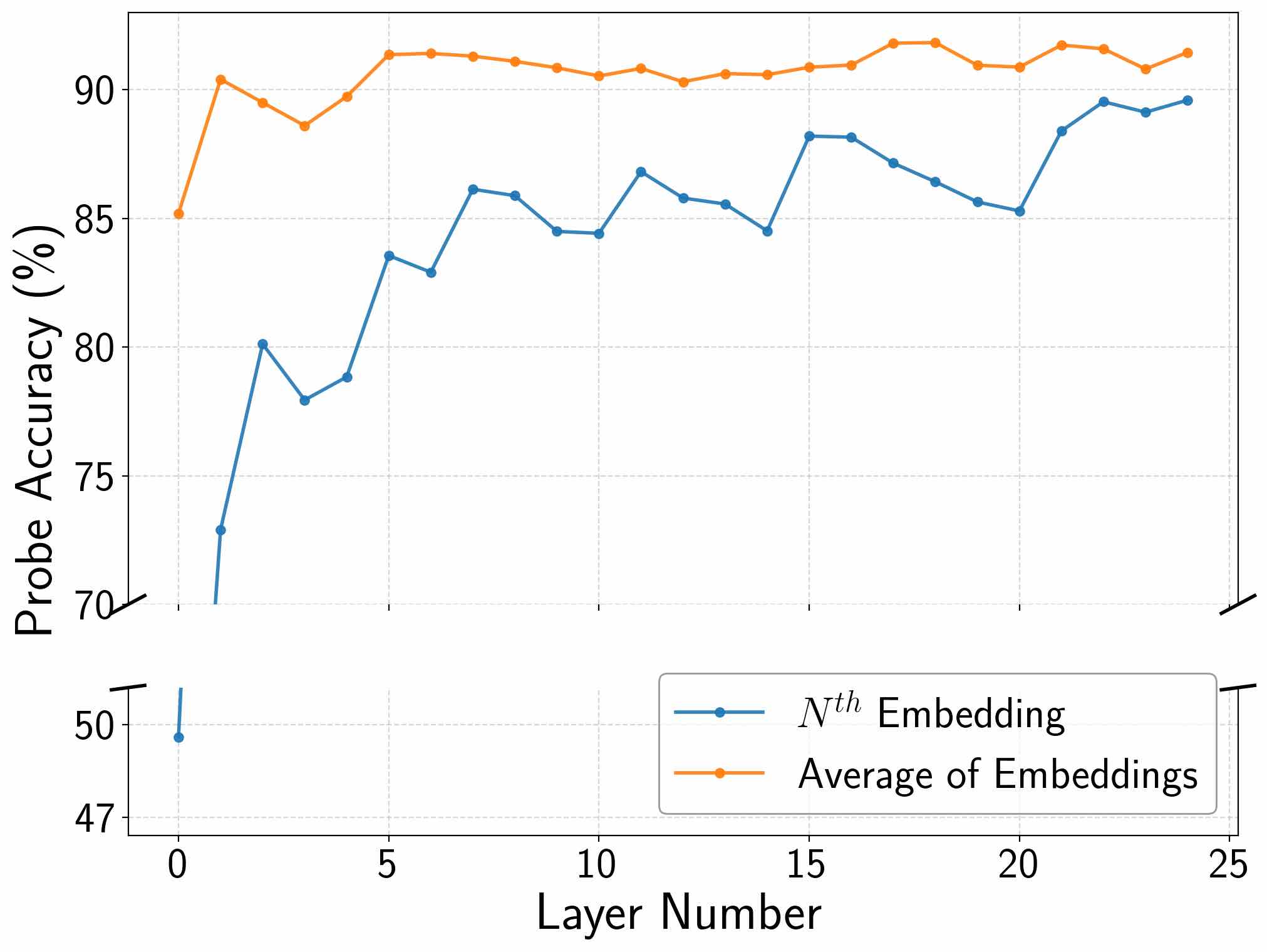}
    \caption{\textbf{Investigation} probe accuracy for \texttt{Qwen2.5-0.5B} using average and $N^{th}$ embeddings vs. layer}
    \label{fig:Investigation_qwen0p5b_right_most_vs_average}
\end{figure}

\begin{figure}[H]
    \centering
    \includegraphics[width=\linewidth]{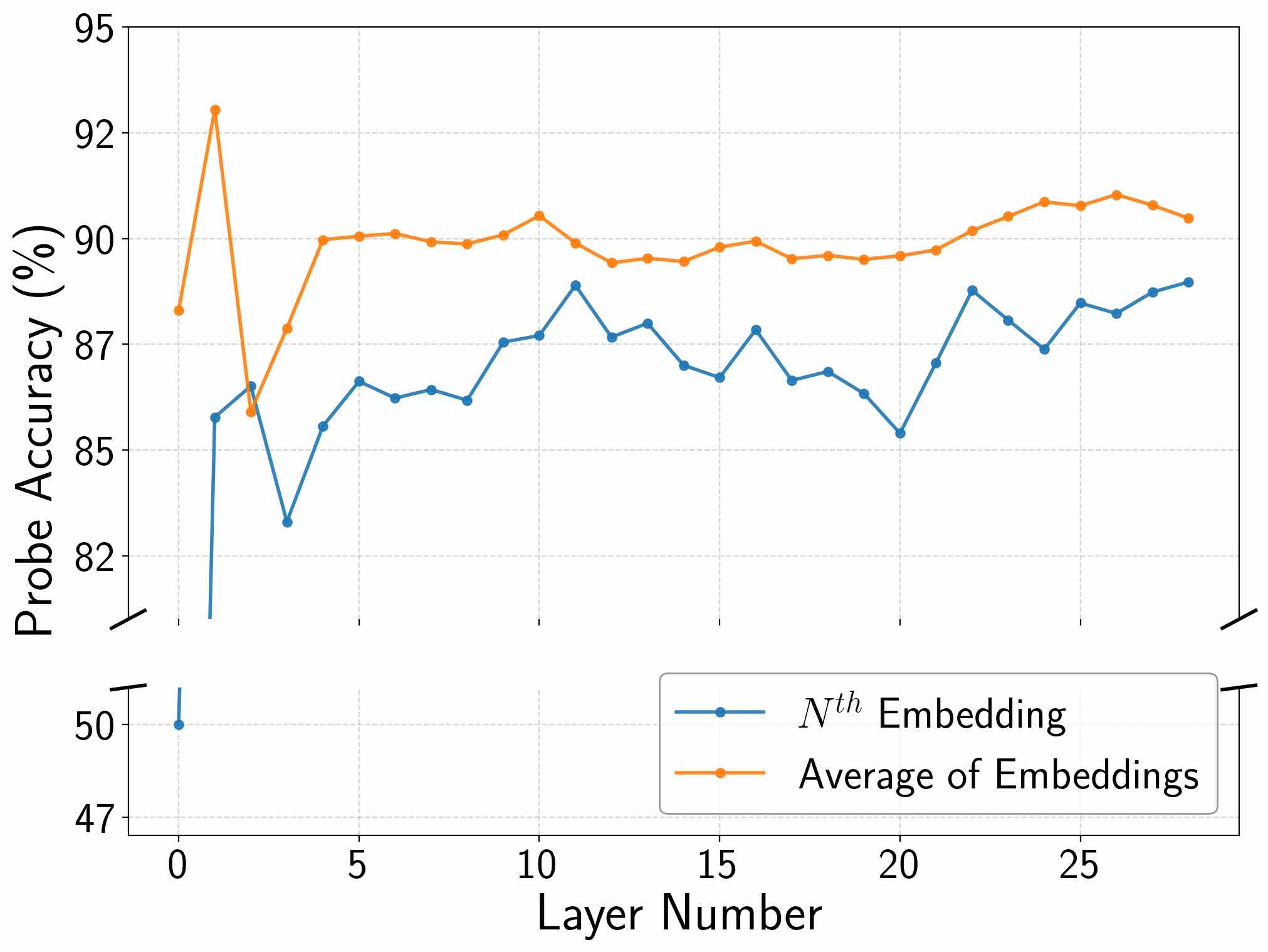}
    \caption{\textbf{Investigation} probe accuracy for \texttt{Qwen2.5-1.5B} using average and $N^{th}$ embeddings vs. layer}
    \label{fig:Investigation_qwen1p5b_right_most_vs_average}
\end{figure}

\begin{figure}[H]
    \centering
    \includegraphics[width=\linewidth]{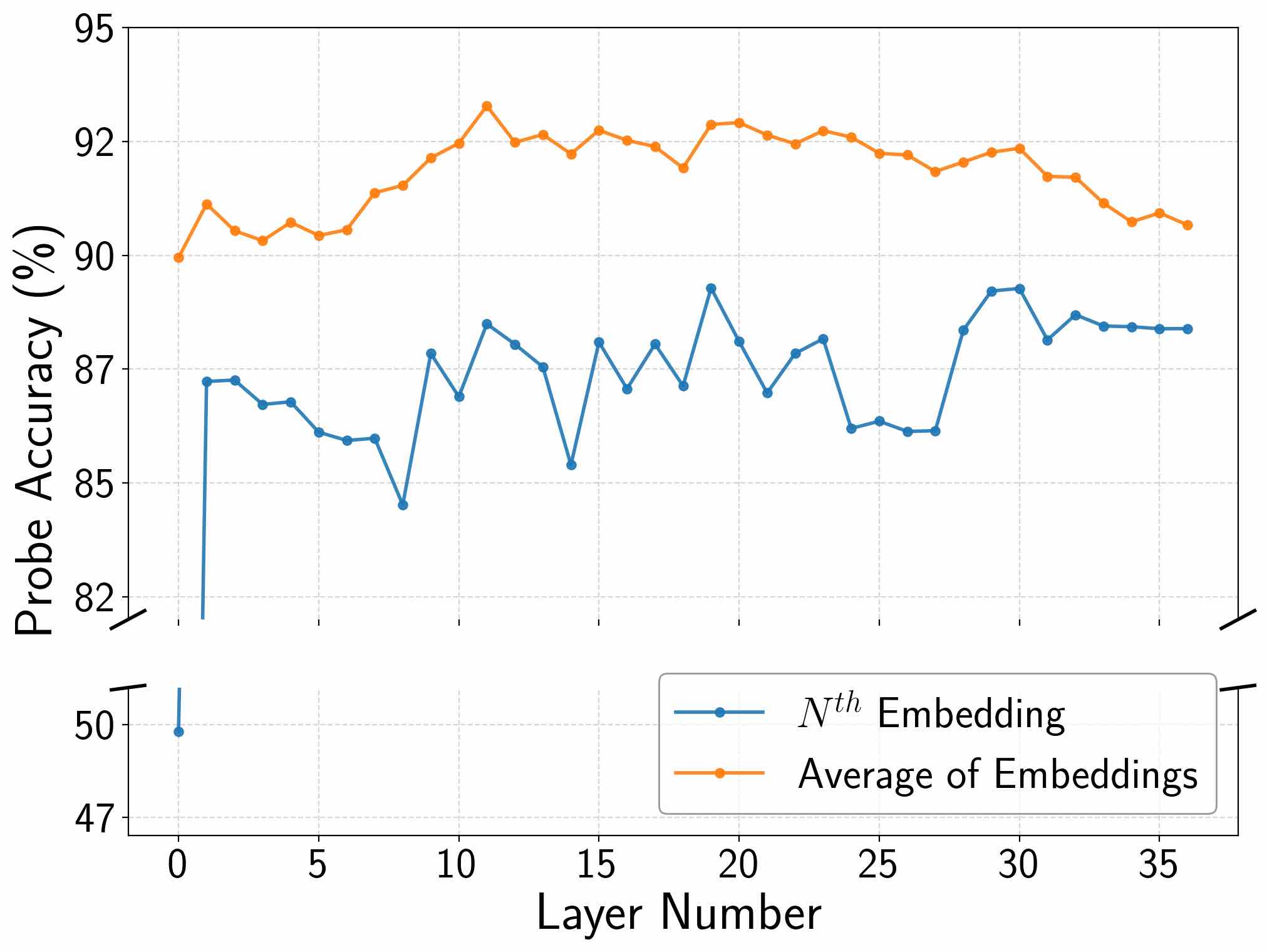}
    \caption{\textbf{Investigation} probe accuracy for \texttt{Qwen2.5-3B} using average and $N^{th}$ embeddings vs. layer}
    \label{fig:Investigation_qwen3b_right_most_vs_average}
\end{figure}

\begin{figure}[H]
    \centering
    \includegraphics[width=\linewidth]{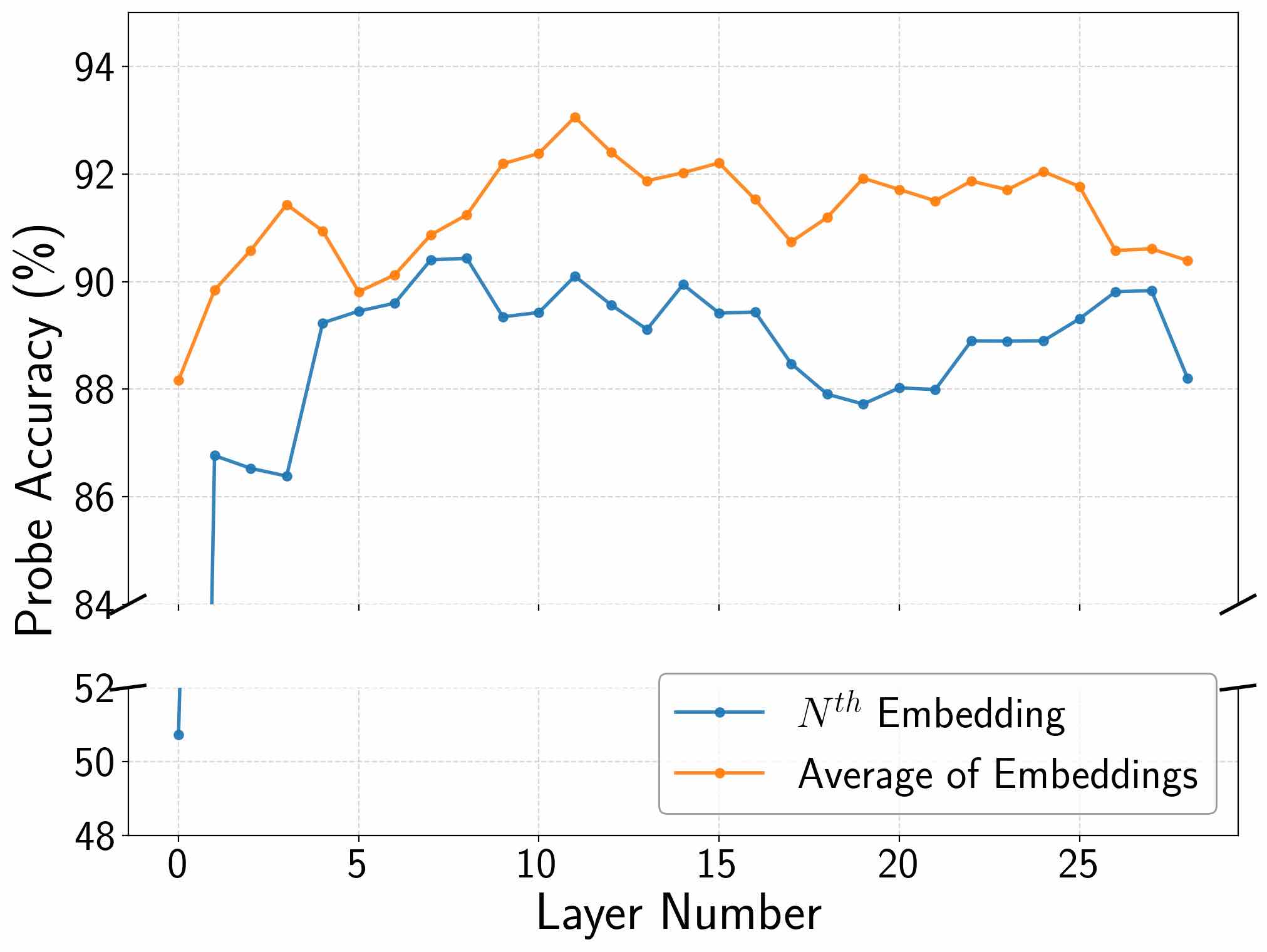}
    \caption{\textbf{Investigation} probe accuracy for \texttt{Qwen2.5-7B} using average and $N^{th}$ embeddings vs. layer}
    \label{fig:Investigation_qwen7b_right_most_vs_average}
\end{figure}

Figures \ref{fig:Investigation_llama_right_most_vs_average}--\ref{fig:Investigation_qwen7b_right_most_vs_average} illustrate the \textbf{Investigation} probe accuracies across layers of all LLMs using both the $N^{th}$ embedding and the average of all embeddings in the respective layer.

\subsubsection{Investigation Probe Cross-Check}
\begin{figure}[H]
    \centering
    \includegraphics[width=\linewidth]{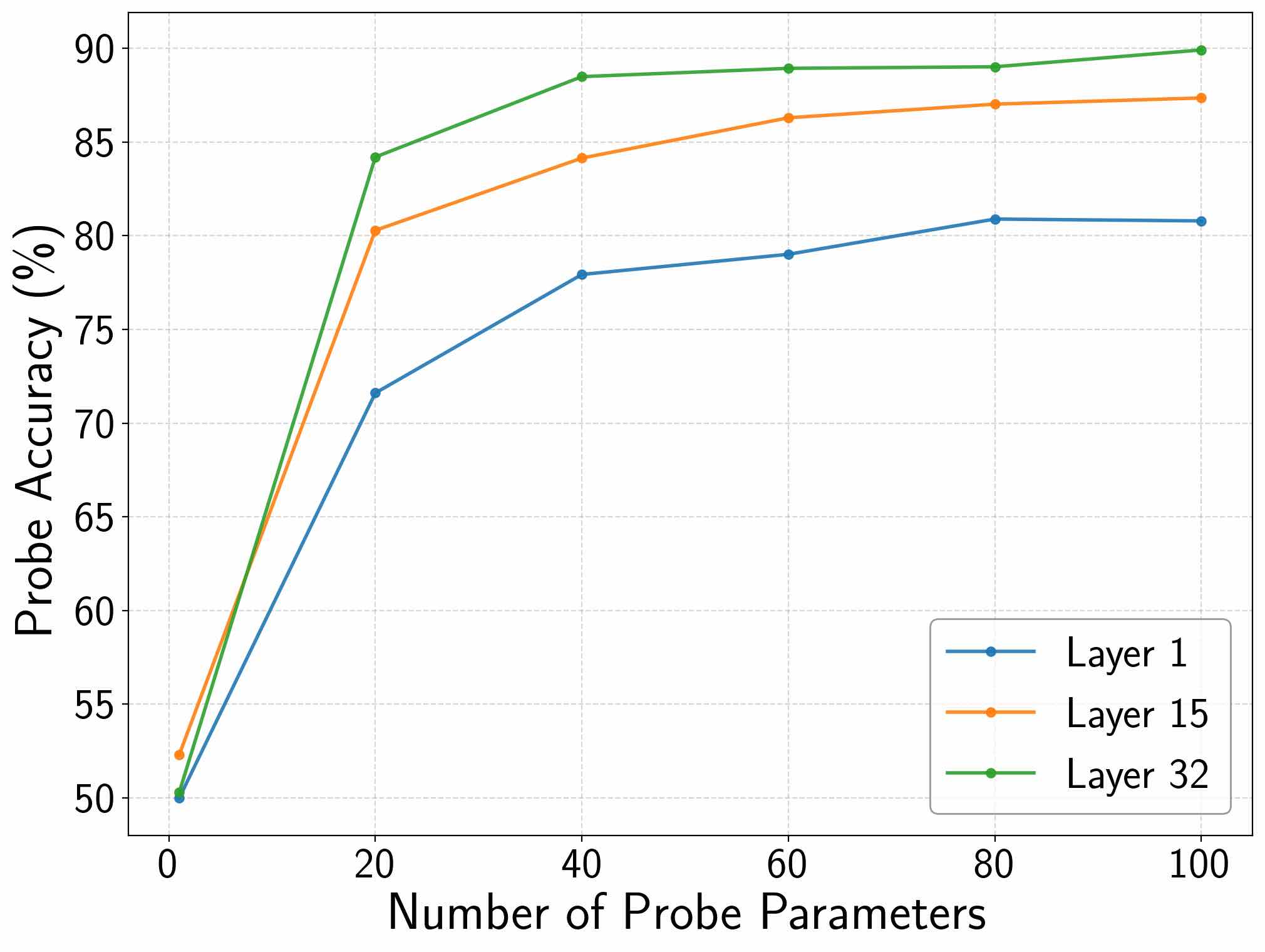}
    \caption{\textbf{Investigation} probe accuracy for \texttt{Llama-3-8B} as a function of probe size}
    \label{fig:Investigation_llama_vs_probe_params}
\end{figure}

\begin{figure}[H]
    \centering
    \includegraphics[width=\linewidth]{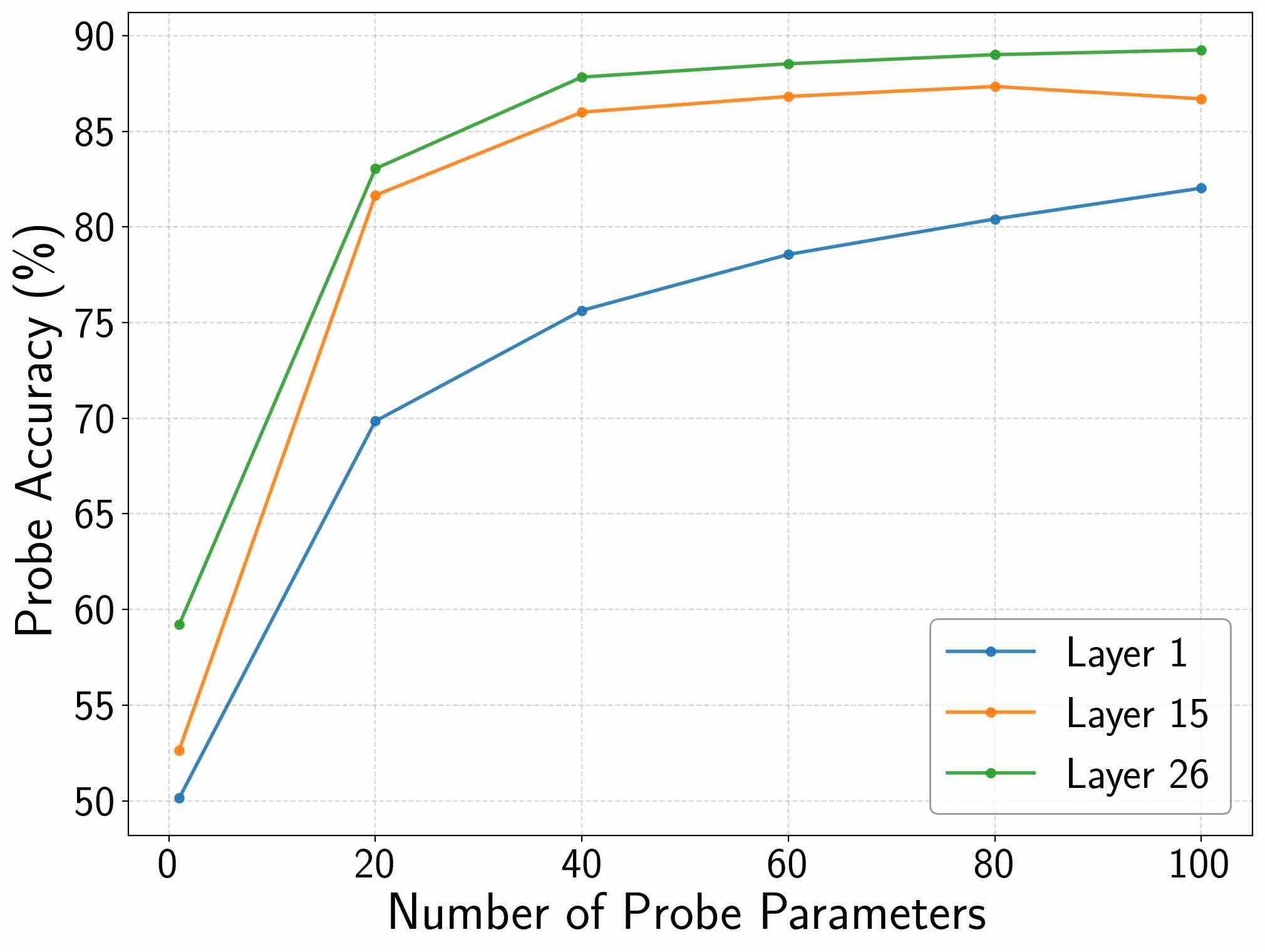}
    \caption{\textbf{Investigation} probe accuracy for \texttt{Gemma-2-2B} as a function of probe size}
    \label{fig:Investigation_gemma2b_vs_probe_params}
\end{figure}

\begin{figure}[H]
    \centering
    \includegraphics[width=\linewidth]{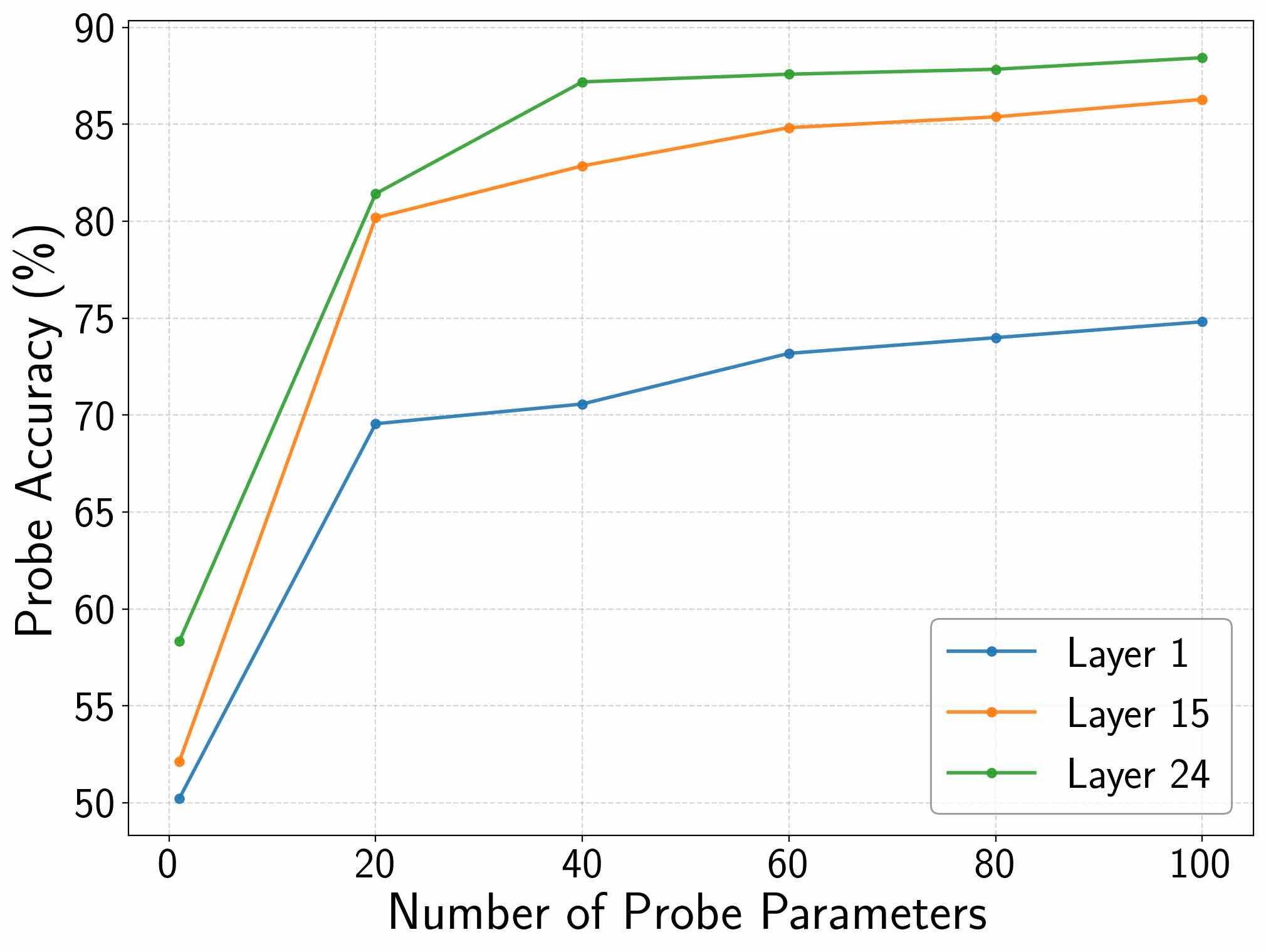}
    \caption{\textbf{Investigation} probe accuracy for \texttt{Qwen2.5-0.5B} as a function of probe size}
    \label{fig:Investigation_qwen0p5b_vs_probe_params}
\end{figure}

\begin{figure}[H]
    \centering
    \includegraphics[width=\linewidth]{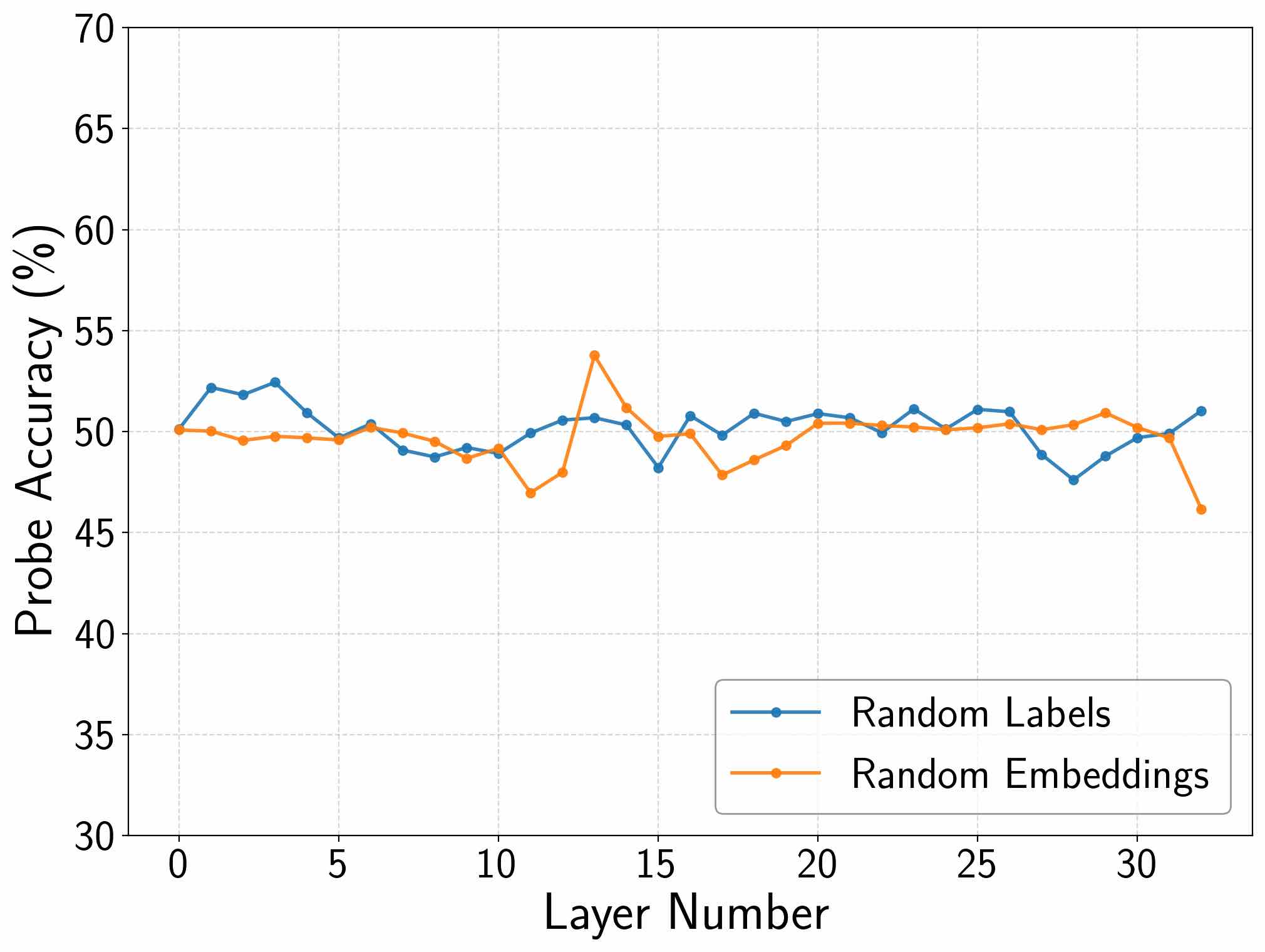}
    \caption{\textbf{Investigation} probe accuracy across layers in \texttt{Llama-3-8B} using random embeddings or random labels during probe training}
    \label{fig:Investigation_meta3b_control}
\end{figure}

\begin{figure}[H]
    \centering
    \includegraphics[width=\linewidth]{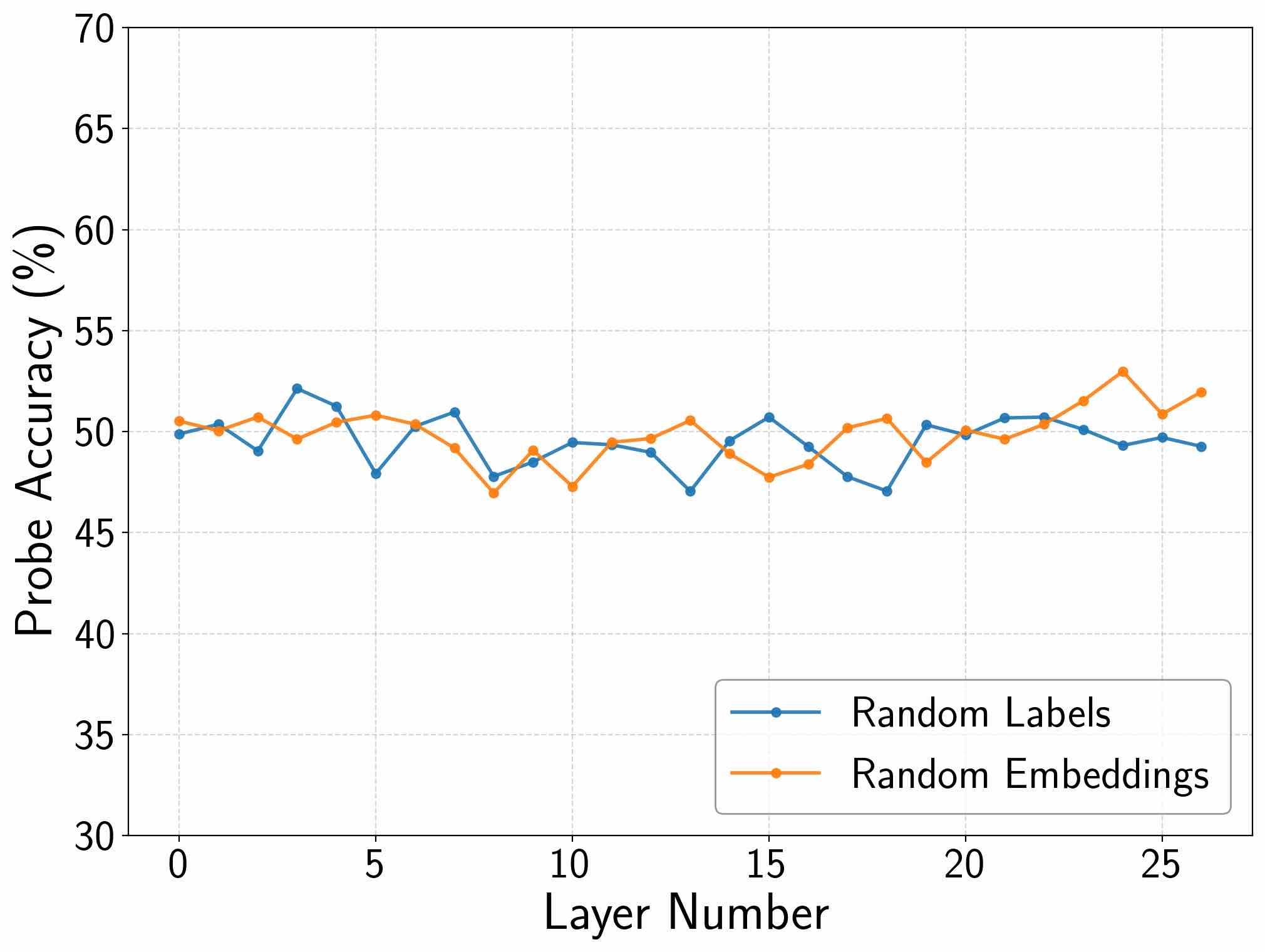}
    \caption{\textbf{Investigation} probe accuracy across layers in \texttt{Gemma-2-2B} using random embeddings or random labels during probe training}
    \label{fig:Investigation_gemma2b_control}
\end{figure}

\begin{figure}[H]
    \centering
    \includegraphics[width=\linewidth]{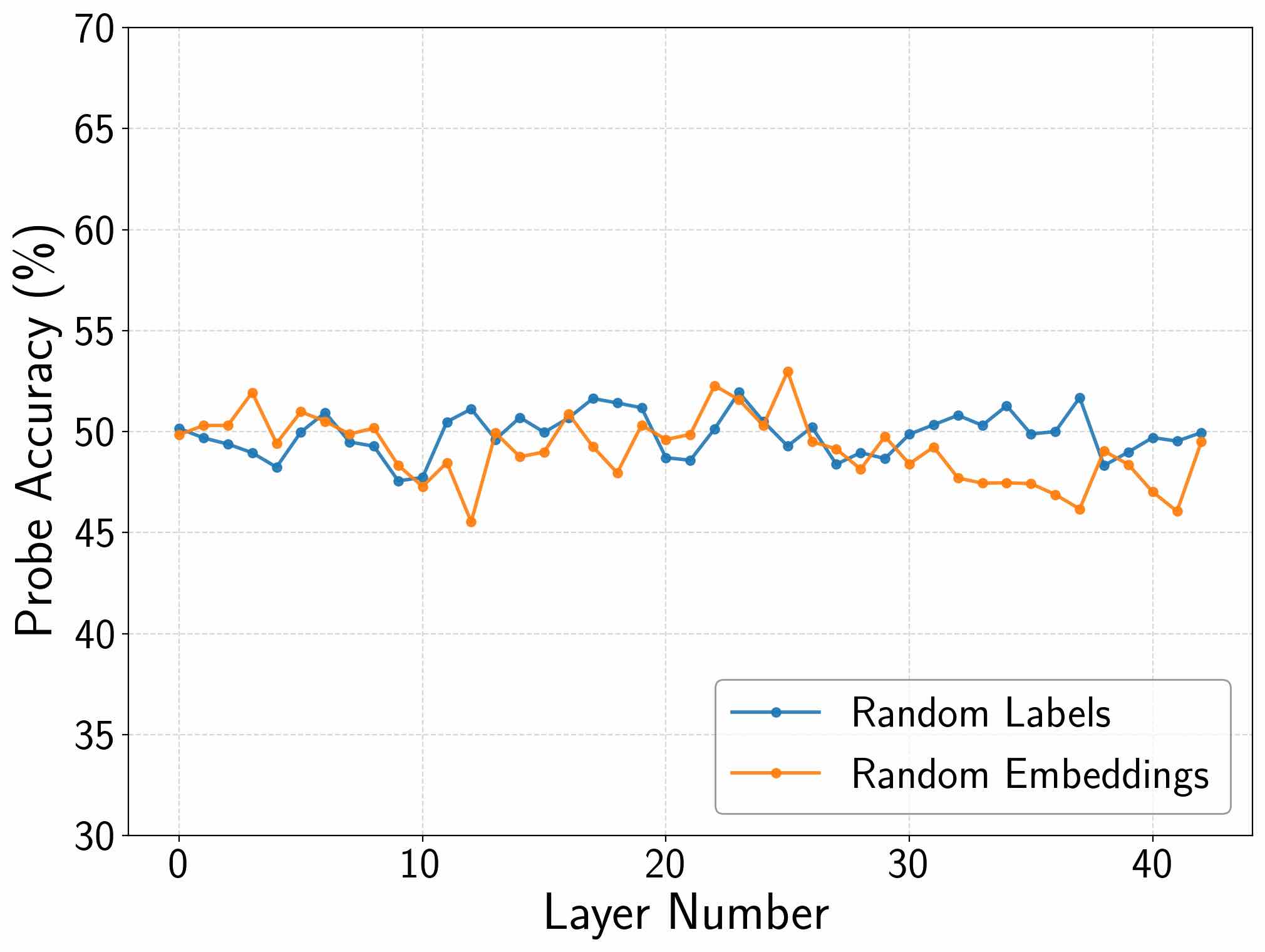}
    \caption{\textbf{Investigation} probe accuracy across layers in \texttt{Gemma-2-9B} using random embeddings or random labels during probe training}
    \label{fig:Investigation_gemma9b_control}
\end{figure}

\begin{figure}[H]
    \centering
    \includegraphics[width=\linewidth]{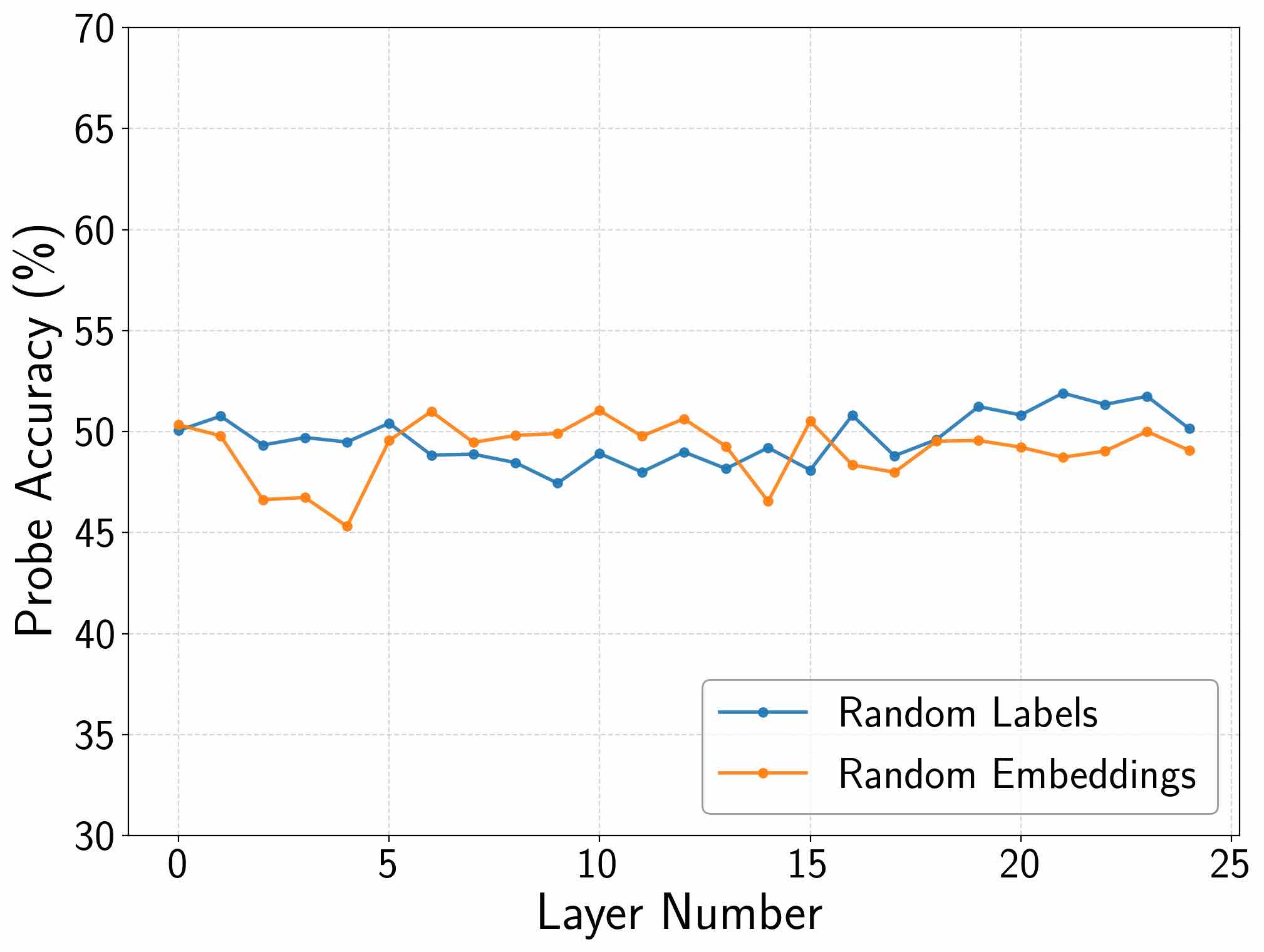}
    \caption{\textbf{Investigation} probe accuracy across layers in \texttt{Qwen2.5-0.5B} using random embeddings or random labels during probe training}
    \label{fig:Investigation_qwen0p5b_control}
\end{figure}

\begin{figure}[H]
    \centering
    \includegraphics[width=\linewidth]{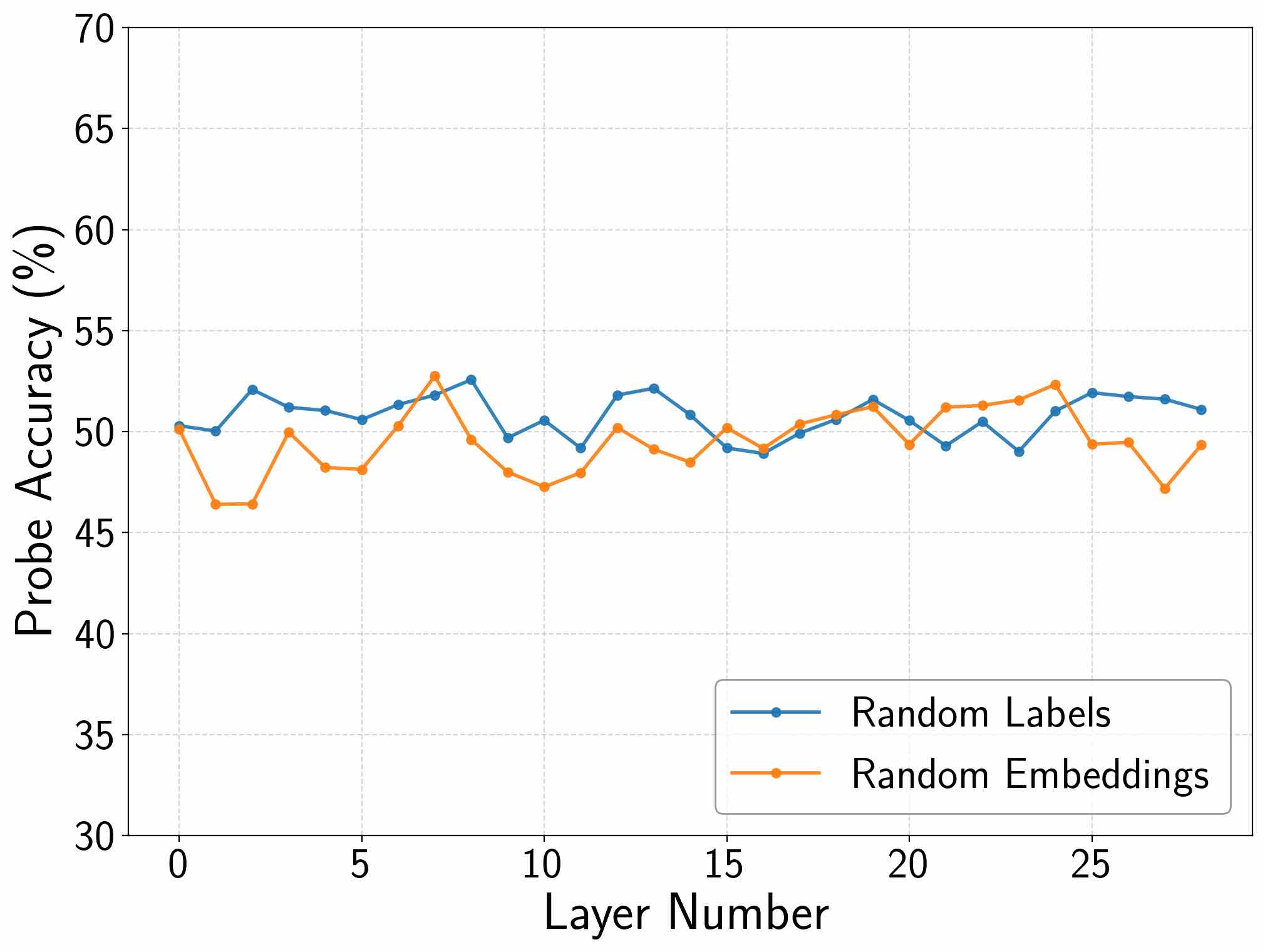}
    \caption{\textbf{Investigation} probe accuracy across layers in \texttt{Qwen2.5-1.5B} using random embeddings or random labels during probe training}
    \label{fig:Investigation_qwen1p5b_control}
\end{figure}

\begin{figure}[H]
    \centering
    \includegraphics[width=\linewidth]{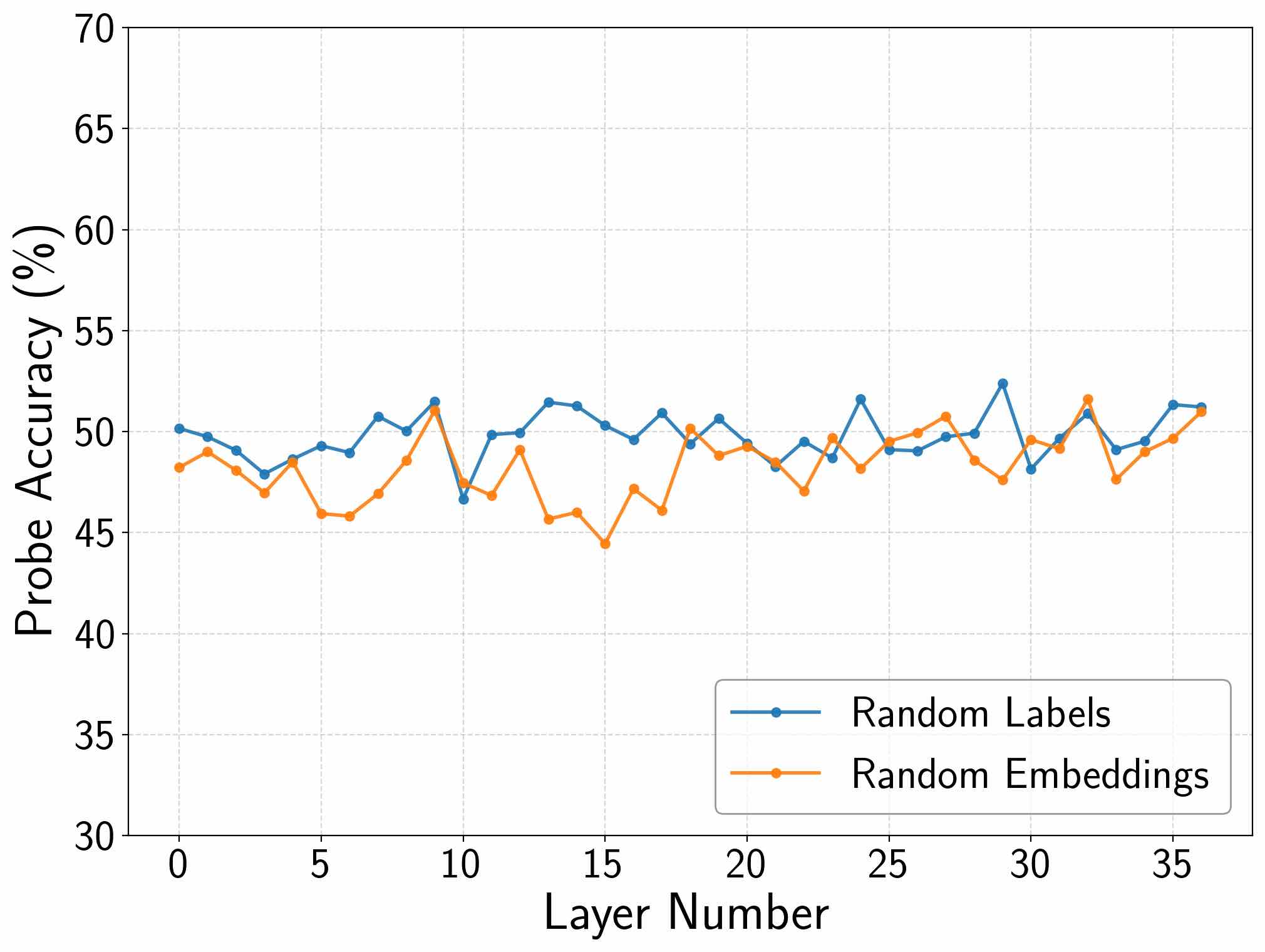}
    \caption{\textbf{Investigation} probe accuracy across layers in \texttt{Qwen2.5-3B} using random embeddings or random labels during probe training}
    \label{fig:Investigation_qwen3b_control}
\end{figure}

\begin{figure}[H]
    \centering
    \includegraphics[width=\linewidth]{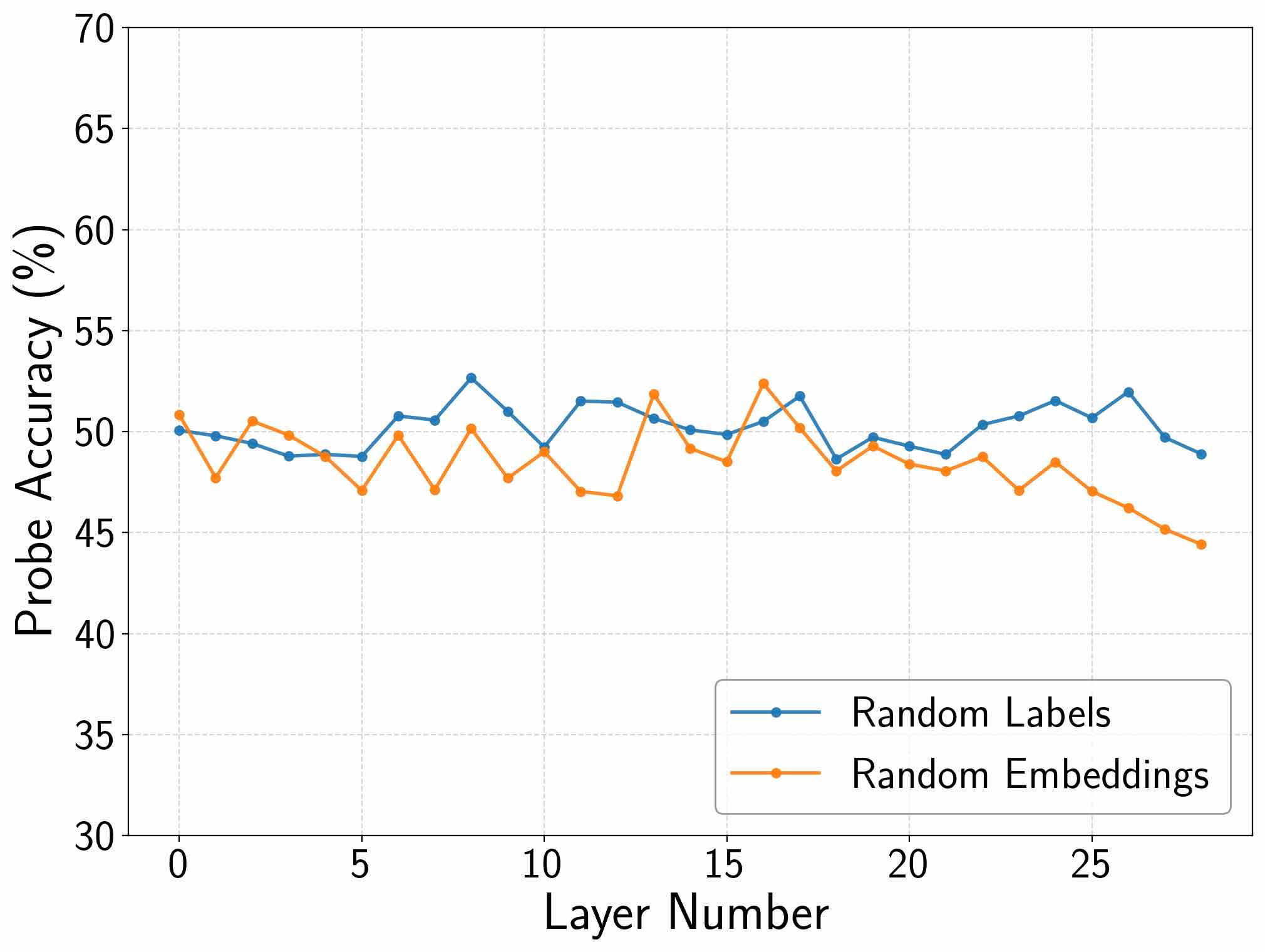}
    \caption{\textbf{Investigation} probe accuracy across layers in \texttt{Qwen2.5-7B} using random embeddings or random labels during probe training}
    \label{fig:Investigation_qwen7b_control}
\end{figure}

Figures \ref{fig:Investigation_llama_vs_probe_params}, \ref{fig:Investigation_gemma2b_vs_probe_params}, and \ref{fig:Investigation_qwen0p5b_vs_probe_params} show the \textbf{Investigation} probe accuracy versus probe size for \texttt{Llama-3-8B}, \texttt{Gemma-2-2B}, and \texttt{Qwen2.5-0.5B}, respectively, and Table \ref{table:investigation_vs_params} shows a summary of these results. Figures \ref{fig:Investigation_meta3b_control}--\ref{fig:Investigation_qwen7b_control} show the probe accuracies across layers for all LLMs when the probes are trained on the control task (randomizing embeddings or labels).

\begin{table}[H]
\centering
\begin{threeparttable}
\begin{tabular}{cc|cccc}
\hline\hline
\multirow{2}{*}{Probed LLM} & \multirow{2}{1cm}{Probed Layer} & \multicolumn{4}{c}{\# Probe parameters} \\
    & & 20 & 40 & 80 & max \\
\hline\hline

\multirow{3}{*}{\texttt{Llama-3-8B}} & 1 & 71 &	73 &	78 &	83 \\
& 15 &	61 &	73 &	79 &	90 \\
& 32 &	77 &	83 &	87 &	92 \\

\hline
                            
\multirow{3}{*}{\texttt{Gemma-2-2B}} & 1 &	68 &	73 &	77 &	85 \\
& 15 &	67 &	76 &	82 &	88 \\
& 26 &	73 &	81 &	86 &	89 \\

\hline

\multirow{3}{*}{\texttt{Qwen2.5-0.5B}} & 1 &	68 &	72 &	75 &	74 \\
& 15 &	64 &	77 &	83 &	89 \\
& 24 &	73 &	80 &	83 &	90 \\

\hline\hline
\end{tabular}
\begin{tablenotes}
\footnotesize
\item[$\bullet$] All results are in percentage (\%).
\item[$\bullet$] ``max'' denotes 4,096 for \texttt{Llama-3-8B}, 2,304 for \texttt{Gemma-2-2B}, and 896 for \texttt{Qwen2.5-0.5B}.
\item[$\bullet$] standard deviation for each result $\leq$ 2\%.
\end{tablenotes}
\end{threeparttable}
\caption{\textbf{Investigation} probe accuracy across model families, sizes, layers, and probe sizes} 
\label{table:investigation_vs_params}
\end{table}

\subsection{Extended Results for Inference of Democracy}

\subsubsection{Probing for Democracy using Nth Embedding vs. Average Embedding}
\begin{figure}[H]
    \centering
    \includegraphics[width=\linewidth]{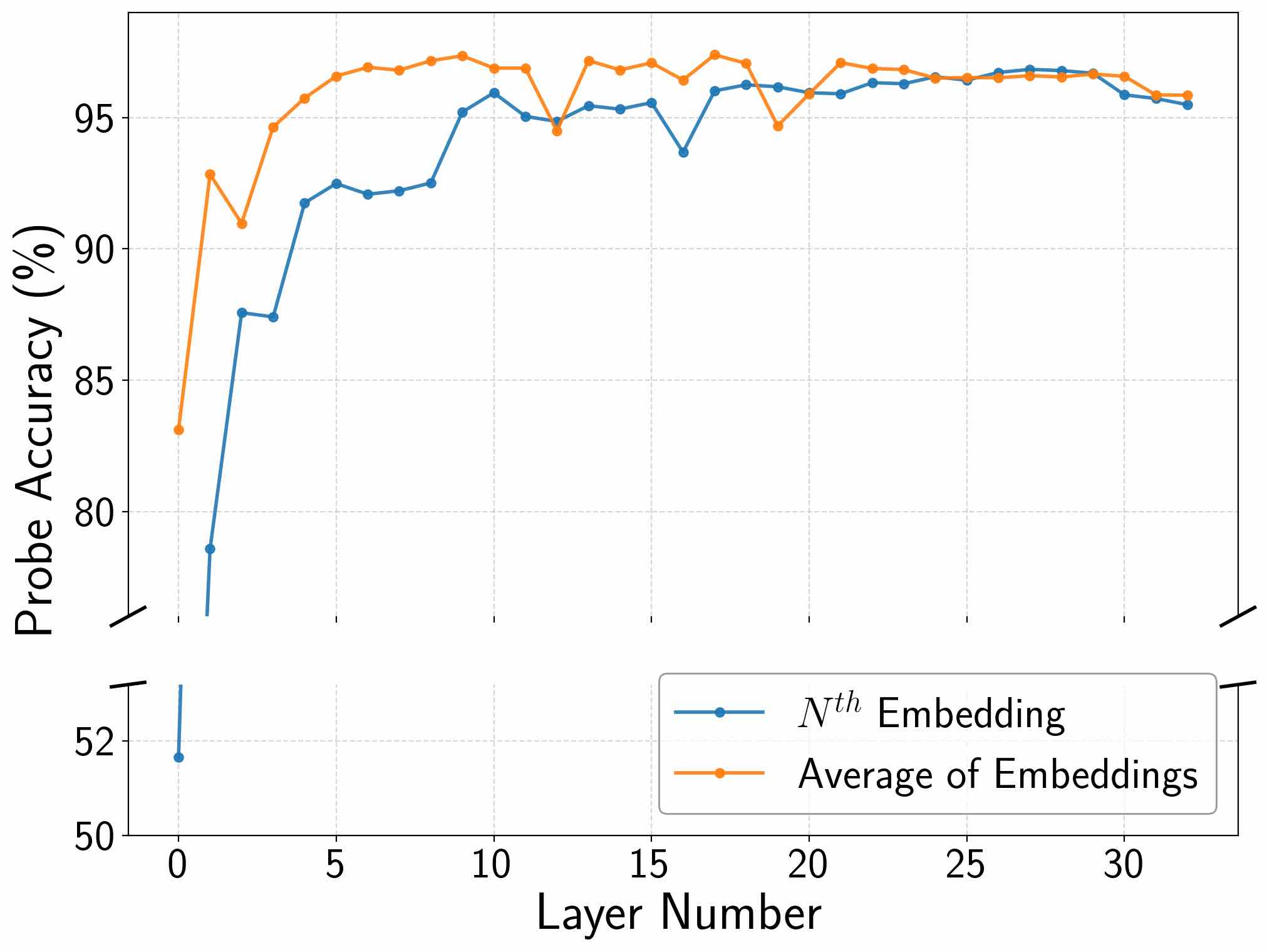}
    \caption{\textbf{Democracy} probe accuracy for \texttt{Llama-3-8B} using average and $N^{th}$ embeddings vs. layer}
    \label{fig:Democracy_llama_right_most_vs_average}
\end{figure}

\begin{figure}[H]
    \centering
    \includegraphics[width=\linewidth]{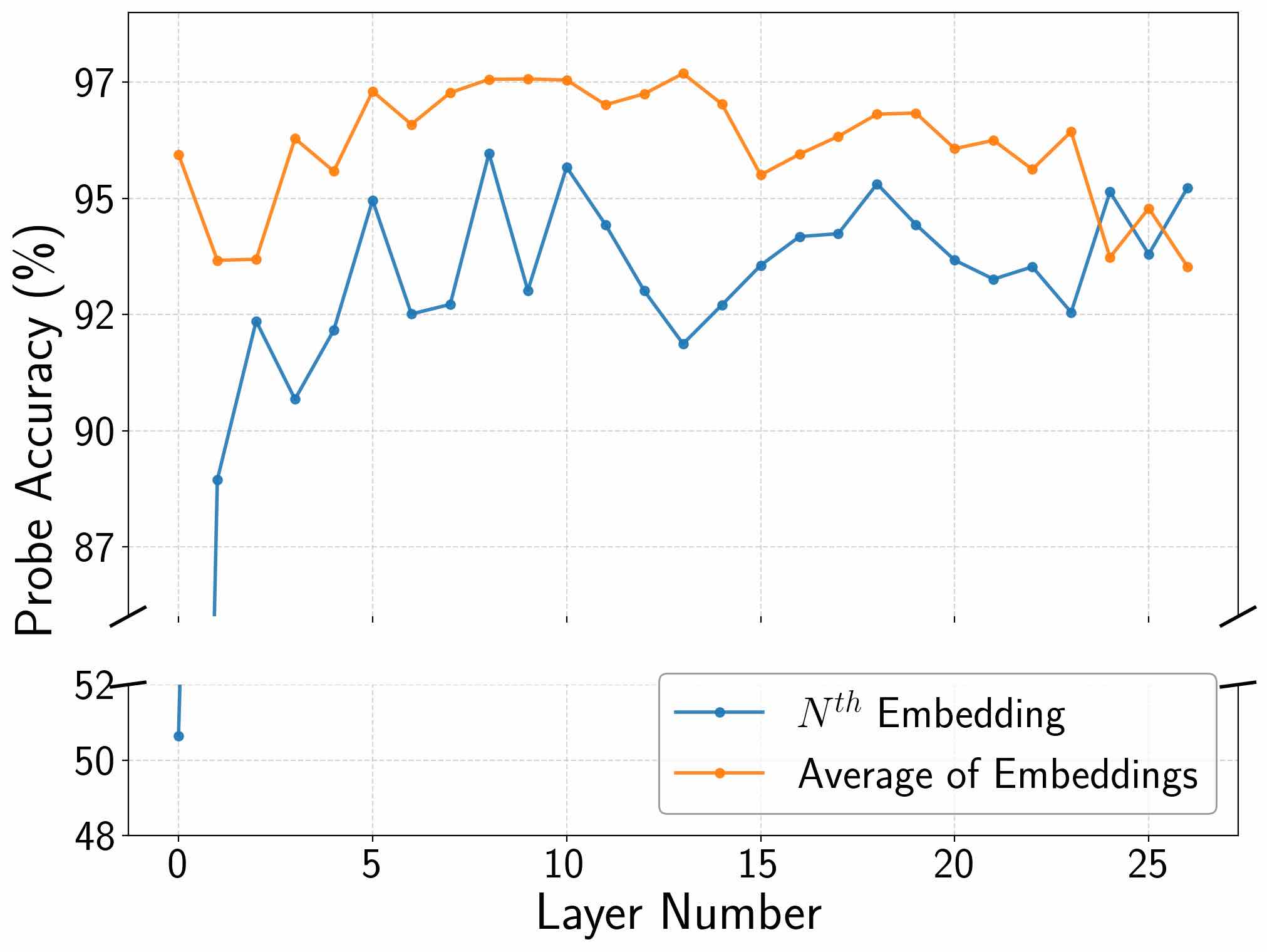}
    \caption{\textbf{Democracy} probe accuracy for \texttt{Gemma-2-2B} using average and $N^{th}$ embeddings vs. layer}
    \label{fig:Democracy_gemma2b_right_most_vs_average}
\end{figure}

\begin{figure}[H]
    \centering
    \includegraphics[width=\linewidth]{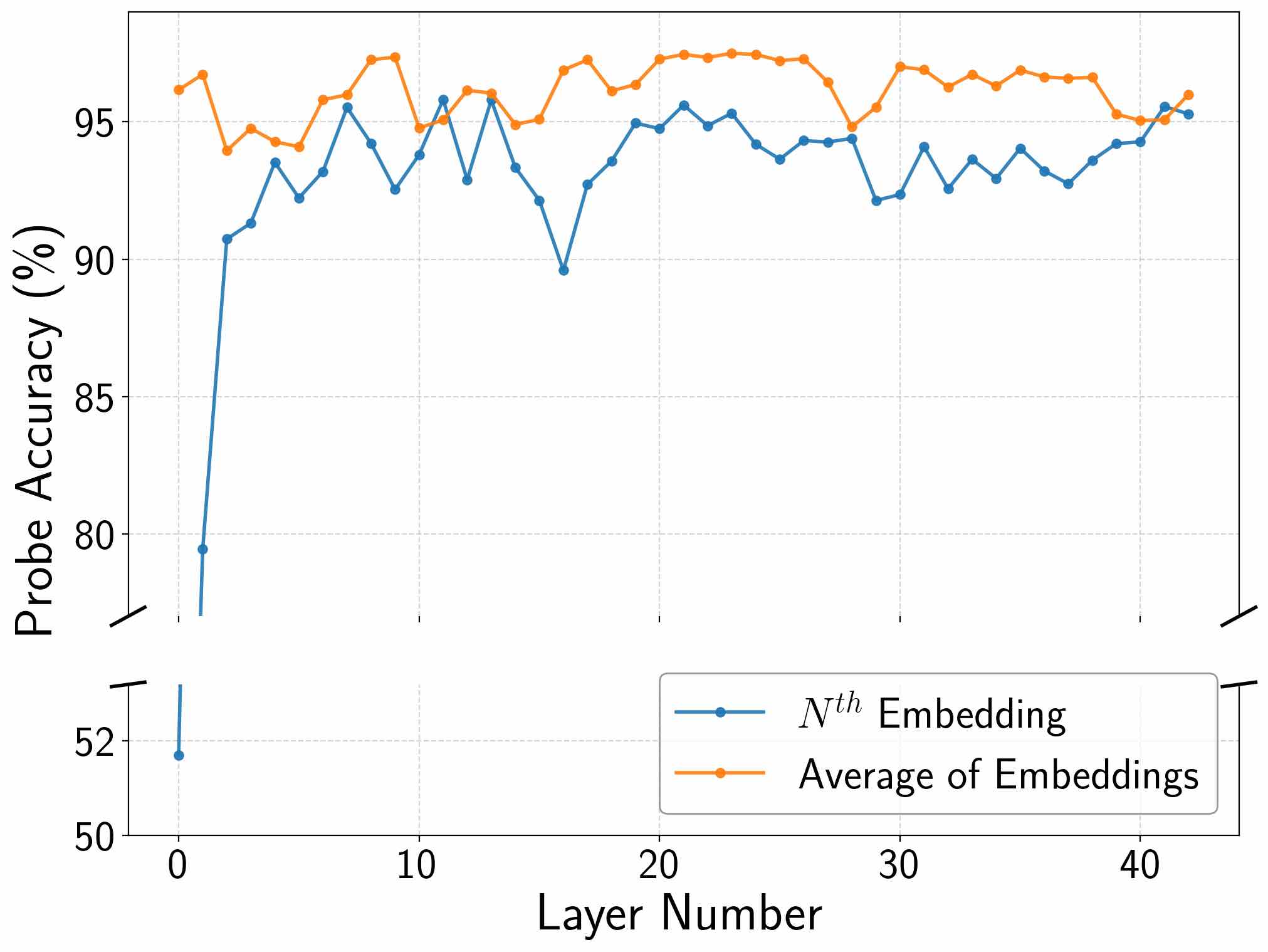}
    \caption{\textbf{Democracy} probe accuracy for \texttt{Gemma-2-9B} using average and $N^{th}$ embeddings vs. layer}
    \label{fig:Democracy_gemma9b_right_most_vs_average}
\end{figure}

\begin{figure}[H]
    \centering
    \includegraphics[width=\linewidth]{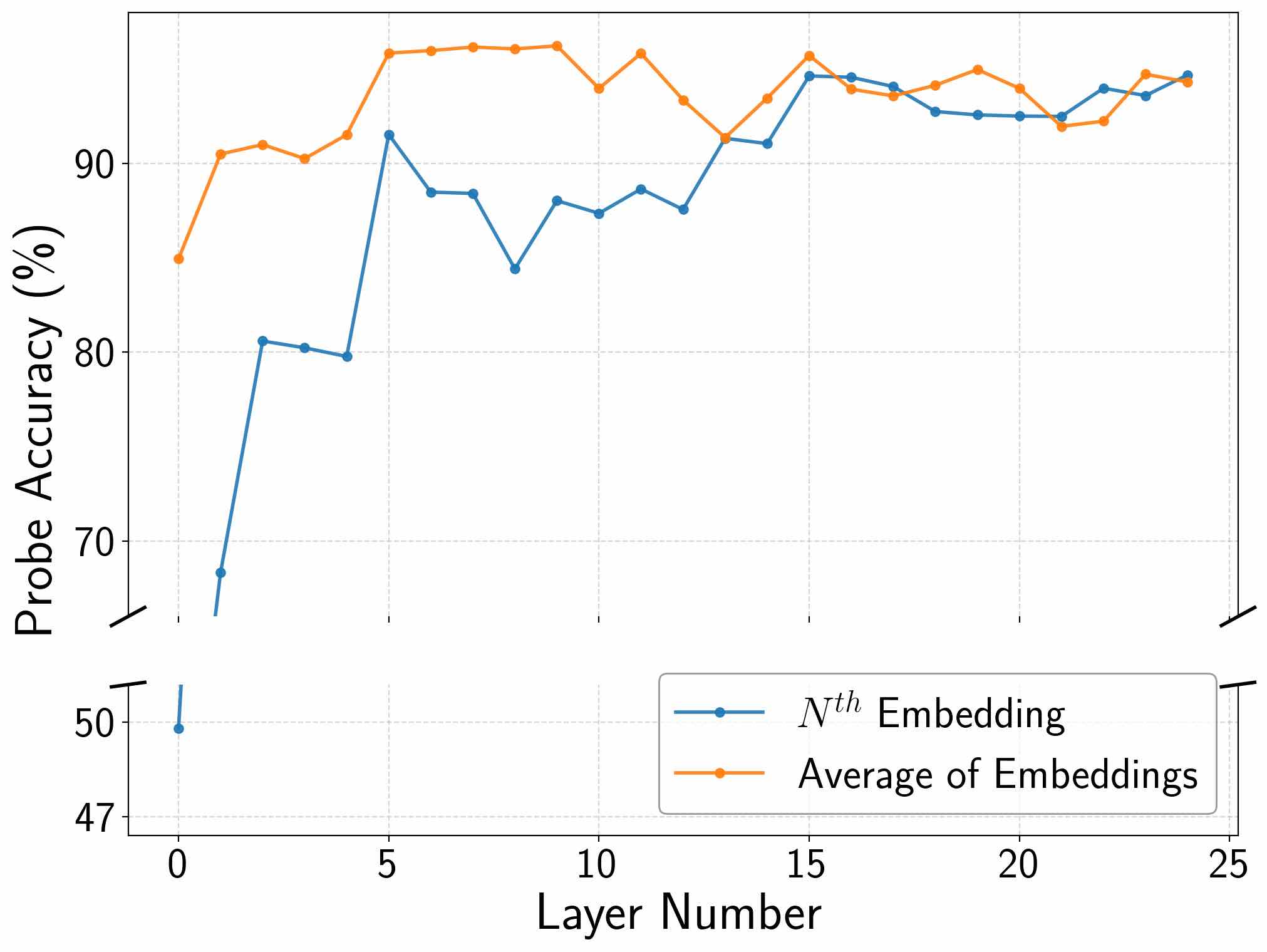}
    \caption{\textbf{Democracy} probe accuracy for \texttt{Qwen2.5-0.5B} using average and $N^{th}$ embeddings vs. layer}
    \label{fig:Democracy_qwen0p5b_right_most_vs_average}
\end{figure}

\begin{figure}[H]
    \centering
    \includegraphics[width=\linewidth]{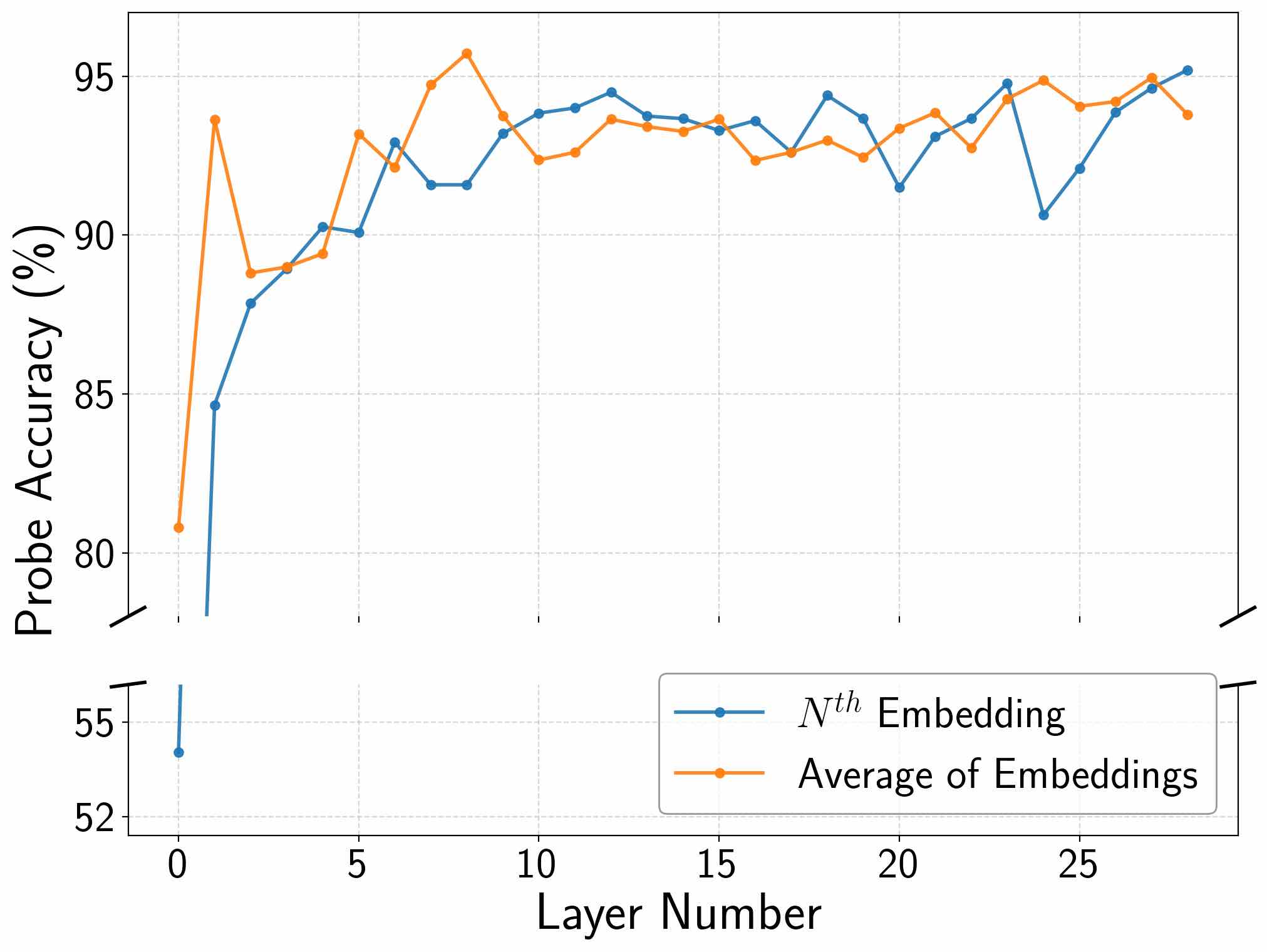}
    \caption{\textbf{Democracy} probe accuracy for \texttt{Qwen2.5-1.5B} using average and $N^{th}$ embeddings vs. layer}
    \label{fig:Democracy_qwen1p5b_right_most_vs_average}
\end{figure}

\begin{figure}[H]
    \centering
    \includegraphics[width=\linewidth]{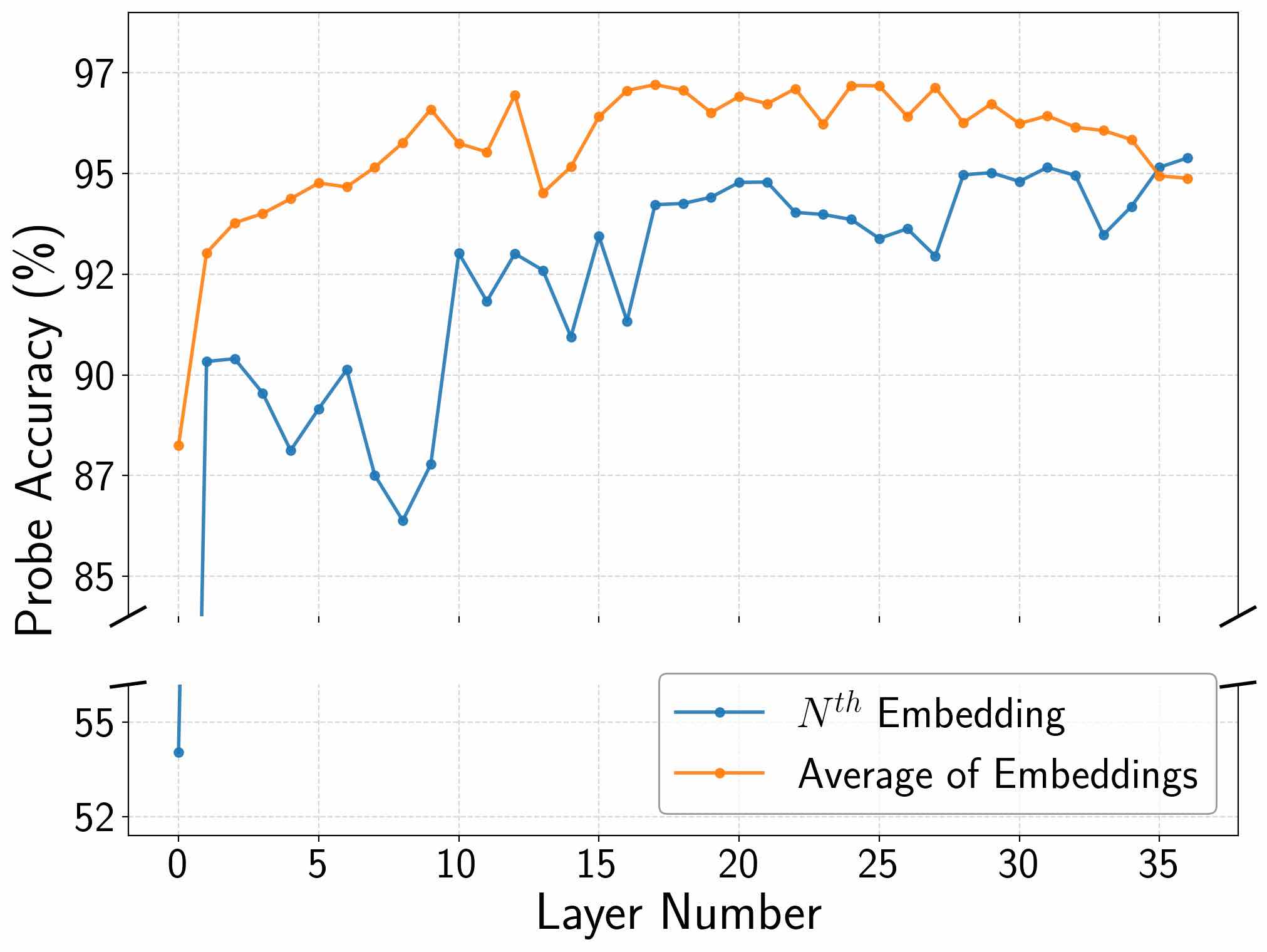}
    \caption{\textbf{Democracy} probe accuracy for \texttt{Qwen2.5-3B} using average and $N^{th}$ embeddings vs. layer}
    \label{fig:Democracy_qwen3b_right_most_vs_average}
\end{figure}

\begin{figure}[H]
    \centering
    \includegraphics[width=\linewidth]{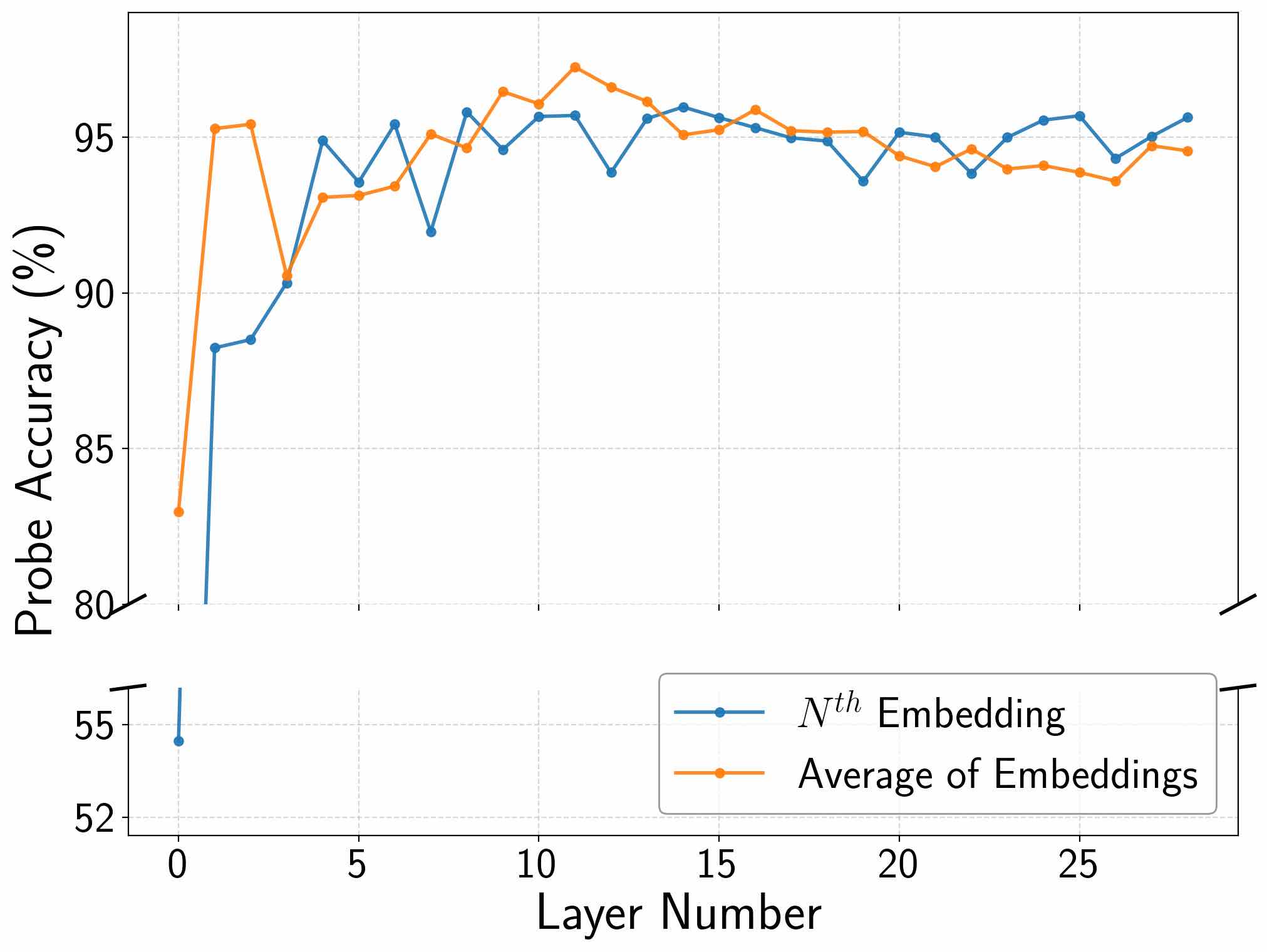}
    \caption{\textbf{Democracy} probe accuracy for \texttt{Qwen2.5-7B} using average and $N^{th}$ embeddings vs. layer}
    \label{fig:Democracy_qwen7b_right_most_vs_average}
\end{figure}

Figures \ref{fig:Democracy_llama_right_most_vs_average}--\ref{fig:Democracy_qwen7b_right_most_vs_average} illustrate the \textbf{Democracy} probe accuracies across layers of all LLMs using both the $N^{th}$ embedding and the average of all embeddings in the respective layer.

\subsubsection{Democracy Probe Cross-Check}
\begin{figure}[H]
    \centering
    \includegraphics[width=\linewidth]{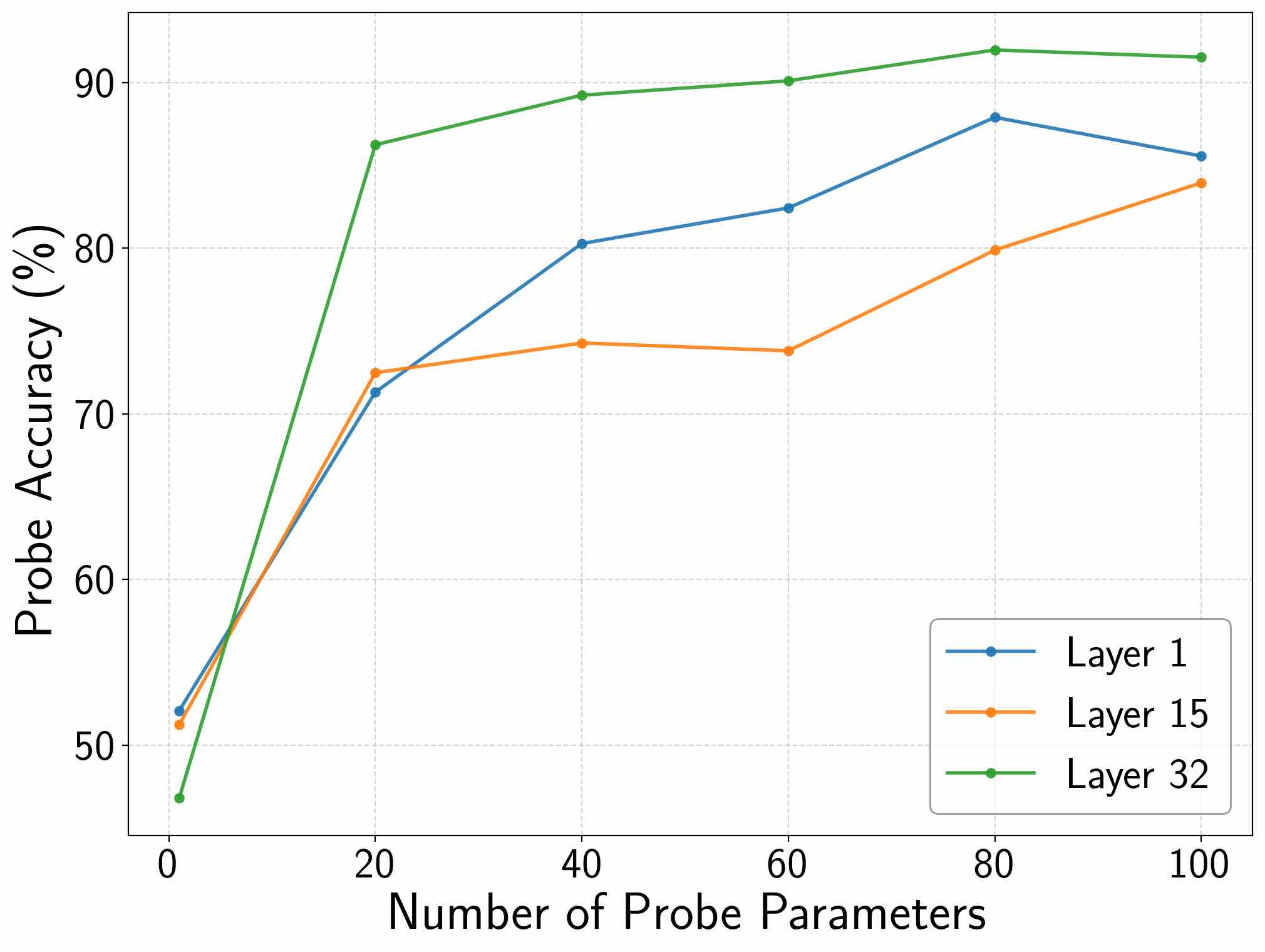}
    \caption{\textbf{Democracy} probe accuracy for \texttt{Llama-3-8B} as a function of probe size}
    \label{fig:Democracy_llama_vs_probe_params}
\end{figure}

\begin{figure}[H]
    \centering
    \includegraphics[width=\linewidth]{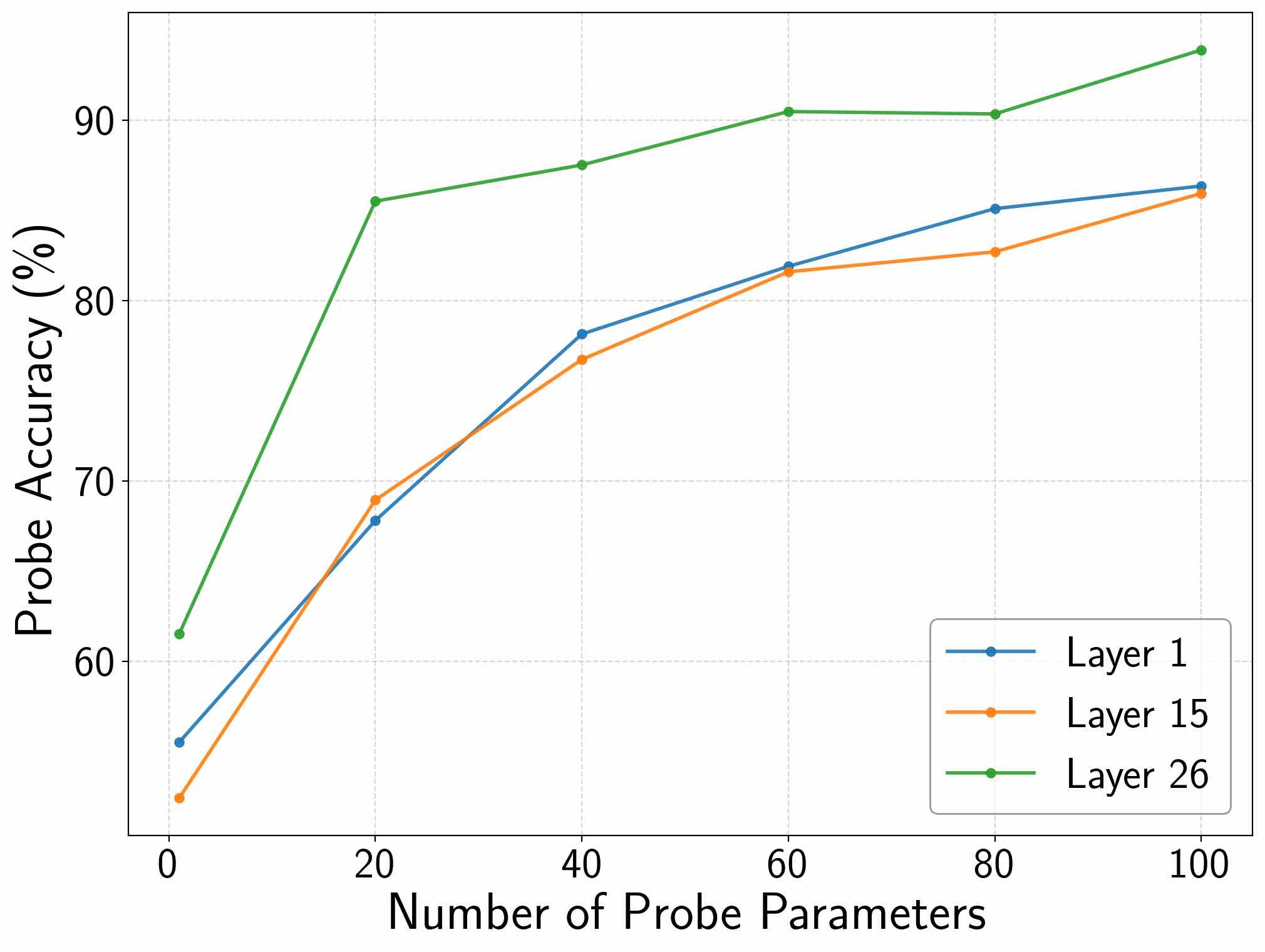}
    \caption{\textbf{Democracy} probe accuracy for \texttt{Gemma-2-2B} as a function of probe size}
    \label{fig:Democracy_gemma2b_vs_probe_params}
\end{figure}

\begin{figure}[H]
    \centering
    \includegraphics[width=\linewidth]{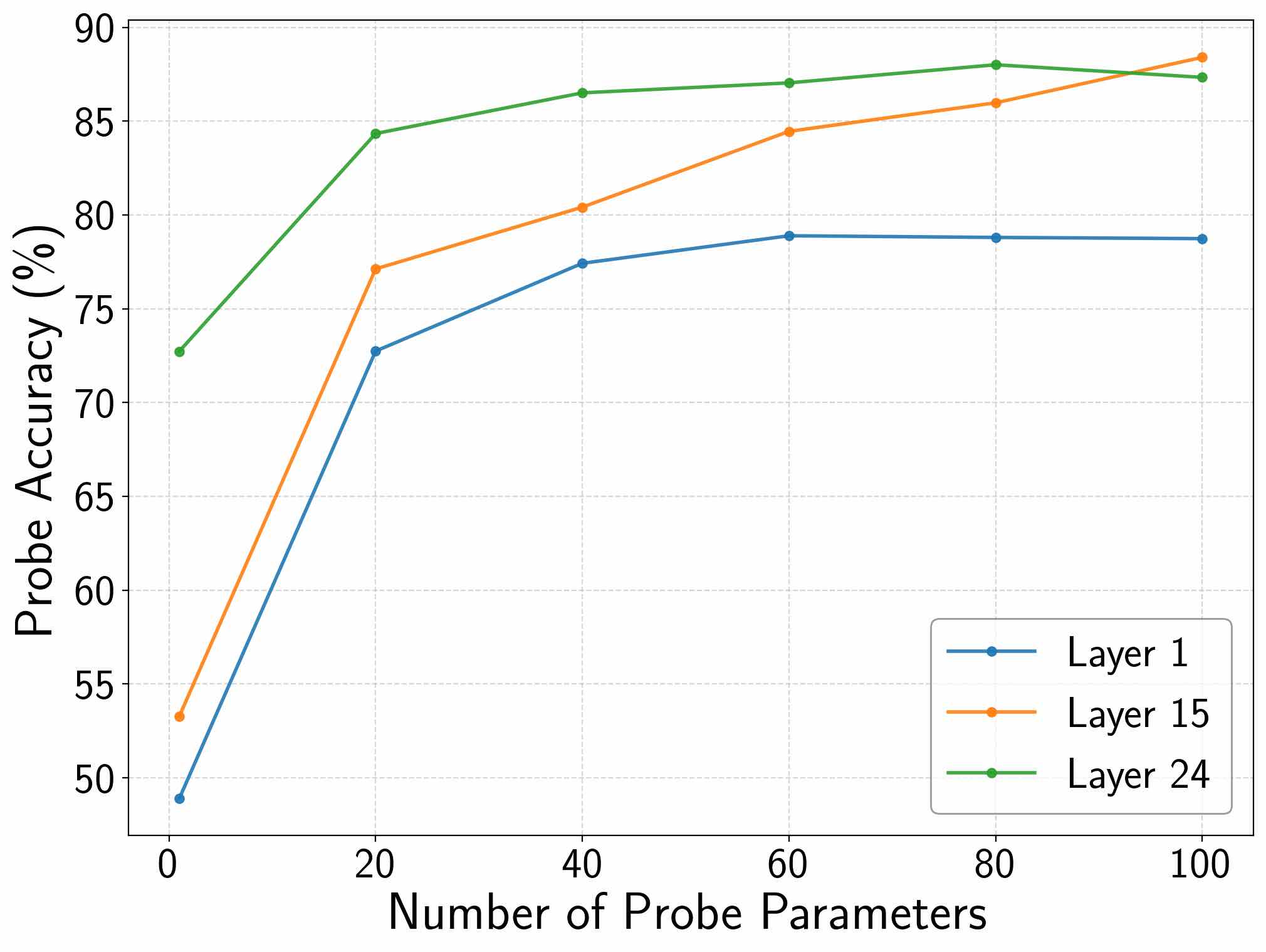}
    \caption{\textbf{Democracy} probe accuracy for \texttt{Qwen2.5-0.5B} as a function of probe size}
    \label{fig:Democracy_qwen0p5b_vs_probe_params}
\end{figure}

\begin{figure}[H]
    \centering
    \includegraphics[width=\linewidth]{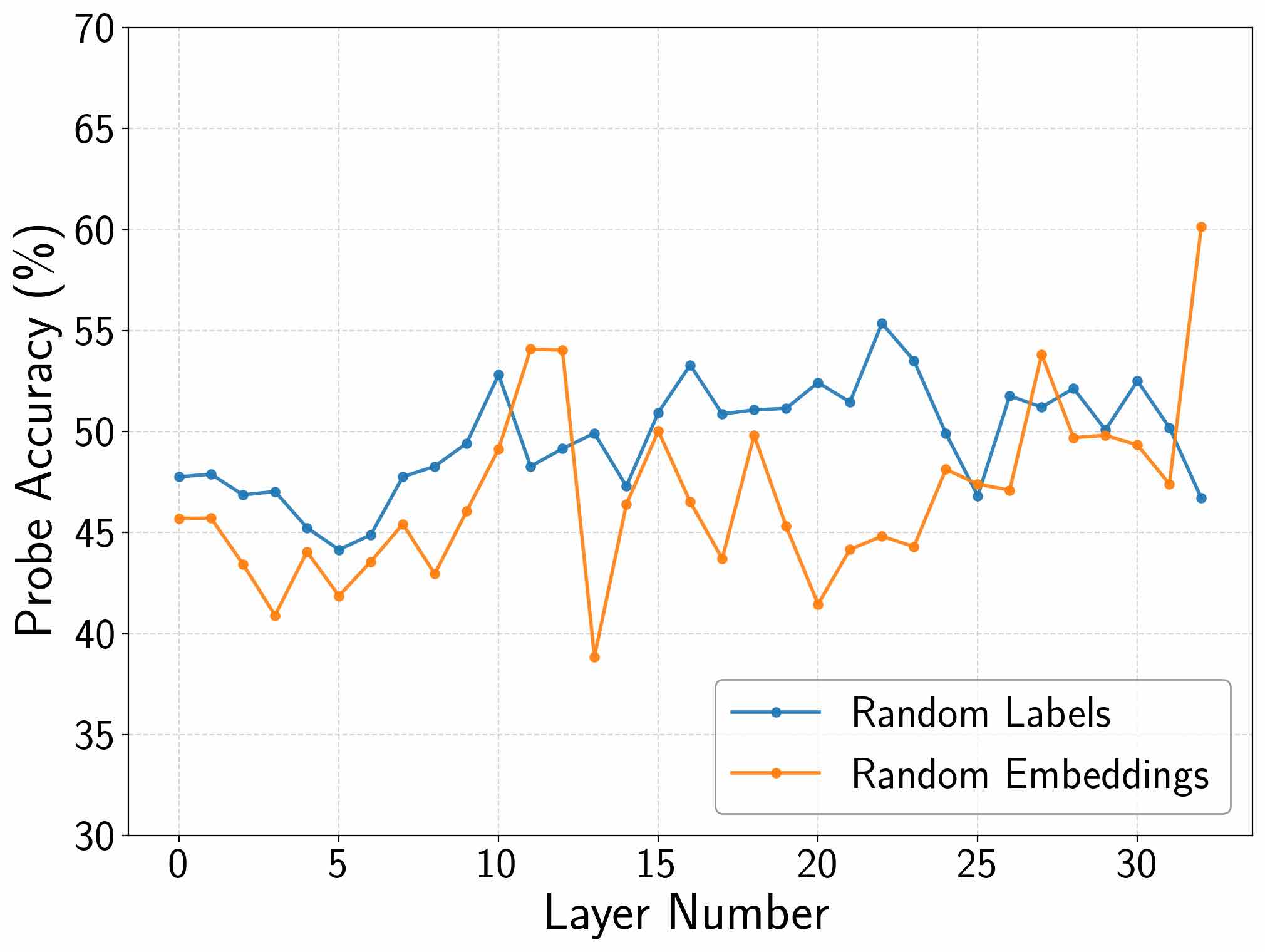}
    \caption{\textbf{Democracy} probe accuracy across layers in \texttt{Llama-3-8B} using random embeddings or random labels during probe training}
    \label{fig:Democracy_meta3b_control}
\end{figure}

\begin{figure}[H]
    \centering
    \includegraphics[width=\linewidth]{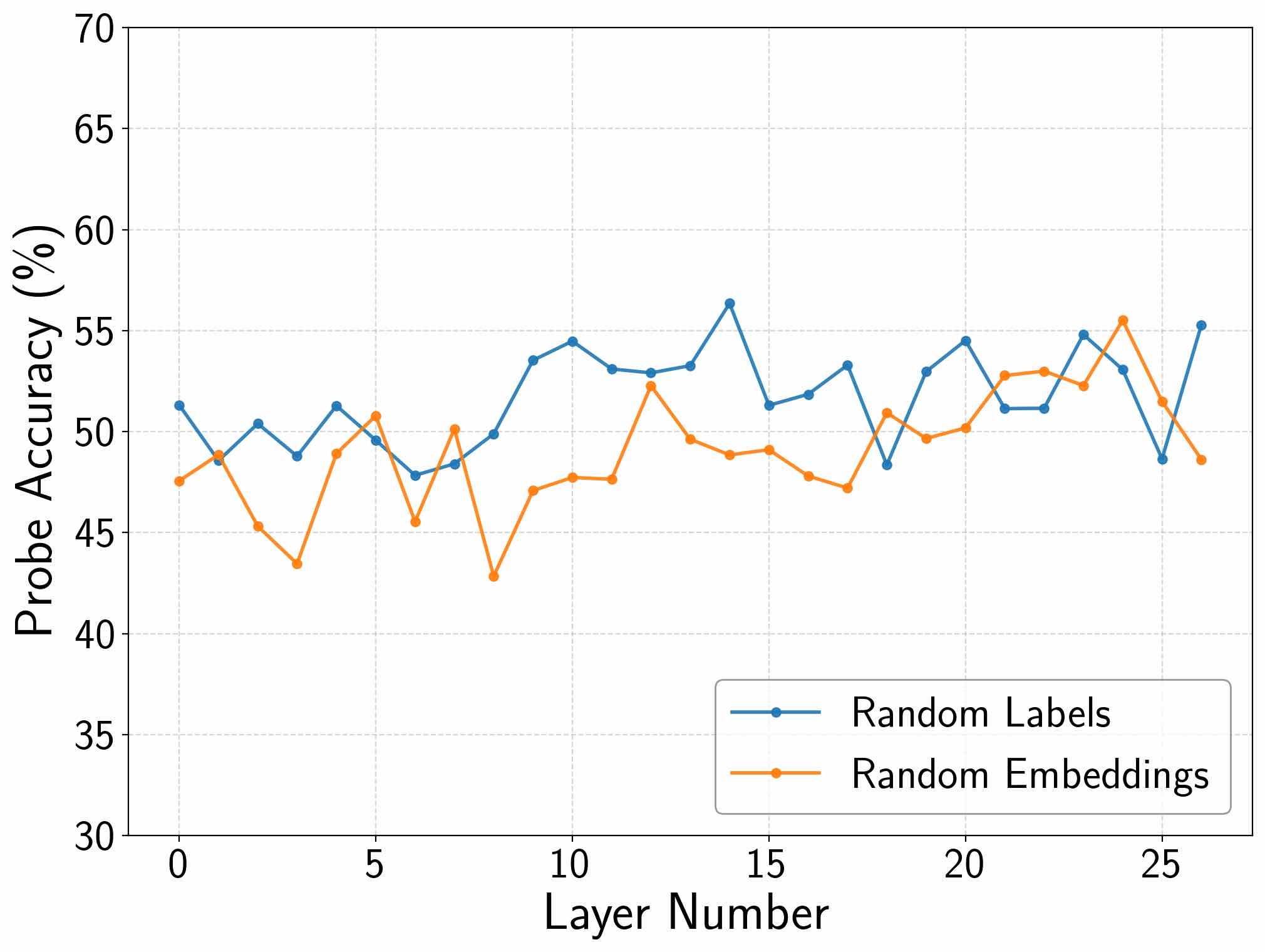}
    \caption{\textbf{Democracy} probe accuracy across layers in \texttt{Gemma-2-2B} using random embeddings or random labels during probe training}
    \label{fig:Democracy_gemma2b_control}
\end{figure}

\begin{figure}[H]
    \centering
    \includegraphics[width=\linewidth]{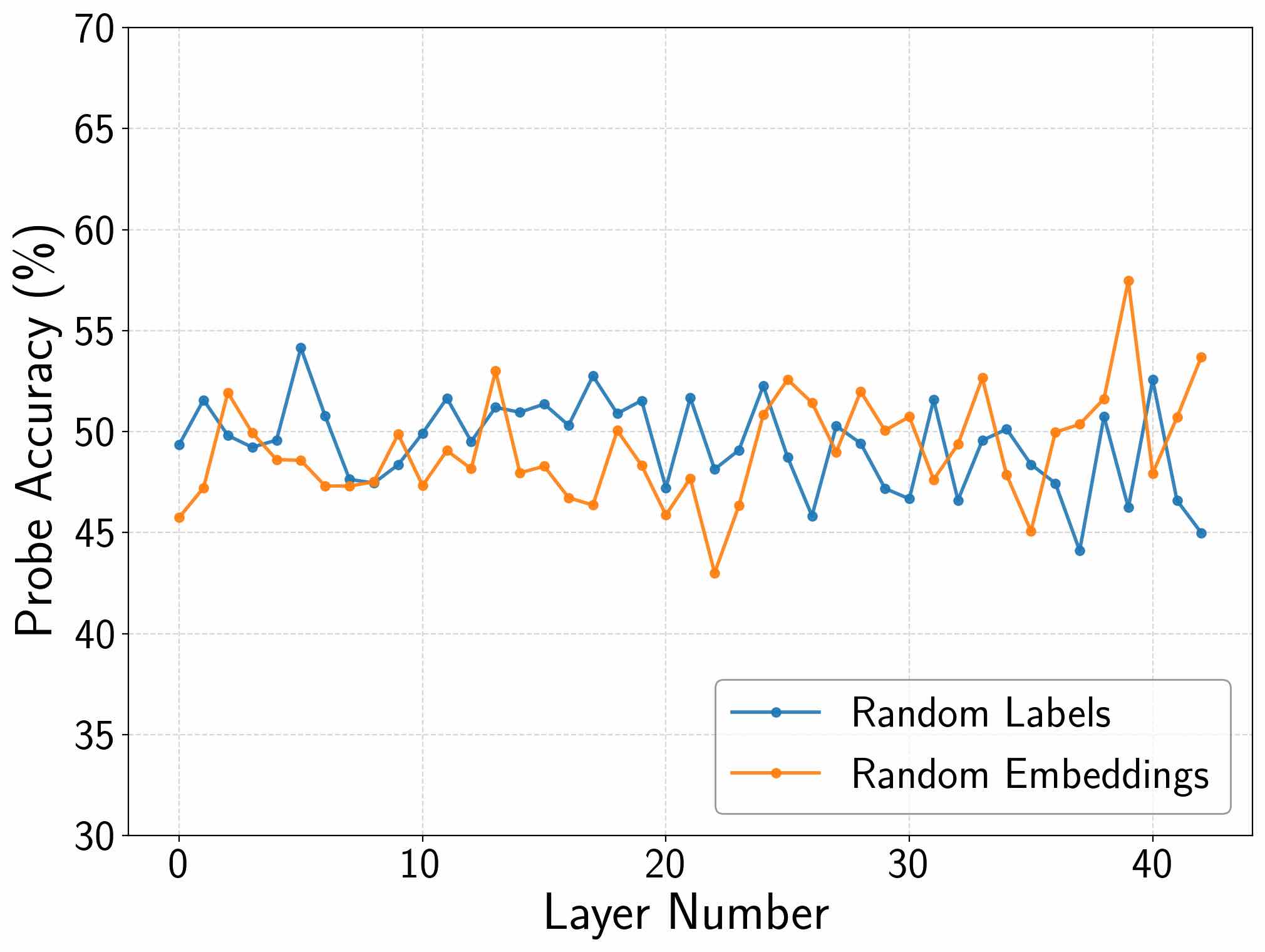}
    \caption{\textbf{Democracy} probe accuracy across layers in \texttt{Gemma-2-9B} using random embeddings or random labels during probe training}
    \label{fig:Democracy_gemma9b_control}
\end{figure}

\begin{figure}[H]
    \centering
    \includegraphics[width=\linewidth]{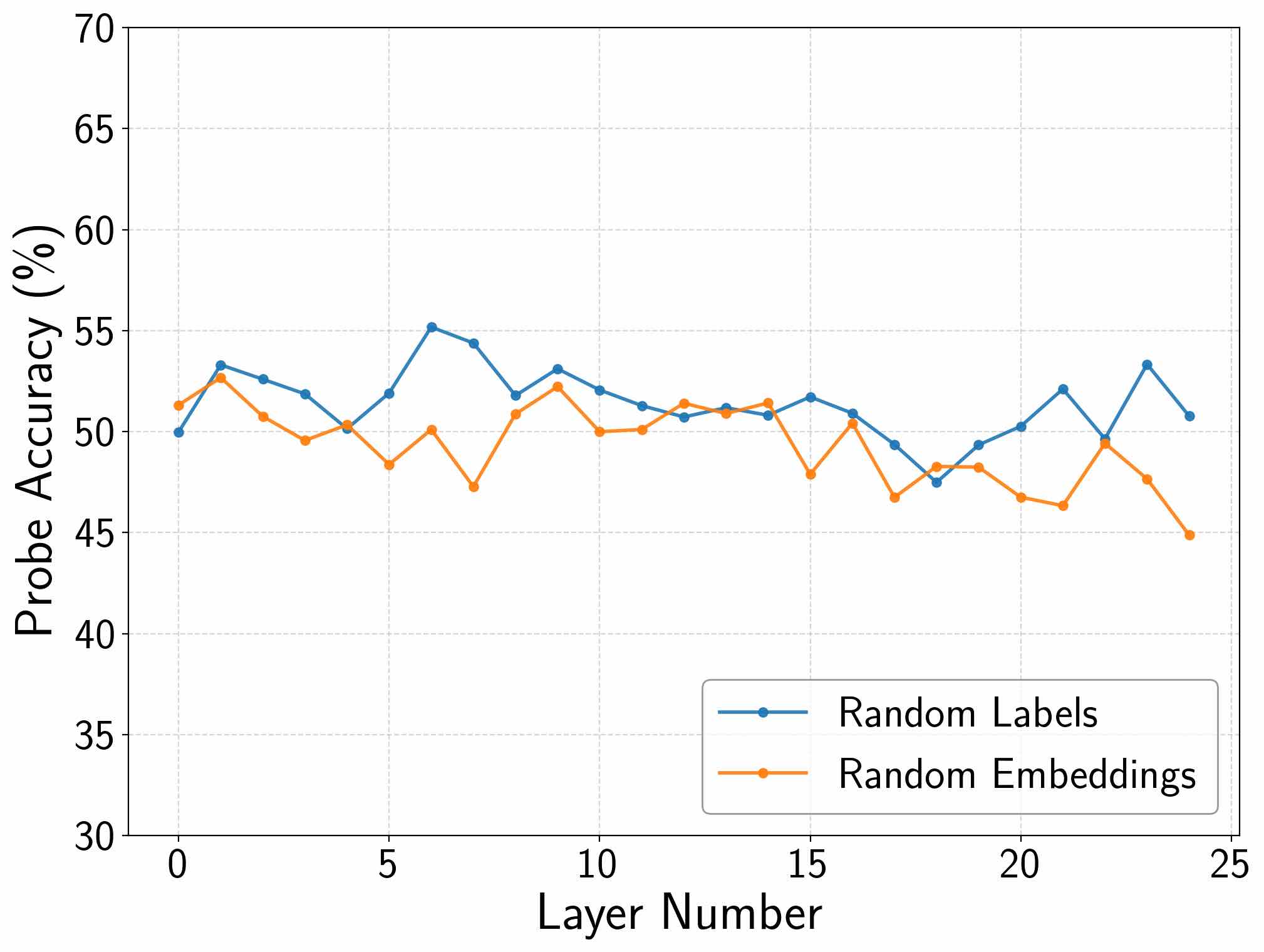}
    \caption{\textbf{Democracy} probe accuracy across layers in \texttt{Qwen2.5-0.5B} using random embeddings or random labels during probe training}
    \label{fig:Democracy_qwen0p5b_control}
\end{figure}

\begin{figure}[H]
    \centering
    \includegraphics[width=\linewidth]{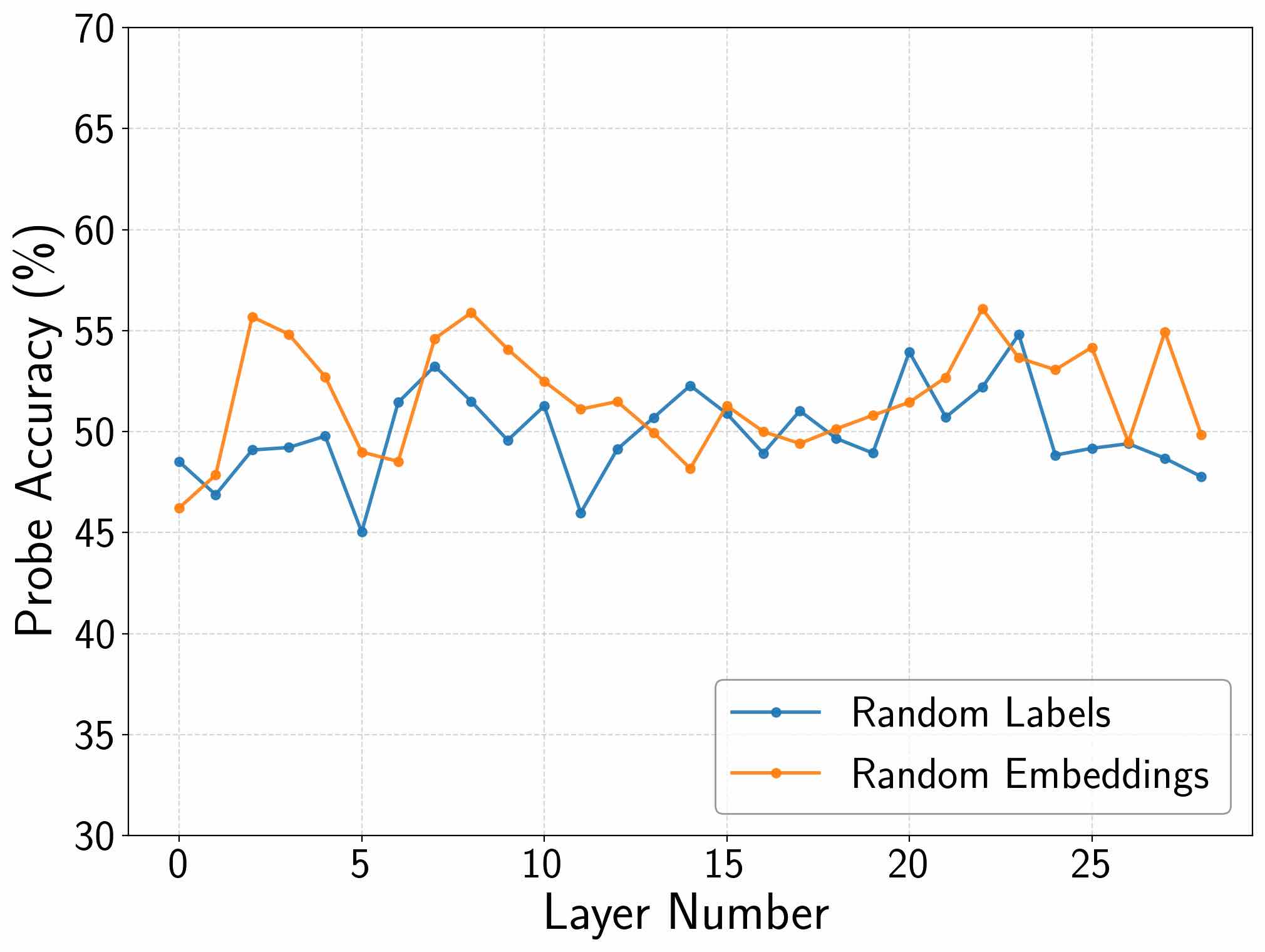}
    \caption{\textbf{Democracy} probe accuracy across layers in \texttt{Qwen2.5-1.5B} using random embeddings or random labels during probe training}
    \label{fig:Democracy_qwen1p5b_control}
\end{figure}

\begin{figure}[H]
    \centering
    \includegraphics[width=\linewidth]{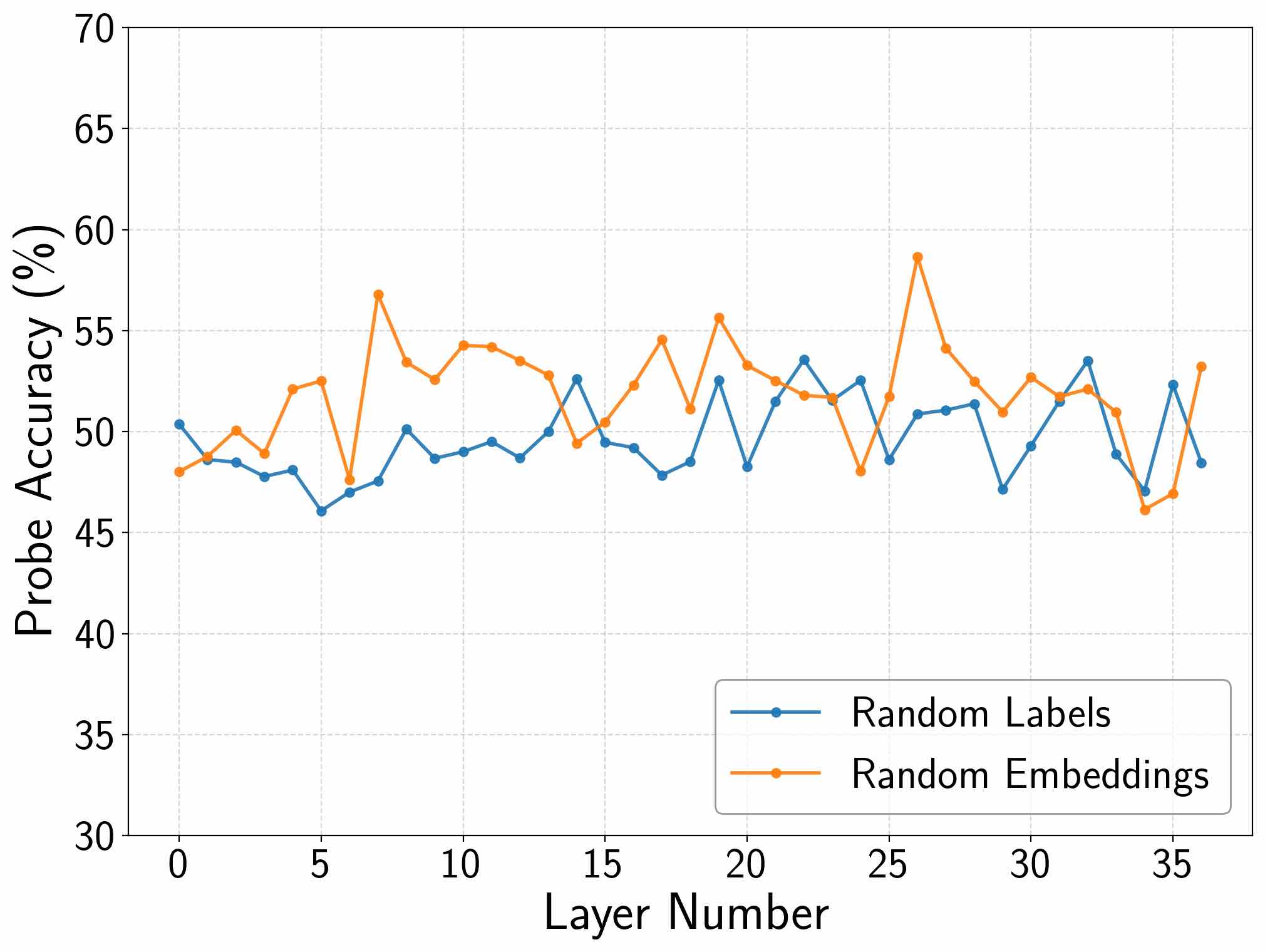}
    \caption{\textbf{Democracy} probe accuracy across layers in \texttt{Qwen2.5-3B} using random embeddings or random labels during probe training}
    \label{fig:Democracy_qwen3b_control}
\end{figure}

\begin{figure}[H]
    \centering
    \includegraphics[width=\linewidth]{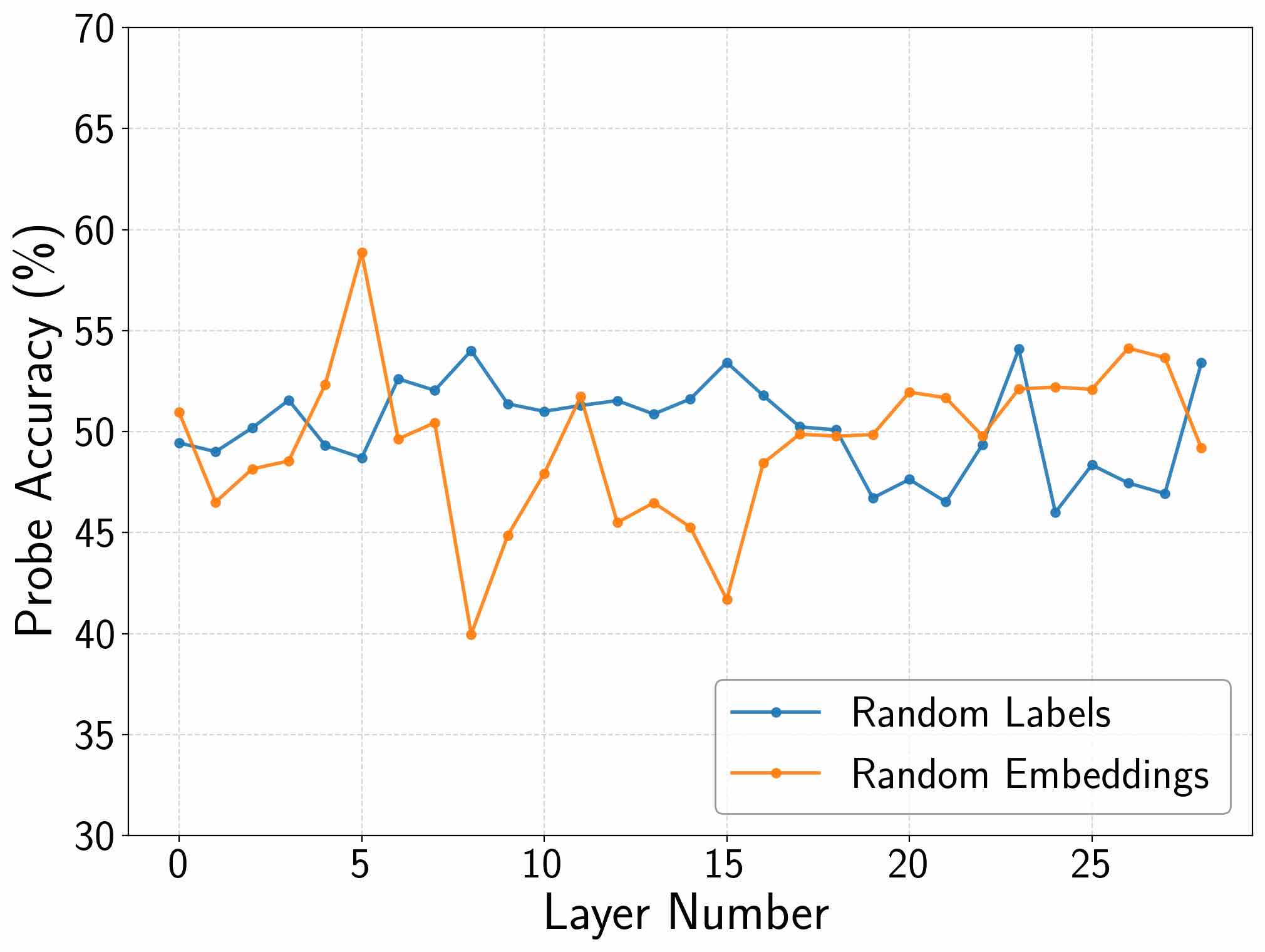}
    \caption{\textbf{Democracy} probe accuracy across layers in \texttt{Qwen2.5-7B} using random embeddings or random labels during probe training}
    \label{fig:Democracy_qwen7b_control}
\end{figure}

Figures \ref{fig:Democracy_llama_vs_probe_params}, \ref{fig:Democracy_gemma2b_vs_probe_params}, and \ref{fig:Democracy_qwen0p5b_vs_probe_params} show the \textbf{Democracy} probe accuracy versus probe size for \texttt{Llama-3-8B}, \texttt{Gemma-2-2B}, and \texttt{Qwen2.5-0.5B}, respectively, and Table \ref{table:Democracy_vs_params} shows a summary of these results. Figures \ref{fig:Democracy_meta3b_control}--\ref{fig:Democracy_qwen7b_control} show the probe accuracies across layers for all LLMs when the probes are trained on the control task (randomizing embeddings or labels).

\begin{table}[H]
\centering
\begin{threeparttable}
\begin{tabular}{cc|cccc}
\hline\hline
\multirow{2}{*}{Probed LLM} & \multirow{2}{1cm}{Probed Layer} & \multicolumn{4}{c}{\# Probe parameters} \\
    & & 20 & 40 & 80 & max \\
\hline\hline

\multirow{3}{*}{\texttt{Llama-3-8B}} & 1 &	71 &	80 &	88 &	79 \\
& 15 &	72 &	74 &	80 &	96 \\
& 32 &	86 &	89 &	92 &	95 \\

\hline
                            
\multirow{3}{*}{\texttt{Gemma-2-2B}} & 1 &	68 &	78 &	85 &	89 \\
& 15 &	69 &	77 &	83 &	94 \\
& 26 &	86 &	88 &	90 &	95 \\

\hline

\multirow{3}{*}{\texttt{Qwen2.5-0.5B}} & 1 &	73 &	77 &	79 &	68 \\
& 15 &	77 &	80 &	86 &	95 \\
& 24 &	84 &	87 &	88 &	95 \\

\hline\hline
\end{tabular}
\begin{tablenotes}
\footnotesize
\item[$\bullet$] All results are in percentage (\%).
\item[$\bullet$] ``max'' denotes 4,096 for \texttt{Llama-3-8B}, 2,304 for \texttt{Gemma-2-2B}, and 896 for \texttt{Qwen2.5-0.5B}.
\item[$\bullet$] standard deviation for each result $\leq$ 3\%.
\end{tablenotes}
\end{threeparttable}
\caption{\textbf{Democracy} probe accuracy across model families, sizes, layers, and probe sizes} 
\label{table:Democracy_vs_params}
\end{table}

\subsection{Extended Results for Inference of Envy}

\subsubsection{Probing for Envy using Nth Embedding vs. Average Embedding}
\begin{figure}[H]
    \centering
    \includegraphics[width=\linewidth]{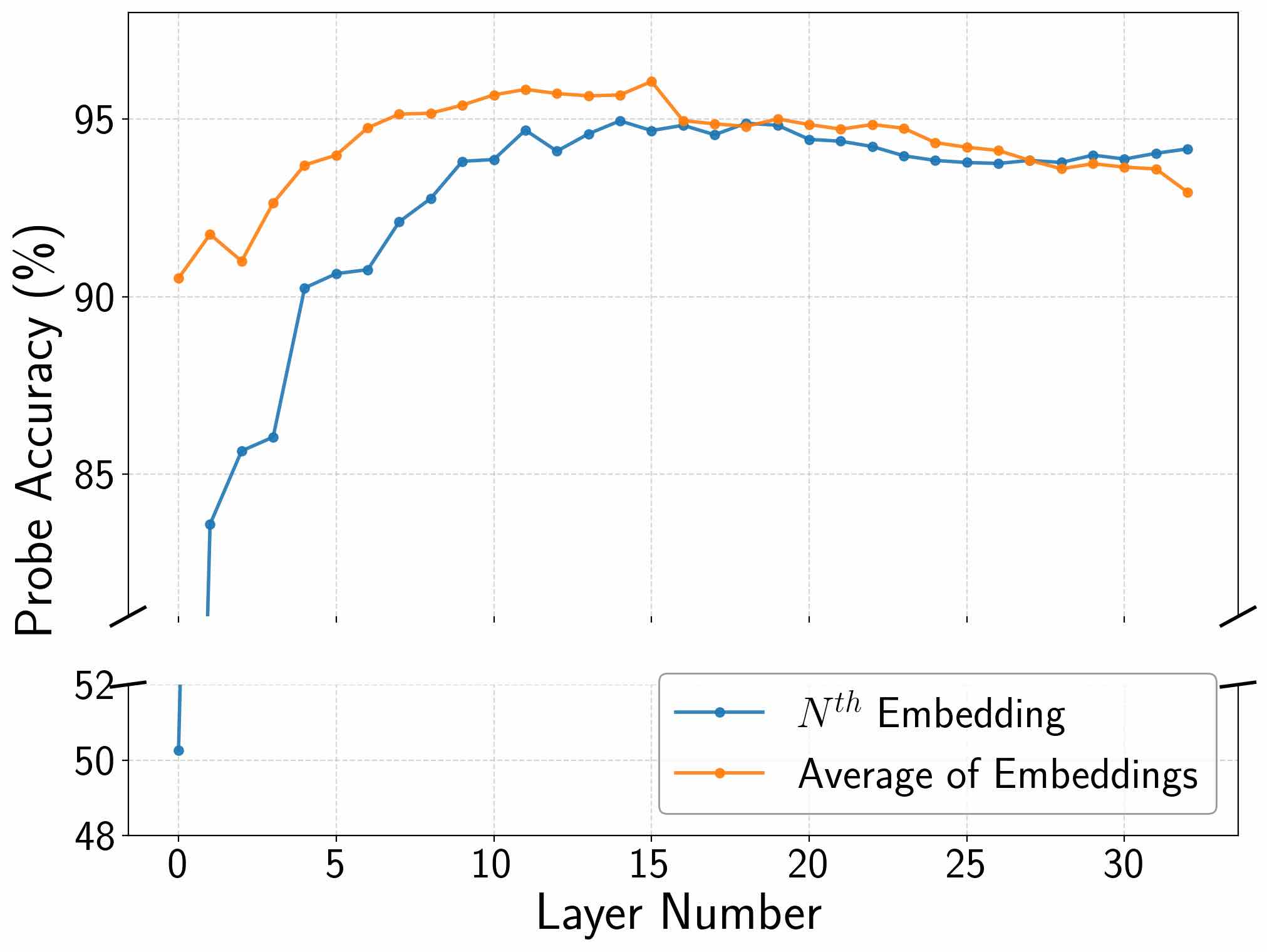}
    \caption{\textbf{Envy} probe accuracy for \texttt{Llama-3-8B} using average and $N^{th}$ embeddings vs. layer}
    \label{fig:Envy_llama_right_most_vs_average}
\end{figure}

\begin{figure}[H]
    \centering
    \includegraphics[width=\linewidth]{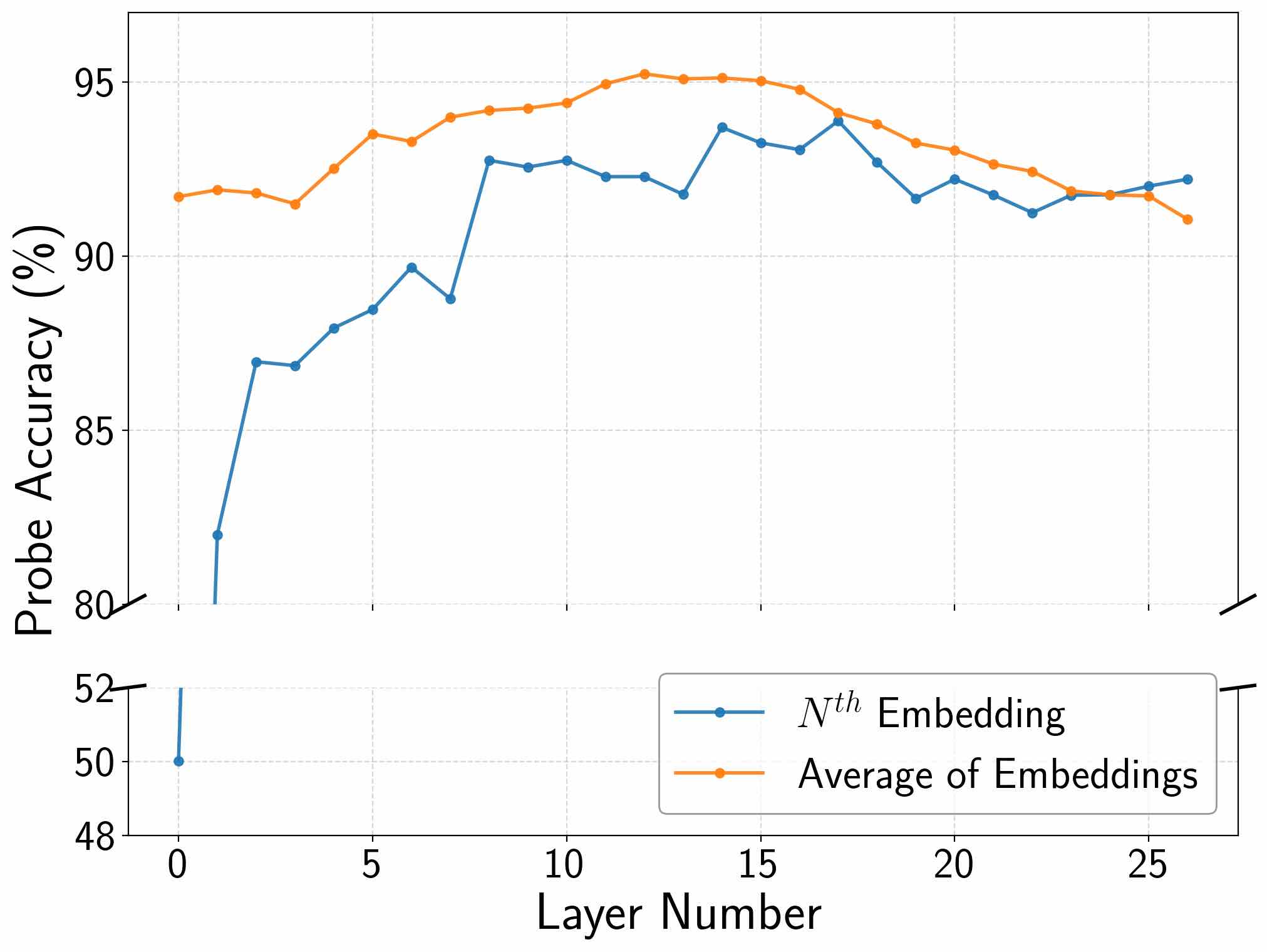}
    \caption{\textbf{Envy} probe accuracy for \texttt{Gemma-2-2B} using average and $N^{th}$ embeddings vs. layer}
    \label{fig:Envy_gemma2b_right_most_vs_average}
\end{figure}

\begin{figure}[H]
    \centering
    \includegraphics[width=\linewidth]{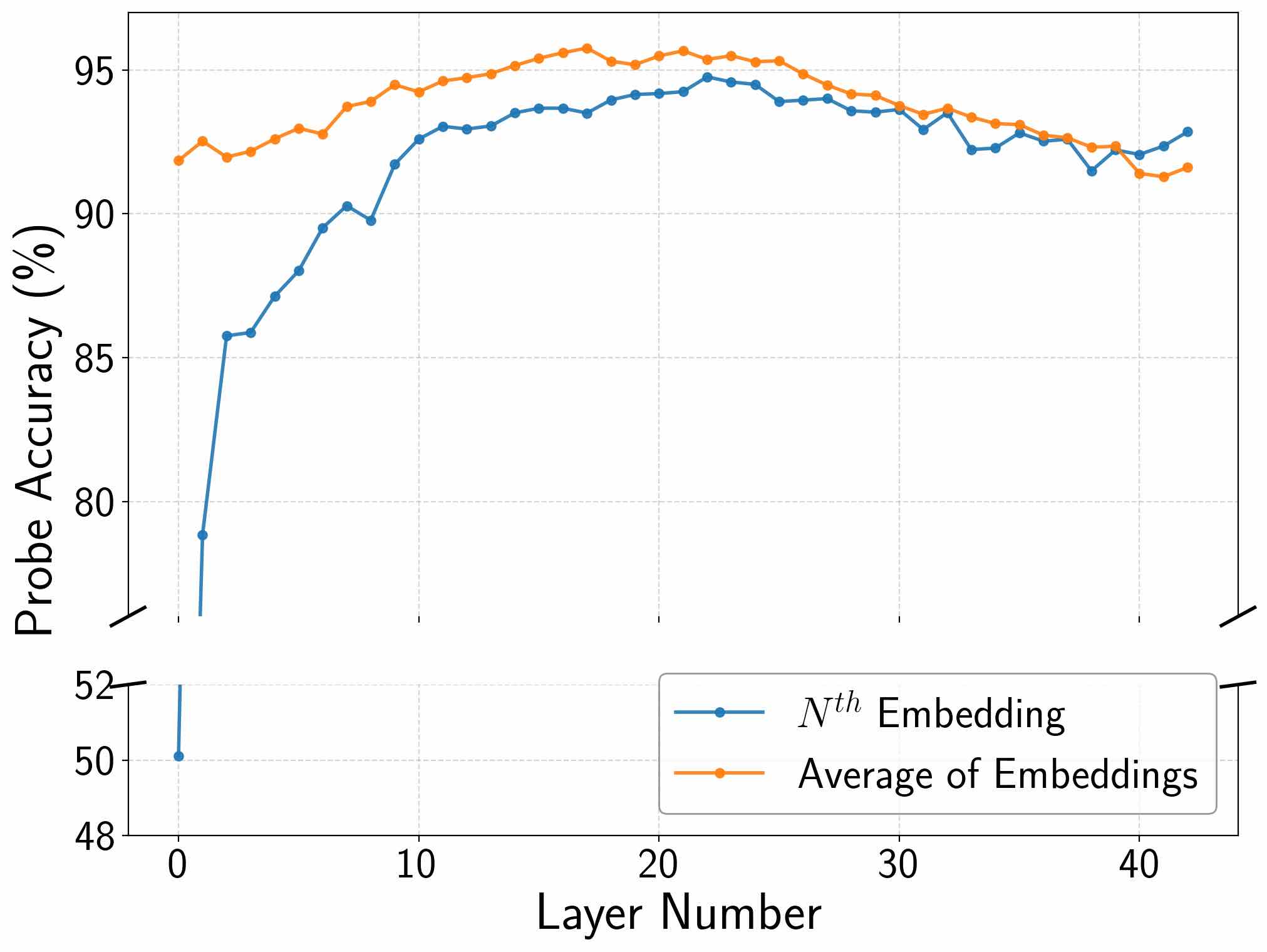}
    \caption{\textbf{Envy} probe accuracy for \texttt{Gemma-2-9B} using average and $N^{th}$ embeddings vs. layer}
    \label{fig:Envy_gemma9b_right_most_vs_average}
\end{figure}

\begin{figure}[H]
    \centering
    \includegraphics[width=\linewidth]{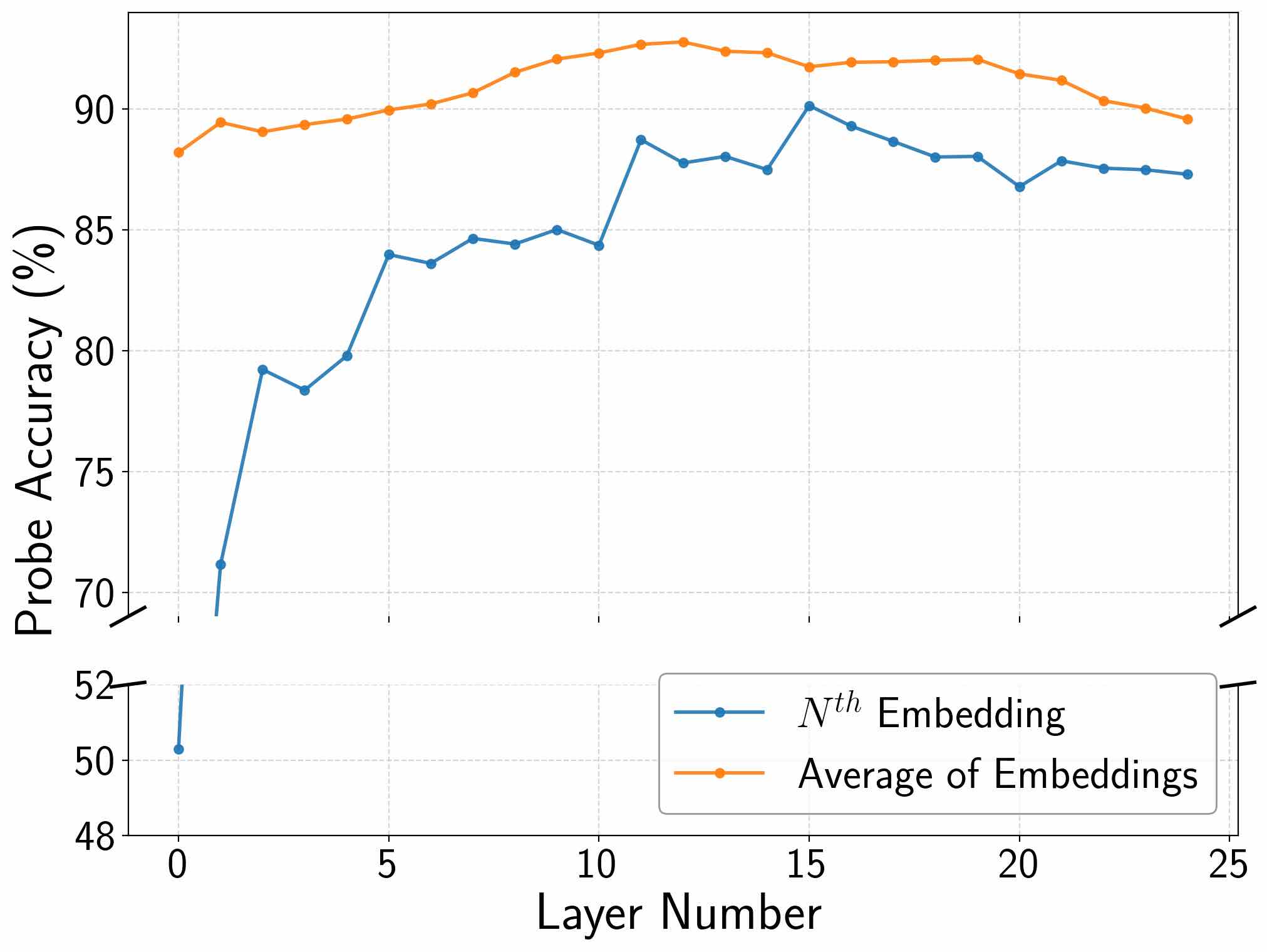}
    \caption{\textbf{Envy} probe accuracy for \texttt{Qwen2.5-0.5B} using average and $N^{th}$ embeddings vs. layer}
    \label{fig:Envy_qwen0p5b_right_most_vs_average}
\end{figure}

\begin{figure}[H]
    \centering
    \includegraphics[width=\linewidth]{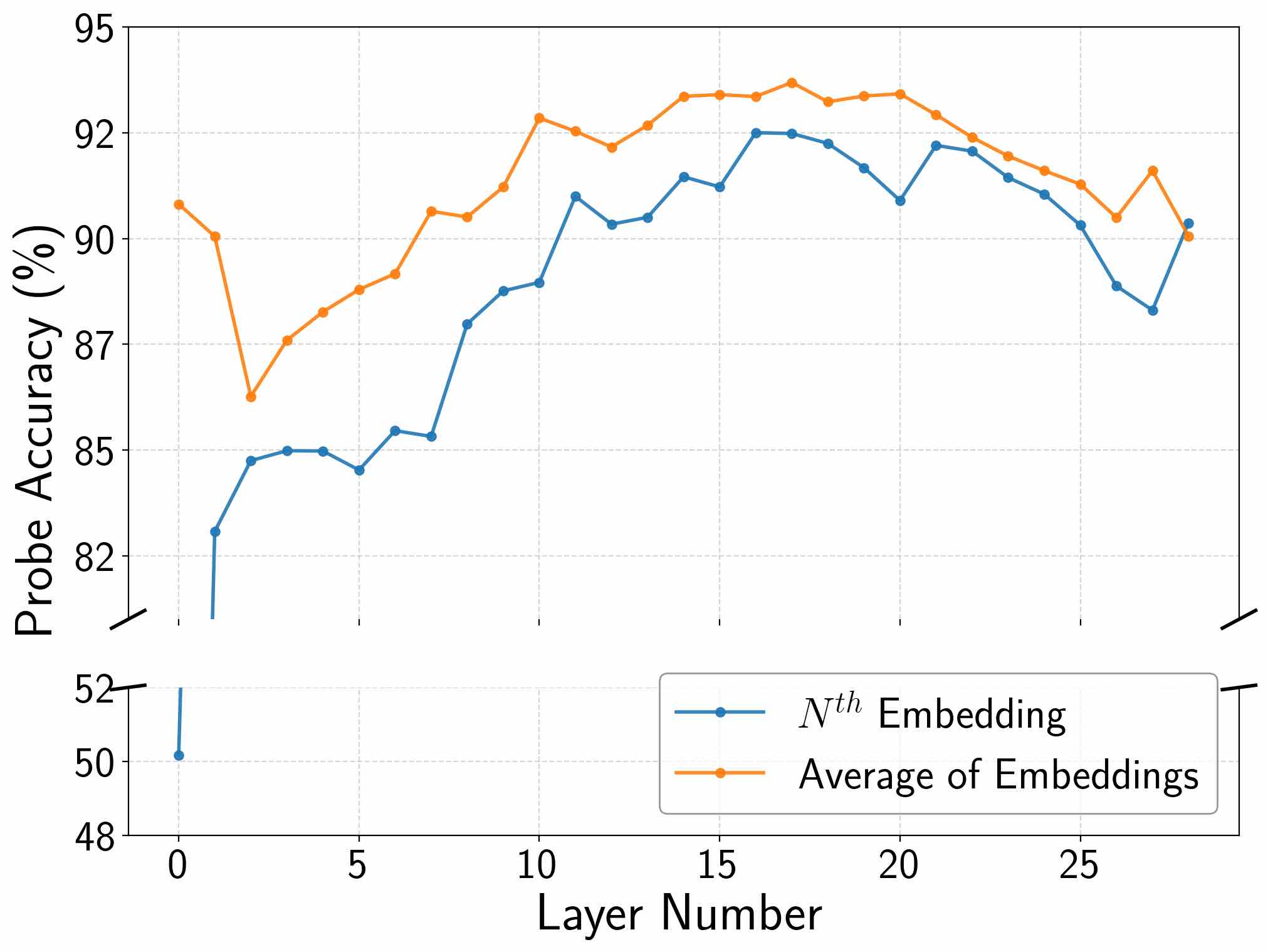}
    \caption{\textbf{Envy} probe accuracy for \texttt{Qwen2.5-1.5B} using average and $N^{th}$ embeddings vs. layer}
    \label{fig:Envy_qwen1p5b_right_most_vs_average}
\end{figure}

\begin{figure}[H]
    \centering
    \includegraphics[width=\linewidth]{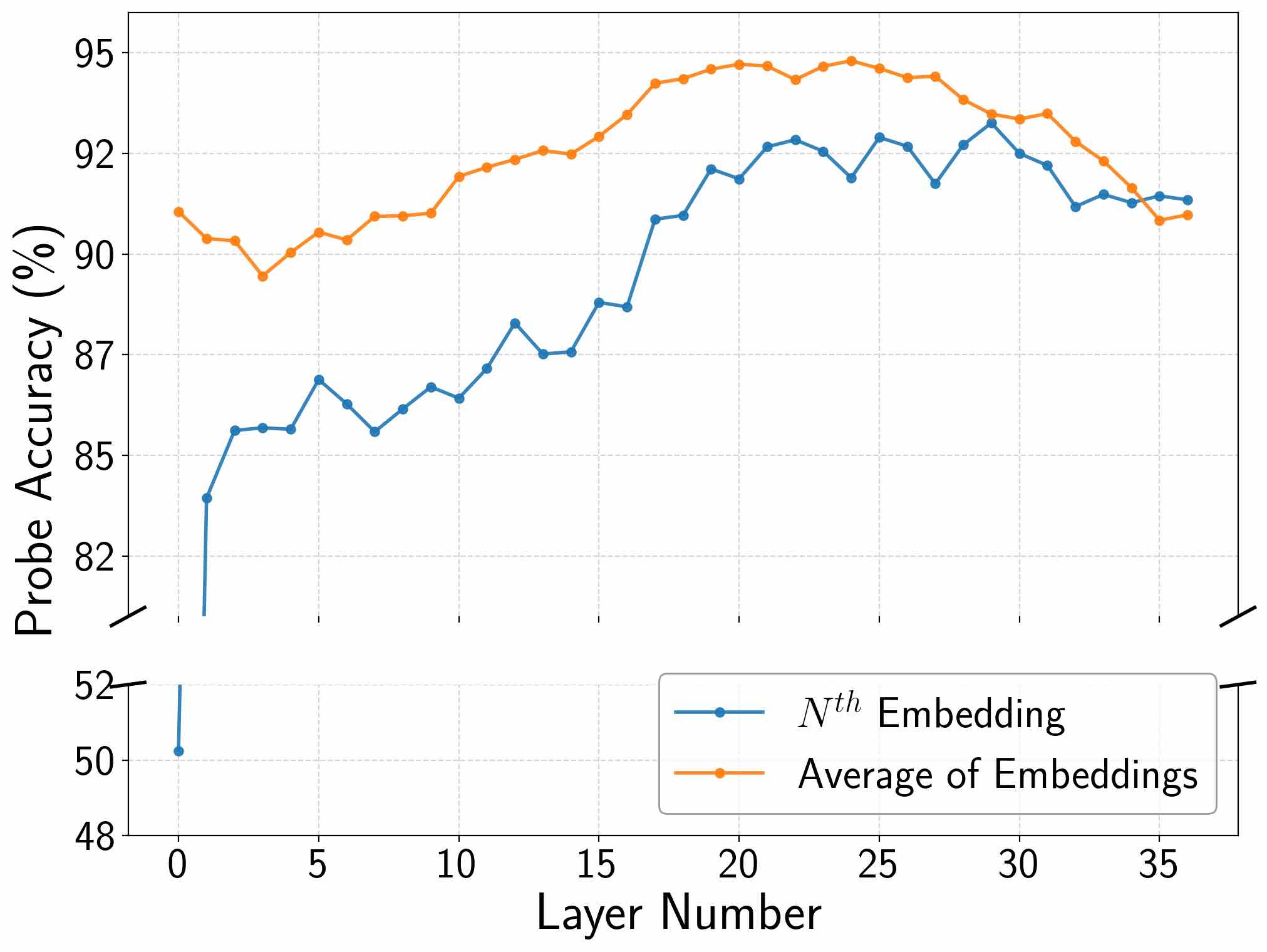}
    \caption{\textbf{Envy} probe accuracy for \texttt{Qwen2.5-3B} using average and $N^{th}$ embeddings vs. layer}
    \label{fig:Envy_qwen3b_right_most_vs_average}
\end{figure}

\begin{figure}[H]
    \centering
    \includegraphics[width=\linewidth]{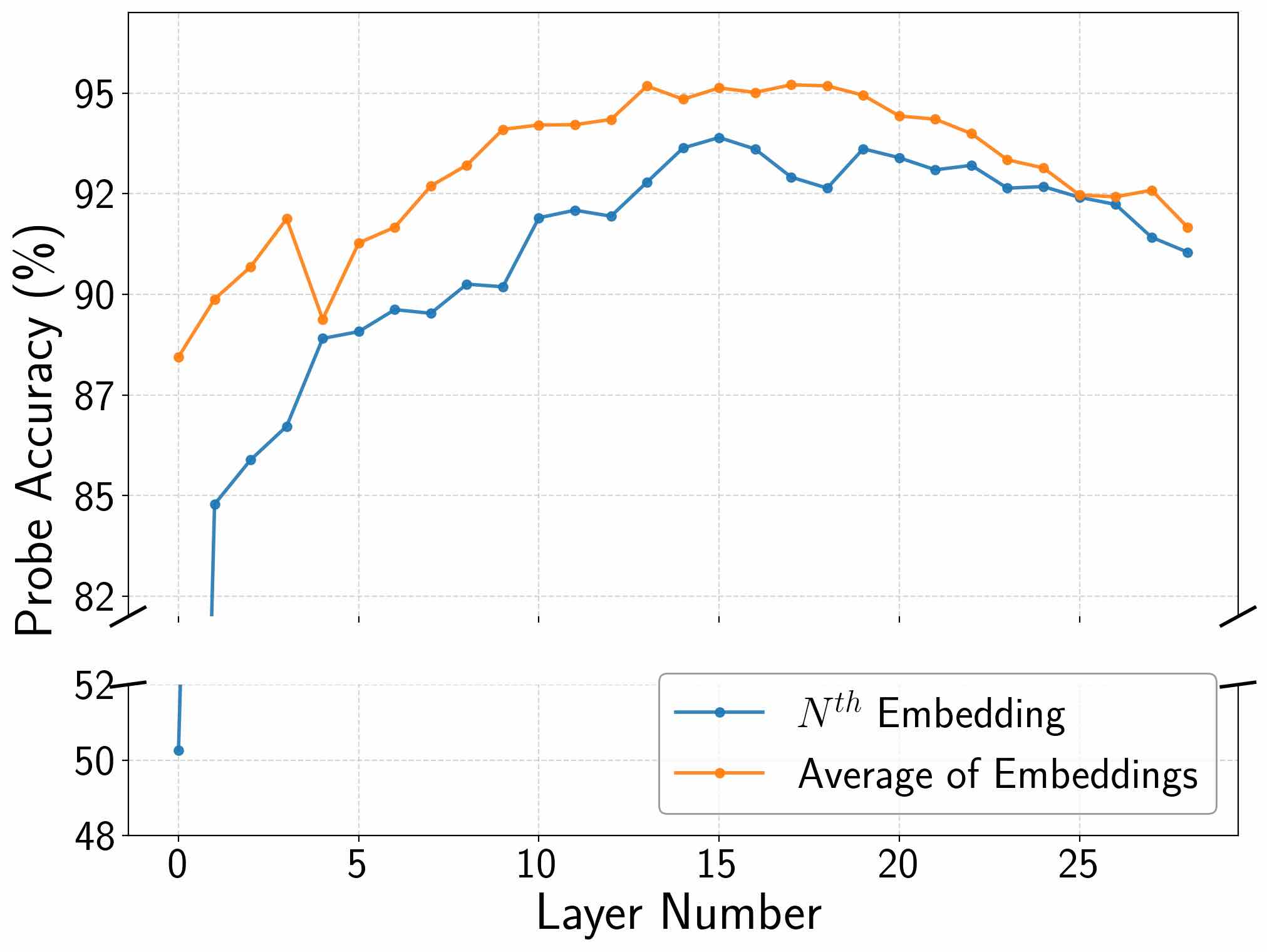}
    \caption{\textbf{Envy} probe accuracy for \texttt{Qwen2.5-7B} using average and $N^{th}$ embeddings vs. layer}
    \label{fig:Envy_qwen7b_right_most_vs_average}
\end{figure}

Figures \ref{fig:Envy_llama_right_most_vs_average}--\ref{fig:Envy_qwen7b_right_most_vs_average} illustrate the \textbf{Envy} probe accuracies across layers of all LLMs using both the $N^{th}$ embedding and the average of all embeddings in the respective layer.

\subsubsection{Envy Probe Cross-Check}
\begin{figure}[H]
    \centering
    \includegraphics[width=\linewidth]{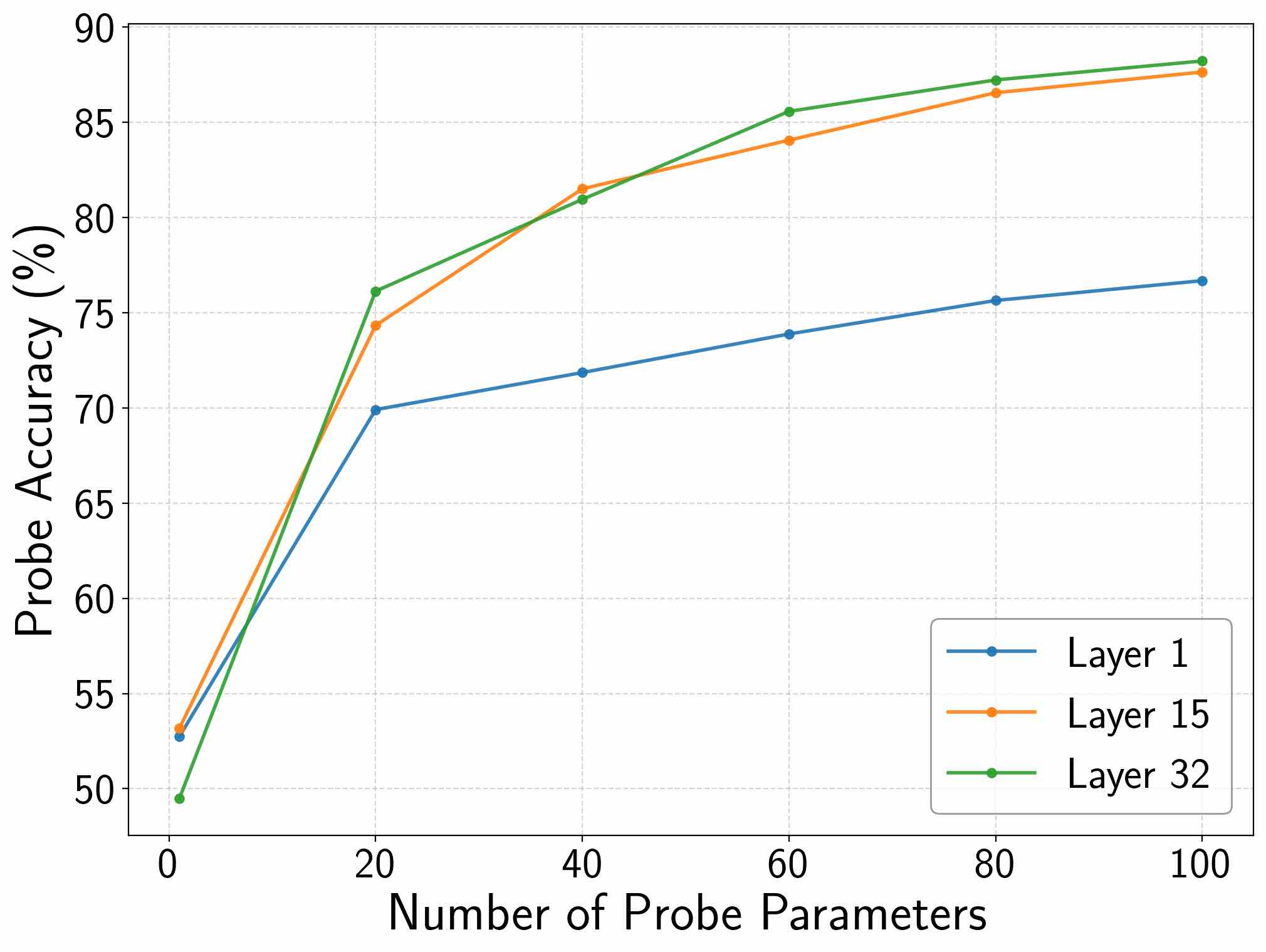}
    \caption{\textbf{Envy} probe accuracy for \texttt{Llama-3-8B} as a function of probe size}
    \label{fig:Envy_llama_vs_probe_params}
\end{figure}

\begin{figure}[H]
    \centering
    \includegraphics[width=\linewidth]{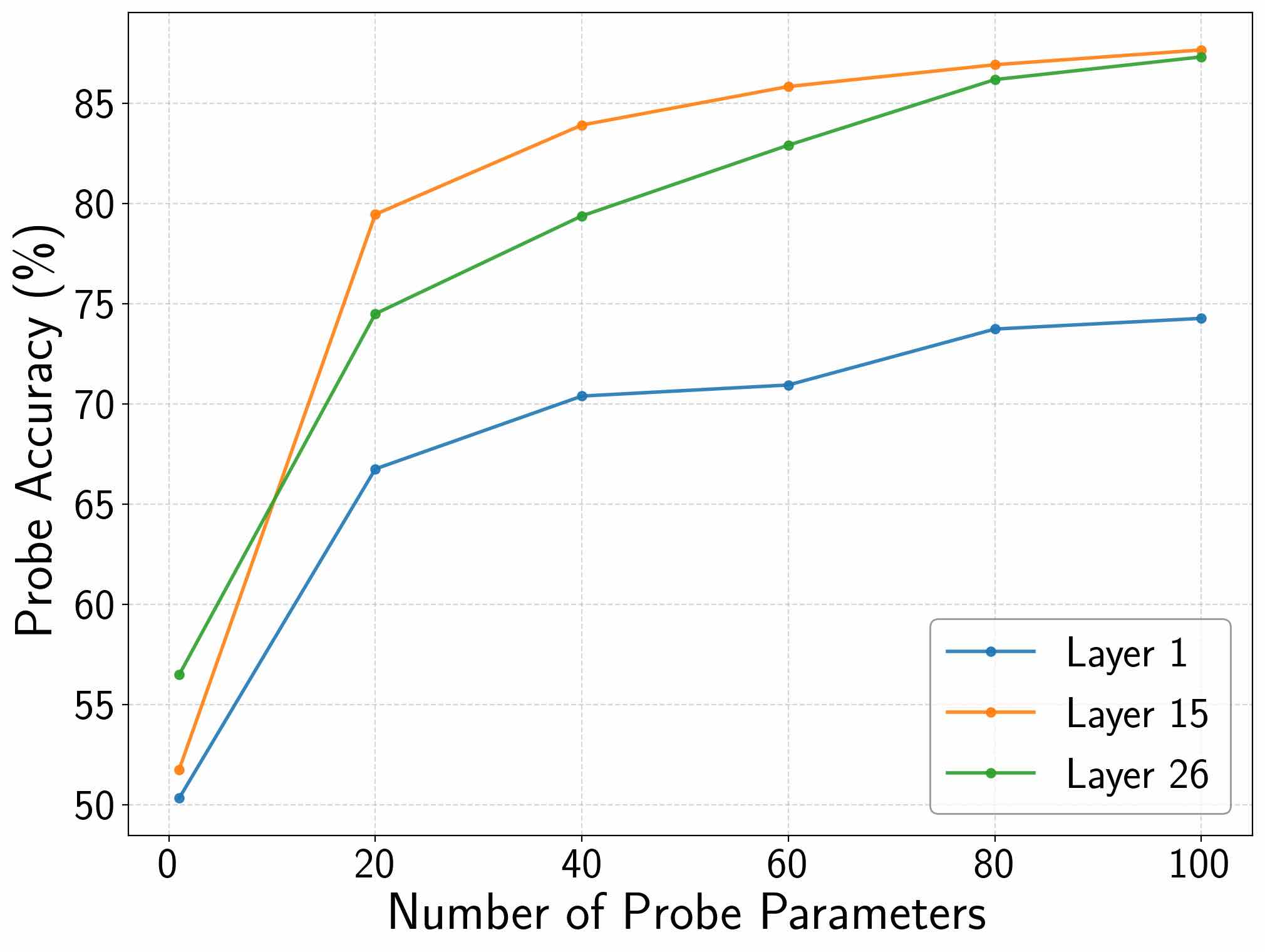}
    \caption{\textbf{Envy} probe accuracy for \texttt{Gemma-2-2B} as a function of probe size}
    \label{fig:Envy_gemma2b_vs_probe_params}
\end{figure}

\begin{figure}[H]
    \centering
    \includegraphics[width=\linewidth]{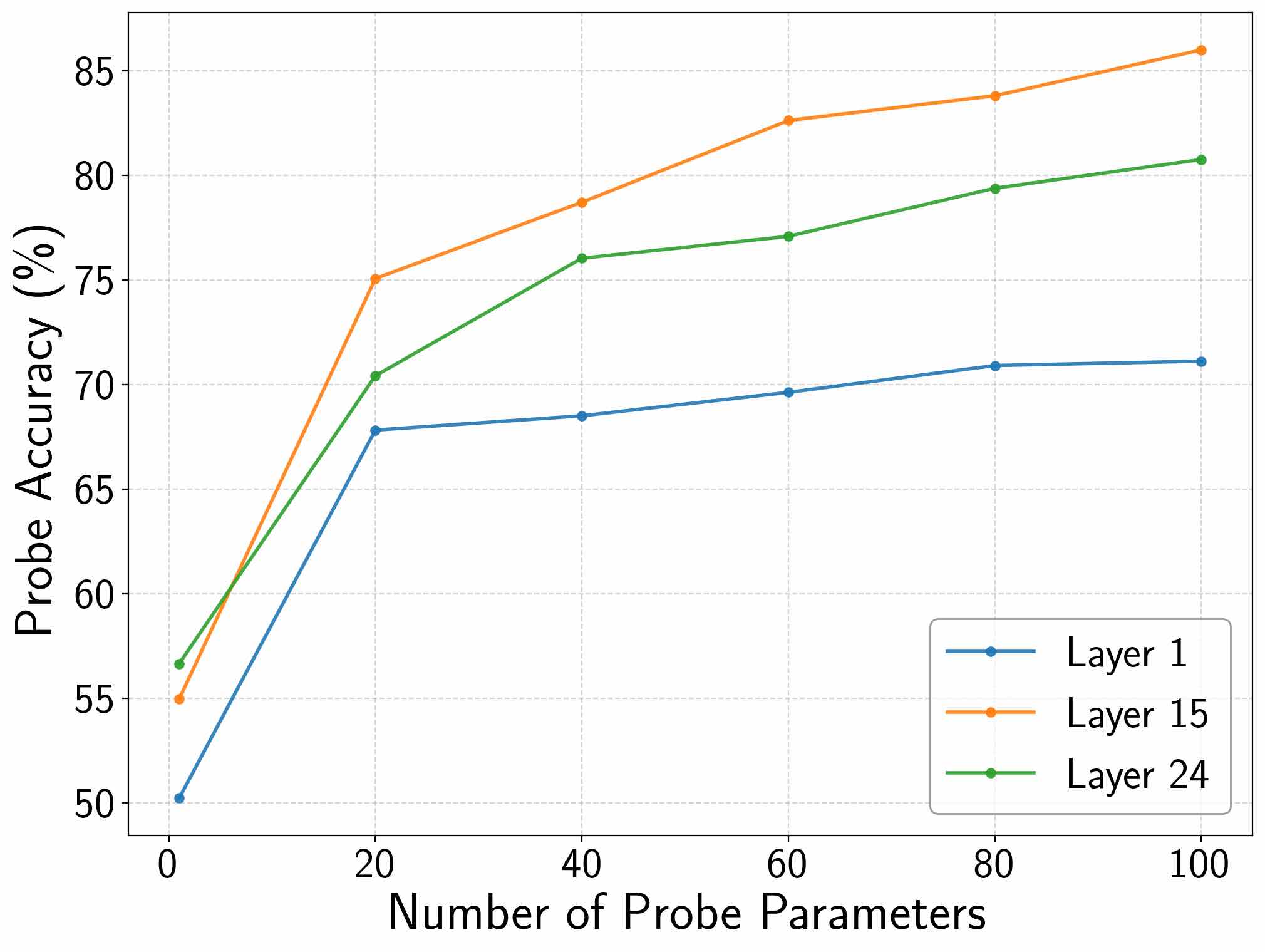}
    \caption{\textbf{Envy} probe accuracy for \texttt{Qwen2.5-0.5B} as a function of probe size}
    \label{fig:Envy_qwen0p5b_vs_probe_params}
\end{figure}

\begin{figure}[H]
    \centering
    \includegraphics[width=\linewidth]{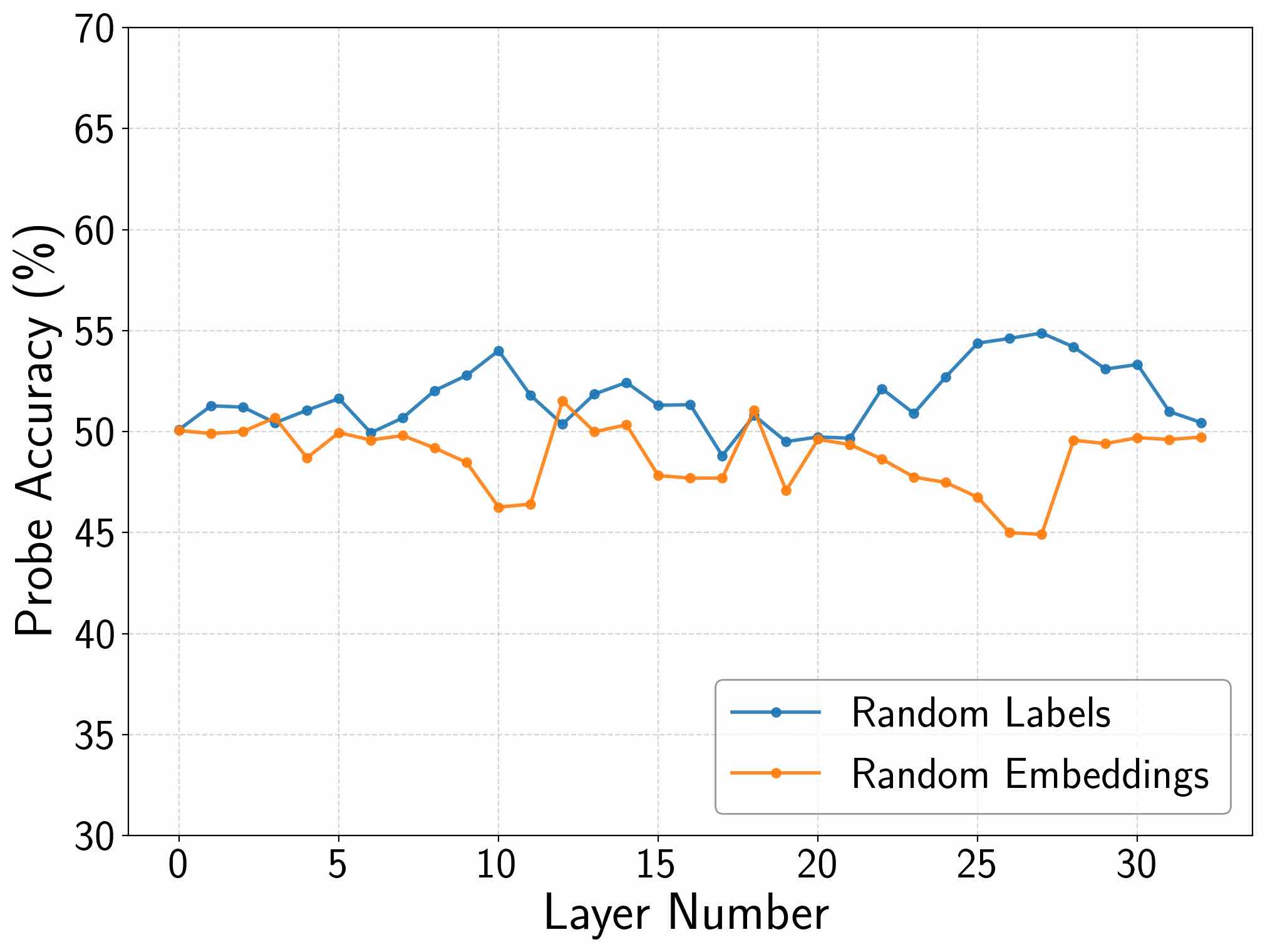}
    \caption{\textbf{Envy} probe accuracy across layers in \texttt{Llama-3-8B} using random embeddings or random labels during probe training}
    \label{fig:Envy_meta3b_control}
\end{figure}

\begin{figure}[H]
    \centering
    \includegraphics[width=\linewidth]{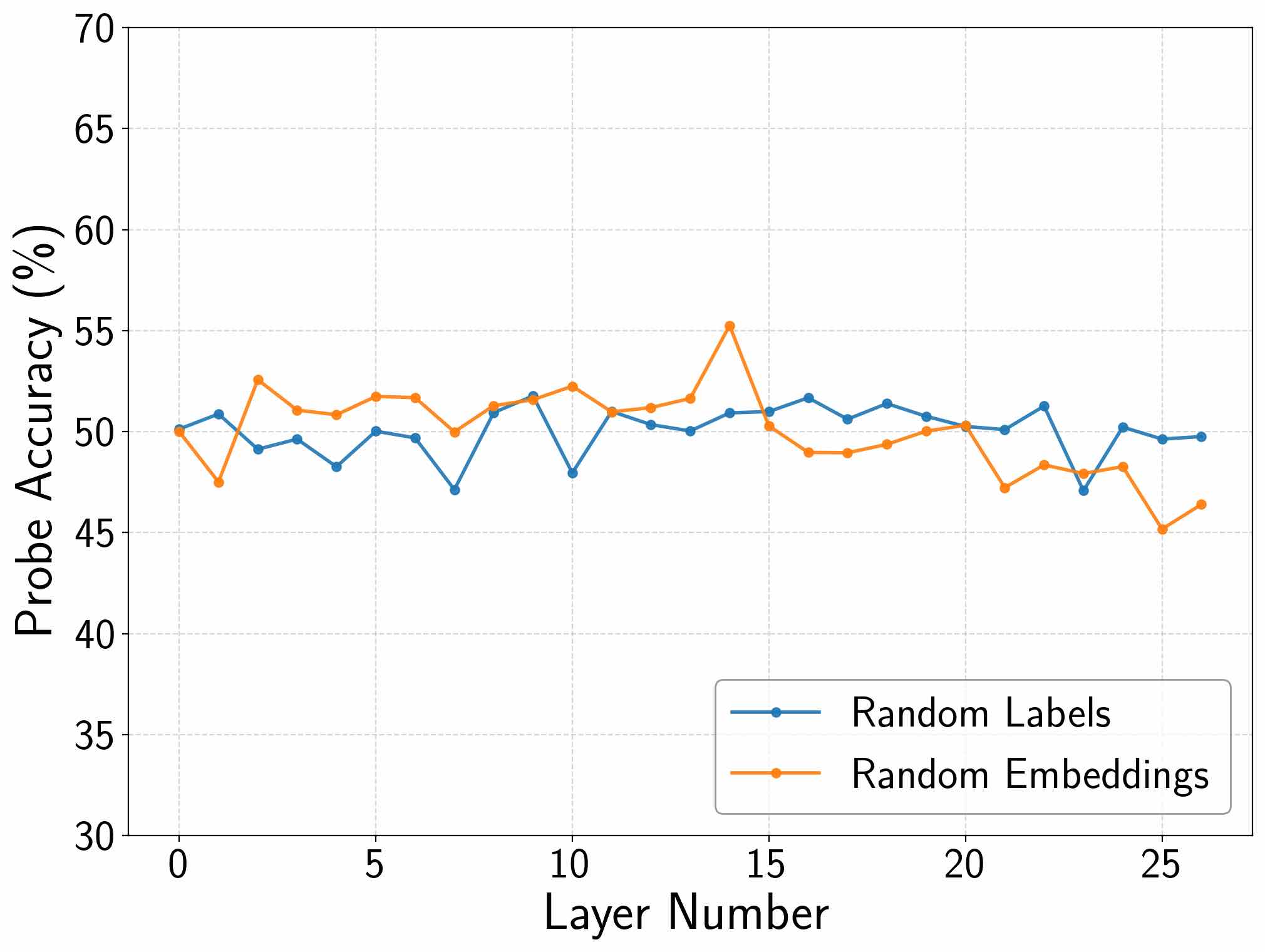}
    \caption{\textbf{Envy} probe accuracy across layers in \texttt{Gemma-2-2B} using random embeddings or random labels during probe training}
    \label{fig:Envy_gemma2b_control}
\end{figure}

\begin{figure}[H]
    \centering
    \includegraphics[width=\linewidth]{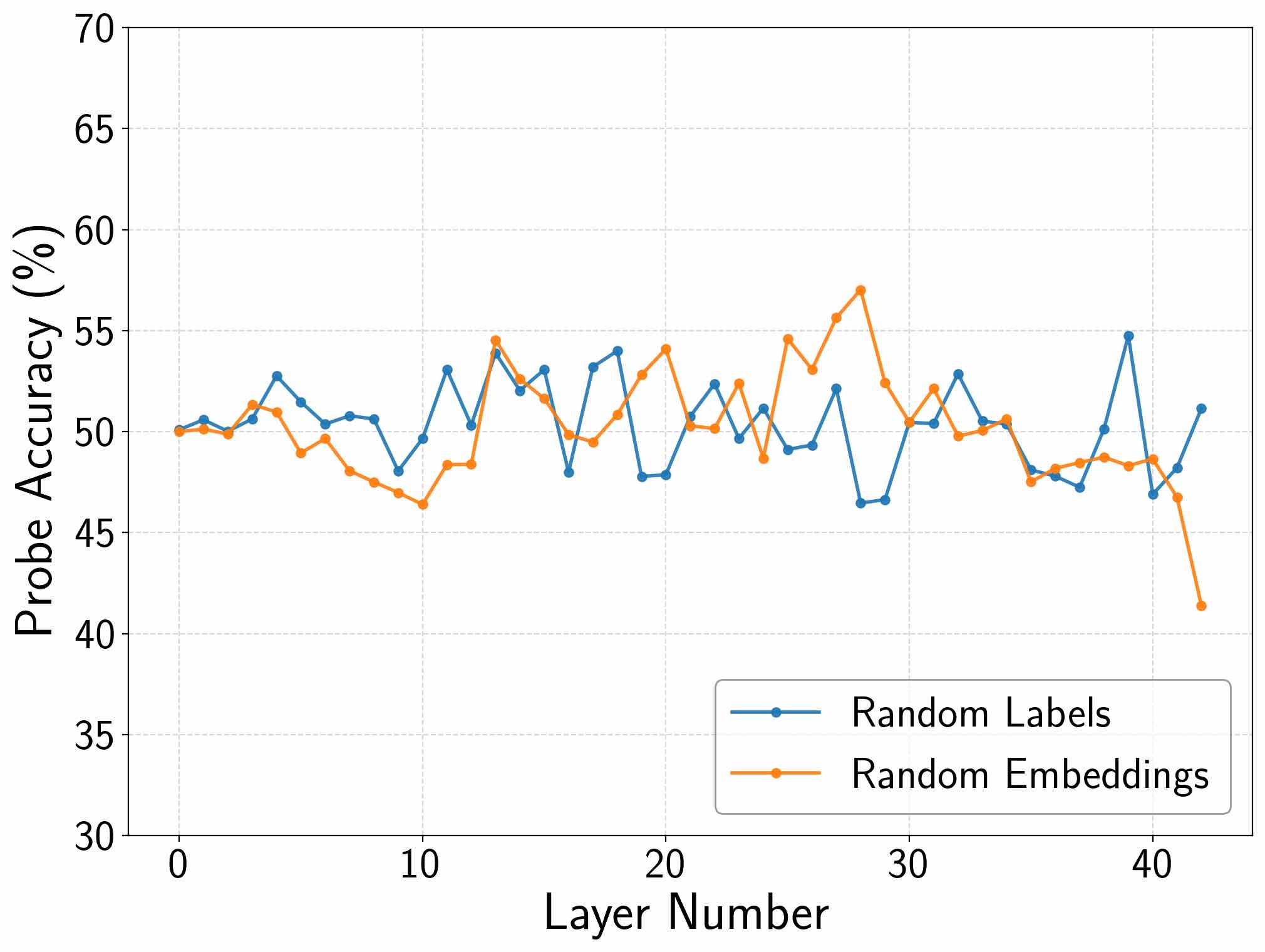}
    \caption{\textbf{Envy} probe accuracy across layers in \texttt{Gemma-2-9B} using random embeddings or random labels during probe training}
    \label{fig:Envy_gemma9b_control}
\end{figure}

\begin{figure}[H]
    \centering
    \includegraphics[width=\linewidth]{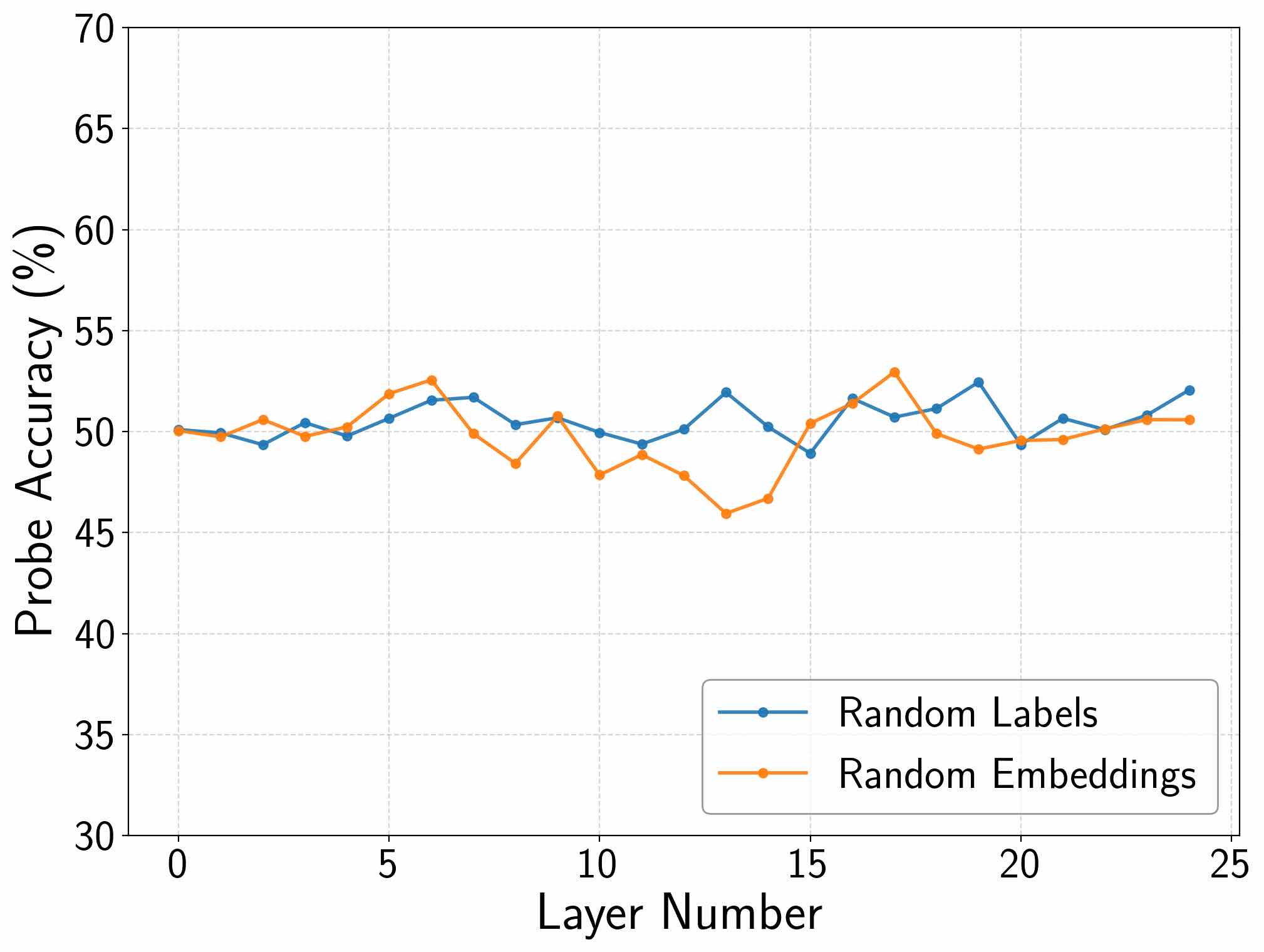}
    \caption{\textbf{Envy} probe accuracy across layers in \texttt{Qwen2.5-0.5B} using random embeddings or random labels during probe training}
    \label{fig:Envy_qwen0p5b_control}
\end{figure}

\begin{figure}[H]
    \centering
    \includegraphics[width=\linewidth]{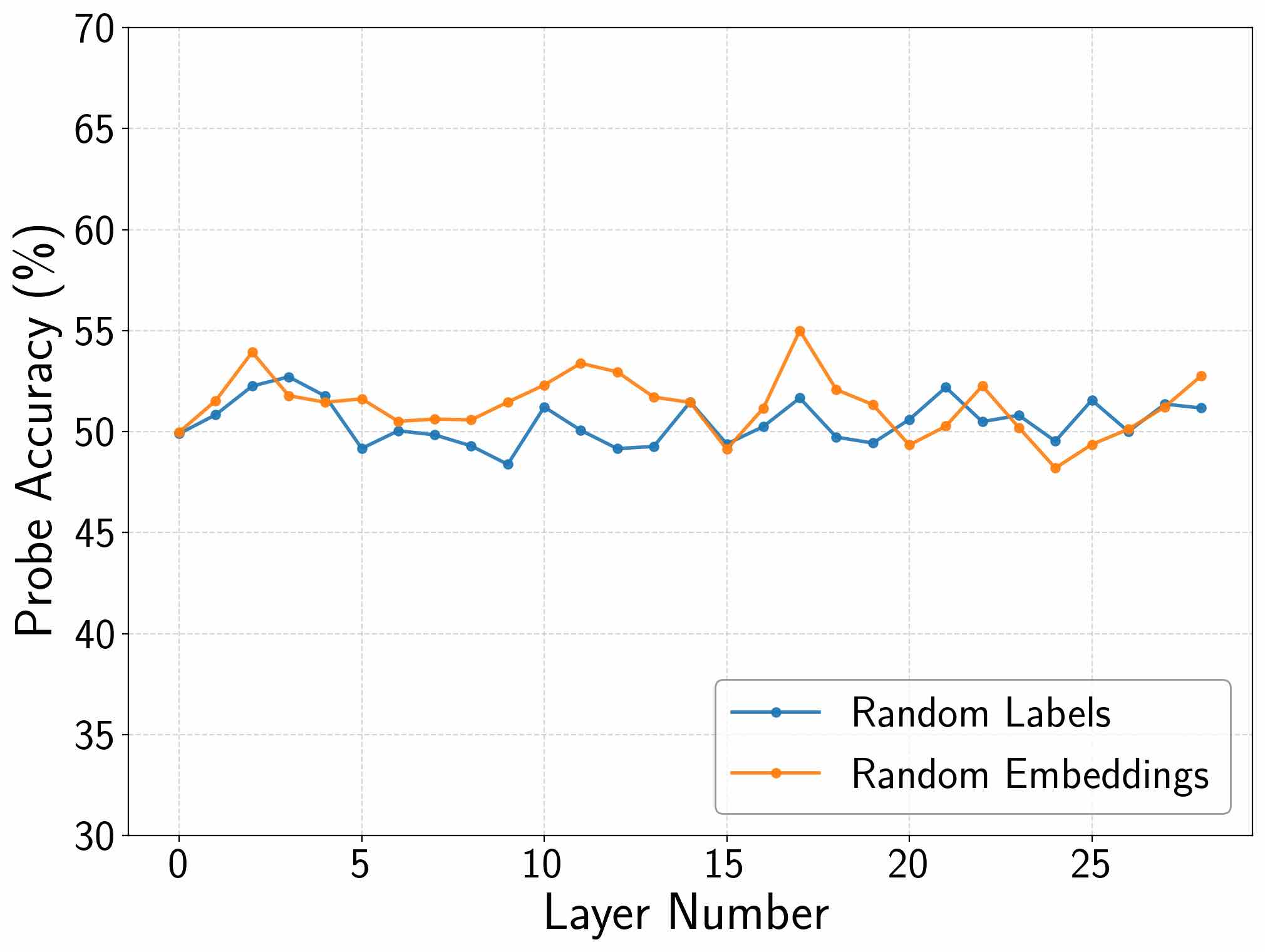}
    \caption{\textbf{Envy} probe accuracy across layers in \texttt{Qwen2.5-1.5B} using random embeddings or random labels during probe training}
    \label{fig:Envy_qwen1p5b_control}
\end{figure}

\begin{figure}[H]
    \centering
    \includegraphics[width=\linewidth]{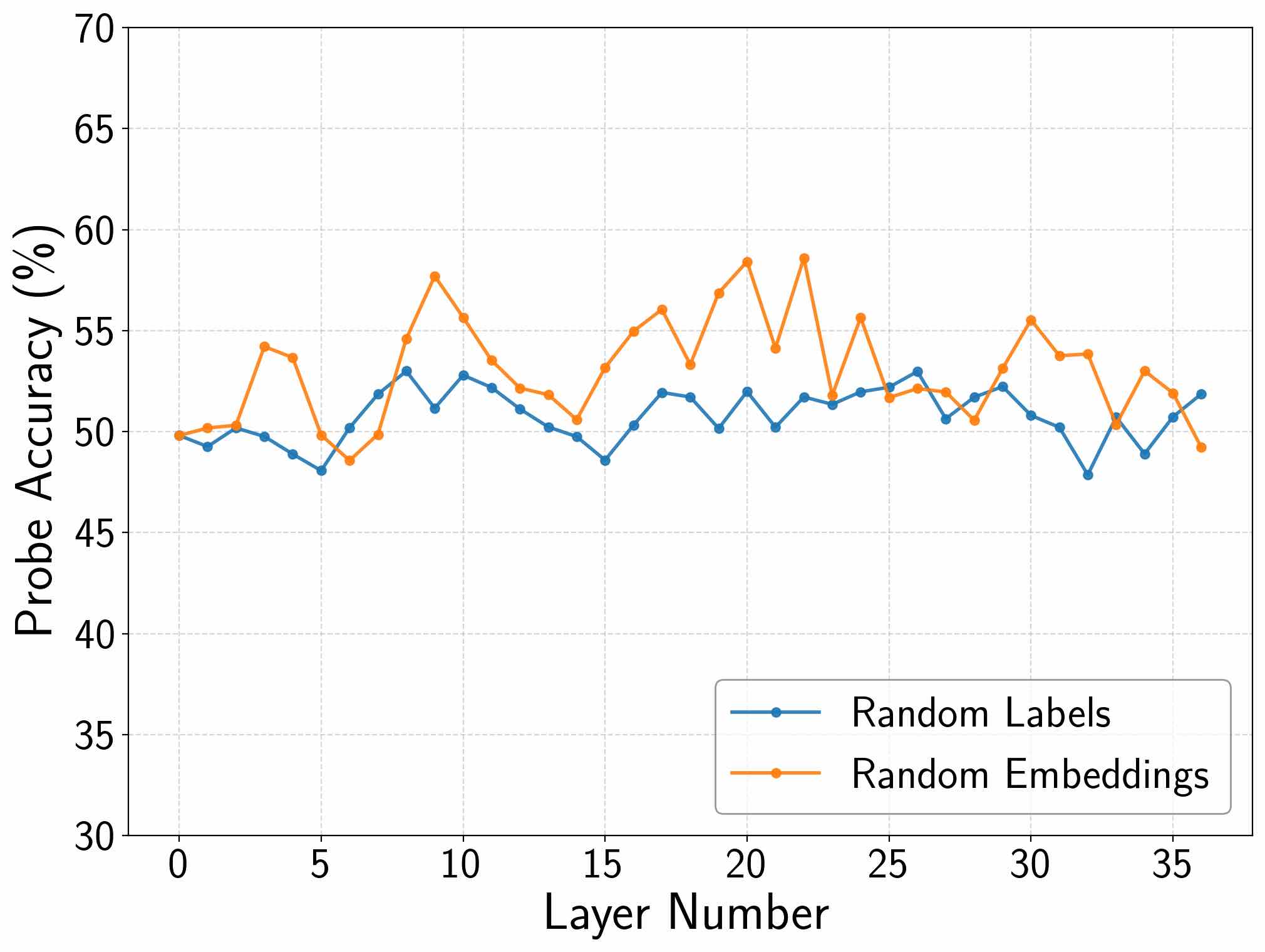}
    \caption{\textbf{Envy} probe accuracy across layers in \texttt{Qwen2.5-3B} using random embeddings or random labels during probe training}
    \label{fig:Envy_qwen3b_control}
\end{figure}

\begin{figure}[H]
    \centering
    \includegraphics[width=\linewidth]{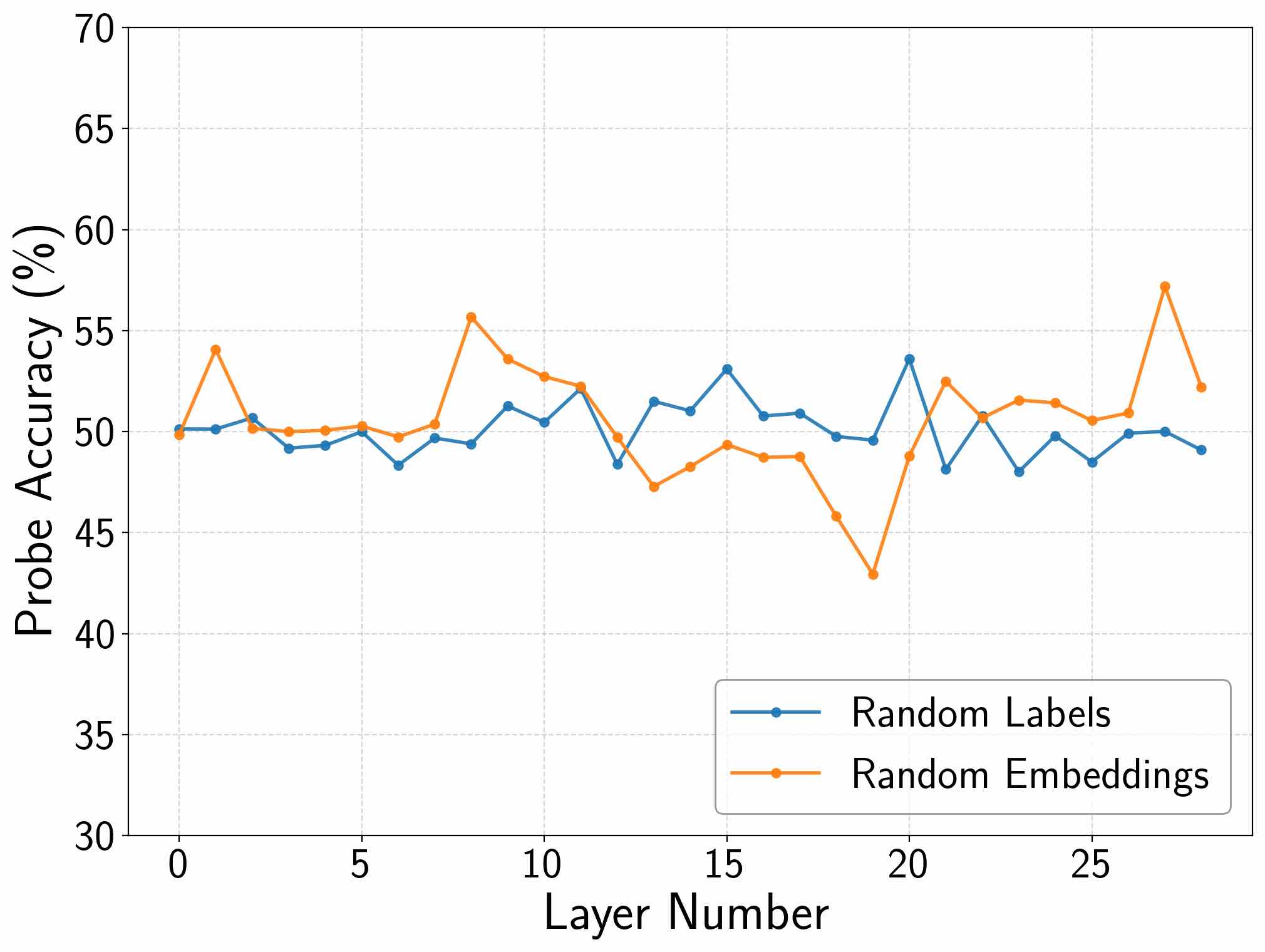}
    \caption{\textbf{Envy} probe accuracy across layers in \texttt{Qwen2.5-7B} using random embeddings or random labels during probe training}
    \label{fig:Envy_qwen7b_control}
\end{figure}

Figures \ref{fig:Envy_llama_vs_probe_params}, \ref{fig:Envy_gemma2b_vs_probe_params}, and \ref{fig:Envy_qwen0p5b_vs_probe_params} show the \textbf{Envy} probe accuracy versus probe size for \texttt{Llama-3-8B}, \texttt{Gemma-2-2B}, and \texttt{Qwen2.5-0.5B}, respectively, and Table \ref{table:Envy_vs_params} shows a summary of these results. Figures \ref{fig:Envy_meta3b_control}--\ref{fig:Envy_qwen7b_control} show the probe accuracies across layers for all LLMs when the probes are trained on the control task (randomizing embeddings or labels).

\begin{table}[H]
\centering
\begin{threeparttable}
\begin{tabular}{cc|cccc}
\hline\hline
\multirow{2}{*}{Probed LLM} & \multirow{2}{1cm}{Probed Layer} & \multicolumn{4}{c}{\# Probe parameters} \\
    & & 20 & 40 & 80 & max \\
\hline\hline

\multirow{3}{*}{\texttt{Llama-3-8B}} & 1 &	70 &	72 &	76 &	84 \\
& 15 &	74 &	82 &	87 &	95 \\
& 32 &	76 &	81 &	87 &	94 \\

\hline
                            
\multirow{3}{*}{\texttt{Gemma-2-2B}} & 1 &	67 &	70 &	74 &	82 \\
& 15 &	79 &	84 &	87 &	93 \\
& 26 &	74 &	79 &	86 &	92 \\

\hline

\multirow{3}{*}{\texttt{Qwen2.5-0.5B}} & 1 &	68 &	69 &	71 &	71 \\
& 15 &	75 &	79 &	84 &	90 \\
& 24 &	70 &	76 &	79 &	87 \\

\hline\hline
\end{tabular}
\begin{tablenotes}
\footnotesize
\item[$\bullet$] All results are in percentage (\%).
\item[$\bullet$] ``max'' denotes 4,096 for \texttt{Llama-3-8B}, 2,304 for \texttt{Gemma-2-2B}, and 896 for \texttt{Qwen2.5-0.5B}.
\item[$\bullet$] standard deviation for each result $\leq$ 1\%.
\end{tablenotes}
\end{threeparttable}
\caption{\textbf{Envy} probe accuracy across model families, sizes, layers, and probe sizes} 
\label{table:Envy_vs_params}
\end{table}
\section{Extended Results for Waxing and Waning of Concepts} \label{appendix:exp2_extended}
\setcounter{figure}{0}
\setcounter{table}{0}

\begin{figure}[H]
    \centering
    \includegraphics[width=\linewidth]{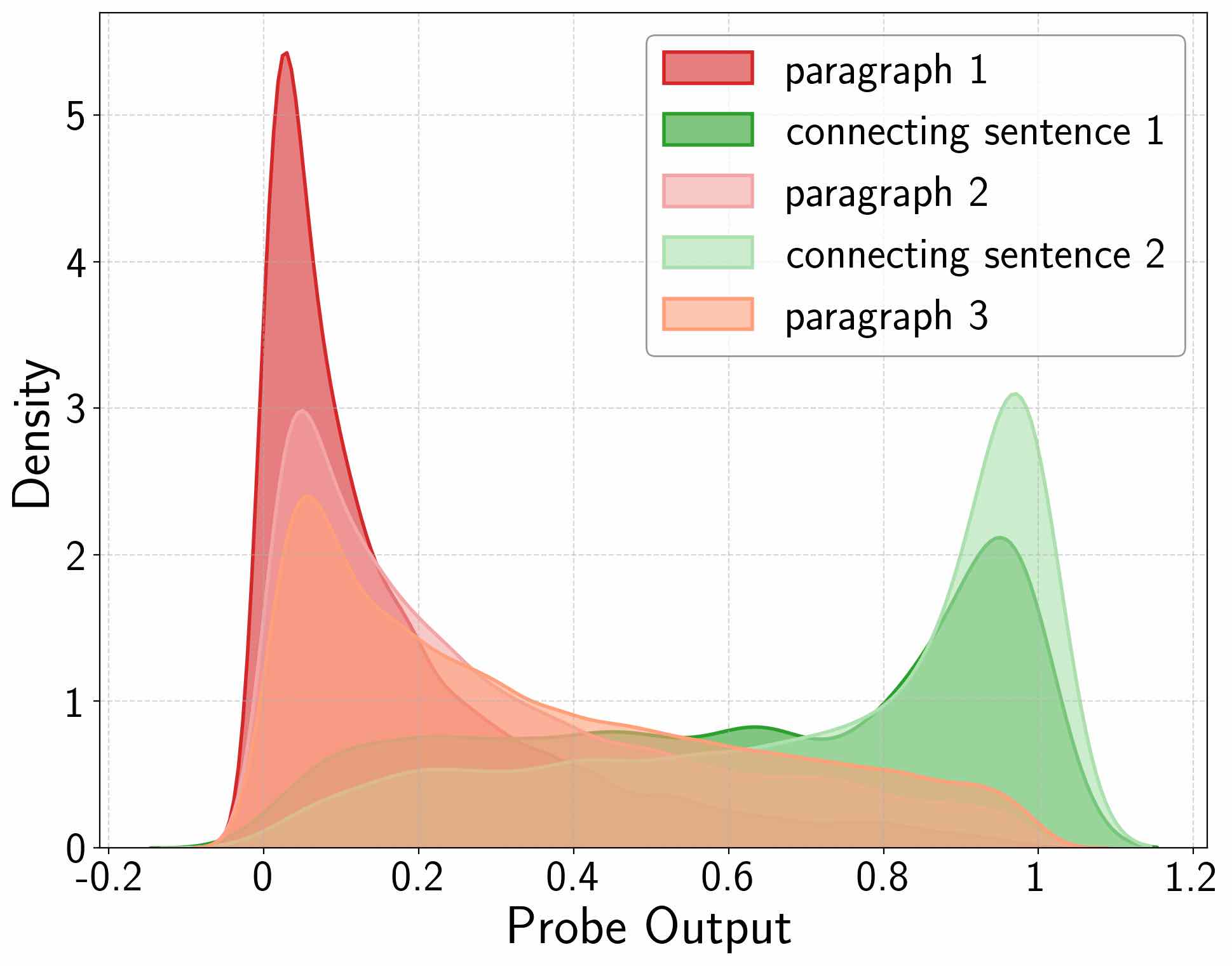}
    \caption{Kernel density estimation of the \textbf{Ambition} probe’s output values across story segments, aggregated over all stories, from layer 13 of \texttt{Llama-3-8B} (using final subword token embeddings)}
    \label{fig:kde_example}
\end{figure}

In addition to investigating whether concepts wax and wane in an LLM's embeddings as its context expands, we also studied how this behavior varies across layers to identify which layer best captures this change. For each layer, we generated kernel density estimation (KDE) plots showing the distribution of the probe’s output values for the target concept across different story segments (e.g., paragraph 1, transition sentence 1, paragraph 2), aggregated over all stories in the concept dataset. Figure \ref{fig:kde_example} shows an example for the \textbf{ambition} probe at layer 13 of \texttt{Llama-3-8B} (using the final subword token embedding), with KDEs for the transition sentences shown in shades of green and those for paragraphs in shades of red.

\begin{figure}[H]
    \centering
    \includegraphics[width=\linewidth]{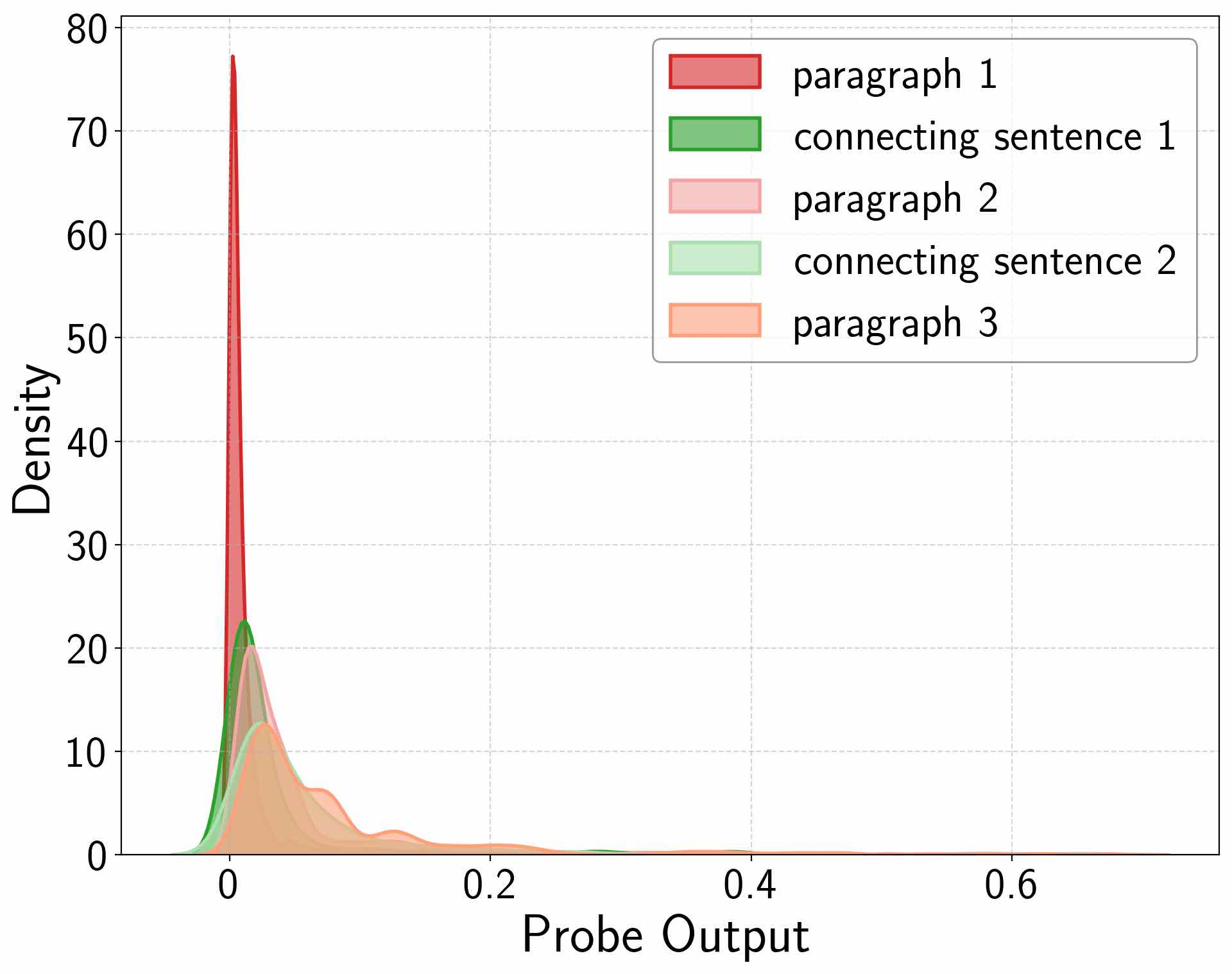}
    \caption{Kernel density estimation of the \textbf{Ambition} probe’s output values across story segments, aggregated over all stories, from layer 13 of \texttt{Llama-3-8B} (using cumulative mean embeddings)}
    \label{fig:kde_example_avg}
\end{figure}

A layer that accurately captures waxing and waning should show the transition KDEs concentrated above 0.5 and paragraph KDEs below 0.5. In contrast, a layer that does not capture the waxing and waning would either have all KDEs for paragraphs and transitions concentrated around the same point, as shown in Figure \ref{fig:kde_example_avg} (using cumulative mean embeddings), or have the KDEs for each segment widely distributed.

We stacked these plots for all LLM layers to identify the one that best tracks the waxing and waning, as shown ahead. Table \ref{table:prominence_best_layer} shows the layer in each LLM that best captures the waxing and waning for each investigated concept.

We use the probes for the corresponding layers to track the model embedding's word-level behavior. Using the story datasets, we obtained an aggregate view of each probe's outputs. This was computed as the average probe output for each word position across all 50 stories. For example, the aggregate value for word number 3 in sentence 15 is determined by the average of the corresponding probe outputs for all the words in that position across all stories. To obtain correct average values, all sentences at the same sentence position across stories must have the same number of words, which might not be the case. To alleviate this issue, we applied left-padding to each sentence to align its length with longest sentence at that position. The padded tokens were excluded from the averaging computation.

\begin{table*}[htbp]
\centering
\begin{tabular}{c|cccc}
\hline\hline
\multirow{2}{*}{Probed LLM} & \multicolumn{4}{c}{Layer that best captures concept waxing and waning} \\
 & Ambition & Investigation & Democracy & Envy \\
\hline\hline
\texttt{Llama-3-8B} & 13 & 31 & 7 & 10 \\
\texttt{Gemma-2-2B} & 12 & 6 & 16 & 14 \\
\texttt{Gemma-2-9B} & 22 & 11 & 11 & 19 \\
\texttt{Qwen2.5-0.5B} & 20 & 16 & 5 & 15 \\
\texttt{Qwen2.5-1.5B} & 15 & 18 & 14 & 20 \\
\texttt{Qwen2.5-3B} & 27 & 25 & 14 & 27 \\
\texttt{Qwen2.5-7B} & 19 & 12 & 12 & 24 \\
\hline\hline
\end{tabular}
\caption{LLM layers that best capture the waxing and waning of the investigated concepts}
\label{table:prominence_best_layer}
\end{table*}

\subsection{Tracking Waxing and Waning of Ambition}

\subsubsection{Layer-Wise KDEs for Ambition Probe Outputs}

\begin{figure}[H]
    \centering
    \includegraphics[width=\linewidth]{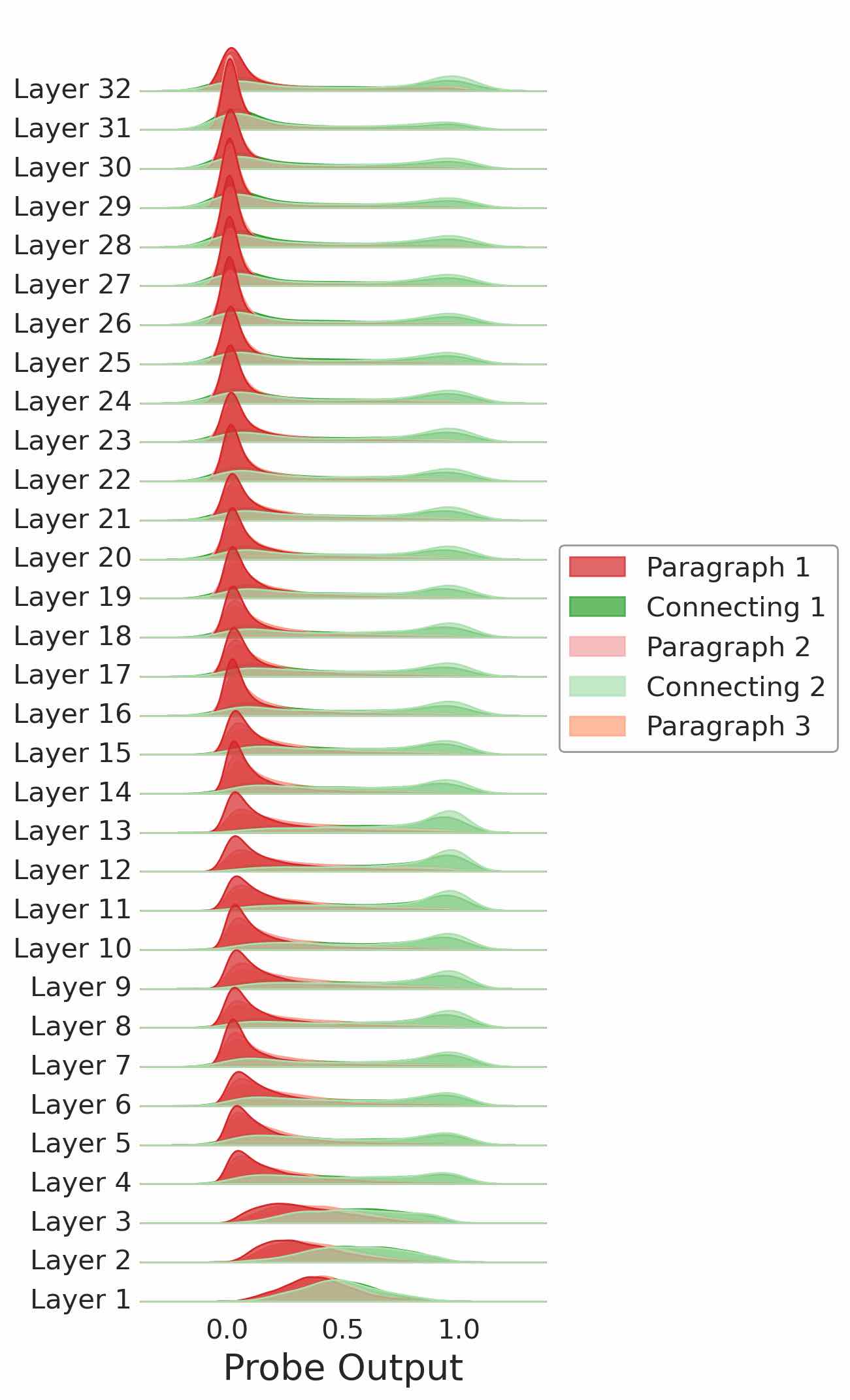}
    \caption{Layer-wise KDEs for \textbf{ambition} probe outputs in \texttt{Llama-3-8B}}
    \label{fig:ambition_KDE_llama}
\end{figure}

\begin{figure}[H]
    \centering
    \includegraphics[width=\linewidth]{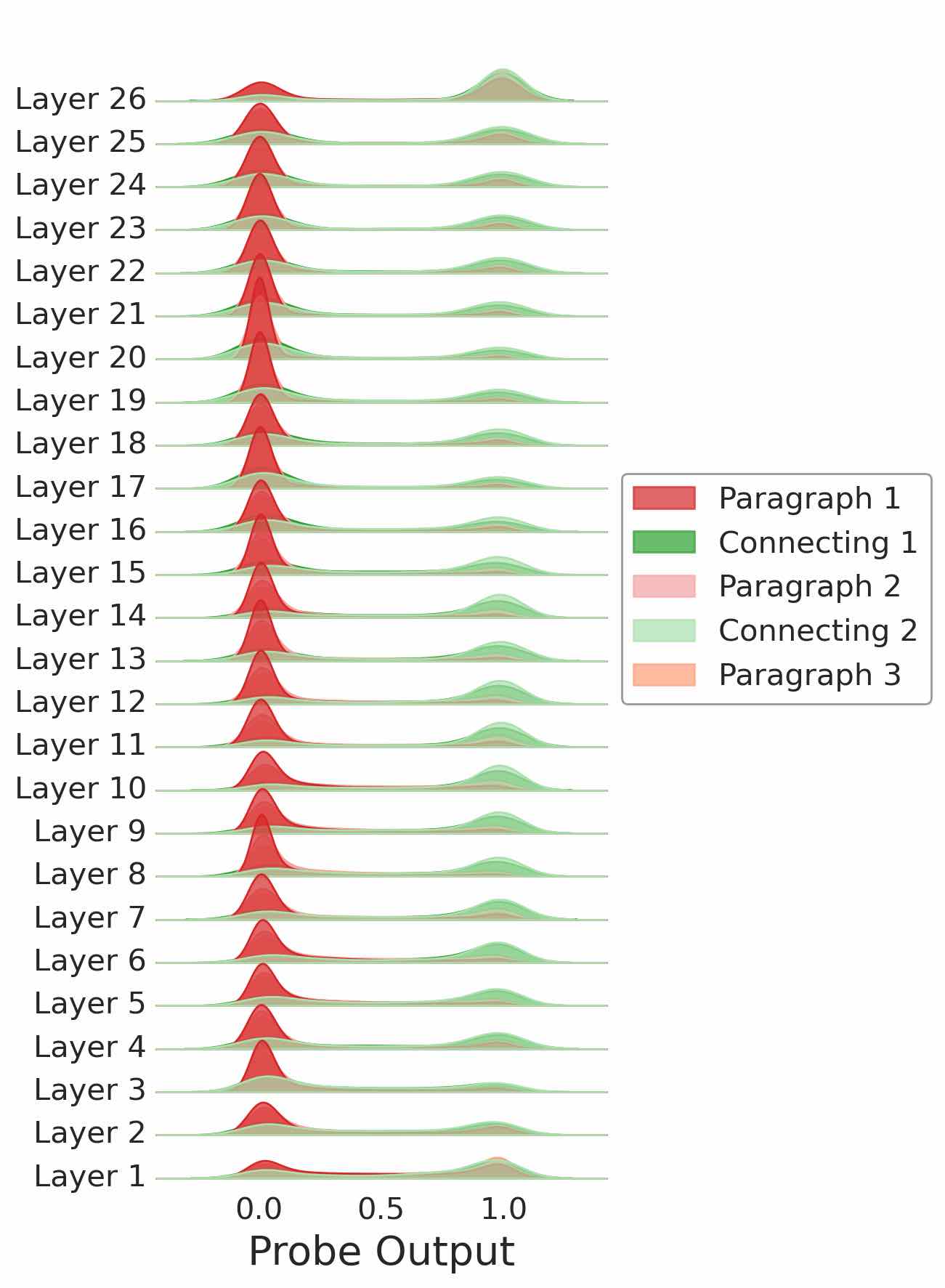}
    \caption{Layer-wise KDEs for \textbf{ambition} probe outputs in \texttt{Gemma-2-2B}}
    \label{fig:ambition_KDE_gemma2b}
\end{figure}

\begin{figure}[H]
    \centering
    \includegraphics[width=\linewidth]{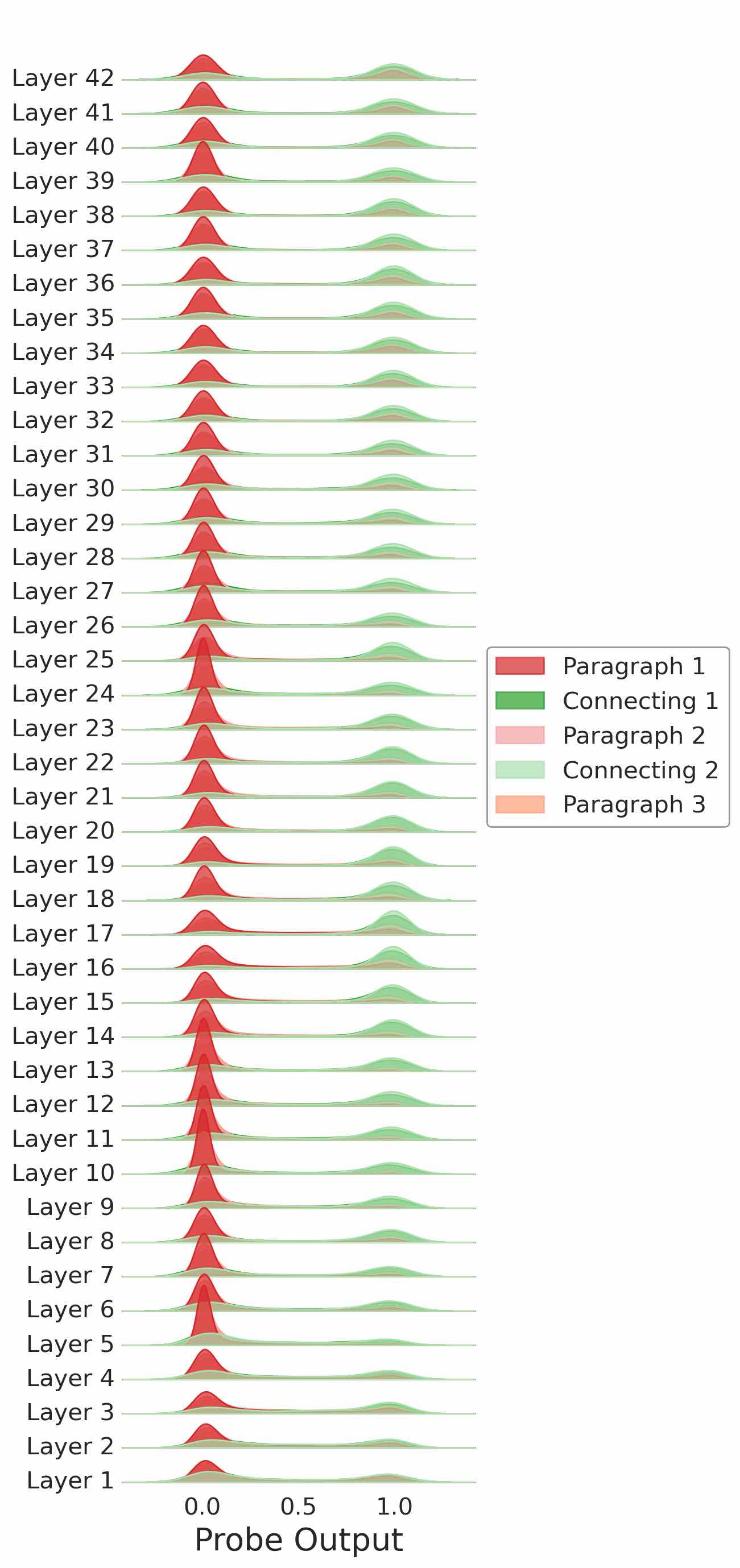}
    \caption{Layer-wise KDEs for \textbf{ambition} probe outputs in \texttt{Gemma-2-9B}}
    \label{fig:ambition_KDE_gemma9b}
\end{figure}

\begin{figure}[H]
    \centering
    \includegraphics[width=\linewidth]{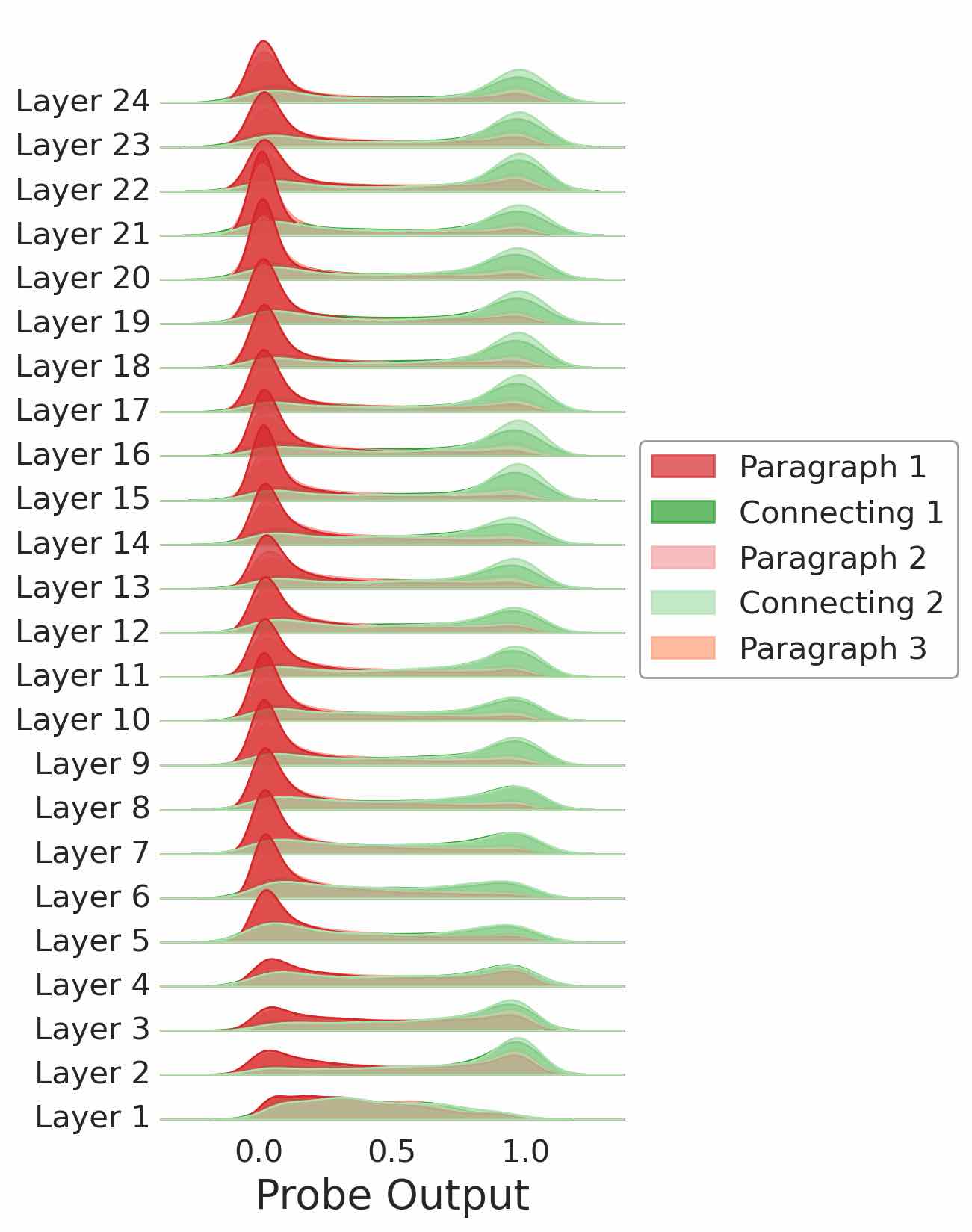}
    \caption{Layer-wise KDEs for \textbf{ambition} probe outputs in \texttt{Qwen2.5-0.5B}}
    \label{fig:ambition_KDE_qwen0p5b}
\end{figure}

\begin{figure}[H]
    \centering
    \includegraphics[width=\linewidth]{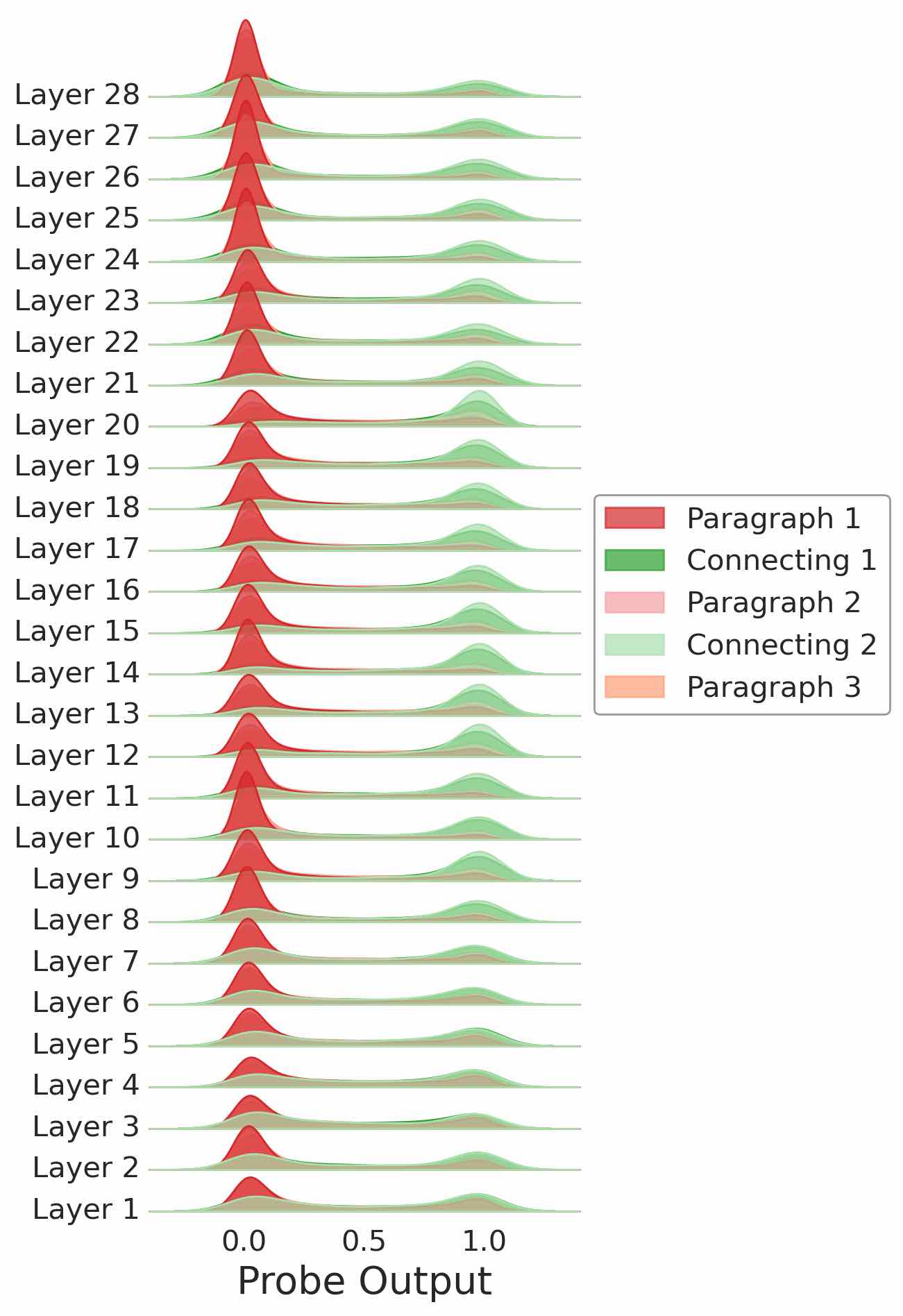}
    \caption{Layer-wise KDEs for \textbf{ambition} probe outputs in \texttt{Qwen2.5-1.5B}}
    \label{fig:ambition_KDE_qwen1p5b}
\end{figure}

\begin{figure}[H]
    \centering
    \includegraphics[width=\linewidth]{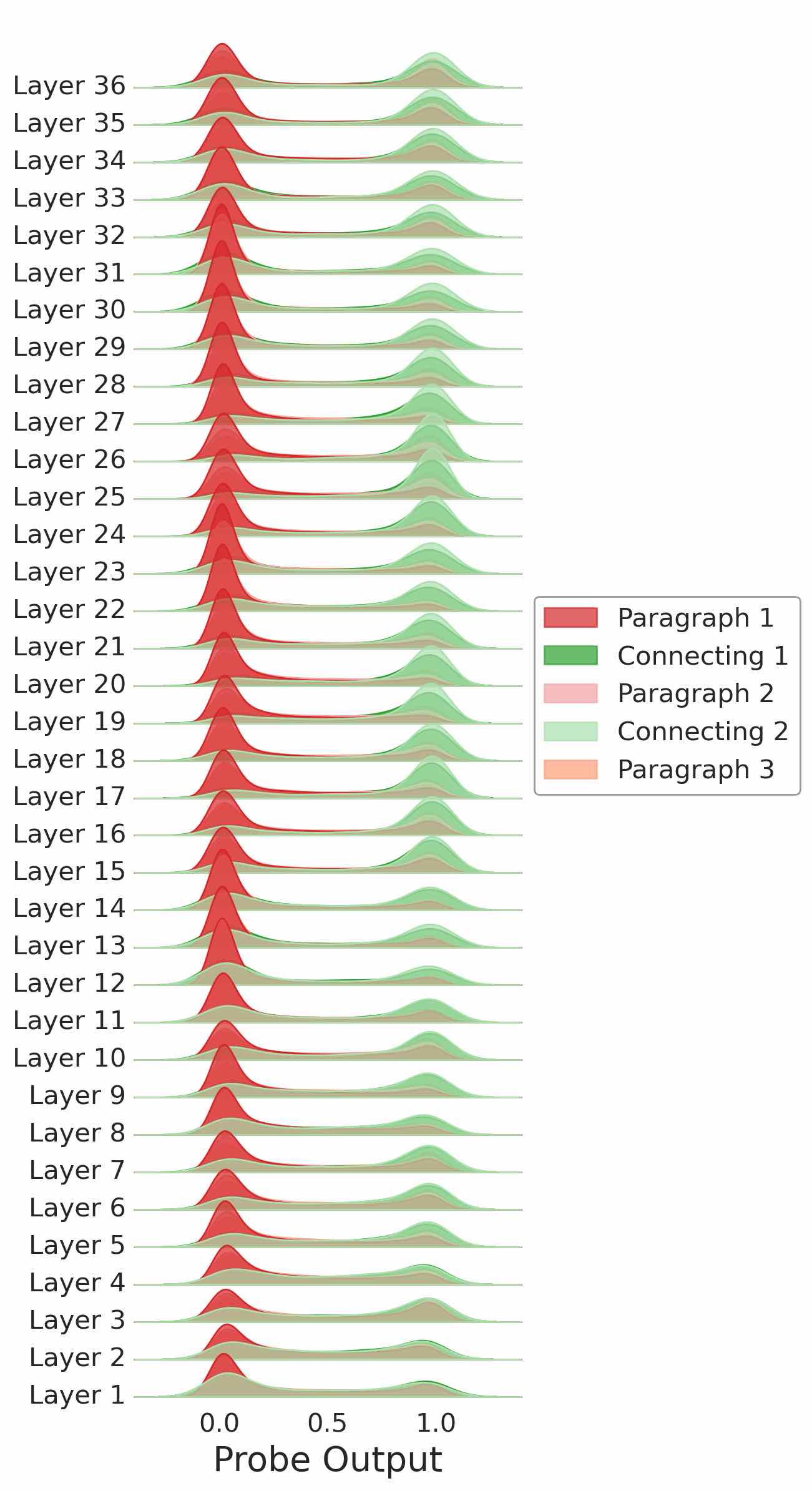}
    \caption{Layer-wise KDEs for \textbf{ambition} probe outputs in \texttt{Qwen2.5-3B}}
    \label{fig:ambition_KDE_qwen3b}
\end{figure}

\begin{figure}[H]
    \centering
    \includegraphics[width=\linewidth]{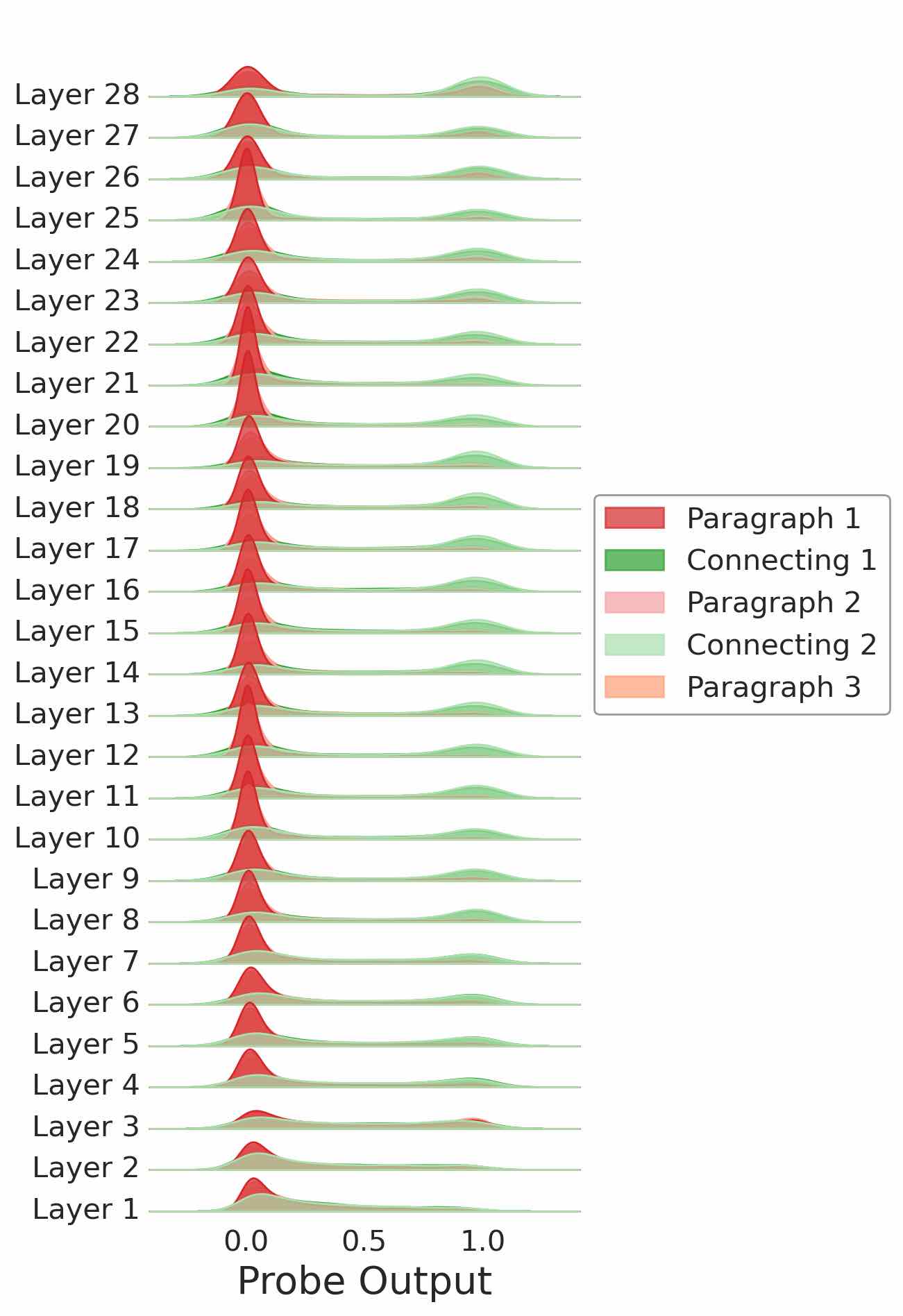}
    \caption{Layer-wise KDEs for \textbf{ambition} probe outputs in \texttt{Qwen2.5-7B}}
    \label{fig:ambition_KDE_qwen7b}
\end{figure}

\subsubsection{Ambition Probe Results for Best Layers}

\begin{figure*}[h]
    \centering
    \includegraphics[width=\linewidth]{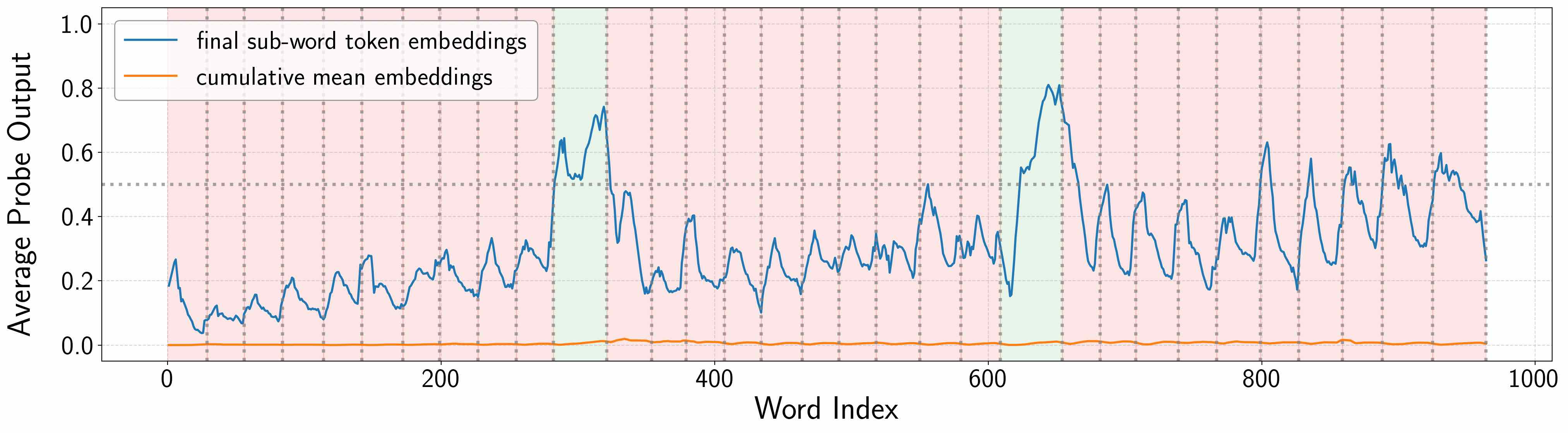}
    \caption{\textbf{Ambition} probe outputs across words using both representative embeddings in \texttt{Gemma-2-2B}}
    \label{fig:ambition_prom_gemma2b_last_aggregate}
\end{figure*}

\begin{figure*}[h]
    \centering
    \includegraphics[width=\linewidth]{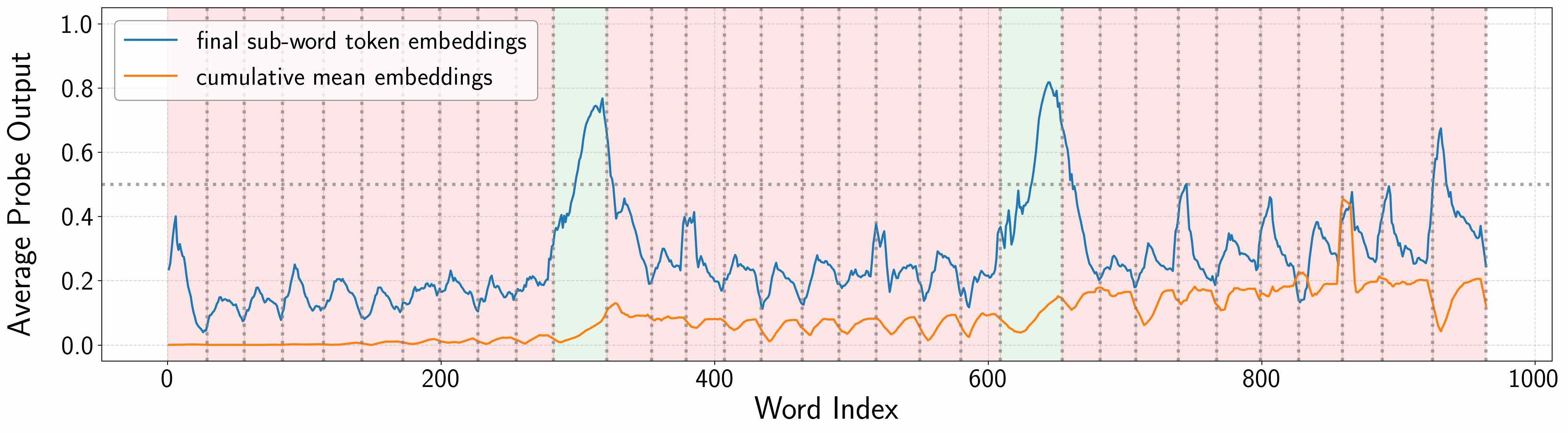}
    \caption{\textbf{Ambition} probe outputs across words using both representative embeddings in \texttt{Gemma-2-9B}}
    \label{fig:ambition_prom_gemma9b_last_aggregate}
\end{figure*}

\begin{figure*}[h]
    \centering
    \includegraphics[width=\linewidth]{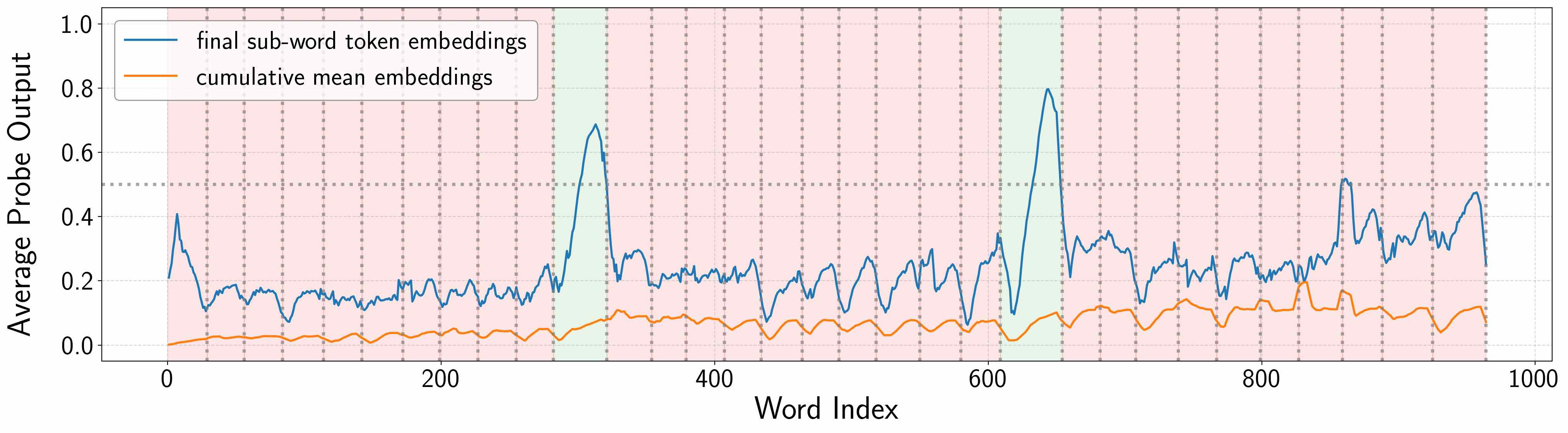}
    \caption{\textbf{Ambition} probe outputs across words using both representative embeddings in \texttt{Qwen2.5-0.5B}}
    \label{fig:ambition_prom_qwen0p5b_last_aggregate}
\end{figure*}

\begin{figure*}[h]
    \centering
    \includegraphics[width=\linewidth]{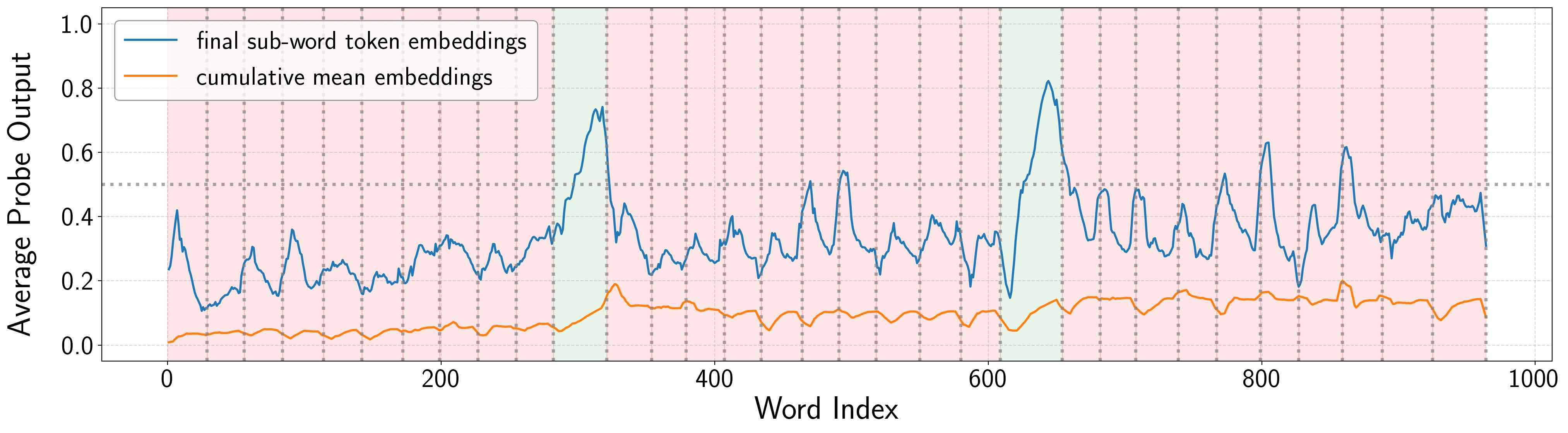}
    \caption{\textbf{Ambition} probe outputs across words using both representative embeddings in \texttt{Qwen2.5-1.5B}}
    \label{fig:ambition_prom_qwen1p5b_last_aggregate}
\end{figure*}

\begin{figure*}[h]
    \centering
    \includegraphics[width=\linewidth]{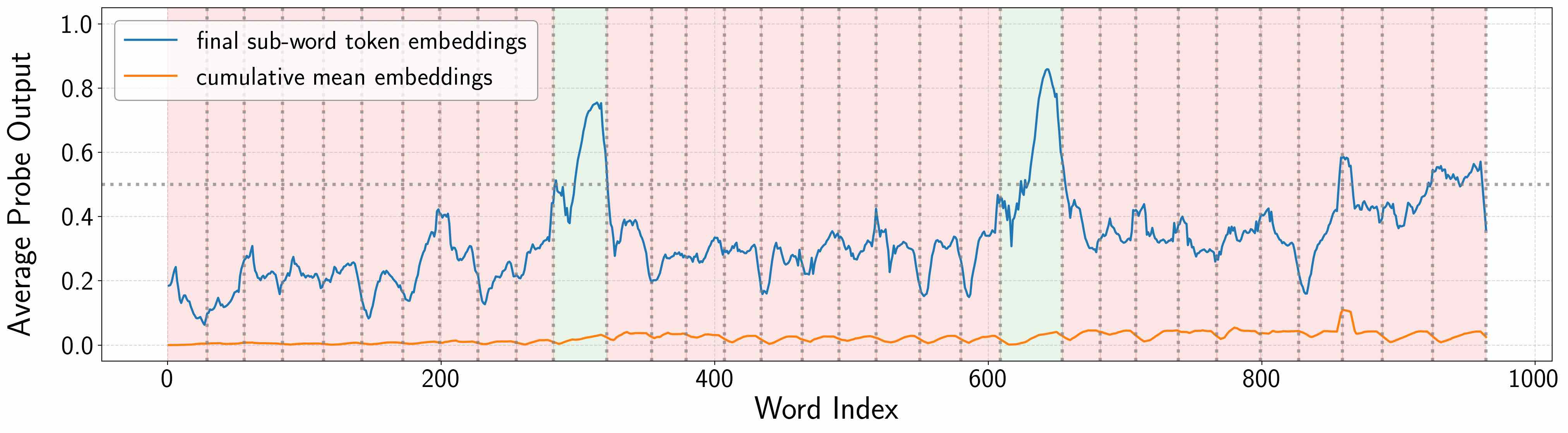}
    \caption{\textbf{Ambition} probe outputs across words using both representative embeddings in \texttt{Qwen2.5-3B}}
    \label{fig:ambition_prom_qwen3b_last_aggregate}
\end{figure*}

\begin{figure*}[h]
    \centering
    \includegraphics[width=\linewidth]{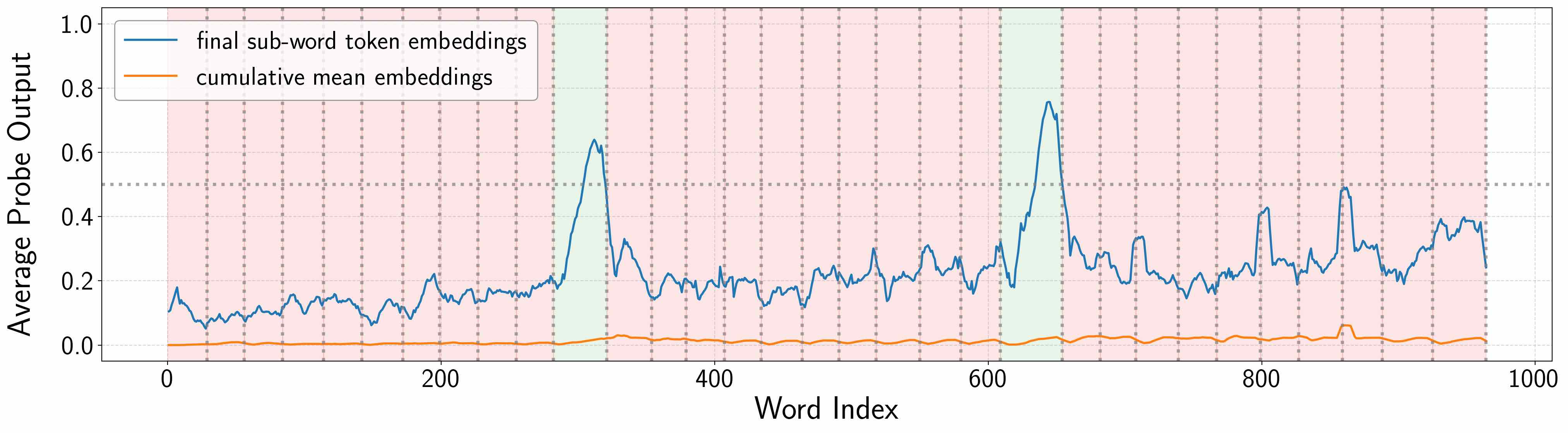}
    \caption{\textbf{Ambition} probe outputs across words using both representative embeddings in \texttt{Qwen2.5-7B}}
    \label{fig:ambition_prom_qwen7b_last_aggregate}
\end{figure*}

\FloatBarrier
\clearpage

\subsection{Tracking Waxing and Waning of Investigation}

\subsubsection{Layer-Wise KDEs for Investigation Probe Outputs}

\begin{figure}[H]
    \centering
    \includegraphics[width=\linewidth]{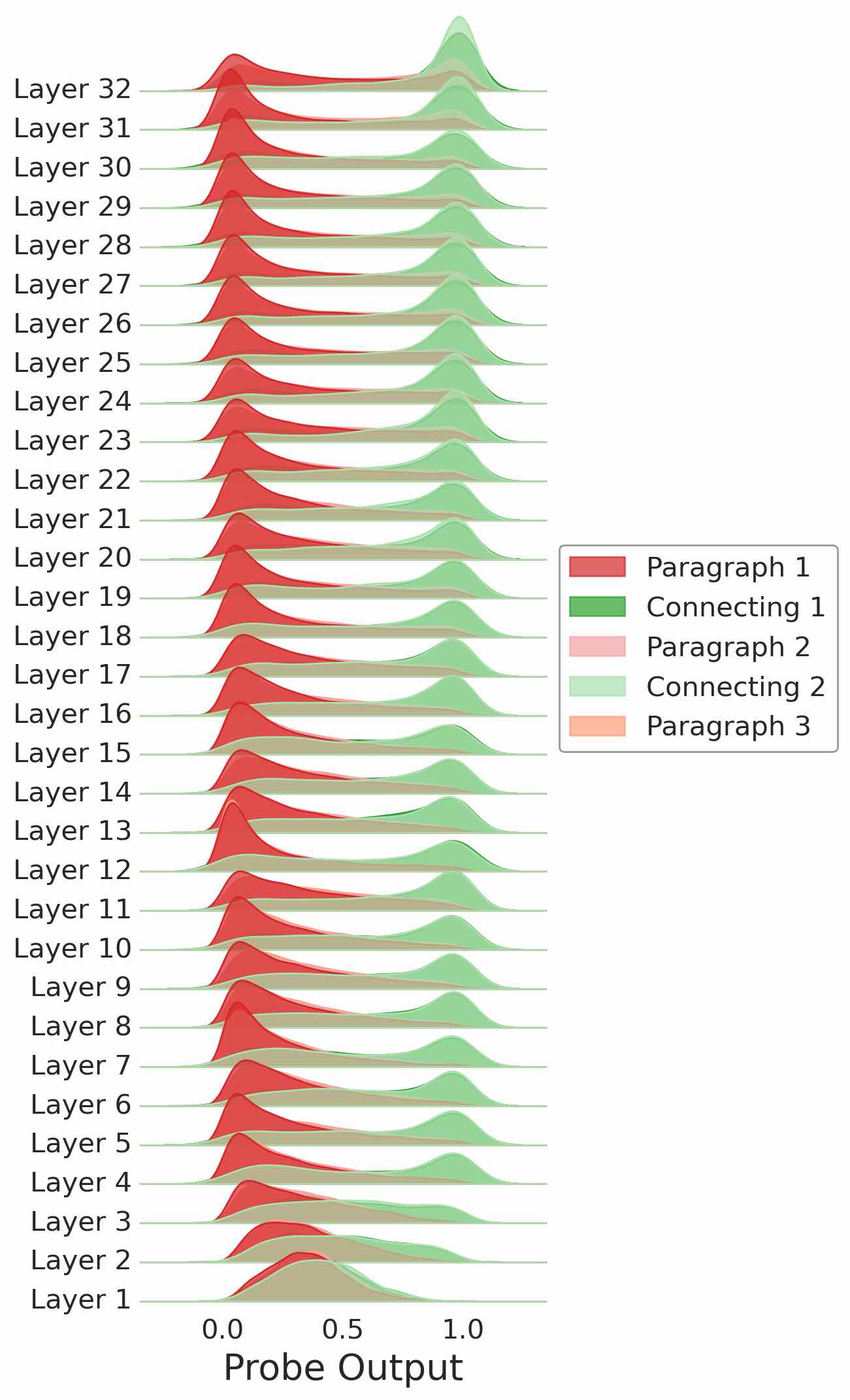}
    \caption{Layer-wise KDEs for \textbf{investigation} probe outputs in \texttt{Llama-3-8B}}
    \label{fig:investigation_KDE_llama}
\end{figure}

\begin{figure}[H]
    \centering
    \includegraphics[width=\linewidth]{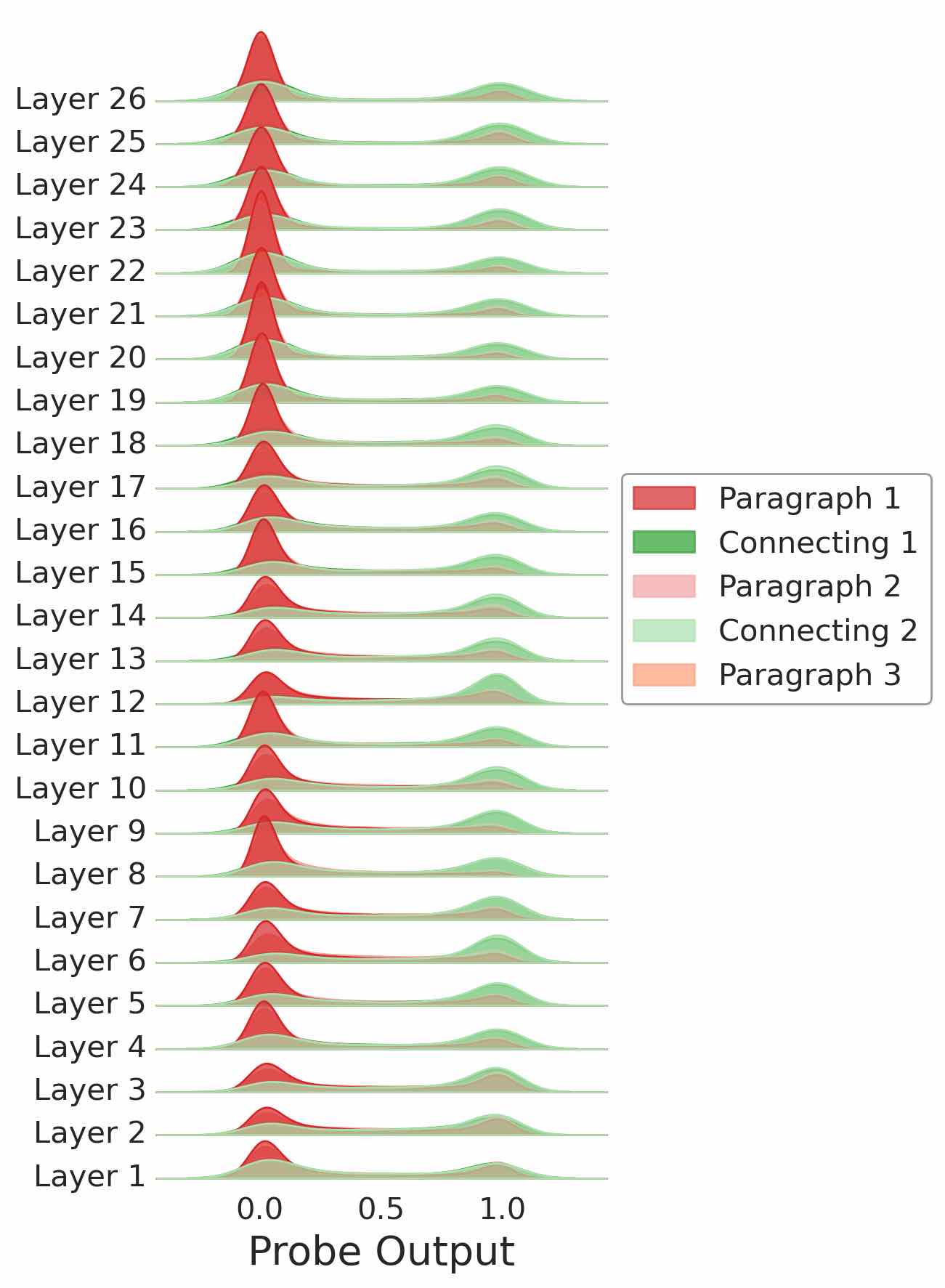}
    \caption{Layer-wise KDEs for \textbf{investigation} probe outputs in \texttt{Gemma-2-2B}}
    \label{fig:investigation_KDE_gemma2b}
\end{figure}

\begin{figure}[H]
    \centering
    \includegraphics[width=\linewidth]{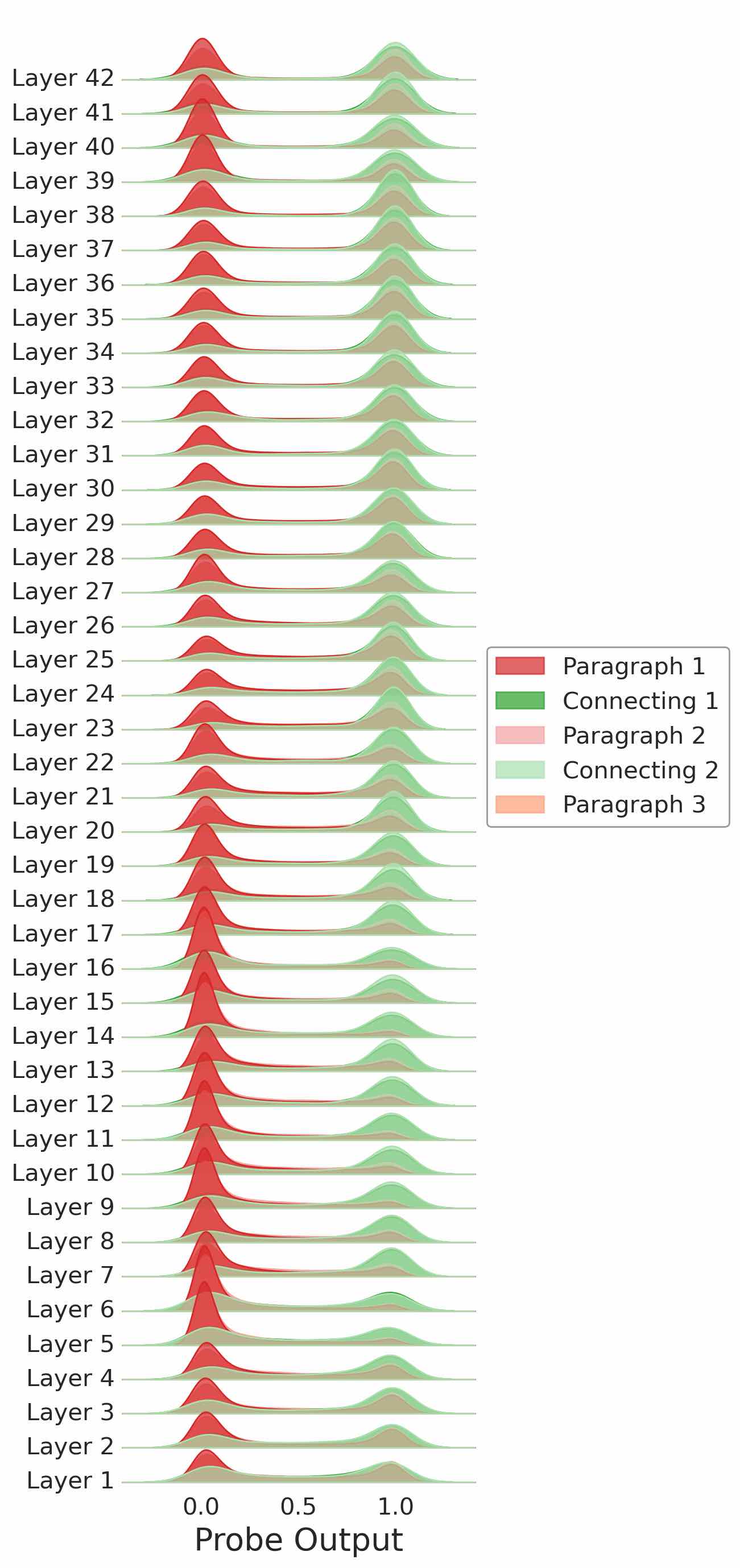}
    \caption{Layer-wise KDEs for \textbf{investigation} probe outputs in \texttt{Gemma-2-9B}}
    \label{fig:investigation_KDE_gemma9b}
\end{figure}

\begin{figure}[H]
    \centering
    \includegraphics[width=\linewidth]{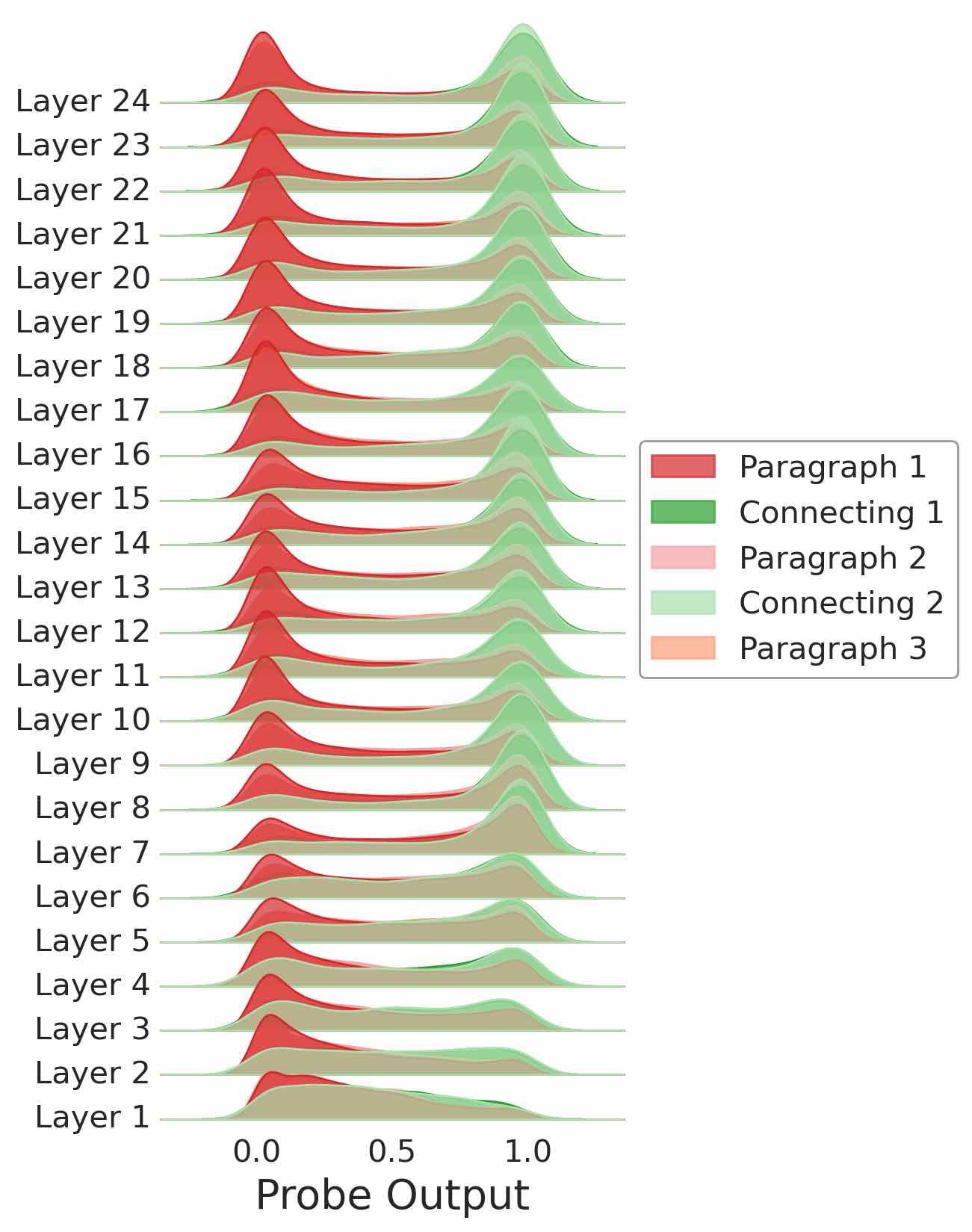}
    \caption{Layer-wise KDEs for \textbf{investigation} probe outputs in \texttt{Qwen2.5-0.5B}}
    \label{fig:investigation_KDE_qwen0p5b}
\end{figure}

\begin{figure}[H]
    \centering
    \includegraphics[width=\linewidth]{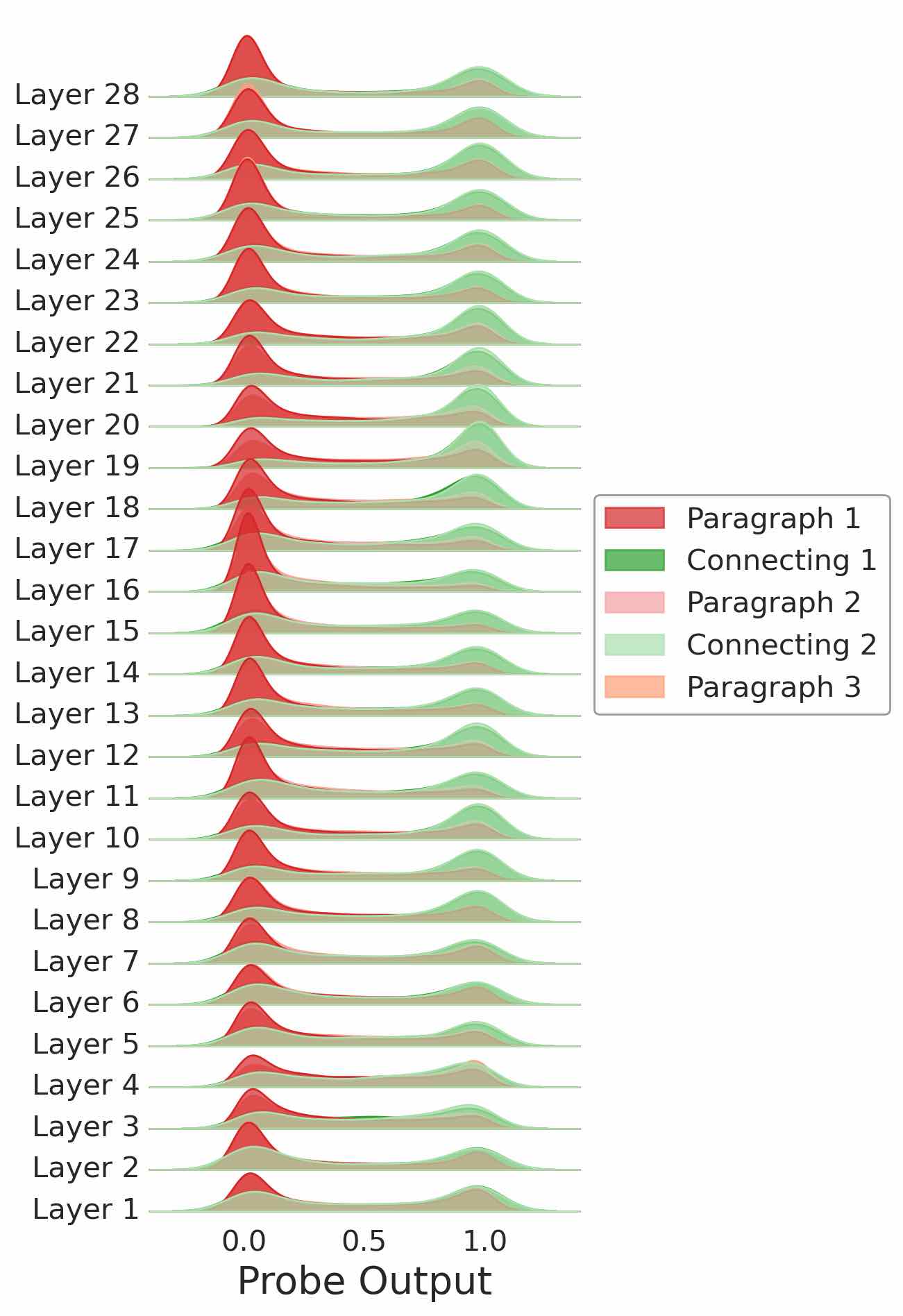}
    \caption{Layer-wise KDEs for \textbf{investigation} probe outputs in \texttt{Qwen2.5-1.5B}}
    \label{fig:investigation_KDE_qwen1p5b}
\end{figure}

\begin{figure}[H]
    \centering
    \includegraphics[width=\linewidth]{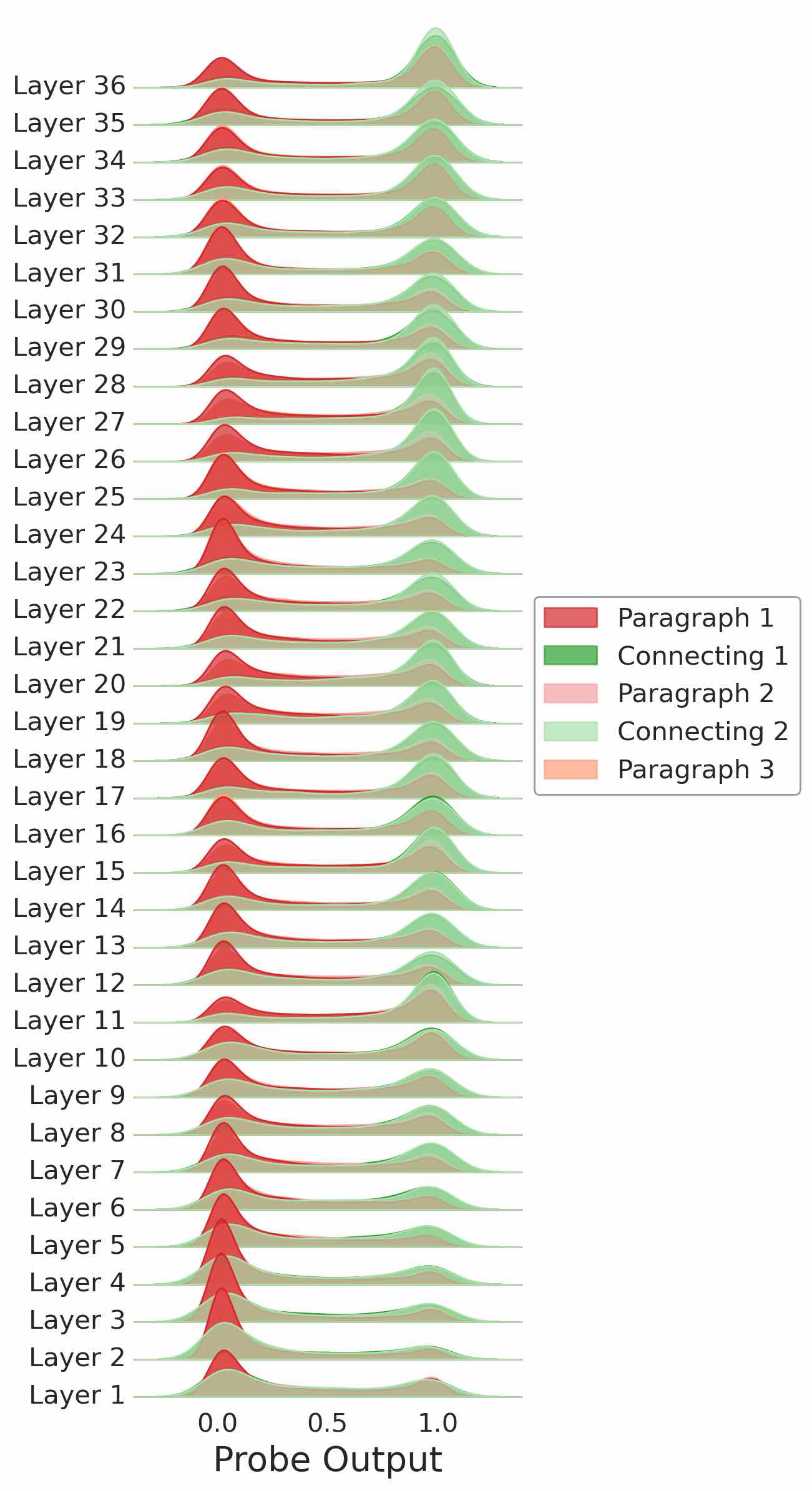}
    \caption{Layer-wise KDEs for \textbf{investigation} probe outputs in \texttt{Qwen2.5-3B}}
    \label{fig:investigation_KDE_qwen3b}
\end{figure}

\begin{figure}[H]
    \centering
    \includegraphics[width=\linewidth]{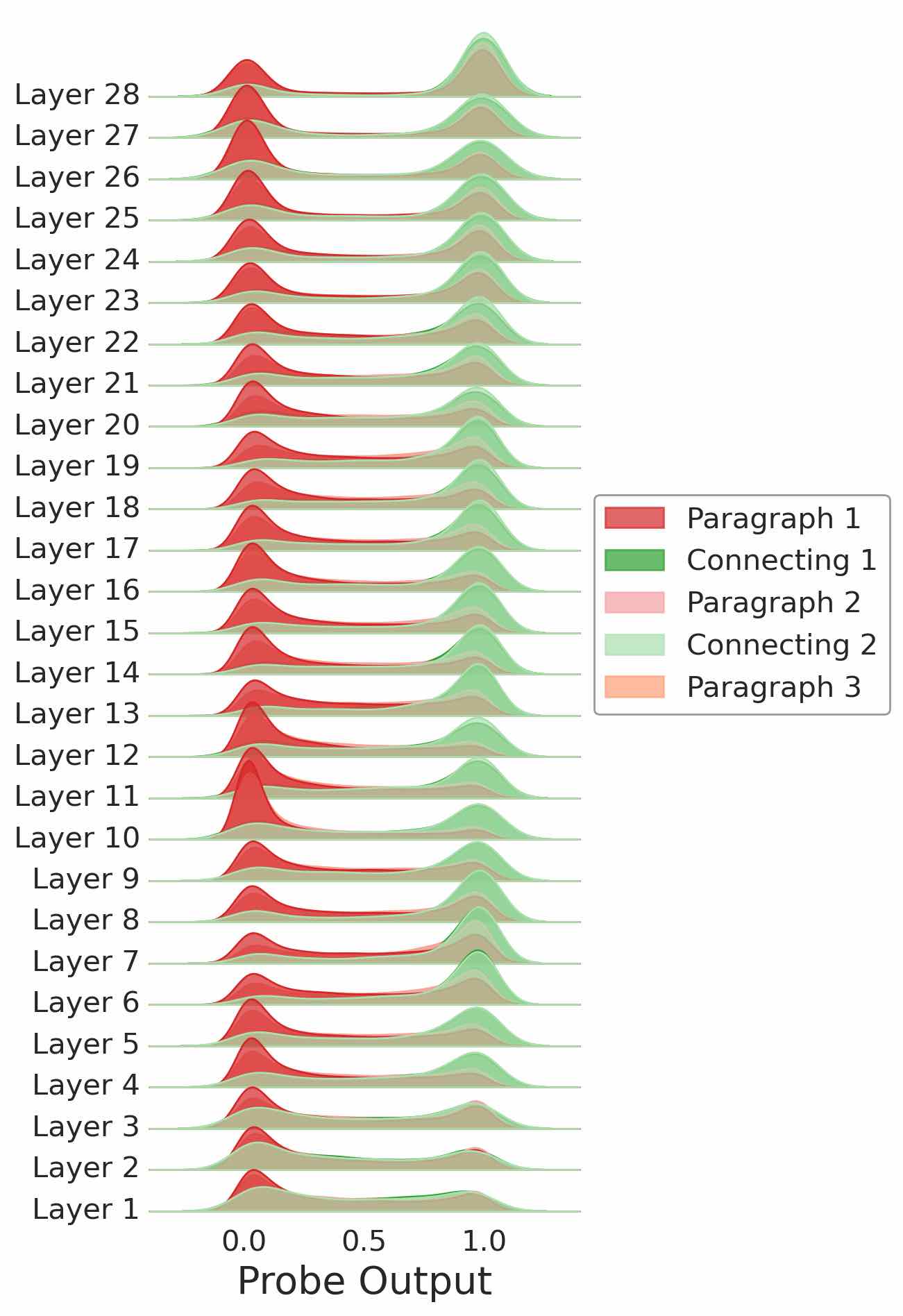}
    \caption{Layer-wise KDEs for \textbf{investigation} probe outputs in \texttt{Qwen2.5-7B}}
    \label{fig:investigation_KDE_qwen7b}
\end{figure}

\subsubsection{Investigation Probe Results for Best Layers}

\begin{figure*}[h]
    \centering
    \includegraphics[width=\linewidth]{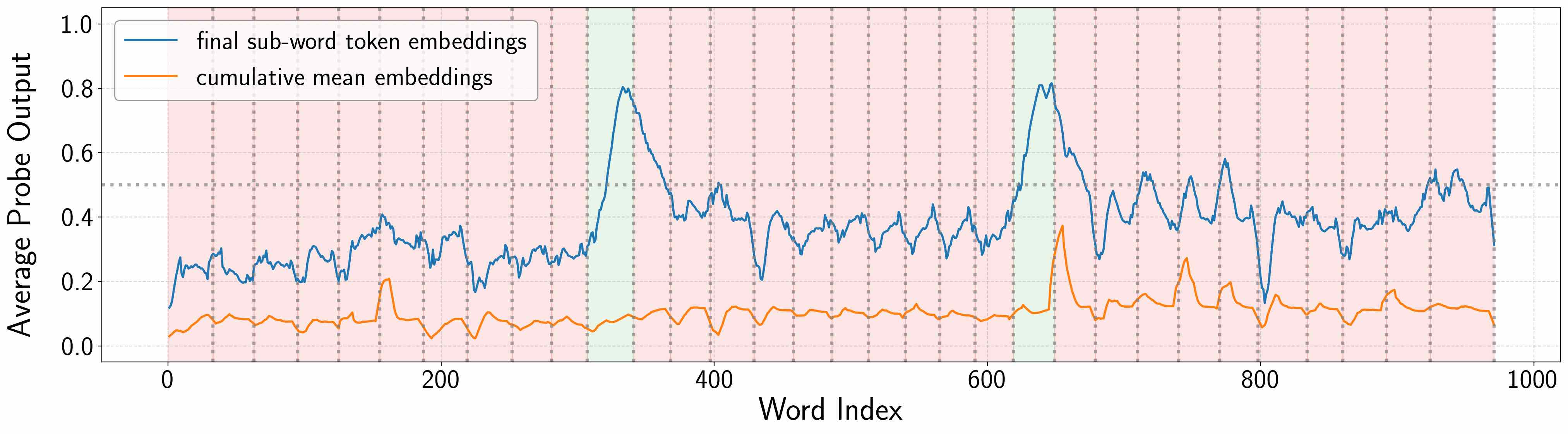}
    \caption{\textbf{Investigation} probe outputs across words using both representative embeddings in \texttt{Llama-3-8B}}
    \label{fig:Investigation_prom_llama_last_aggregate}
\end{figure*}

\begin{figure*}[h]
    \centering
    \includegraphics[width=\linewidth]{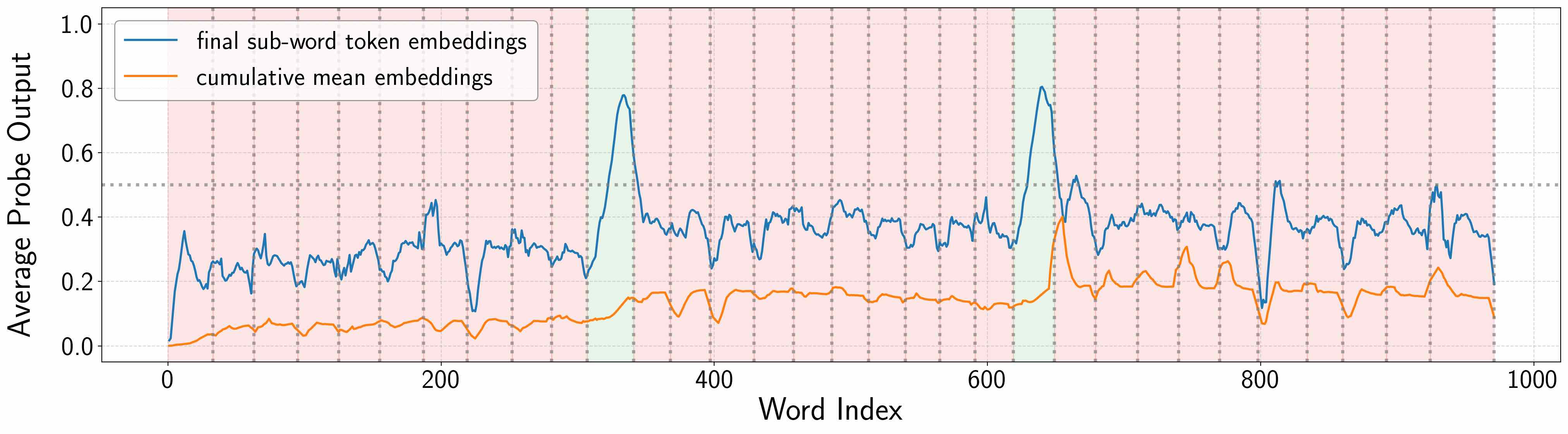}
    \caption{\textbf{Investigation} probe outputs across words using both representative embeddings in \texttt{Gemma-2-2B}}
    \label{fig:Investigation_prom_gemma2b_last_aggregate}
\end{figure*}

\begin{figure*}[h]
    \centering
    \includegraphics[width=\linewidth]{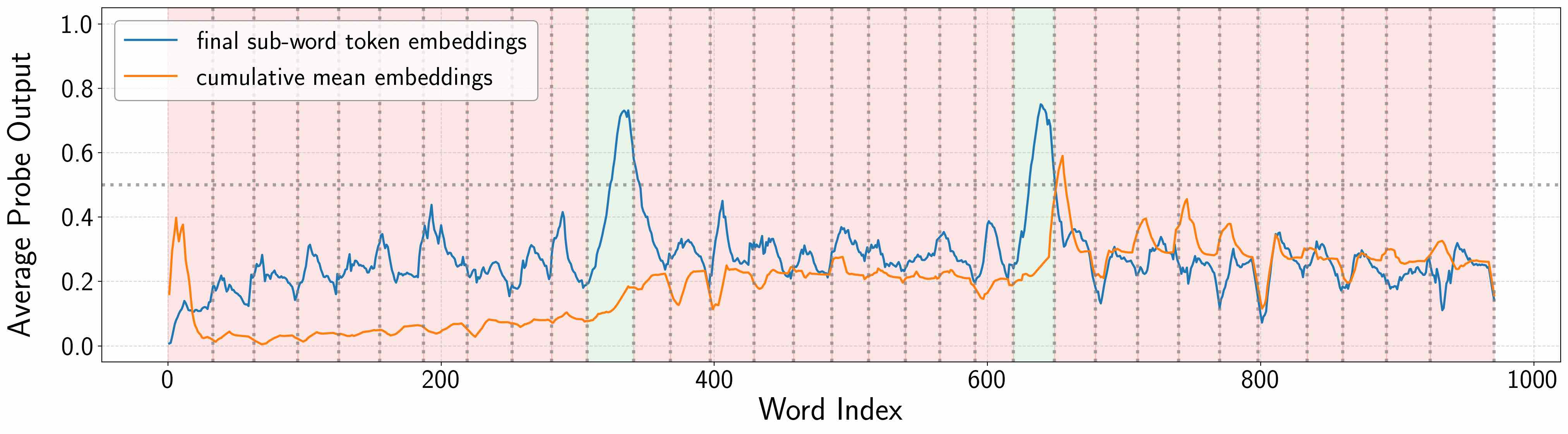}
    \caption{\textbf{Investigation} probe outputs across words using both representative embeddings in \texttt{Gemma-2-9B}}
    \label{fig:Investigation_prom_gemma9b_last_aggregate}
\end{figure*}

\begin{figure*}[h]
    \centering
    \includegraphics[width=\linewidth]{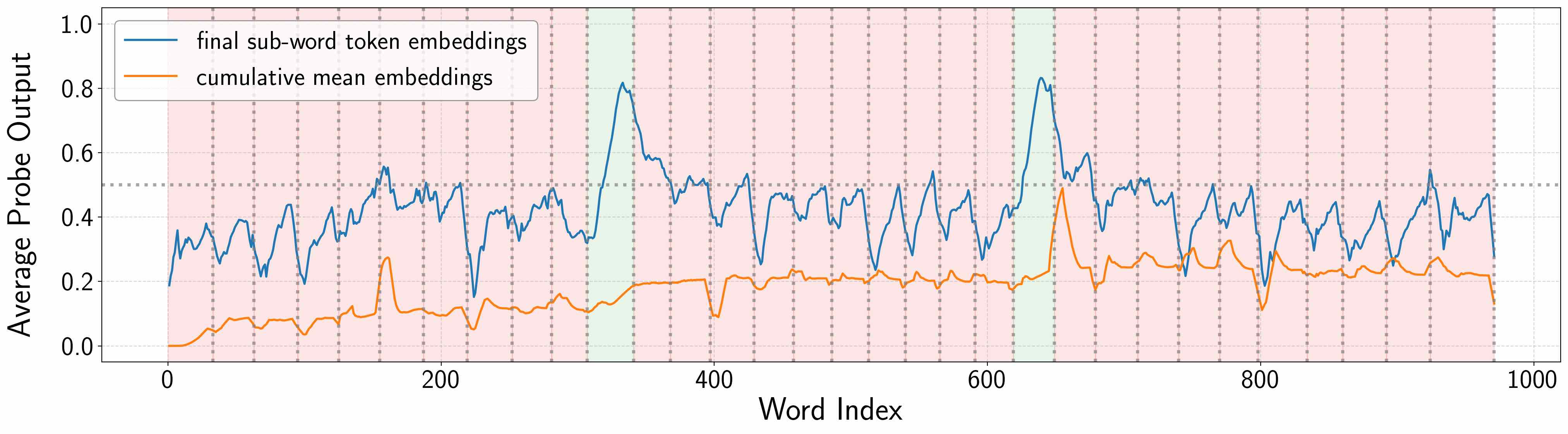}
    \caption{\textbf{Investigation} probe outputs across words using both representative embeddings in \texttt{Qwen2.5-0.5B}}
    \label{fig:Investigation_prom_qwen0p5b_last_aggregate}
\end{figure*}

\begin{figure*}[h]
    \centering
    \includegraphics[width=\linewidth]{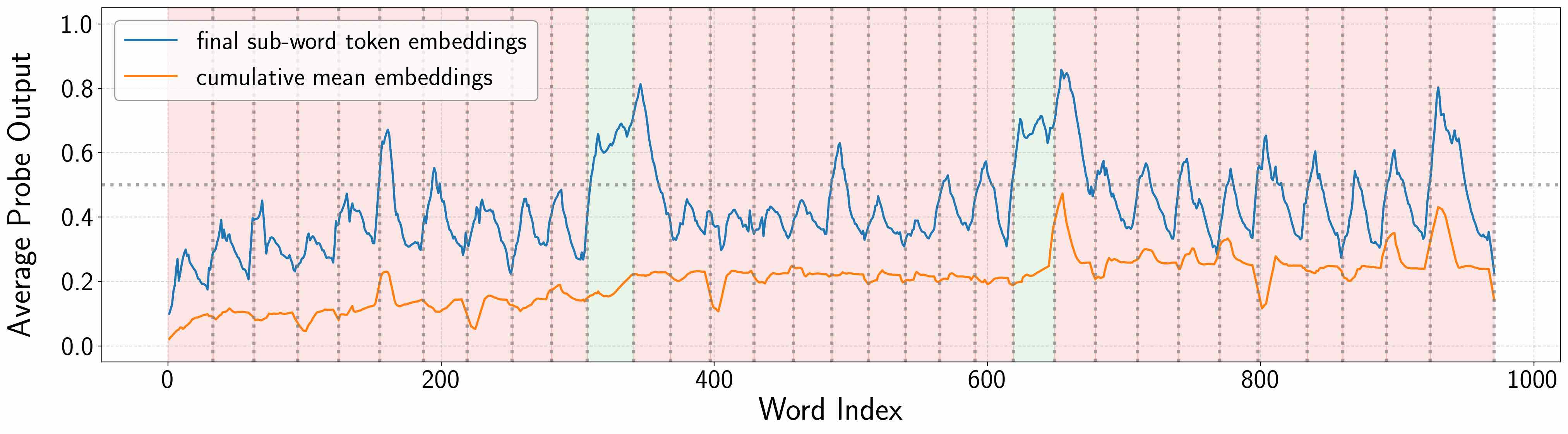}
    \caption{\textbf{Investigation} probe outputs across words using both representative embeddings in \texttt{Qwen2.5-1.5B}}
    \label{fig:Investigation_prom_qwen1p5b_last_aggregate}
\end{figure*}

\begin{figure*}[h]
    \centering
    \includegraphics[width=\linewidth]{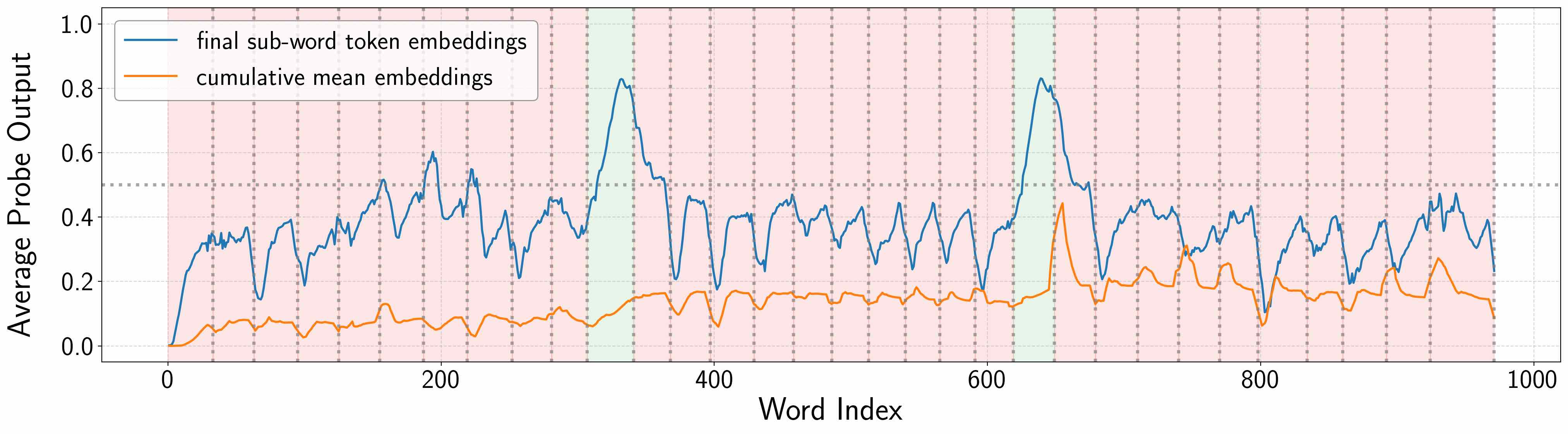}
    \caption{\textbf{Investigation} probe outputs across words using both representative embeddings in \texttt{Qwen2.5-3B}}
    \label{fig:Investigation_prom_qwen3b_last_aggregate}
\end{figure*}

\begin{figure*}[h]
    \centering
    \includegraphics[width=\linewidth]{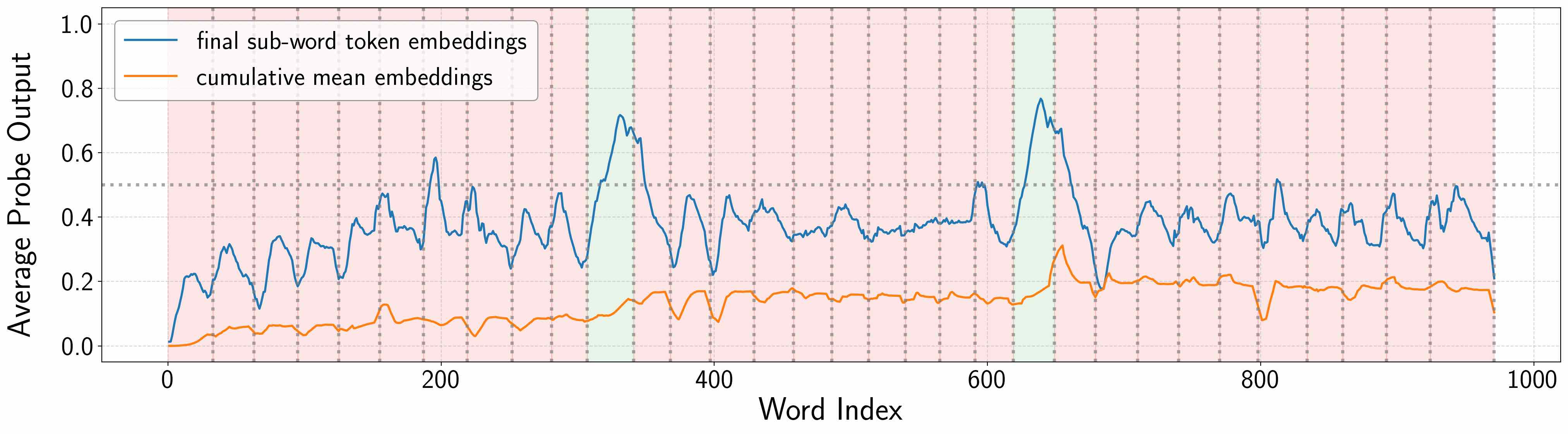}
    \caption{\textbf{Investigation} probe outputs across words using both representative embeddings in \texttt{Qwen2.5-7B}}
    \label{fig:Investigation_prom_qwen7b_last_aggregate}
\end{figure*}

\FloatBarrier
\clearpage

\subsection{Tracking Waxing and Waning of Democracy}

\subsubsection{Layer-Wise KDEs for Democracy Probe Outputs}

\begin{figure}[H]
    \centering
    \includegraphics[width=\linewidth]{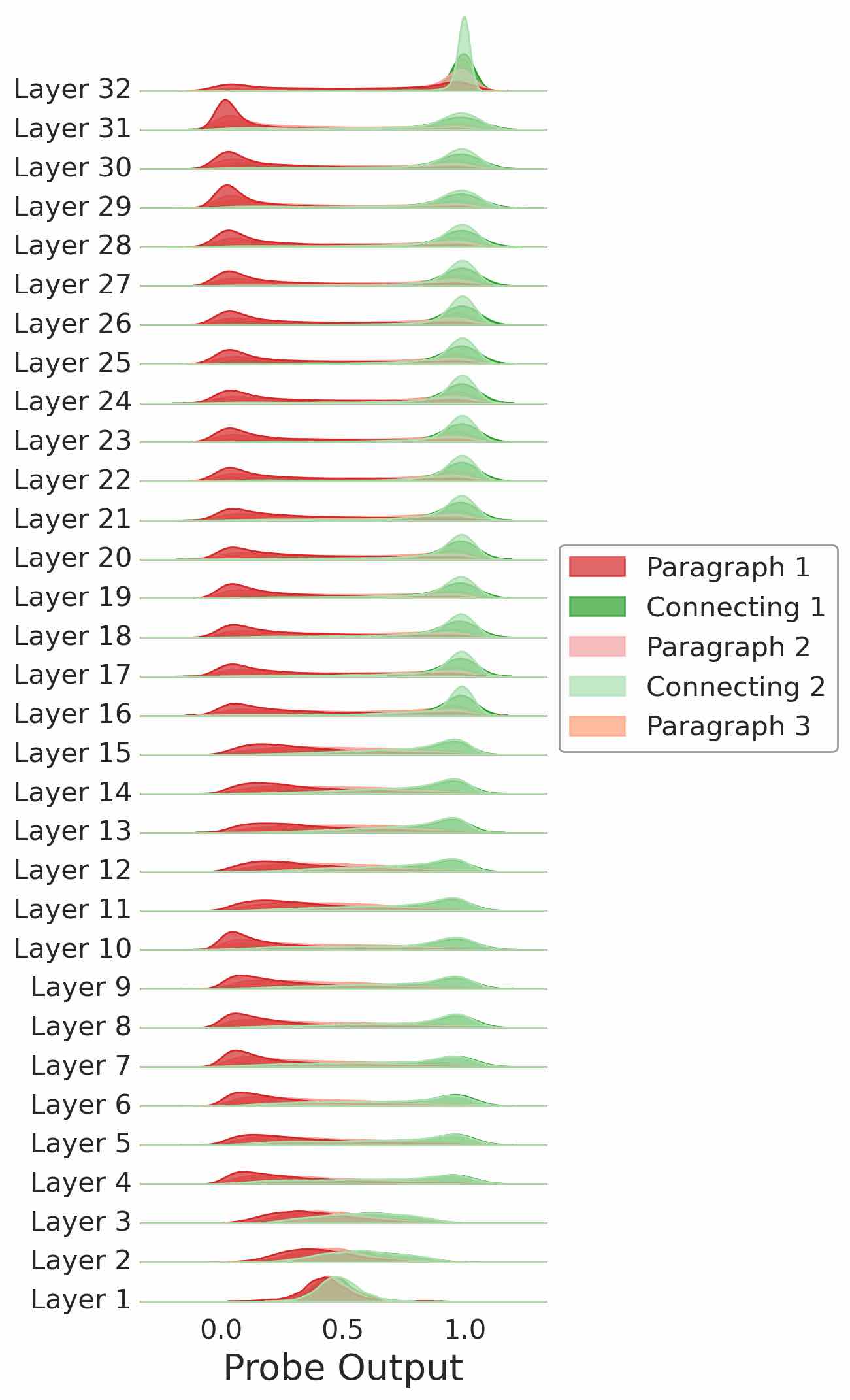}
    \caption{Layer-wise KDEs for \textbf{democracy} probe outputs in \texttt{Llama-3-8B}}
    \label{fig:democracy_KDE_llama}
\end{figure}

\begin{figure}[H]
    \centering
    \includegraphics[width=\linewidth]{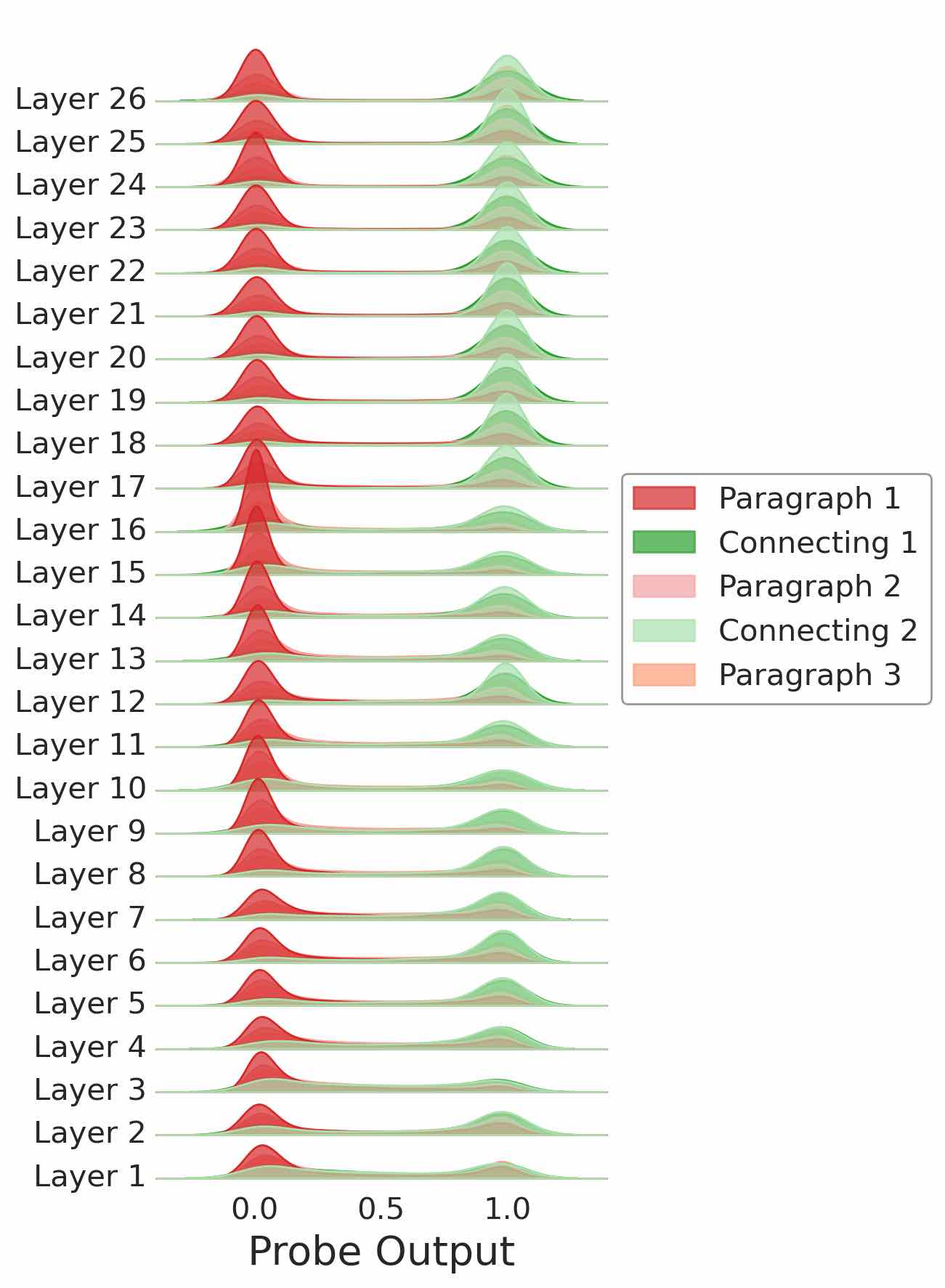}
    \caption{Layer-wise KDEs for \textbf{democracy} probe outputs in \texttt{Gemma-2-2B}}
    \label{fig:democracy_KDE_gemma2b}
\end{figure}

\begin{figure}[H]
    \centering
    \includegraphics[width=\linewidth]{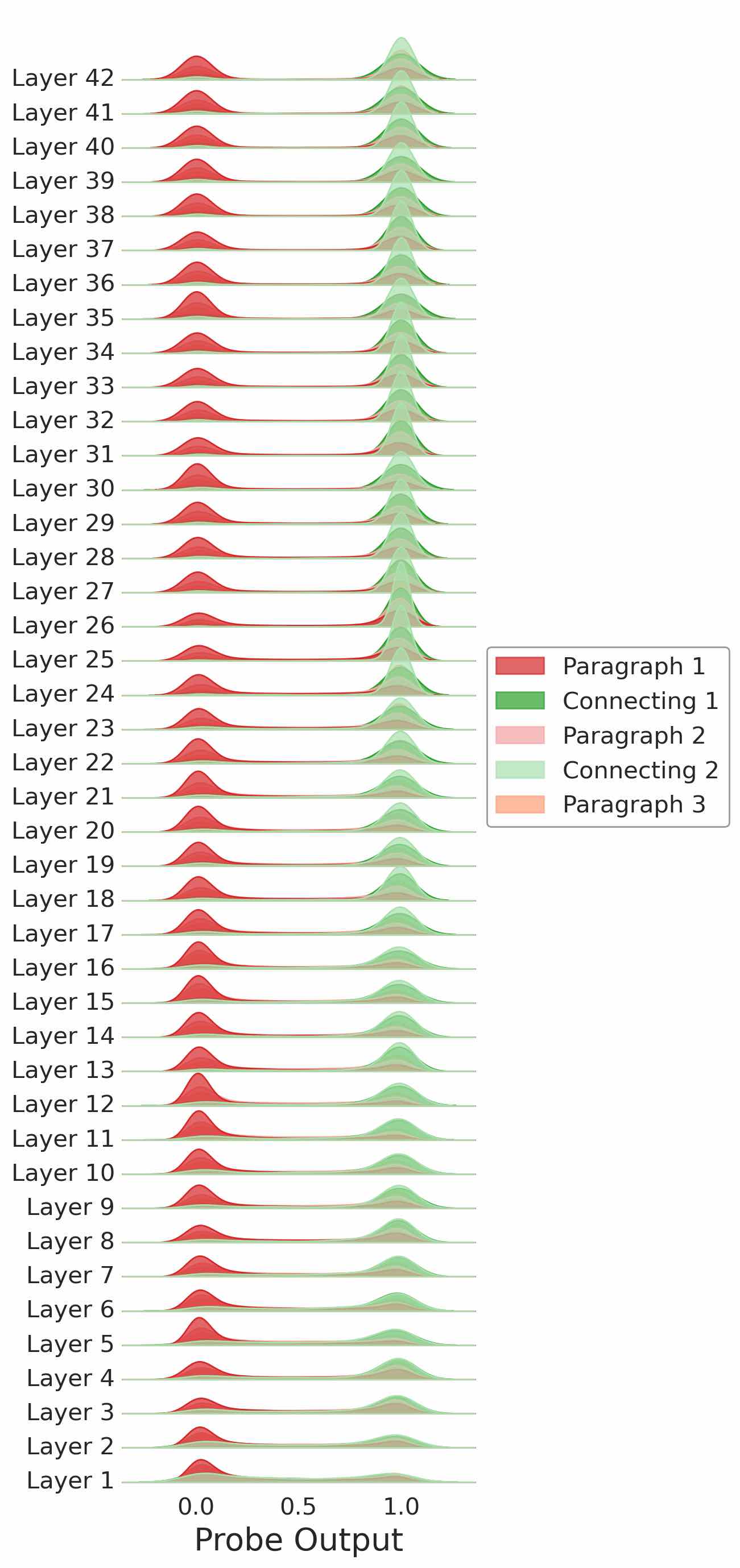}
    \caption{Layer-wise KDEs for \textbf{democracy} probe outputs in \texttt{Gemma-2-9B}}
    \label{fig:democracy_KDE_gemma9b}
\end{figure}

\begin{figure}[H]
    \centering
    \includegraphics[width=\linewidth]{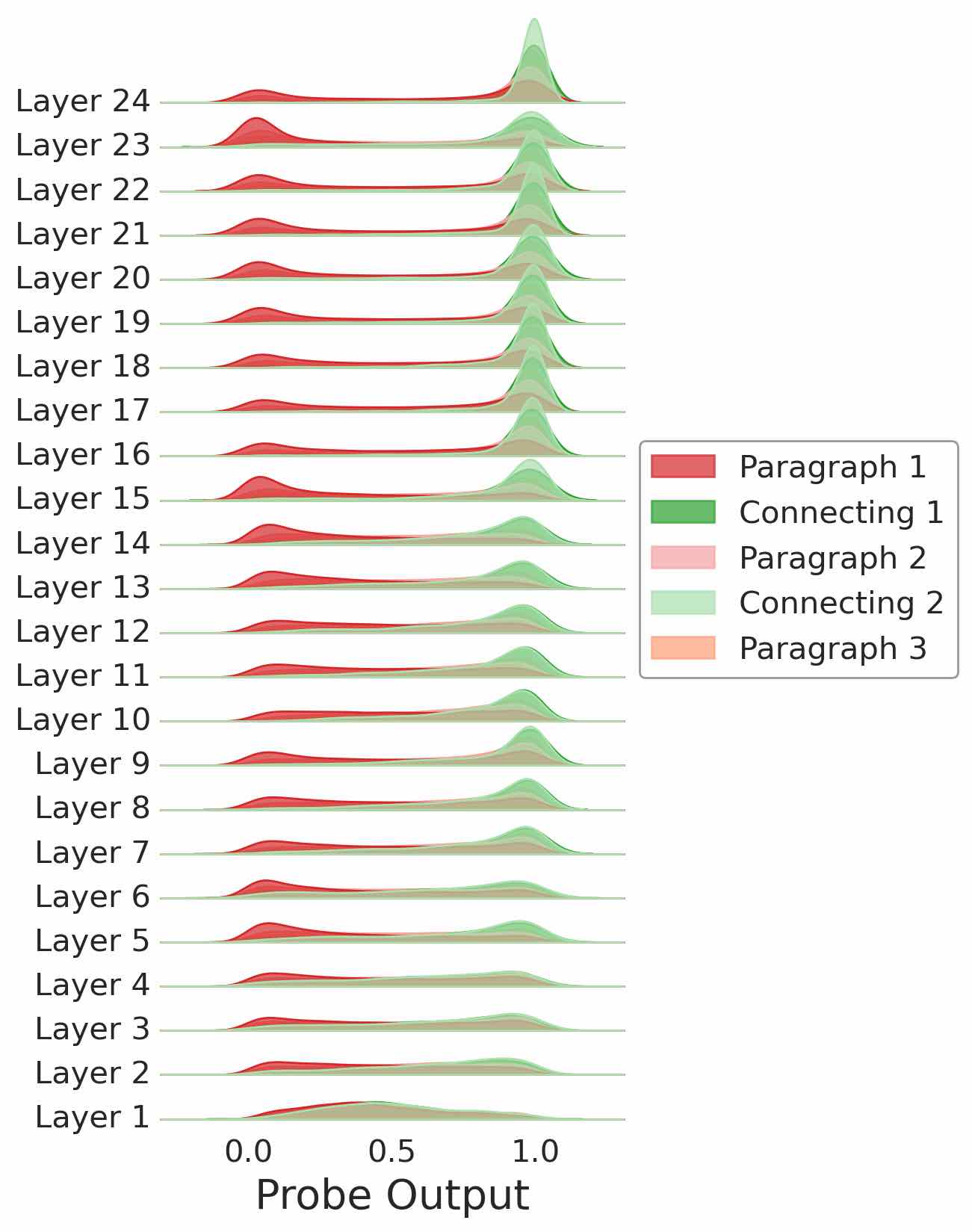}
    \caption{Layer-wise KDEs for \textbf{democracy} probe outputs in \texttt{Qwen2.5-0.5B}}
    \label{fig:democracy_KDE_qwen0p5b}
\end{figure}

\begin{figure}[H]
    \centering
    \includegraphics[width=\linewidth]{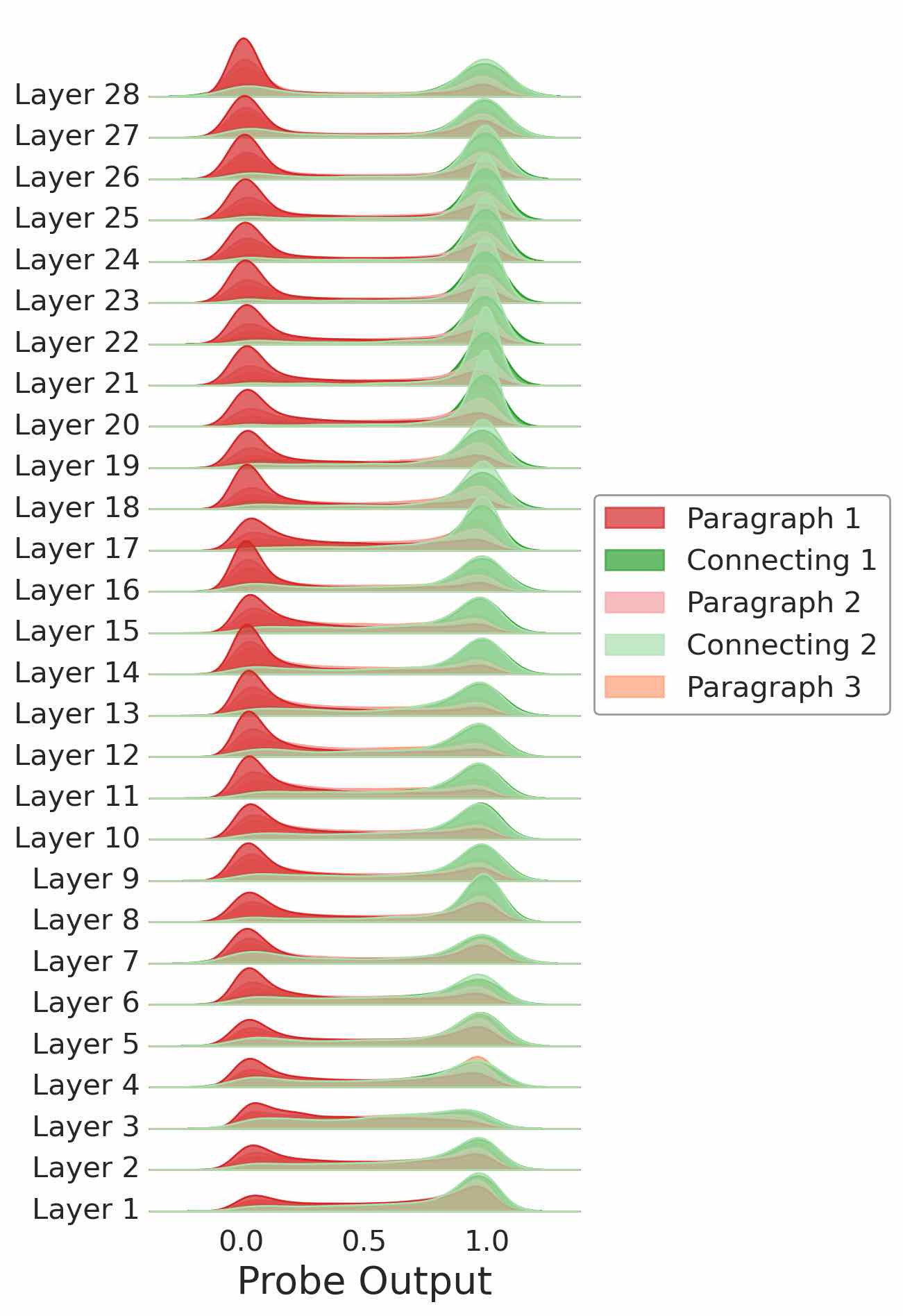}
    \caption{Layer-wise KDEs for \textbf{democracy} probe outputs in \texttt{Qwen2.5-1.5B}}
    \label{fig:democracy_KDE_qwen1p5b}
\end{figure}

\begin{figure}[H]
    \centering
    \includegraphics[width=\linewidth]{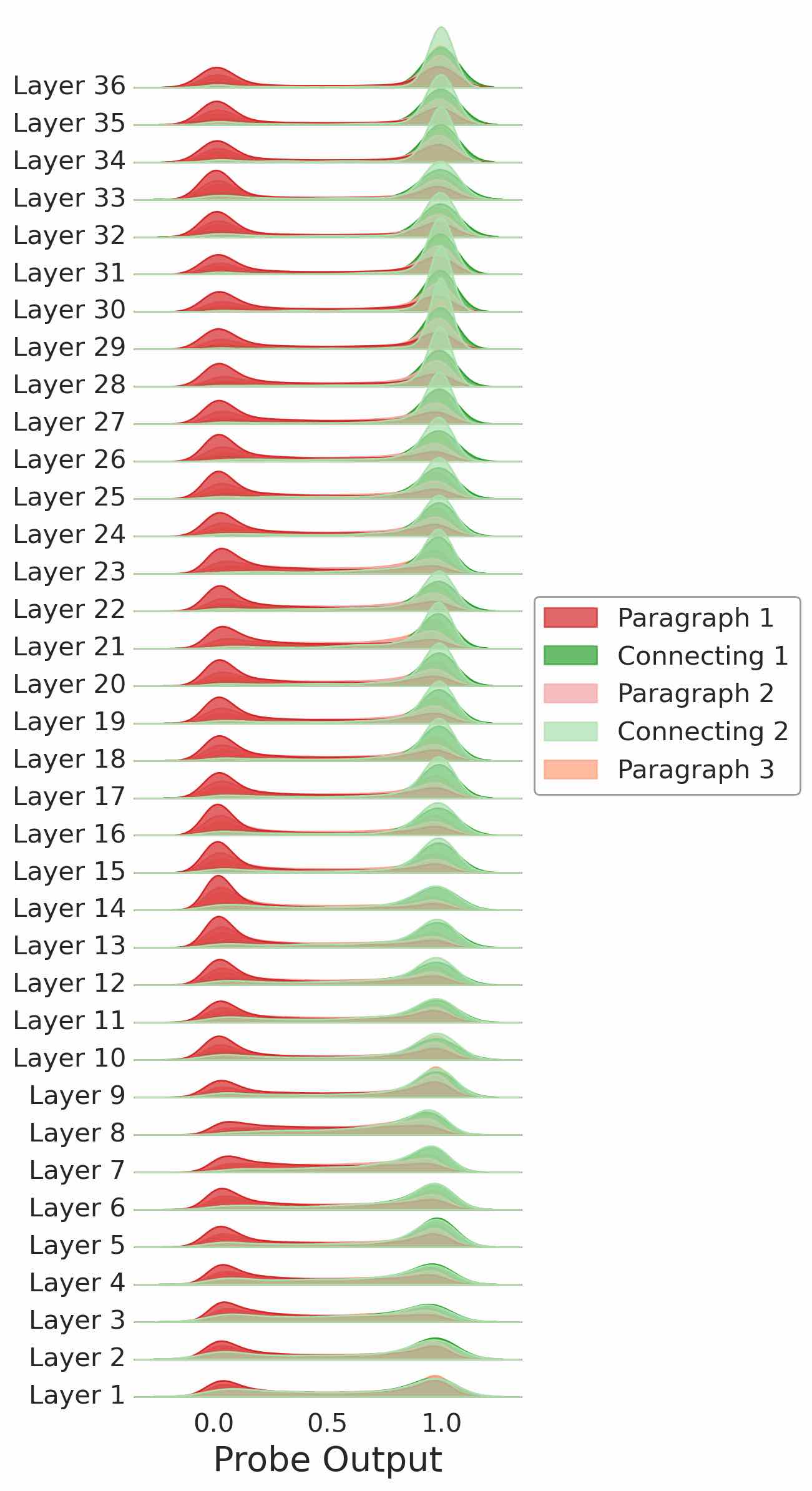}
    \caption{Layer-wise KDEs for \textbf{democracy} probe outputs in \texttt{Qwen2.5-3B}}
    \label{fig:democracy_KDE_qwen3b}
\end{figure}

\begin{figure}[H]
    \centering
    \includegraphics[width=\linewidth]{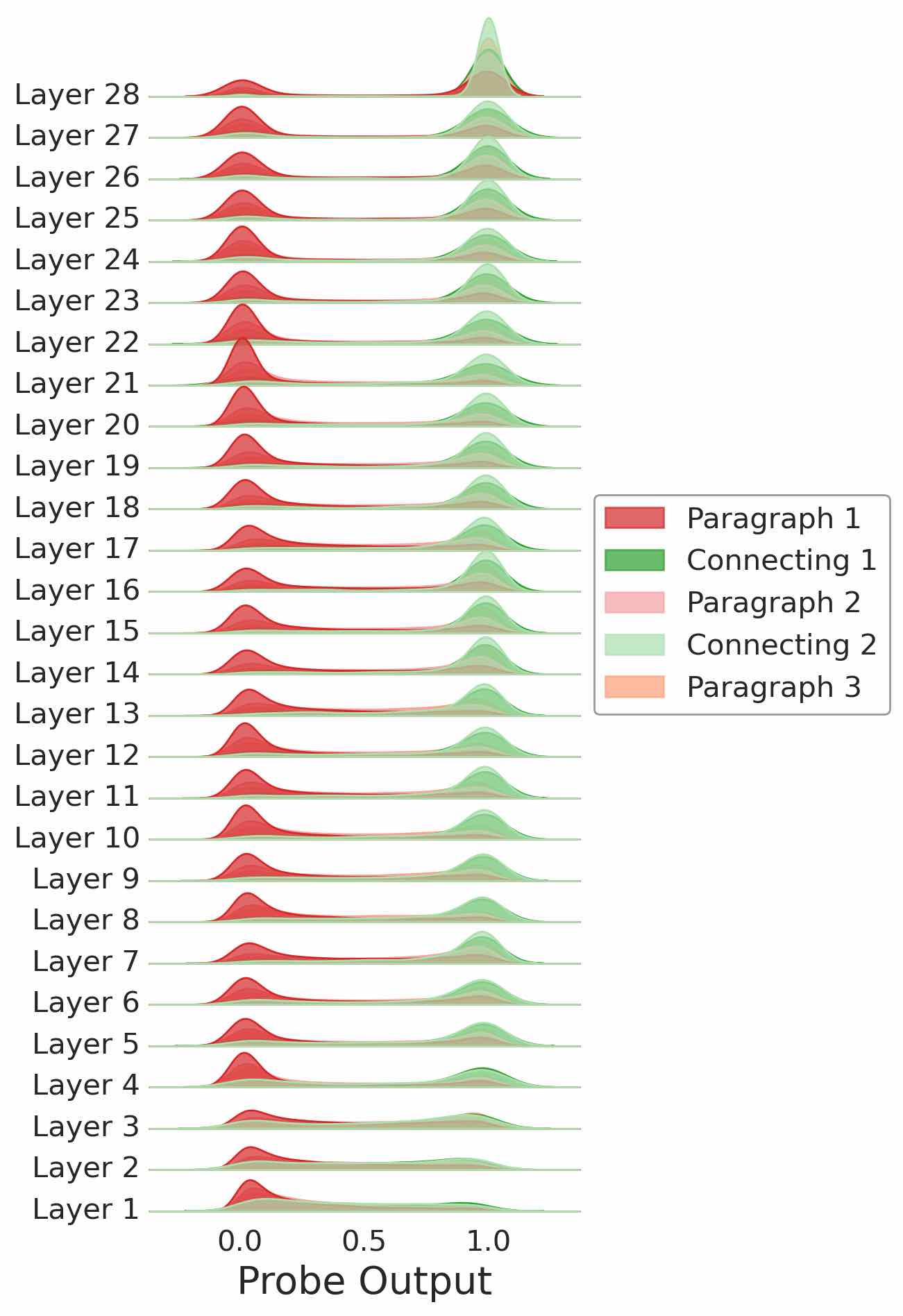}
    \caption{Layer-wise KDEs for \textbf{democracy} probe outputs in \texttt{Qwen2.5-7B}}
    \label{fig:democracy_KDE_qwen7b}
\end{figure}

\subsubsection{Democracy Probe Results for Best Layers}

\begin{figure*}[h]
    \centering
    \includegraphics[width=\linewidth]{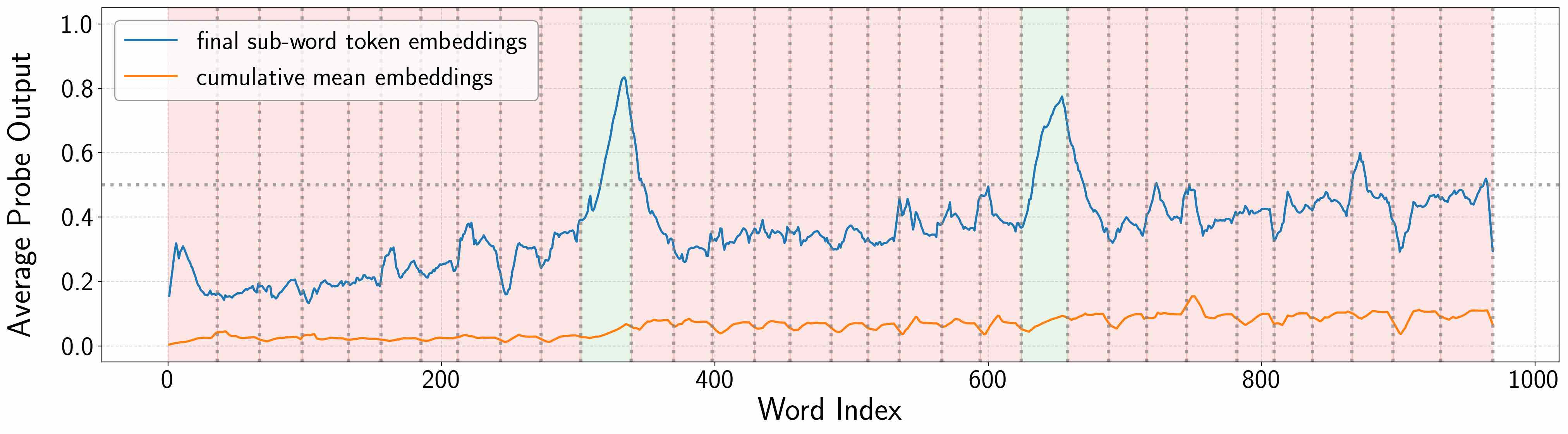}
    \caption{\textbf{Democracy} probe outputs across words using both representative embeddings in \texttt{Llama-3-8B}}
    \label{fig:Democracy_prom_llama_last_aggregate}
\end{figure*}

\begin{figure*}[h]
    \centering
    \includegraphics[width=\linewidth]{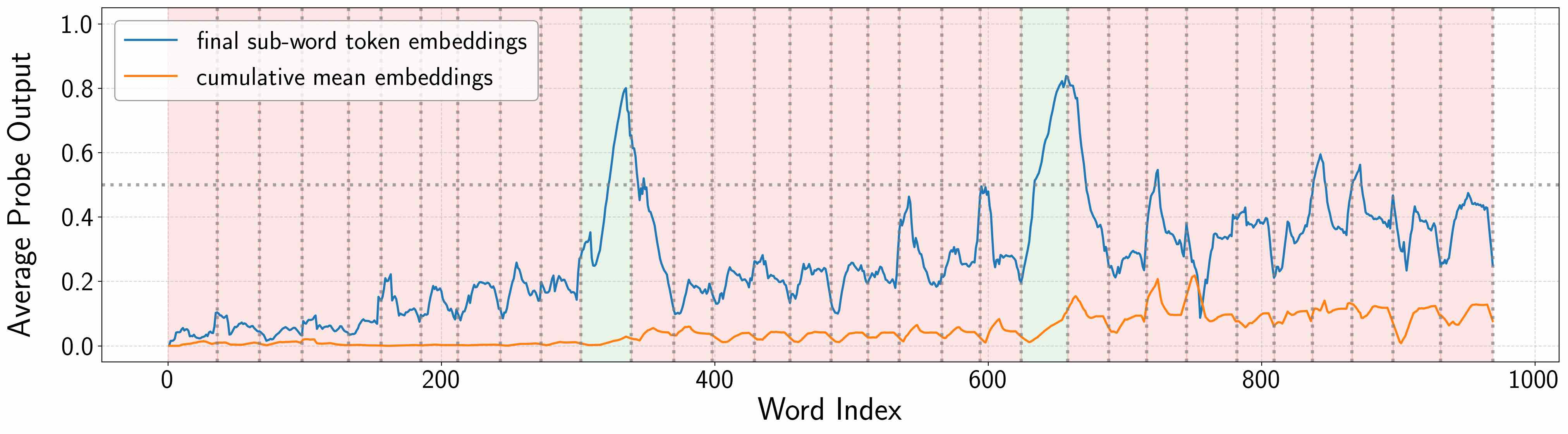}
    \caption{\textbf{Democracy} probe outputs across words using both representative embeddings in \texttt{Gemma-2-2B}}
    \label{fig:Democracy_prom_gemma2b_last_aggregate}
\end{figure*}

\begin{figure*}[h]
    \centering
    \includegraphics[width=\linewidth]{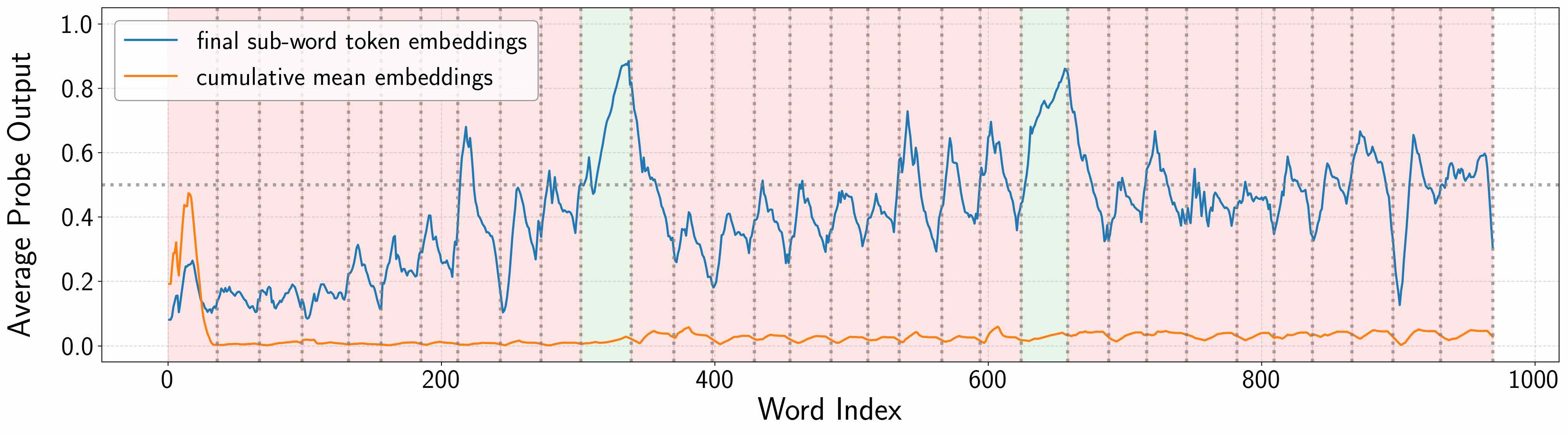}
    \caption{\textbf{Democracy} probe outputs across words using both representative embeddings in \texttt{Gemma-2-9B}}
    \label{fig:Democracy_prom_gemma9b_last_aggregate}
\end{figure*}

\begin{figure*}[h]
    \centering
    \includegraphics[width=\linewidth]{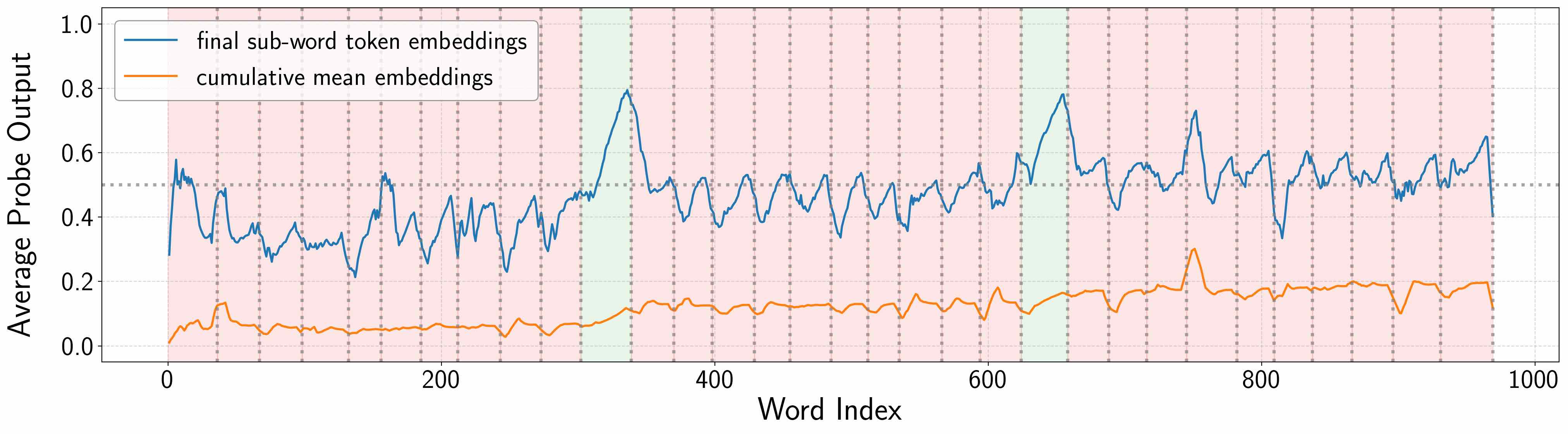}
    \caption{\textbf{Democracy} probe outputs across words using both representative embeddings in \texttt{Qwen2.5-0.5B}}
    \label{fig:Democracy_prom_qwen0p5b_last_aggregate}
\end{figure*}

\begin{figure*}[h]
    \centering
    \includegraphics[width=\linewidth]{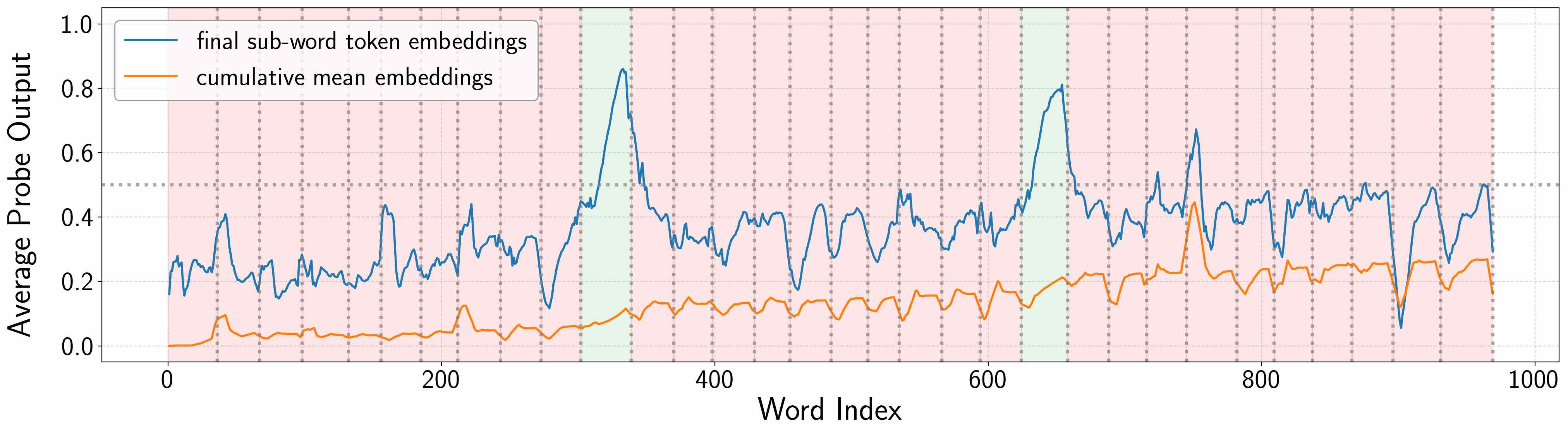}
    \caption{\textbf{Democracy} probe outputs across words using both representative embeddings in \texttt{Qwen2.5-1.5B}}
    \label{fig:Democracy_prom_qwen1p5b_last_aggregate}
\end{figure*}

\begin{figure*}[h]
    \centering
    \includegraphics[width=\linewidth]{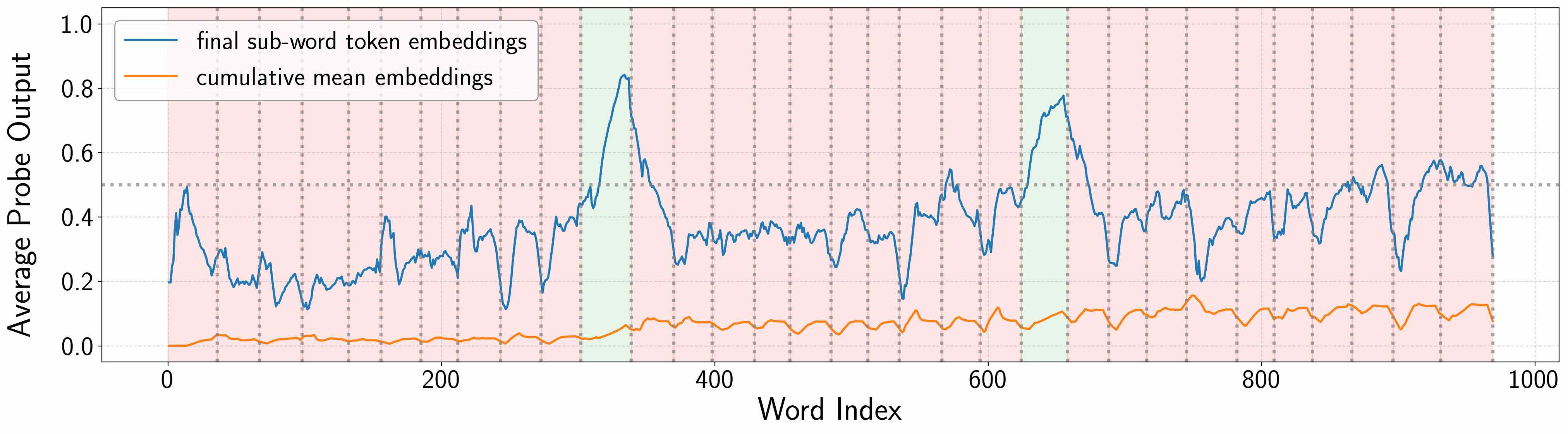}
    \caption{\textbf{Democracy} probe outputs across words using both representative embeddings in \texttt{Qwen2.5-3B}}
    \label{fig:Democracy_prom_qwen3b_last_aggregate}
\end{figure*}

\begin{figure*}[h]
    \centering
    \includegraphics[width=\linewidth]{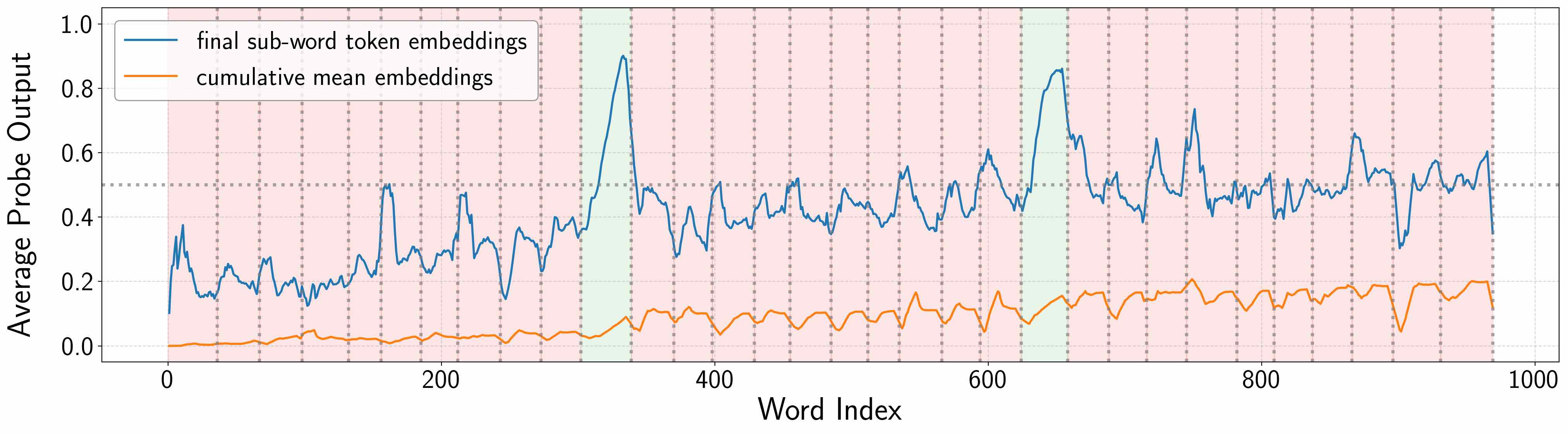}
    \caption{\textbf{Democracy} probe outputs across words using both representative embeddings in \texttt{Qwen2.5-7B}}
    \label{fig:Democracy_prom_qwen7b_last_aggregate}
\end{figure*}

\FloatBarrier
\clearpage

\subsection{Tracking Waxing and Waning of Envy}

\subsubsection{Layer-Wise KDEs for Envy Probe Outputs}

\begin{figure}[H]
    \centering
    \includegraphics[width=\linewidth]{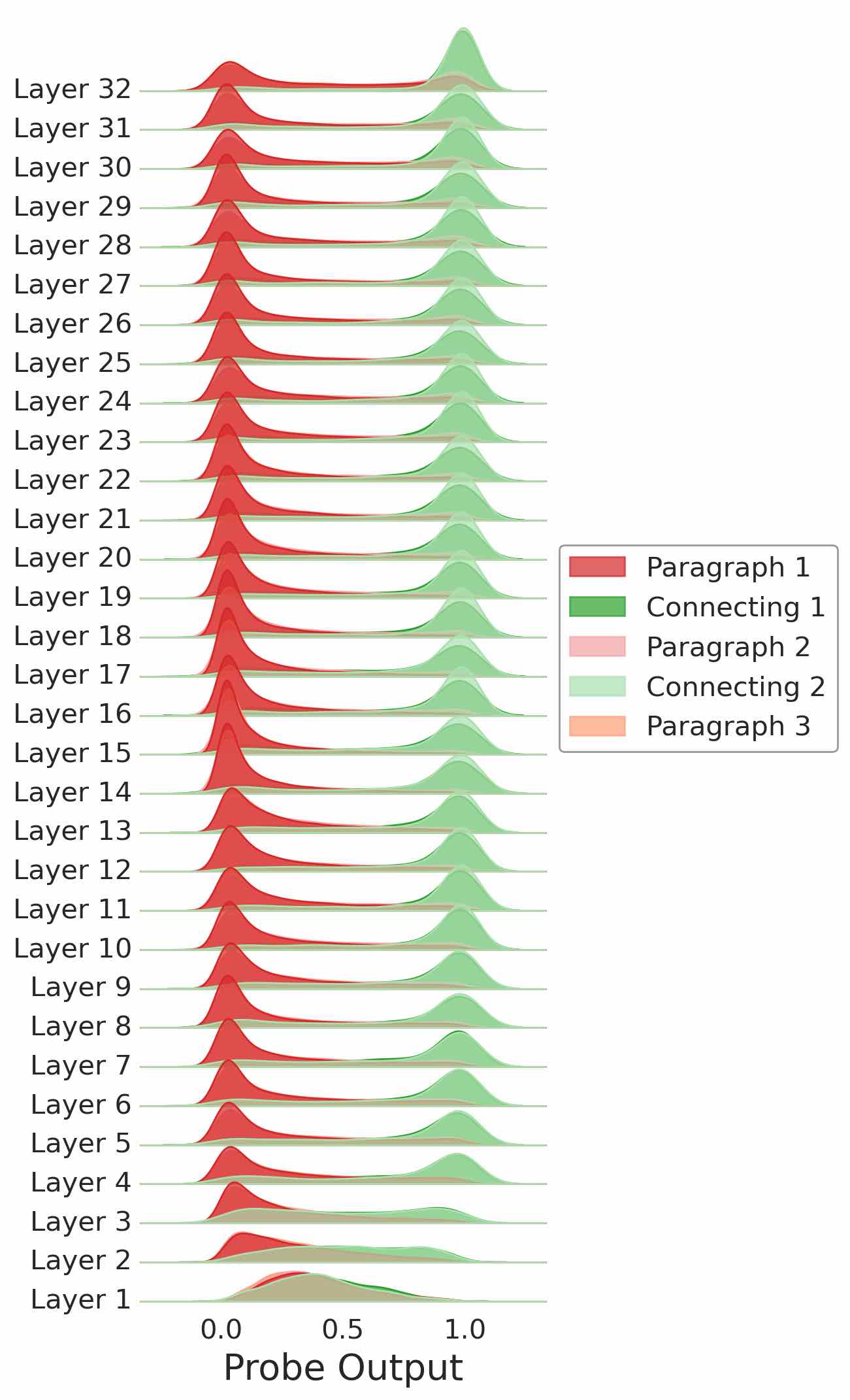}
    \caption{Layer-wise KDEs for \textbf{envy} probe outputs in \texttt{Llama-3-8B}}
    \label{fig:envy_KDE_llama}
\end{figure}

\begin{figure}[H]
    \centering
    \includegraphics[width=\linewidth]{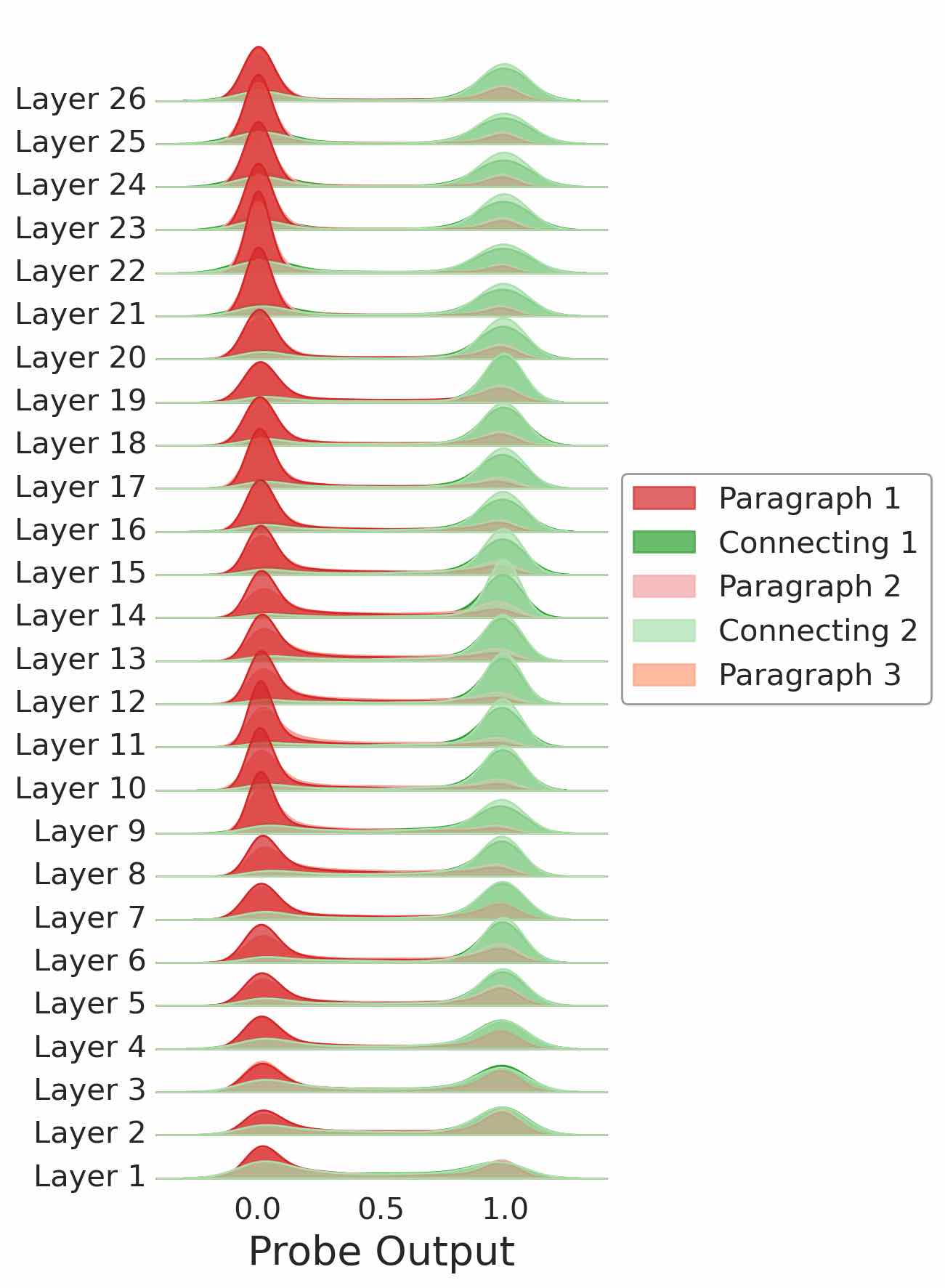}
    \caption{Layer-wise KDEs for \textbf{envy} probe outputs in \texttt{Gemma-2-2B}}
    \label{fig:envy_KDE_gemma2b}
\end{figure}

\begin{figure}[H]
    \centering
    \includegraphics[width=\linewidth]{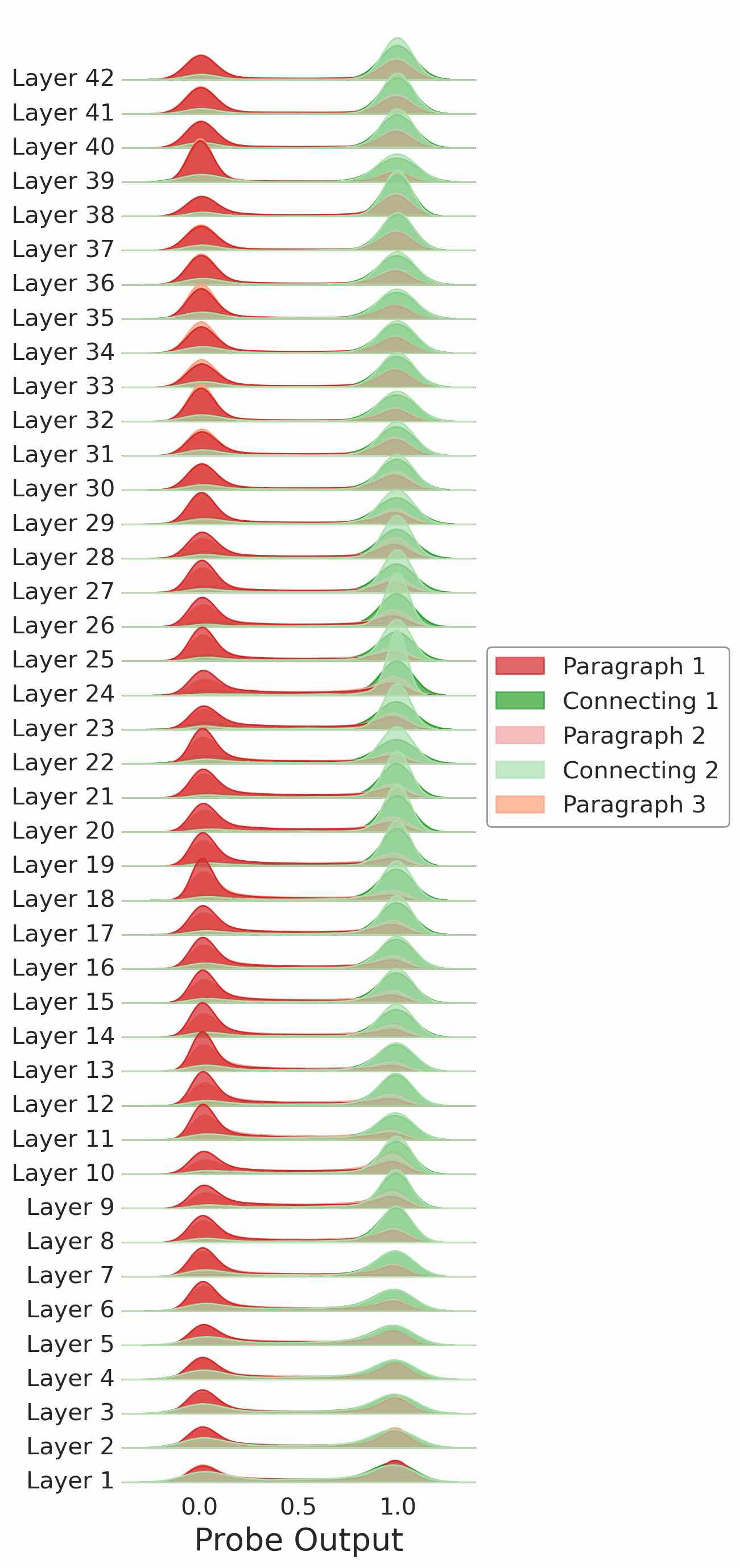}
    \caption{Layer-wise KDEs for \textbf{envy} probe outputs in \texttt{Gemma-2-9B}}
    \label{fig:envy_KDE_gemma9b}
\end{figure}

\begin{figure}[H]
    \centering
    \includegraphics[width=\linewidth]{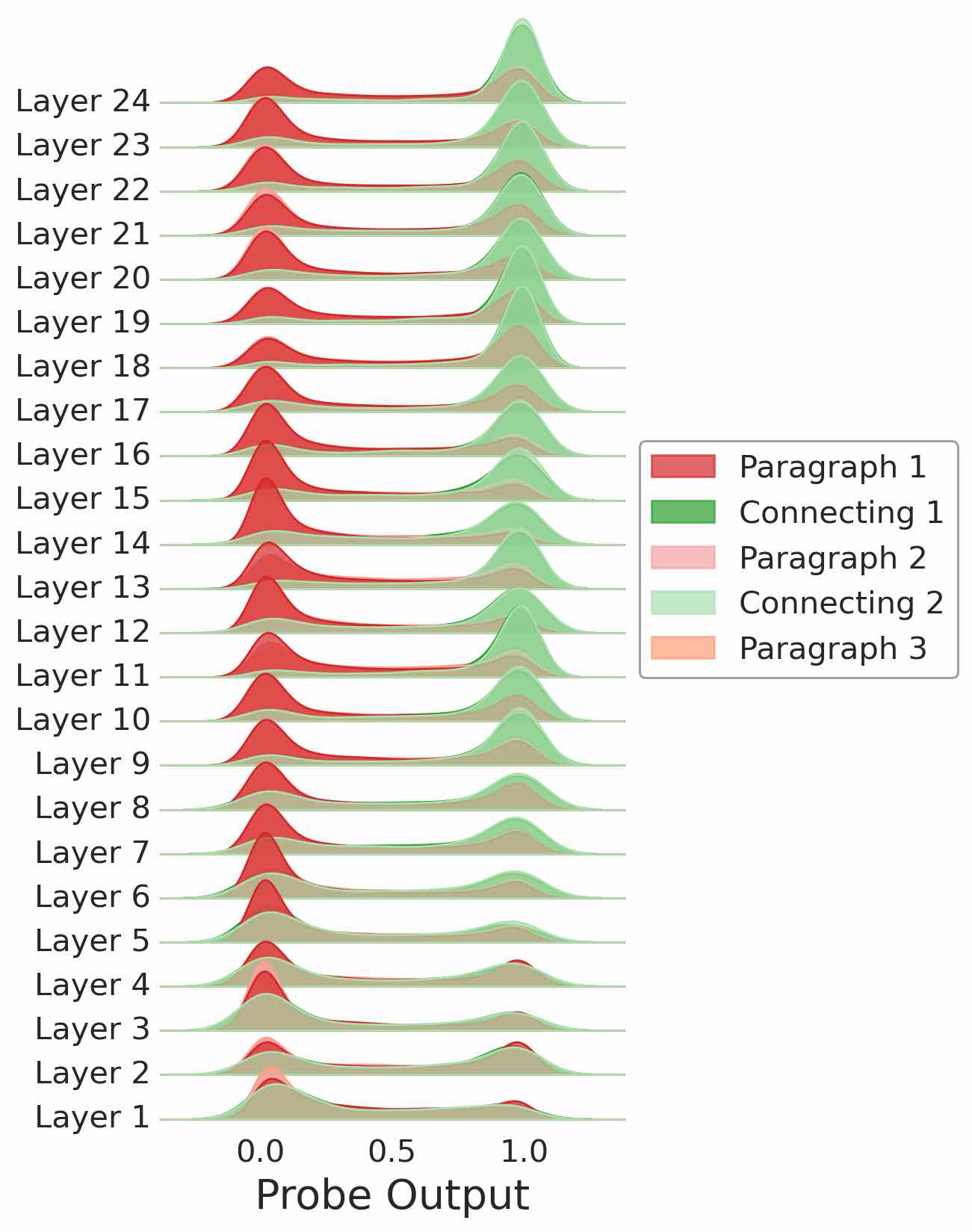}
    \caption{Layer-wise KDEs for \textbf{envy} probe outputs in \texttt{Qwen2.5-0.5B}}
    \label{fig:envy_KDE_qwen0p5b}
\end{figure}

\begin{figure}[H]
    \centering
    \includegraphics[width=\linewidth]{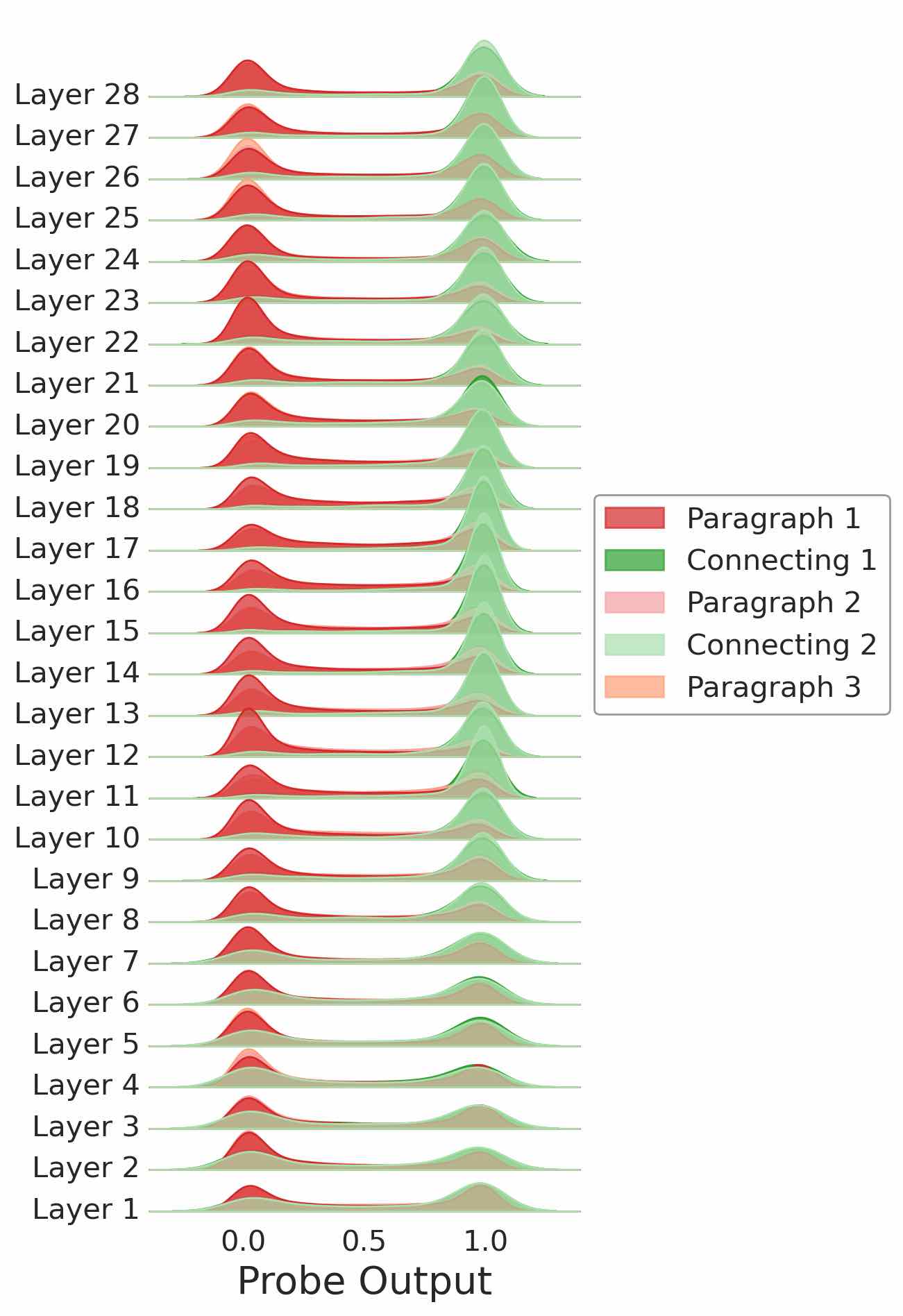}
    \caption{Layer-wise KDEs for \textbf{envy} probe outputs in \texttt{Qwen2.5-1.5B}}
    \label{fig:envy_KDE_qwen1p5b}
\end{figure}

\begin{figure}[H]
    \centering
    \includegraphics[width=\linewidth]{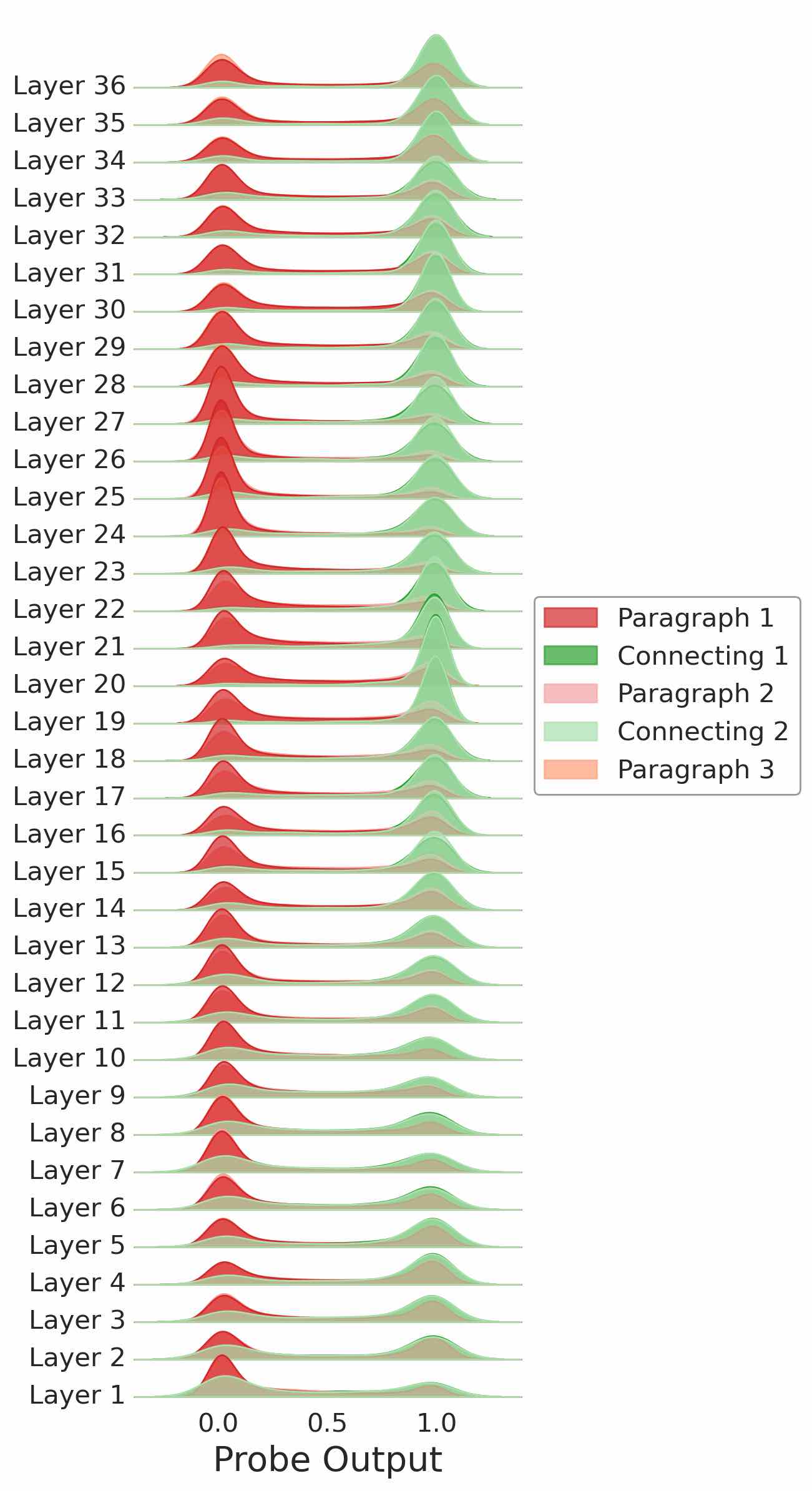}
    \caption{Layer-wise KDEs for \textbf{envy} probe outputs in \texttt{Qwen2.5-3B}}
    \label{fig:envy_KDE_qwen3b}
\end{figure}

\begin{figure}[H]
    \centering
    \includegraphics[width=\linewidth]{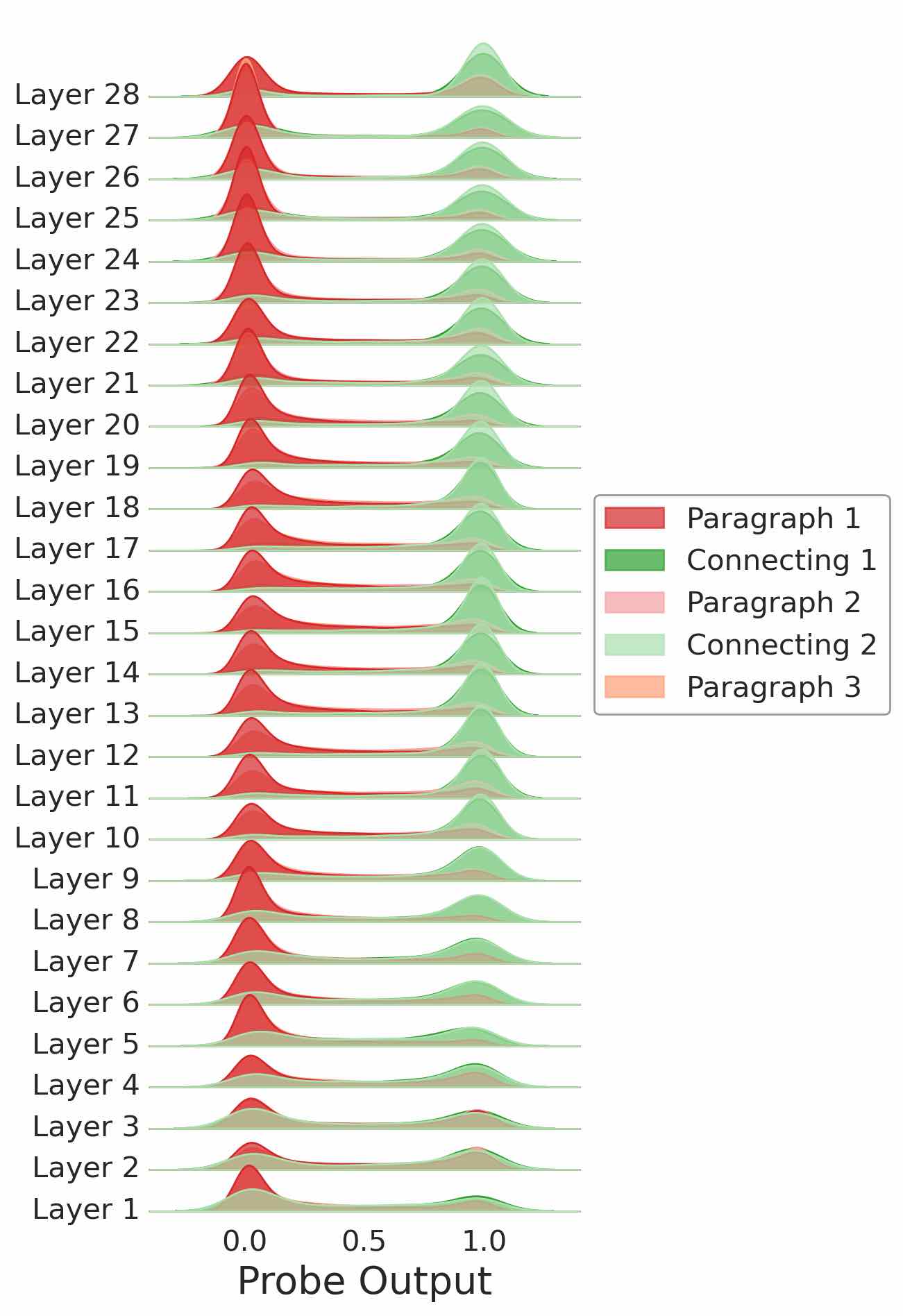}
    \caption{Layer-wise KDEs for \textbf{envy} probe outputs in \texttt{Qwen2.5-7B}}
    \label{fig:envy_KDE_qwen7b}
\end{figure}

\subsubsection{Envy Probe Results for Best Layers}

\begin{figure*}[h]
    \centering
    \includegraphics[width=\linewidth]{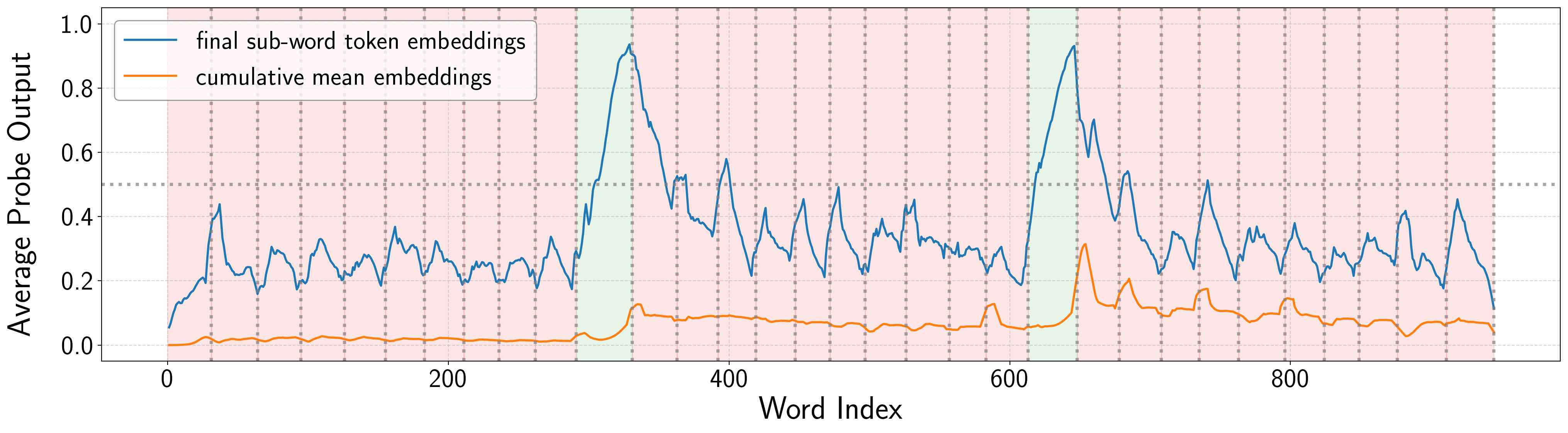}
    \caption{\textbf{Envy} probe outputs across words using both representative embeddings in \texttt{Llama-3-8B}}
    \label{fig:Envy_prom_llama_last_aggregate}
\end{figure*}

\begin{figure*}[h]
    \centering
    \includegraphics[width=\linewidth]{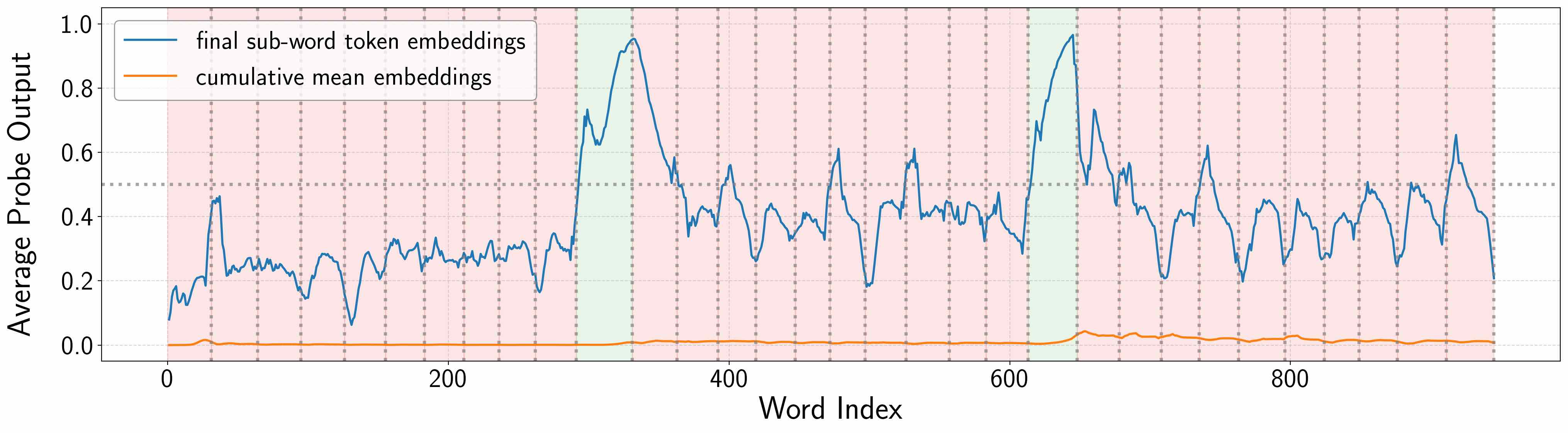}
    \caption{\textbf{Envy} probe outputs across words using both representative embeddings in \texttt{Gemma-2-2B}}
    \label{fig:Envy_prom_gemma2b_last_aggregate}
\end{figure*}

\begin{figure*}[h]
    \centering
    \includegraphics[width=\linewidth]{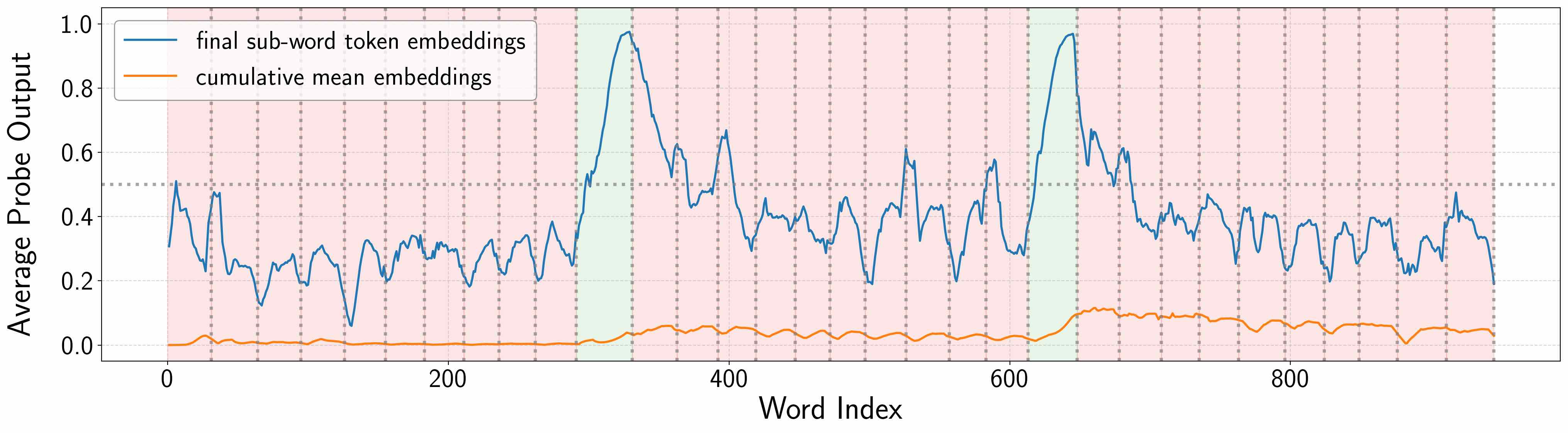}
    \caption{\textbf{Envy} probe outputs across words using both representative embeddings in \texttt{Gemma-2-9B}}
    \label{fig:Envy_prom_gemma9b_last_aggregate}
\end{figure*}

\begin{figure*}[h]
    \centering
    \includegraphics[width=\linewidth]{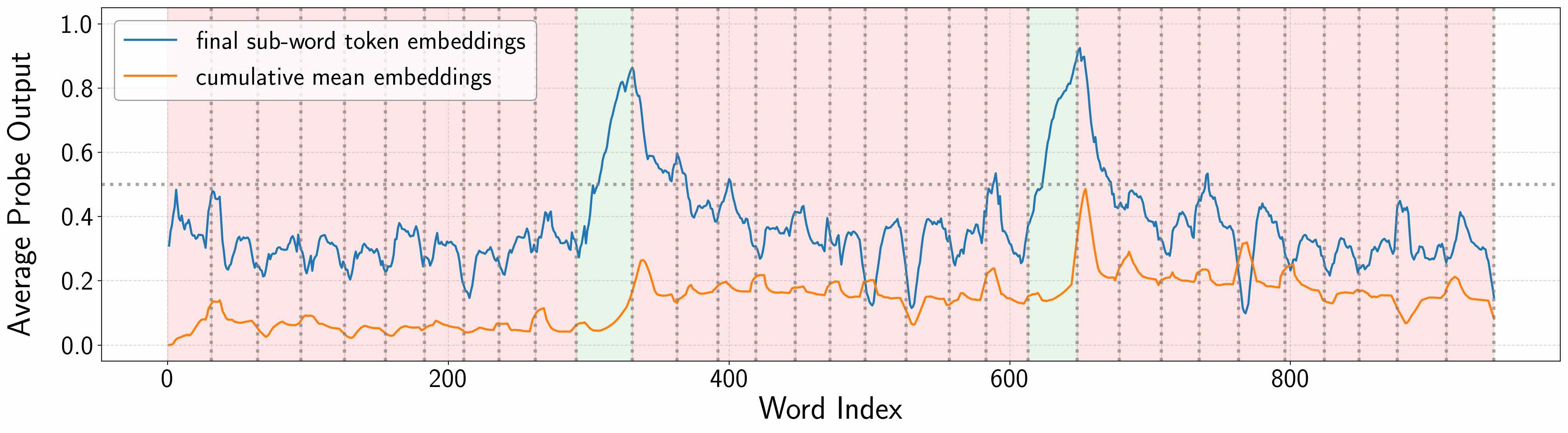}
    \caption{\textbf{Envy} probe outputs across words using both representative embeddings in \texttt{Qwen2.5-0.5B}}
    \label{fig:Envy_prom_qwen0p5b_last_aggregate}
\end{figure*}

\begin{figure*}[h]
    \centering
    \includegraphics[width=\linewidth]{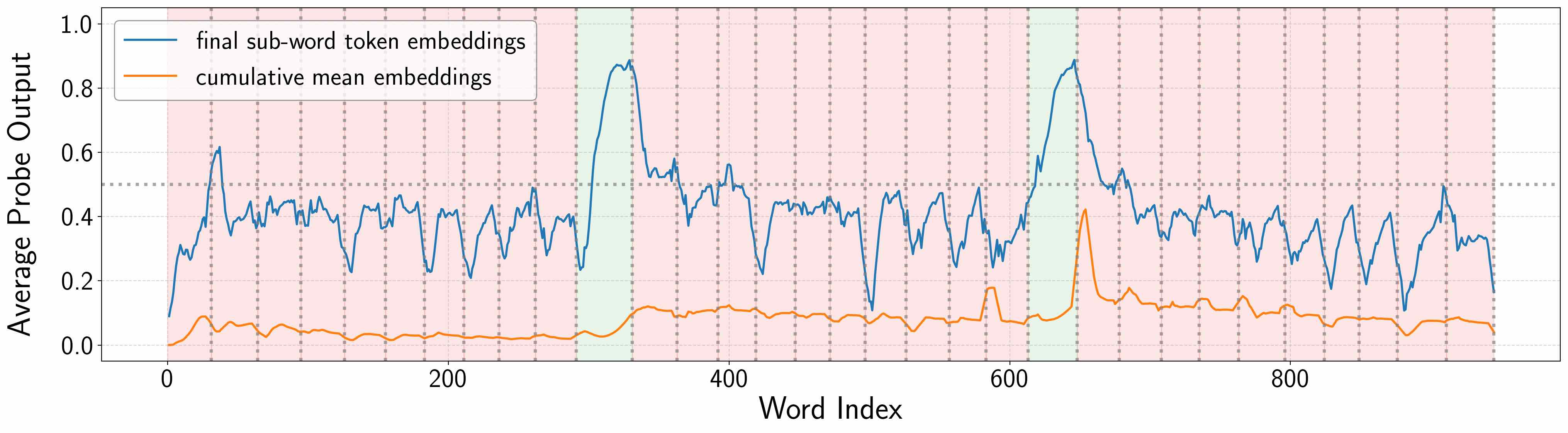}
    \caption{\textbf{Envy} probe outputs across words using both representative embeddings in \texttt{Qwen2.5-1.5B}}
    \label{fig:Envy_prom_qwen1p5b_last_aggregate}
\end{figure*}

\begin{figure*}[h]
    \centering
    \includegraphics[width=\linewidth]{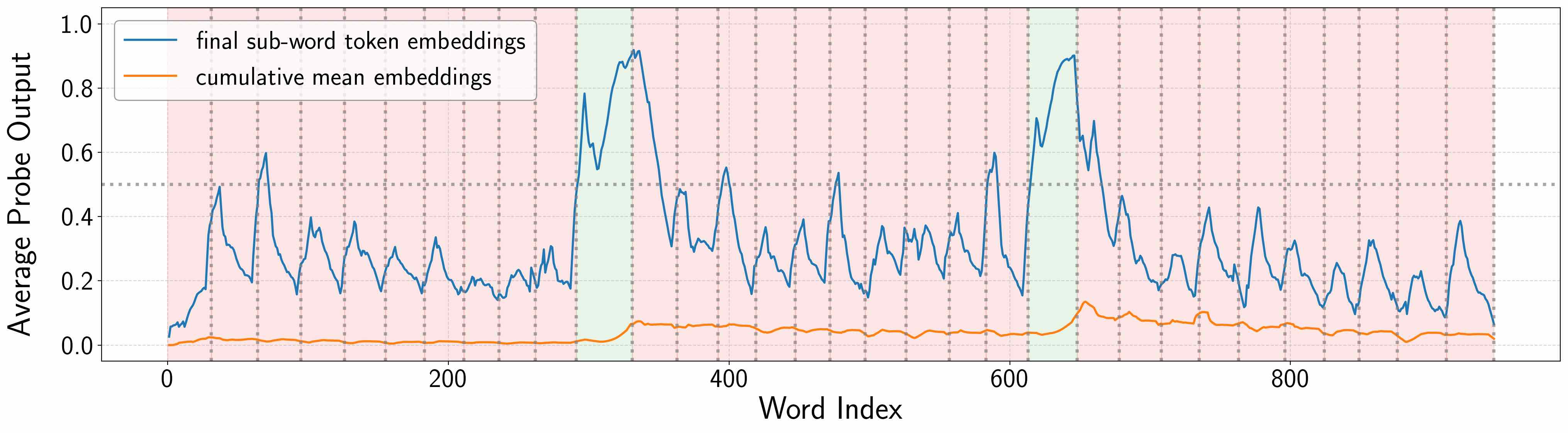}
    \caption{\textbf{Envy} probe outputs across words using both representative embeddings in \texttt{Qwen2.5-3B}}
    \label{fig:Envy_prom_qwen3b_last_aggregate}
\end{figure*}

\begin{figure*}[h]
    \centering
    \includegraphics[width=\linewidth]{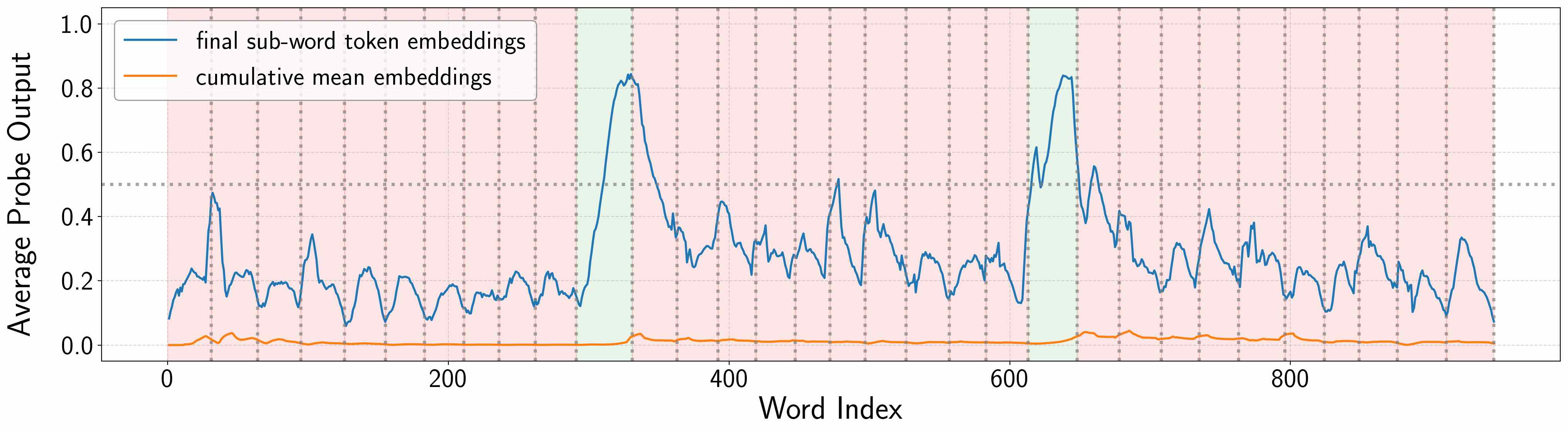}
    \caption{\textbf{Envy} probe outputs across words using both representative embeddings in \texttt{Qwen2.5-7B}}
    \label{fig:Envy_prom_qwen7b_last_aggregate}
\end{figure*}
\clearpage
\section{Dataset Overview} \label{appendix:datasets_included}
\setcounter{figure}{0}
\setcounter{table}{0}

We release all the datasets created and used in this paper to be used by others for concept exploration in LLMs. Table \ref{table:dataset_description} on the next page describes the contents of each file.

\begin{table*}
    \centering
    \begin{tabular}{|p{0.17\linewidth}|p{0.35\linewidth}|p{0.45\linewidth}|}
        \hline
        File name & General file description & File contents \\
        \hline\hline
        templates.txt & contains example templates used to create concept datasets & contains one template per line \\
        \hline
        ambition.csv & contains examples where \textbf{ambition} is either present or absent & file has 2 columns:
            \begin{itemize}
                \item input\_text: contains examples
                \item label: assigned label for each example
            \end{itemize} \\
        \hline
        investigation.csv & contains examples where \textbf{investigation} is either present or absent & file has 2 columns:
            \begin{itemize}
                \item input\_text: contains examples
                \item label: assigned label for each example
            \end{itemize} \\
        \hline
        democracy.csv & contains examples where \textbf{democracy} is either present or absent & file has 2 columns:
            \begin{itemize}
                \item input\_text: contains examples
                \item label: assigned label for each example
            \end{itemize} \\
        \hline
        envy.csv & contains examples where \textbf{envy} is either present or absent & file has 2 columns:
            \begin{itemize}
                \item input\_text: contains examples
                \item label: assigned label for each example
            \end{itemize} \\
        \hline
        amb\_strength.csv & contains 32-sentence stories where \textbf{ambition} is present in only two distant sentences per story & file has 2 columns:
            \begin{itemize}
                \item input\_text: contains stories
                \item label: list of assigned labels for each sentence in each story
            \end{itemize} \\
        \hline
        inv\_strength.csv & contains 32-sentence stories where \textbf{investigation} is present in only two distant sentences per story & file has 2 columns:
            \begin{itemize}
                \item input\_text: contains stories
                \item label: list of assigned labels for each sentence in each story
            \end{itemize} \\
        \hline
        dem\_strength.csv & contains 32-sentence stories where \textbf{democracy} is present in only two distant sentences per story & file has 2 columns:
            \begin{itemize}
                \item input\_text: contains stories
                \item label: list of assigned labels for each sentence in each story
            \end{itemize} \\
        \hline
        env\_strength.csv & contains 32-sentence stories where \textbf{envy} is present in only two distant sentences per story & file has 2 columns:
            \begin{itemize}
                \item input\_text: contains stories
                \item label: list of assigned labels for each sentence in each story
            \end{itemize} \\
        \hline
    \end{tabular}
    \caption{Description of released files for the datasets}
    \label{table:dataset_description}
\end{table*}

\clearpage

\end{document}